\documentclass[11pt,letterpaper]{mystyle}
\usepackage{textcomp}
\usepackage{graphicx}
\usepackage{fancyhdr}
\usepackage{tikz}
\usepackage{gensymb}
\usepackage[utf8]{inputenc} 
\usepackage[T1]{fontenc}
\usepackage[numbers]{natbib}
\usepackage{adjustbox}
\usepackage{siunitx}
\usepackage{breakurl}    
\usepackage{fontawesome5}

\usepackage[toc,page]{appendix} 
\usepackage{marvosym}

\usepackage{booktabs} 
\usepackage{multirow} 
\usepackage{geometry} 

\usepackage{setspace}

\usepackage{hyperref}
\usepackage[edges]{forest}
\usepackage{subcaption}
\usepackage{soul}
\usepackage{booktabs}
\usepackage{tabularx}
\usepackage{multirow}
\usepackage[utf8]{inputenc} 
\usepackage{booktabs}       
\usepackage{amsfonts}       
\usepackage{nicefrac}      
\usepackage{microtype}     
\usepackage{xcolor}        
\usepackage{colortbl}
\usepackage{enumitem}
\usepackage[numbers]{natbib}
\usepackage{amssymb}
\usepackage{wrapfig}
\usepackage{courier}
\usepackage{bxcoloremoji}
\usepackage{CJKutf8}
\usepackage{ragged2e}
\usepackage{longtable}
\usepackage{threeparttable}

\fancypagestyle{headstyle}{
    \fancyhead[L]{
        \includegraphics[width=60pt]{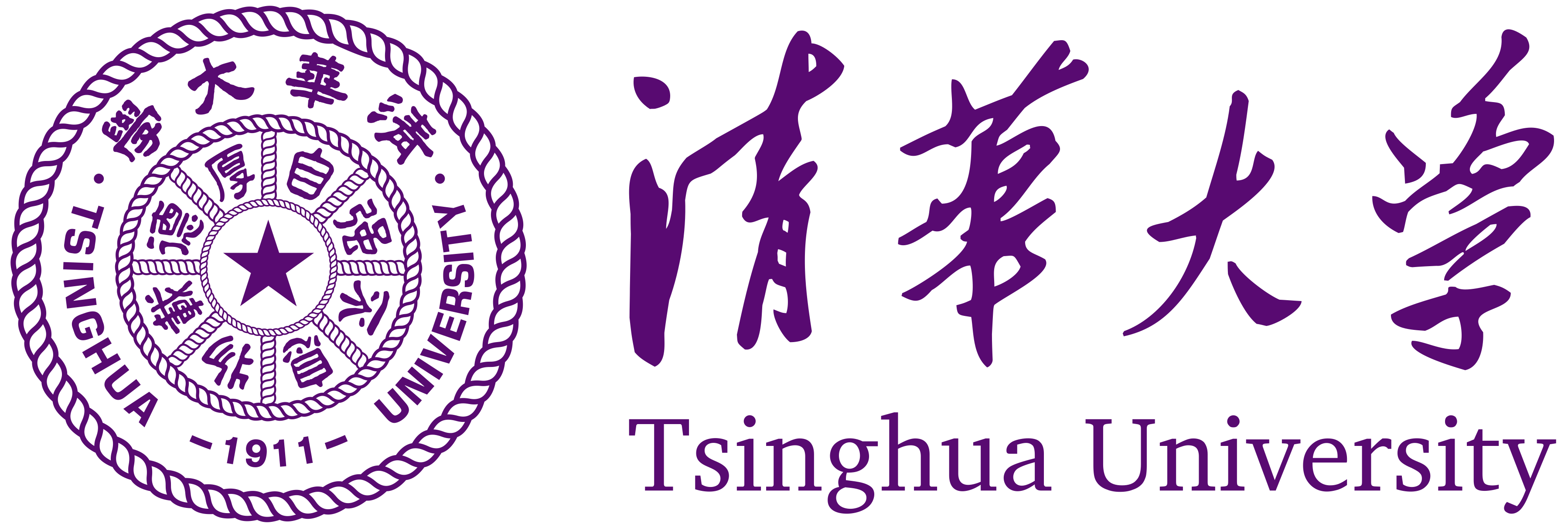}
    }
    \fancyhead[C]{
        \emph{Advanced Long-term Earth System Forecasting}
    }
    \fancyhead[R]{
        \includegraphics[width=100pt]{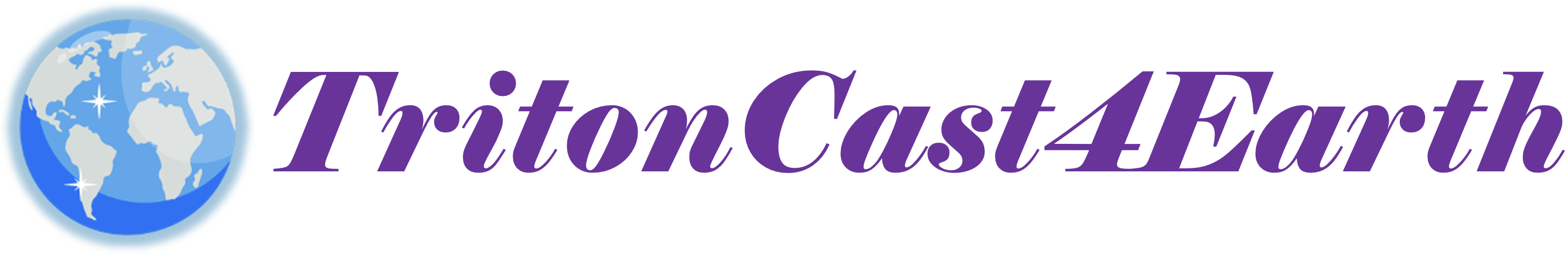}
    }

}

\definecolor{hidden-red}{RGB}{205, 44, 36}
\definecolor{hidden-blue}{RGB}{194,232,247}
\definecolor{hidden-orange}{RGB}{243,202,120}
\definecolor{hidden-green}{RGB}{34,139,34}
\definecolor{hidden-pink}{RGB}{255,245,247}
\definecolor{hidden-black}{RGB}{20,68,106}
\definecolor{purple}{RGB}{144,153,196}
\definecolor{yellow}{RGB}{255,228,123}
\definecolor{hidden-yellow}{RGB}{255,248,203}
\definecolor{tkcolor}{RGB}{224,223,255}
\definecolor{darkblue}{rgb}{0, 0.40, 0.75}

\hypersetup{colorlinks=true, citecolor=darkblue, linkcolor=darkblue, urlcolor=darkblue}
\tcbset{
  aibox/.style={
    width=\linewidth,
    top=8pt,
    bottom=4pt,
    colback=blue!6!white,
    colframe=black,
    colbacktitle=black,
    enhanced,
    center,
    attach boxed title to top left={yshift=-0.1in,xshift=0.15in},
    boxed title style={boxrule=0pt,colframe=white,},
  }
}

\tikzstyle{my-box}=[
rectangle,
draw=hidden-black,
rounded corners,
text opacity=1,
minimum height=1.5em,
minimum width=5em,
inner sep=2pt,
align=center,
fill opacity=.5,
]
\tikzstyle{leaf}=[
my-box,
minimum height=1.5em,
fill=hidden-red!20,
text=black,
align=left,
font=\normalsize,
inner xsep=5pt,
inner ysep=4pt,
]
\tikzstyle{leaf2}=[
my-box,
minimum height=1.5em,
fill=hidden-green!20,
text=black,
align=left,
font=\normalsize,
inner xsep=5pt,
inner ysep=4pt,
]
\tikzstyle{leaf3}=[
my-box,
minimum height=1.5em,
fill=yellow!32,
text=black,
align=left,
font=\normalsize,
inner xsep=5pt,
inner ysep=4pt,
align=left,
text width=45em,
]
\tikzstyle{leaf4}=[
my-box,
minimum height=1.5em,
fill=hidden-blue!57,
text=black,
align=left,
font=\normalsize,
inner xsep=5pt,
inner ysep=4pt,
]
\tikzstyle{leaf5}=[
my-box,
minimum height=1.5em,
fill=darkblue!15,
text=black,
align=left,
font=\normalsize,
inner xsep=5pt,
inner ysep=4pt,
]
\tikzstyle{leaf6}=[
my-box,
minimum height=1.5em,
fill=purple!30,
text=black,
align=left,
font=\normalsize,
inner xsep=5pt,
inner ysep=4pt,
]

\newtcolorbox{AIbox}[2][]{aibox,title=#2,#1}

\tcbset{
  takeawaybox/.style={
    width=\linewidth,
    top=8pt,
    bottom=4pt,
    colback=hidden-yellow,
    colframe=black,
    colbacktitle=black,
    enhanced,
    center,
    attach boxed title to top left={yshift=-0.1in,xshift=0.15in},
    boxed title style={boxrule=0pt,colframe=white,},
  }
}

\newtcolorbox{TakeawayBox}[2][]{takeawaybox,title=#2,#1}

\title{\textit{Advanced Long-term Earth System Forecasting}}
\def\method{TritonCast}

\author{
  Hao Wu$^{1,2, *, \text{\Letter}}$, 
  Yuan Gao$^{1,*, \text{\Letter}}$, 
  Ruijian Gou$^{3,*}$, 
  Xian Wu$^{2,*}$, 
  Chuhan Wu$^{2,*}$, 
  Huahui Yi$^{4,*}$, 
  Johannes Brandstetter$^{5,6}$, 
  Fan Xu$^{7,20}$, 
  Kun Wang$^{8}$,
  Penghao Zhao$^{2}$, 
  Hao Jia$^{2}$, 
  Qi Song$^{7}$, 
  Xinliang Liu$^{3}$, 
  Juncai He$^{1}$, 
  Shuhao Cao$^{9}$, 
  Huanshuo Dong$^{7}$, 
  Yanfei Xiang$^{1}$, 
  Fan Zhang$^{10}$,
  Haixin Wang$^{11}$, 
  Xingjian Shi$^{12}$, 
  Qiufeng Wang$^{13}$, 
  Shuaipeng Li$^{2}$, 
  Ruobing Xie$^{2}$, 
  Feng Tao$^{14}$,
  Yuxu Lu$^{15}$,
  Yu Guo$^{16}$,
  Yuntian Chen$^{17}$, 
  Yuxuan Liang$^{18}$, 
  Qingsong Wen$^{19}$, 
  Wanli Ouyang$^{10,20}$,
  Deliang Chen$^{1}$, 
  Niklas Boers$^{21,22}$,
  Xiaomeng Huang$^{1,\text{\Letter},\ddag}$
\\

\normalfont{
$^1$ Tsinghua University,
$^2$ Tencent,
$^3$ Ocean University of China,
$^4$ West China Biomedical Big Data Center,
$^{5}$ Johannes Kepler University (JKU) Linz,
$^{6}$ Emmi AI
$^7$ University of Science and Technology of China, 
$^8$ Nanyang Technological University,
$^9$ University of Missouri-Kansas City,
$^{10}$ The Chinese University of Hong Kong,
$^{11}$ University of California, Los Angeles,
$^{12}$ Boson AI,
$^{13}$ Southeast University,
$^{14}$ Cornell University, 
$^{15}$ The Hong Kong Polytechnic University
$^{16}$ City University of Hong Kong
$^{17}$ Eastern Institute of Technology, Ningbo
$^{18}$ Hong Kong University of Science and Technology (Guangzhou),
$^{19}$ Squirrel Ai Learning,
$^{20}$ Shenzhen Loop Area Institute
$^{21}$ Technical University of Munich
$^{22}$ Potsdam Institute for Climate Impact Research

}}

\begin{document}

\begin{abstract}
\textbf{\large{Abstract}:} Reliable long-term forecasting of Earth system dynamics with data-driven models is fundamentally limited by instabilities in current artificial intelligence (AI) models during extended autoregressive simulations. These failures often originate from inherent spectral bias, leading to inadequate representation of critical high-frequency, small-scale processes and subsequent uncontrolled error amplification. Inspired by the nested grids used in numerical models to resolve small scales, we present \method{}, an AI-based Earth system model designed for long-term forecasts. At the core of its design is a dedicated latent dynamical core, which ensures the long-term stability of the macro-evolution at a coarse scale. An outer structure then fuses this stable trend with fine-grained local details. This design effectively mitigates the spectral bias caused by cross-scale interactions. In atmospheric science, \method{} achieves state-of-the-art accuracy on state-of-the-art benchmarks, while demonstrating exceptional long-term stability in year-long autoregressive global forecasts and  multi-year, drift-free climate simulations that span the entire available $2500$-day test period. In oceanography, \method{} extends skillful eddy forecast to $120$ days and exhibits unprecedented zero-shot cross-resolution generalization. Ablation studies reveal that this performance stems from the synergistic interplay of the architecture's core components. TritonCast thus offers a promising pathway towards a new generation of trustworthy, AI-driven Earth system simulations. This significant advance has the potential to accelerate discovery in climate and Earth system science, enabling more reliable long-term forecasting and deeper insights into complex geophysical dynamics.

\vspace{1mm}
\vspace{1mm}

\textbf {*: These authors contributed equally}
\vspace{1mm}

\textbf {$\text{\Letter}$: Led technical programming, implementation, and experimentation}
\vspace{1mm}

\textbf {$\ddag$: Corresponding author. E-mail: hxm@tsinghua.edu.cn}
\vspace{1mm}


\end{abstract}
\maketitle

\pagestyle{headstyle}
\thispagestyle{empty}

\newpage
\section*{Introduction}
Modeling the evolution of the Earth system, including atmospheric and ocean circulations with interactions across many spatial and temporal scales, is a fundamental scientific challenge~\cite{bordoni2025futures, gettelman2022future, kochkov2024neural}. Accurate modeling of these systems is crucial for revealing their inherent cross-scale interactions~\cite{li2021towards, berloff2007ocean,stechmann2014multiscale}. Models unable to resolve high-frequency variability or small spatial scales during long-term integrations can suffer from spurious energy cascades to lower frequencies/larger spatial scales~\cite{marati2004energy, leonard1975energy}, leading to exponential growth of initial errors~\cite{biferale2003shell, leonard1975energy}. This uncontrolled error growth can lead to physically unrealistic outcomes and severely limit the reliability of long-term simulations~\cite{raaisaanen2007reliable}. Accurately capturing multi-scale dynamics while suppressing error accumulation over long iterative roll-outs is essential for advancing Earth system science and especially Earth system modeling. This capability is particularly critical for improving forecasts of complex phenomena such as the evolution of ocean eddies~\cite{hao2025deep,kido2023skillful} for near-term maritime operations, the occurrence of persistent extreme weather conditions, and the response of the Earth's climate system to anthropogenic forcing, e.g. in view of climate change caused by greenhouse gas emissions and land-use change.
 
In past decades, the simulation of Earth system dynamics with multi-scale dynamics has primarily relied on numerically discretizing the governing partial differential equations (PDEs)~\cite{trefethen2000spectral, leveque2007finite,ferziger2019computational}. However, attempts to integrate these systems over extended timescales have encountered a fundamental trade-off between efficiency and accuracy~\cite{liu2015systems, lorenz1969predictability}. Accurately capturing critical multi-scale processes generally requires fine spatiotemporal resolutions~\cite{durran2010numerical,pope2001turbulent,gupta2022towards}, leading to significantly increased computational costs that are often prohibitive for wide applications. Conversely, employing coarse-resolution models reduces computational expense but requires parameterization schemes to approximate unresolved subgrid-scale effects~\cite{stensrud2007parameterization}. While these schemes, or simplifying approaches such as quasi-geostrophic approximations~\cite{pedlosky2013geophysical}, might preserve the large-scale mean state, they often struggle to accurately capture cross-scale energy transfers originating from unresolved processes. These unresolved processes are recognized mechanisms for error amplification~\cite{palmer2001nonlinear,lorenz2017deterministic}. These representation errors typically drive the nonlinear accumulation of simulation errors over time, manifesting as significant phase drift and modal structure deviations in long-term simulation results~\cite{meehl2000coupled, manabe1969climate}. Ultimately, this fundamental trade-off between computational cost and physical accuracy is the underlying reason why traditional numerical methods struggle to achieve efficient and accurate long-term forecasting of Earth system~\cite{bauer2015quiet, schneider2017earth}.

AI offers a powerful data-driven paradigm for Earth system modeling, addressing limitations of numerical methods~\cite{reichstein2019deep,andrychowicz2023deep,irrgang2021towards,pfaff2020learning}. Deep neural networks (DNNs), in particular, excel at learning complex spatiotemporal patterns~\cite{lecun2015deep, liang2025foundation} and have achieved notable success in applications like weather forecasting up to the medium range~\cite{pathak2022fourcastnet, lam2023learning, bi2023accurate,price2025probabilistic,alet2025skillful} and ocean eddy forecasting~\cite{cui2025forecasting, wang2024xihe, xiong2023ai}. However, research reveals an inherent spectral bias in mainstream DNN architectures~\cite{rahaman2019spectral,xu2019frequency}: they tend to prioritize learning dominant, large-scale, low-frequency modes while struggling to represent the less energetic, yet dynamically critical, small to mesoscale high-frequency signals~\cite{fridovich2022spectral, hess2022physically}. Crucially, deterministic AI models often exhibit a coarse effective resolution due to over-smoothing, which undermines long-term stability~\cite{selz2025effective}. This deficiency becomes particularly problematic in long-term autoregressive forecasts, where inaccuracies in high-frequency details can accumulate rapidly as model outputs are repeatedly fed back as inputs. This spectral bias and the resulting blurring critically undermines long-term autoregressive forecasts of data-driven weather models, leading to spurious cross-scale energy transfers and phase-space trajectory distortions due to poorly represented high-frequency dynamics, ultimately causing prediction failure~\cite{rahaman2019spectral}. To address these challenges, the field has recently expanded toward subseasonal and climate timescales. For instance, FengWu-W2S~\cite{ling2024fengwu} leverages multi-layer coupling for seamless forecasting, while emulators~\cite{dheeshjith2025samudra} such as ACE2~\cite{watt2025ace2} and the coupled Ola model~\cite{wang2024coupled} have demonstrated stable decadal-scale simulation. Notably, recent generative and diffusion-based models, such as GenCast~\cite{price2025probabilistic} and GAP~\cite{yang2025generative}, have shown that accurate spatial spectra can be maintained up to the smallest scales, supporting stable rollouts even over millennial periods. Nevertheless, these probabilistic approaches often incur high computational costs due to iterative sampling or the need for large ensembles~\cite{lang2024aifs}. Therefore, achieving an accurate yet computationally efficient representation of these multi-scale dynamics is essential, as their cumulative nonlinear effects govern the system's long-term evolution.

\begin{figure}
\centering
\includegraphics[width=\linewidth]{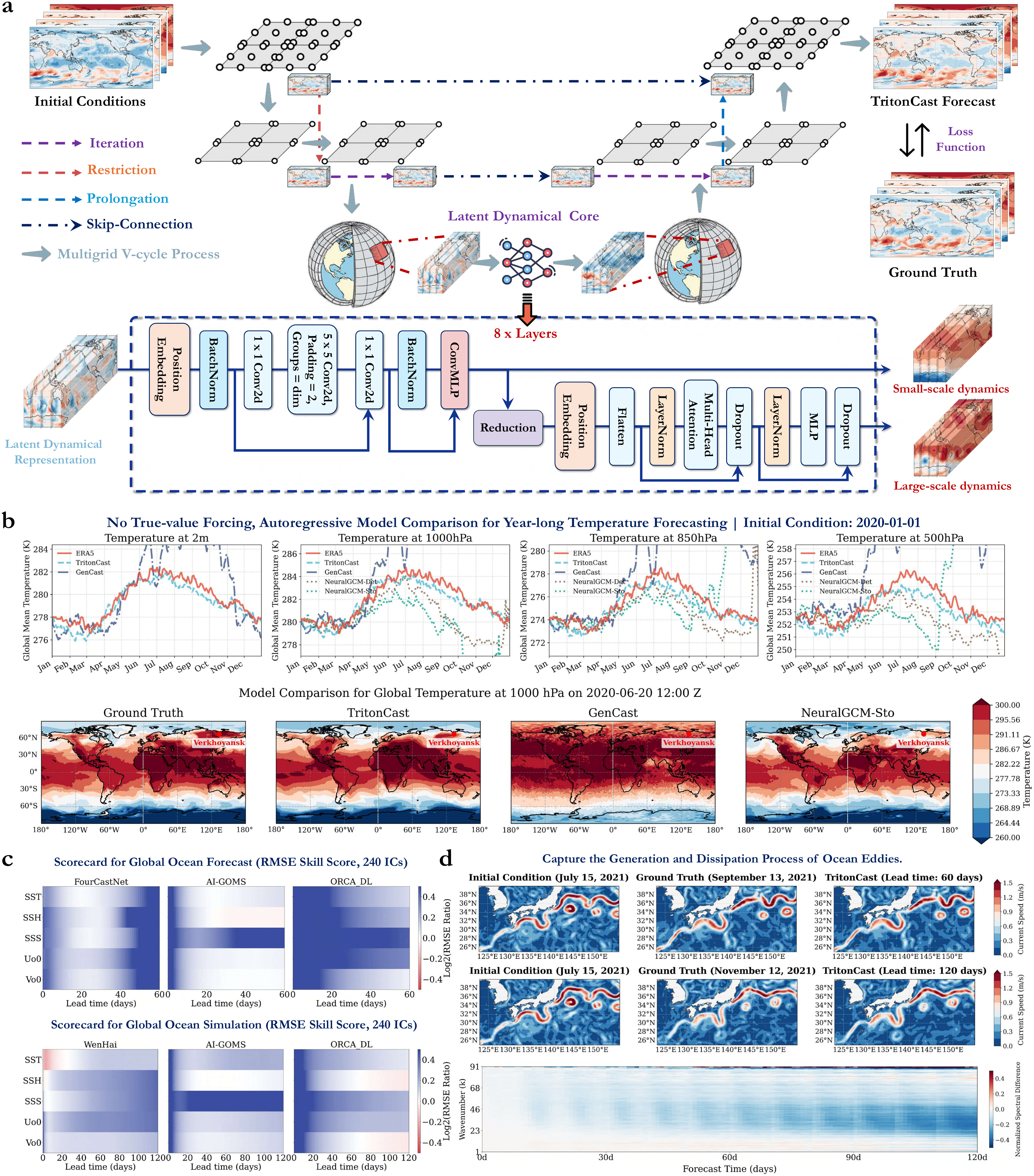}
\caption{\textbf{\method{}'s architecture and long-term forecasting performance.}
\textbf{a,} \method{}'s V-cycle architecture, with a latent core for large-scale dynamics and skip-connections for small-scale fidelity.
\textbf{b,} Stable autoregressive forecast of the 2020 seasonal temperature cycle against ERA5 (top), and the spatial pattern on June 20, 2020 (bottom).
\textbf{c,} Skill scorecard showing significantly lower RMSE than baselines for key ocean variables (e.g., SST, SSH).
\textbf{d,} A 120-day forecast initialized on July 15, 2021, accurately capturing the evolution of Kuroshio eddy.
}
\label{fig1}
\end{figure}

Here we present \method{}, a deep learning framework for long-term Earth system forecasting, designed to fundamentally address spectral bias in AI models. Inspired by multigrid methods in numerical computation, \method{} learns to accurately represent dynamical processes across multiple scales, enabling it to resolve the long-standing conflict between short-term forecast accuracy and long-term simulation/forecast stability within a single architecture. Specifically, \method{} establishes a new state of the art across several critical domains:

\noindent\ding{182} \textbf{Reliable long-term atmospheric stability.} A lightweight model achieves stable, year-long, purely autoregressive global forecasts, accurately reproducing the seasonal cycle and outperforming specialized models (e.g., GenCast, NeuralGCM) prone to climate drift. Notably, as shown in \textbf{Fig.~\ref{fig1}b}, it successfully predicts the record-breaking Siberian heatwave of June 2020 nearly six months in advance, using only initial condition from the beginning of the year. This exceptional long-term stability is corroborated by a year-long quantitative evaluation (\textbf{Fig.~\ref{TritonCast_weather}b}) and physical field comparisons (\textbf{Fig.~\ref{TritonCast_weather}c}), which show significantly lower error accumulation than baseline models (See Supplementary~\ref{appendix:year_long} for more results). The model also remains stable in multi-year climate simulations, accurately replicating seasonal and interannual trends in global mean temperature (\textbf{Fig.~\ref{TritonCast_weather}d}, see Supplementary~\ref{appendix:Climate} for more results).


\noindent\ding{183} \textbf{High-fidelity ocean simulation and forecasting.} In the demanding task of simulation \& forecasting ocean currents, \method{} consistently outperforms strong baselines like AI-GOMS, Wenhai, ORCA\_DL, and PDE-Refiner \textbf{(Fig.~\ref{fig1}c, Fig.~\ref{Figure_4_ocean_kurrent}a)} by accurately preserving the theoretical energy spectrum of ocean and suppressing spectral error for up to $90$ days \textbf{(Fig.~\ref{Figure_4_ocean_kurrent}b,c)}. This is exemplified in a $60$-day North Atlantic forecast, where \method{} accurately captures the intricate thermal structures of the Gulf Stream that are direct signatures of marine heatwaves, while baseline models collapse into overly smoothed or noisy fields (\textbf{Fig.~\ref{Figure2_Ocean}c,d}), see Supplementary~\ref{appendix:ocean_sim_fore} for more results.

\noindent\ding{184} \textbf{State-of-the-art medium-range weather forecasting.} On the comprehensive WeatherBench 2 benchmark, \method{} achieves state-of-the-art forecast skill, matching or exceeding representative AI baselines and the global gold-standard IFS-HRES operational system (\textbf{Fig.~\ref{TritonCast_weather}a}).

\noindent\ding{185} \textbf{Zero-shot cross-resolution adaptability.} Finally, \method{} exhibits an inherent capacity for cross-resolution generalization. When applied directly to unseen, finer $0.125\degree$ grids after training exclusively on $0.25\degree$ data, the model remains stable and continues to resolve detailed mesoscale structures without additional fine-tuning \textbf{(Fig.~\ref{Figure_4_ocean_kurrent}f)}. See Supplementary~\ref{appendix:zeroshot} for further analysis.

By suppressing spectral bias, \method{} advances reliable long-term Earth system forecasting, achieving high fidelity with remarkable computational efficiency. (e.g. a $365$-day global forecast at a $1 \degree$ daily spatiotemporal resolution is produced in $56$~s on a single NVIDIA A100 GPU.) This development unlocks significant potential for next-generation operational forecasting systems, enhancing predictive capabilities of the Earth system.

\section*{\method{}: A Long-term Earth System Forecast Model}

\begin{table*}[h!]
\footnotesize
\centering
\caption{\textbf{Overview of baseline models categorized by experimental domain and task.} This table provides a comprehensive summary of the state-of-the-art models used as baselines for evaluating TritonCast's performance. The models are systematically organized by their respective scientific domains: Atmospheric science, Oceanography, and Turbulence.}
\label{tab:models_manual_width}
\setlength{\tabcolsep}{8pt}
\begin{tabular}{
    >{\raggedright}p{1.7cm} 
    >{\raggedright}p{3.4cm} 
    l 
    >{\raggedright}p{4.7cm} 
    >{\raggedright\arraybackslash}p{2.5cm} 
}
\toprule
\textbf{Domain} & \textbf{Task} & \textbf{Model} & \textbf{Core Methodology} & \textbf{Model Type} \\
\midrule

\multirow{10}{*}{Atmospheric} & \multirow{5}{*}{Medium-range Forecasting}
& IFS-HRES~\cite{rasp2024weatherbench} & NWP System & Numerical\\
& & Keisler~\cite{keisler2022forecasting} & Graph Neural Network & AI / Deterministic \\
& & Pangu~\cite{bi2023accurate} & 3D Swin Transformer & AI / Deterministic \\
& & GraphCast~\cite{lam2023learning} & Graph Neural Network  & AI / Deterministic \\
& & NeuralGCM~\cite{kochkov2024neural} & Hybrid Physics and ML & Hybrid Model\\  
\cmidrule(l){2-5}
& \multirow{3}{*}{Long-term Forecasting} 
& GraphCast~\cite{lam2023learning} & Graph Neural Network & AI / Deterministic \\
& & GenCast~\cite{price2025probabilistic} & Conditional Diffusion Model & AI / Probabilistic \\
& & NeuralGCM~\cite{kochkov2024neural} & Hybrid Physics and ML & Hybrid Model\\
\midrule

\multirow{8}{*}{Oceanography} & \multirow{4}{*}{Global Ocean Simulation} 
& FourCastNet~\cite{pathak2022fourcastnet} & Adaptive Fourier Neural Operator & AI / Deterministic \\
& & AI-GOMS~\cite{xiong2023ai} & Masked Autoencoder & AI / Deterministic \\
& & ORCA\_DL~\cite{guo2025data} & Data-driven Transformer Model & AI / Deterministic \\
& & WenHai~\cite{cui2025forecasting} & Swin Transformer& AI / Deterministic \\
\cmidrule(l){2-5}
& \multirow{3}{*}{Coupled Forecasting}
& FourCastNet~\cite{pathak2022fourcastnet} & Adaptive Fourier Neural Operator & AI / Deterministic \\
& & AI-GOMS~\cite{xiong2023ai} & Masked Autoencoder & AI / Deterministic \\
& & ORCA\_DL~\cite{guo2025data} & Data-driven Transformer Model & AI / Deterministic \\
\cmidrule(l){2-5}
& \multirow{2}{*}{Sea Surface Currents} 
& ORCA\_DL~\cite{guo2025data} & Data-driven Transformer Model & AI / Deterministic \\
& & PDE-Refiner~\cite{lippe2023pde} & Diffusion Model & AI / Deterministic \\
\midrule

\multirow{2}{*}{Turbulence} & \multirow{2}{*}{Kolmogorov Turbulence} 
& FNO~\cite{li2020fourier} & Fourier Neural Operator  & AI / Deterministic \\
& & PDE-Refiner~\cite{lippe2023pde} & Diffusion Model & AI / Deterministic \\
\bottomrule
\end{tabular}
\end{table*}

\method{} is a machine learning model for long-term, stable Earth system forecasts. It solves the "spectral bias" problem in current AI forecast models. These models tend to learn smooth, large-scale features. This tendency causes small-scale errors to grow quickly in long-term autoregressive simulations,  hindering reliable forecasts at longer time scales. Our model takes inspiration from the multigrid method in numerical computing. It uses a hierarchical learning framework (\textbf{Fig.~\ref{fig1}a}). This framework separates the Earth system's high-resolution state into multiple scales. It only processes the system's main low-frequency trends at the coarsest scale. Then, it merges this stable trend with the high-frequency details layer by layer. This "divide and conquer" strategy stops error accumulation. It ensures the model's long-term stability and physical fidelity (See Supplementary~\ref{appendix:model_details} for model details).

To evaluate \method{}'s performance in a full and rigorous way, we systematically compare it with current state-of-the-art (SOTA) AI models and operational systems. We use a series of benchmarks in atmospheric science, oceanography, and turbulence. \textbf{Tab.~\ref{tab:models_manual_width}} lists the main baseline models for comparison.

\noindent\ding{182}~\textbf{Atmospheric Science.}~\textbf{\textit{Medium-range Weather Forecasts}:} To test the model's medium-range forecast accuracy, we train a $1$-billion-parameter atmospheric model. We train it for $400$ hours on $48$ A100 GPUs. We follow the WeatherBench 2 benchmark protocol to directly compare our model with established AI models such as Pangu and GraphCast. We also compare it with the gold-standard operational system, IFS-HRES. \textbf{\textit{Long-term Stability Tests}:} Long-term integration stability without external forcing is a core challenge for current AI models. To test this, we build a lightweight $0.02$-billion-parameter model. It requires only $24$ hours of training on $8$ A100 GPUs. We use the official open-source weights for all baseline models. Under the same initial conditions, we run a full year of purely autoregressive simulations. We compare our model with Graphcast, GenCast, and the hybrid-physics model NeuralGCM (both its deterministic and stochastic versions). See Supplementary~\ref{appendix:weather_appendix_exp} for experimental details.

\noindent\ding{183}~\textbf{Oceanography.}~\textbf{\textit{Global Ocean Long-term Simulations}:} To assess the model's ability in high-fidelity, long-term ocean simulations, we build a $0.02$-billion-parameter ocean model. We pre-train the model for $150$ hours on $48$ A800 GPUs using $0.25\degree$ resolution daily-mean GLORYS12 data from $1993$–$2017$, followed by multi-step supervised fine-tuning. We compare its performance against several strong baselines, including FourCastNet, AI-GOMS, ORCA\_DL, and WenHai, through continuous simulations of up to $120$ days. \textbf{\textit{High-Fidelity Eddy and Current Forecasts}:}~To test the model's ability to resolve key mesoscale dynamics, we train it further. This extra training takes $36$ hours on $8$ A100 GPUs. We stress-test the model on $0.125\degree$ high-resolution data from strong eddy regions, like the Kuroshio and Gulf Stream. Its performance surpasses the advanced spatiotemporal models (\textbf{Fig.~\ref{Figure_4_ocean_kurrent}d}).~\textbf{\textit{Coupled Ocean-Atmosphere Forecasting:}} To test the model's robustness under operational uncertainty, we perform a $60$-day, one-way coupled forecast. In this setup, atmospheric forcing is generated by a $1B$-parameter TritonCast weather model, and its predicted output with inherent uncertainties is used to drive the $0.02B$-parameter ocean model. This provides a stringent, end-to-end AI chain to evaluate the ocean model's practical performance. See Supplementary~\ref{appendix:ocean_appendix_exp} and~\ref{appendix:CMEMS_appendix_exp} for experimental details.

\noindent\ding{184}~\textbf{Turbulence.}~To verify \method{}'s core physical ability to overcome spectral bias, we use the canonical benchmark of two-dimensional Kolmogorov turbulence to assess its physical consistency. In a fair comparison framework with a strict parameter count (about $0.02$ billion), we compare its performance with a comprehensive benchmark library. This library covers many mainstream architectures, including models like FNO and PDE-Refiner. See Supplementary~\ref{appendix:turbulence} for experimental details.

To ensure a fair comparison, we evaluate all baseline models under the same initial conditions. For models with official open-source weights, we run inference directly. For other models, we reproduce them using the same data and methods as TritonCast. We provide the detailed settings for all experiments, including model architectures and training protocols, in the Methods section and the Supplementary~\ref{appendix:EXPSTEUP}.

\section*{Results}

\subsection*{Atmospheric Dynamics Forecasting and Simulation}
\begin{figure}[h!]
\centering
\includegraphics[width=\linewidth]{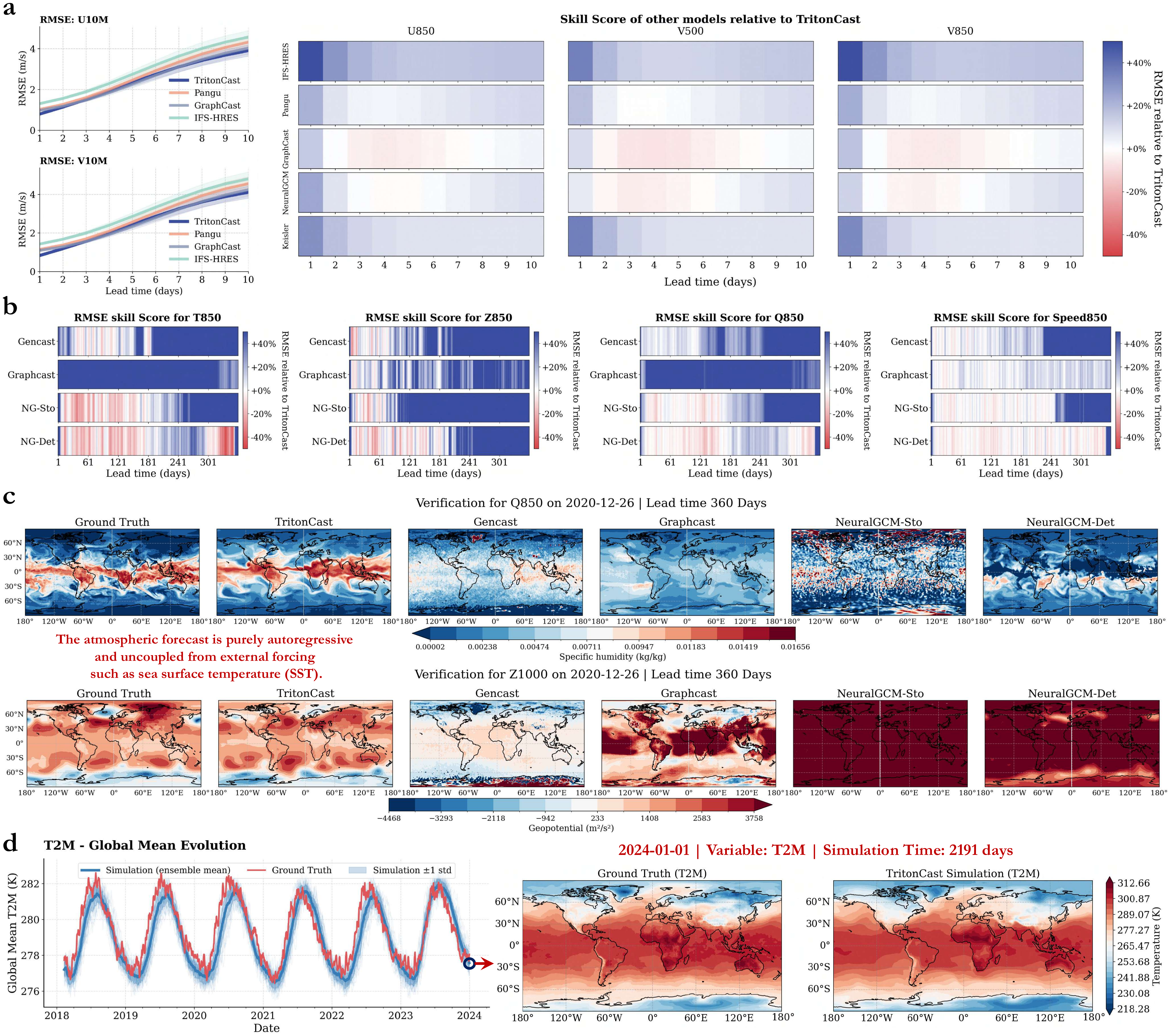}
\caption{\textbf{Comprehensive performance of TritonCast at weather and climate scales.} 
\textbf{a}, In the medium-range forecasts, the forecast error (RMSE) of TritonCast is comparable to state-of-the-art AI and operational models in WeatherBench 2. 
\textbf{b, c}, In year-long, purely autoregressive forecasts, both quantitative scores (\textbf{b}) and physical field comparisons (\textbf{c}) demonstrate the superior long-term stability of TritonCast. Its error accumulation is significantly lower than that of baseline models, and it maintains the physical realism of the fields.
\textbf{d}, In multi-year climate simulations driven by sea surface temperature, TritonCast accurately reproduces the seasonal cycle of global mean temperature and demonstrates long-term integration stability.}
\label{TritonCast_weather}
\end{figure}

To evaluate the performance of TritonCast in atmospheric modeling, we conduct a suite of experiments spanning from standard medium-range weather forecasting to year-long stability tests. We benchmark TritonCast against leading AI and operational physics-based models. The results demonstrate that the model exhibits exceptional performance in both medium-range accuracy and long-term physical fidelity.

\noindent\textbf{\textit{State-of-the-art medium-range weather forecasting.}}~We evaluate TritonCast's medium-range weather forecasting capabilities on the internationally recognized WeatherBench 2 benchmark. The model is benchmarked against leading AI models, including Pangu and GraphCast, and the operational physics-based Integrated Forecasting System (IFS-HRES) from the European Centre for Medium-Range Weather Forecasts (ECMWF). As shown in \textbf{Fig.~\ref{TritonCast_weather}a}, the root-mean-square error (RMSE) for $10$-meter wind speeds (U$10$M and V$10$M) over a $10$-day forecast horizon is comparable to other top-tier models. Furthermore, skill scores for multiple upper-atmospheric variables (\textbf{Fig.~\ref{TritonCast_weather}a, right panels}) indicate that TritonCast exhibits a lower RMSE in most cases, confirming its state-of-the-art accuracy for the $1$-to-$10$ day forecasting task. A more comprehensive multi-variable metric comparison is provided in the Supplementary~\ref{appendix:wb2_full_results}.

\noindent\textbf{\textit{Long-term autoregressive stability.}}~A core design objective of the TritonCast architecture is to enhance long-term stability, overcoming the prediction divergence common in purely data-driven models due to the accumulation of spectral errors. To test this capability, we use a lightweight $0.02$B-parameter TritonCast model to perform a full-year ($360$-day), purely autoregressive forecast without any external forcing, such as sea surface temperature (SST). The results of this test validate TritonCast's stability in long-term forecasting. As shown by the RMSE skill scores in \textbf{Fig.~\ref{TritonCast_weather}b}, although the performance of all models is comparable in the short term, TritonCast's relative error advantage grows with the forecast lead time, indicating a significantly lower rate of error accumulation compared to baseline models like GenCast, Graphcast, and NeuralGCM, including both its deterministic (NG-Det) and stochastic (NG-Sto) variants. To ensure a fair comparison, all baseline models are evaluated using their officially released weights under identical initial conditions (see Supplementary~\ref{appendix:Open_source_Weights} for detailed settings).

This stability is also reflected in the spatio-temporal evolution of the physical fields. As depicted in \textbf{Fig.~\ref{TritonCast_weather}c}, after $360$ days of continuous integration, TritonCast's predicted fields for specific humidity (Q$850$) and geopotential height (Z$1000$) maintain physically plausible large-scale circulation patterns that are consistent with the ground truth. In stark contrast, other baseline models show significant field distortions, drift, or even numerical collapse, reflecting their inability to maintain physical conservation laws during long-term integration. A more detailed analysis of the integration stability for each model, including the time of first numerical divergence, is provided in the Supplementary~\ref{appendix:time_Stability}.

\noindent\textbf{\textit{Potential for multi-year climate simulation.}}~The stability demonstrated by TritonCast in long-term integrations unlocks its potential for application in multi-year climate simulations. To explore this, we use a $0.1$B-parameter TritonCast model to perform a multi-year ensemble simulation with an integration length of $2500$ days, covering the available ERA5 test period. The simulation follows the experimental design of the Atmospheric Model Intercomparison Project (AMIP), using prescribed SST as boundary conditions to drive the atmospheric model's evolution. The simulation results show that TritonCast successfully captures the fundamental spatio-temporal features of the climate system. Over the course of the multi-year simulation, the model's ensemble mean (\textbf{Fig.~\ref{TritonCast_weather}d, left panel}) reproduces the seasonal cycle and inter-annual trends of the global mean $2$-meter temperature. The ensemble spread remains stable; notably, none of the $40$ initial conditions (spaced at 1 days from the start from Jan. 1, 2018) result in systematic drift. After $2191$ days of simulation, the model's output for the global temperature spatial distribution (\textbf{Fig.~\ref{TritonCast_weather}d, right panel}) remains physically realistic at the macroscopic scale (see Supplementary~\ref{appendix:Climate} for detailed results). This demonstrates that the TritonCast architecture provides an effective pathway for developing AI climate models capable of stable, long-term integration (see Supplementary~\ref{appendix:Climate_Simulations} for detailed settings of the ensemble simulation).

\subsection*{Ocean Simulation and Forecasting}

\begin{figure}[h!]
\centering
\includegraphics[width=1\linewidth]{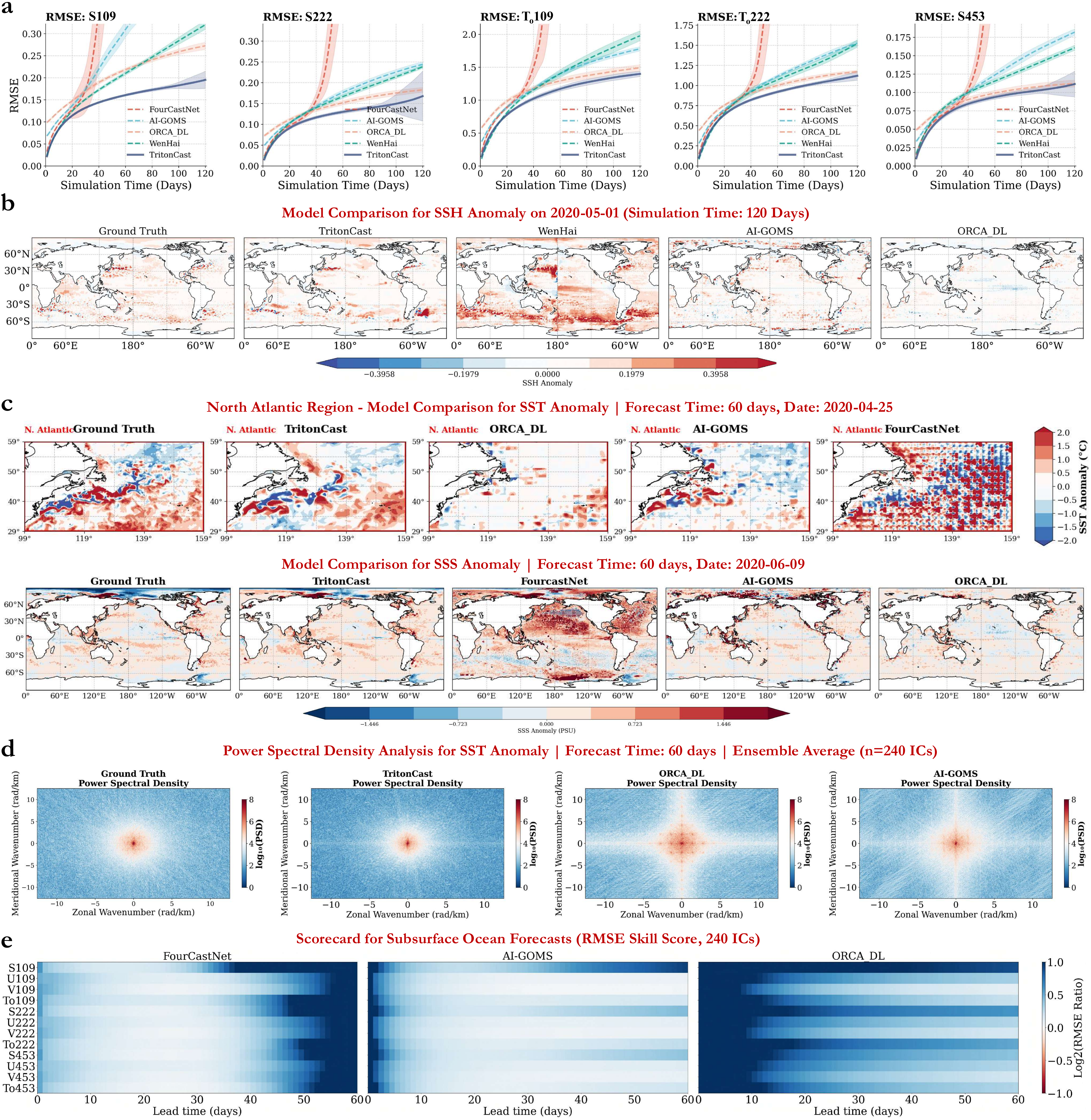}
\caption{\textbf{TritonCast enables high-fidelity, long-term ocean simulation and forecasting.}
\textbf{a}, TritonCast demonstrates lower and more stable RMSE in 120-day deep ocean simulations versus baselines.
\textbf{b}, TritonCast preserves realistic sea surface height anomaly (SSHa) features after 120 days, unlike distorted baseline results.
\textbf{c}, In 60-day forecasts, TritonCast accurately captures fine-scale SSTa and SSSa, while baselines fail.
\textbf{d}, Power spectral density (PSD) shows TritonCast retains fine-scale energy in 60-day SSTa forecasts, which baselines lose.
\textbf{e}, RMSE skill scorecard for 60-day subsurface forecasts.
}
\label{Figure2_Ocean}
\end{figure}

\noindent\textbf{\textit{Long-term Ocean Simulations.}}~To rigorously test the long-term stability and physical fidelity of our architecture, we conduct a series of $120$-day global ocean simulations driven by ERA5 reanalysis. To ensure a fair and high-standard comparison, we retrain several state-of-the-art baseline models, including FourCastNet, AI-GOMS, and ORCA\_DL, under identical conditions, while using the officially released version for the WenHai model (see Supplementary~\ref{appendix:Open_source_Weights} for details). The quantitative results, averaged over a robust ensemble of 240 initial conditions, are unequivocal (\textbf{Fig.~\ref{Figure2_Ocean}a}). Although all models exhibit comparable performance in the initial days, the baseline models show a rapid, often exponential, growth in RMSE for deep ocean variables (e.g., S$222$, T$109$) after $30$-$40$ days, signaling a loss of stability. In stark contrast, TritonCast demonstrates remarkably low and controlled error accumulation throughout the entire $120$-day period. This quantitative superiority translates into a striking difference in physical realism (\textbf{Fig.~\ref{Figure2_Ocean}b}). After $120$ days, TritonCast's simulation of the sea surface height anomaly (SSHa) remains physically coherent and closely matches the ground truth, preserving key oceanic features. Conversely, the outputs of other models are either fields with severe artifacts and noise (e.g., WenHai) or overly smoothed states that have lost key dynamical information (e.g., AI-GOMS, ORCA\_DL), confirming their failure to maintain long-term physical consistency. 

\noindent\textbf{\textit{Coupled Ocean-Atmosphere Forecasting.}}~We advance to a more challenging coupled forecasting scenario, where the ocean model is driven by forecasts generated by our TritonCast's atmospheric model. The $60$-day forecast results, focused on the dynamically complex North Atlantic region, highlight TritonCast's exceptional capability in capturing fine-scale phenomena (\textbf{Fig.~\ref{Figure2_Ocean}c}). In the sea surface temperature anomaly (SSTa) field, TritonCast accurately reproduces the intricate, meandering thermal structures that serve as the direct surface signatures of the Gulf Stream's path and the mesoscale eddies it sheds. In stark contrast, the baseline models fail to capture these dynamics: the forecasts from ORCA\_DL and AI-GOMS are overly smoothed, blurring the sharp thermal fronts into indistinct gradients, while FourCastNet’s output degenerates entirely into a field of unphysical noise. This dramatic difference in visual fidelity is quantitatively explained by a power spectral density analysis, averaged over an ensemble of $240$ initial conditions (\textbf{Fig.~\ref{Figure2_Ocean}d}). TritonCast's two-dimensional spatial power spectrum is nearly indistinguishable from the ground truth, indicating that it realistically preserves energy across the full range of spatial scales. The baseline models, however, exhibit a pronounced energy decay at high wavenumbers. This spectral dissipation is the direct quantitative signature of the blurring and loss of detail observed in \textbf{Fig.~\ref{Figure2_Ocean}c}, providing clear evidence that TritonCast's success lies in its effective mitigation of the spectral bias that plagues long-term autoregressive forecasting in other AI models (Additional variable comparisons are available in the \textbf{Fig.~\ref{Figure2_Ocean}e} and Supplementary~\ref{appendix:ocean_sim_fore}).

\subsection*{High-fidelity Ocean Eddy Forecast}
\begin{figure}[h!]
\centering
\includegraphics[width=\linewidth]{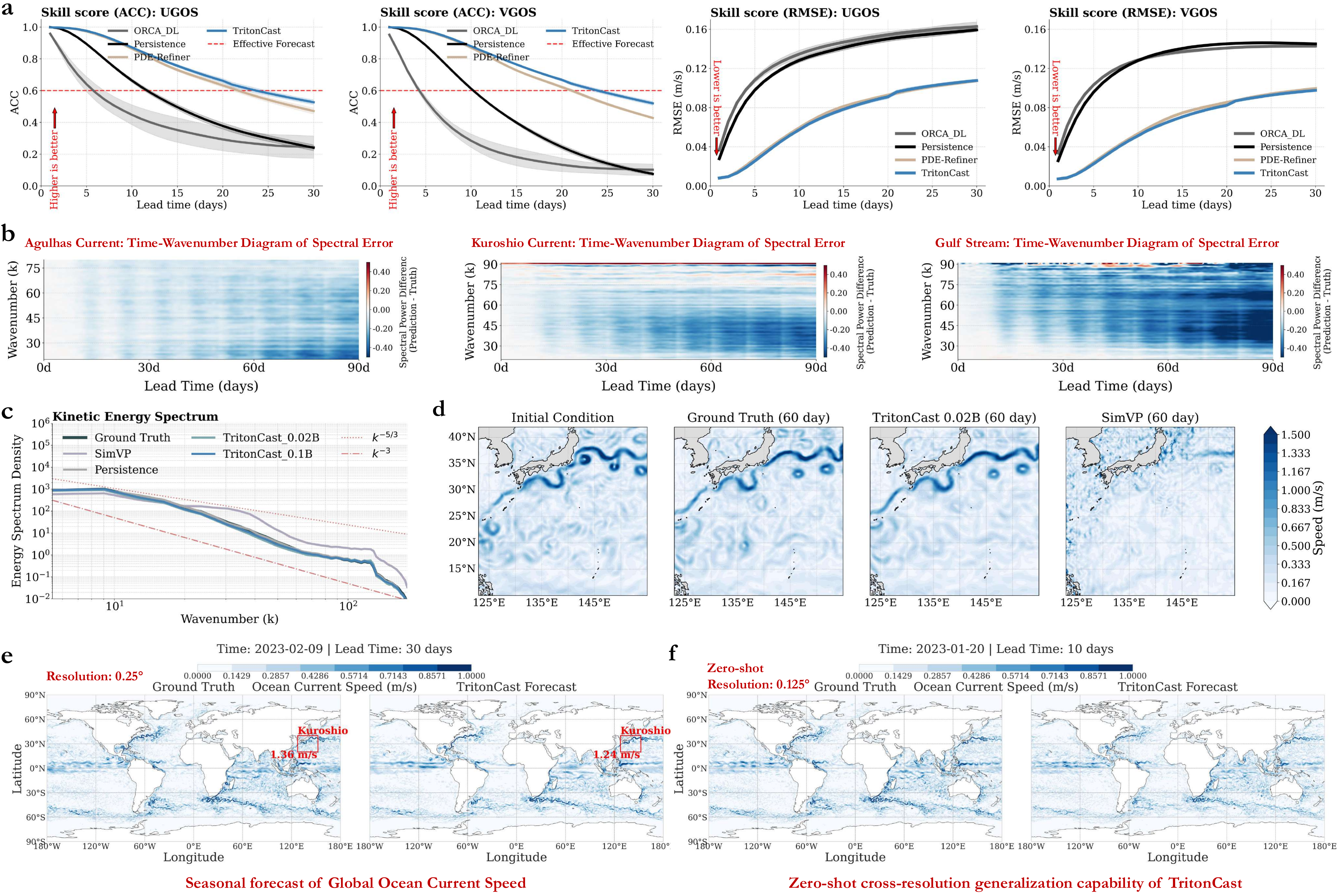}
\caption{\textbf{\method{} demonstrates superior skill and zero-shot generalization in ocean current forecasting.}
\textbf{a}, Quantitative skill scores for $30$-day global forecasts at $0.25\degree$ resolution. \method{} significantly outperforms all baselines, including ORCA\_DL, in both ACC and RMSE.
\textbf{b}, Time-wavenumber diagrams of spectral error for $90$-day forecasts in three major western boundary current regions (Agulhas Current, Kuroshio, Gulf Stream). \method{} suppresses spectral error to very low levels across nearly all scales and lead times.
\textbf{c}, Kinetic energy spectra at $0.125\degree$ resolution. The spectrum of \method{} aligns closely with the ground truth and correctly reproduces theoretical scaling laws.
\textbf{d}, A $60$-day forecast visualization of Kuroshio Extension eddies at $0.125\degree$ resolution. \method{} successfully captures the long-term evolution of eddies, whereas baseline models like SimVP exhibit excessive smoothing.
\textbf{e}, A $30$-day forecast of the global ocean current speed at $0.25\degree$ resolution.
\textbf{f}, Zero-shot cross-resolution generalization capability of \method{}. This panel shows a $10$-day forecast on $0.125\degree$ high-resolution grids using a model trained on $0.25\degree$ data. Even on an unseen resolution, the model generates physically realistic and detailed eddy structures.}
\label{Figure_4_ocean_kurrent}
\end{figure}

To rigorously evaluate \method{}’s capabilities in a challenging, eddy-rich environment, we conduct a series of long-term global ocean forecasting experiments. The results show a significant advance in predictive skill and physical fidelity over current state-of-the-art models.

\noindent\textbf{\textit{Quantitative skill and physical fidelity.}}
~On a global scale, \method{} consistently outperforms strong baselines, including ORCA\_DL, over a $30$-day forecast period, as measured by standard metrics like Anomaly Correlation Coefficient (ACC) and Root Mean Square Error (RMSE) (\textbf{Fig.~\ref{Figure_4_ocean_kurrent}a}). The success of \method{} primarily lies in its effective mitigation of the energy spectral bias common in AI models. We test this in the Kuroshio region, a critical testbed for long-term fidelity due to its intense eddy activity, which transports vast amounts of heat and regulates regional climate. Spectral analysis across this and other major current systems, including the Gulf Stream and the Agulhas Current, reveals that \method{} consistently suppresses spectral error for up to $90$ days (\textbf{Fig.~\ref{Figure_4_ocean_kurrent}b}). Furthermore, its kinetic energy spectrum correctly reproduces the theoretical $k^{-3}$ scaling law for oceanic turbulence, while baseline models suffer from unrealistic energy decay at high wavenumbers (\textbf{Fig.~\ref{Figure_4_ocean_kurrent}c}). This demonstrates that \method{} accurately simulates the cross-scale energy transfers that are essential for long-term forecasting (Additional variable comparisons are available in the Supplementary~\ref{appendix:full_ocean_stream_exp}).

\noindent\textbf{\textit{High-fidelity forecast and generalization.}} 
~In a $60$-day forecast of the Kuroshio Extension, \method{} accurately preserves the intricate structures of ocean eddies regarding their position, morphology, and intensity, while standard AI architectures exhibit excessive smoothing that obliterates these critical features (\textbf{Fig.~\ref{Figure_4_ocean_kurrent}d}). This high-fidelity forecast translates into a groundbreaking extension of predictability. \method{} extends the effective forecast horizon for Kuroshio eddies to approximately $120$ days, an order of magnitude improvement over the $10$-$30$ days reported in prior research~\cite{cui2025forecasting, wang2024xihe}, with anomaly correlation coefficients remaining above $0.85$ (See Supplementary~\ref{appendix:Key_Regions} for detailed results). Most critically, \method{} exhibits an unprecedented zero-shot cross-resolution generalization capability. A model trained exclusively on coarse-resolution $0.25$° data, when applied directly to unseen finer $0.125$° grids, generates forecasts with remarkable detail and physical realism (\textbf{Fig.~\ref{Figure_4_ocean_kurrent}f}). This ability to generalize across scales provides the strongest evidence for its architectural novelty, demonstrating that \method{} has learned the underlying, scale-invariant physical laws governing ocean dynamics, a crucial step towards building truly versatile and reliable AI models for the Earth system (see Supplementary~\ref{appendix:zeroshot} for detailed results).

\subsection*{Robust Turbulence Forecast}
\begin{figure}[h!]
\centering
\includegraphics[width=\linewidth]{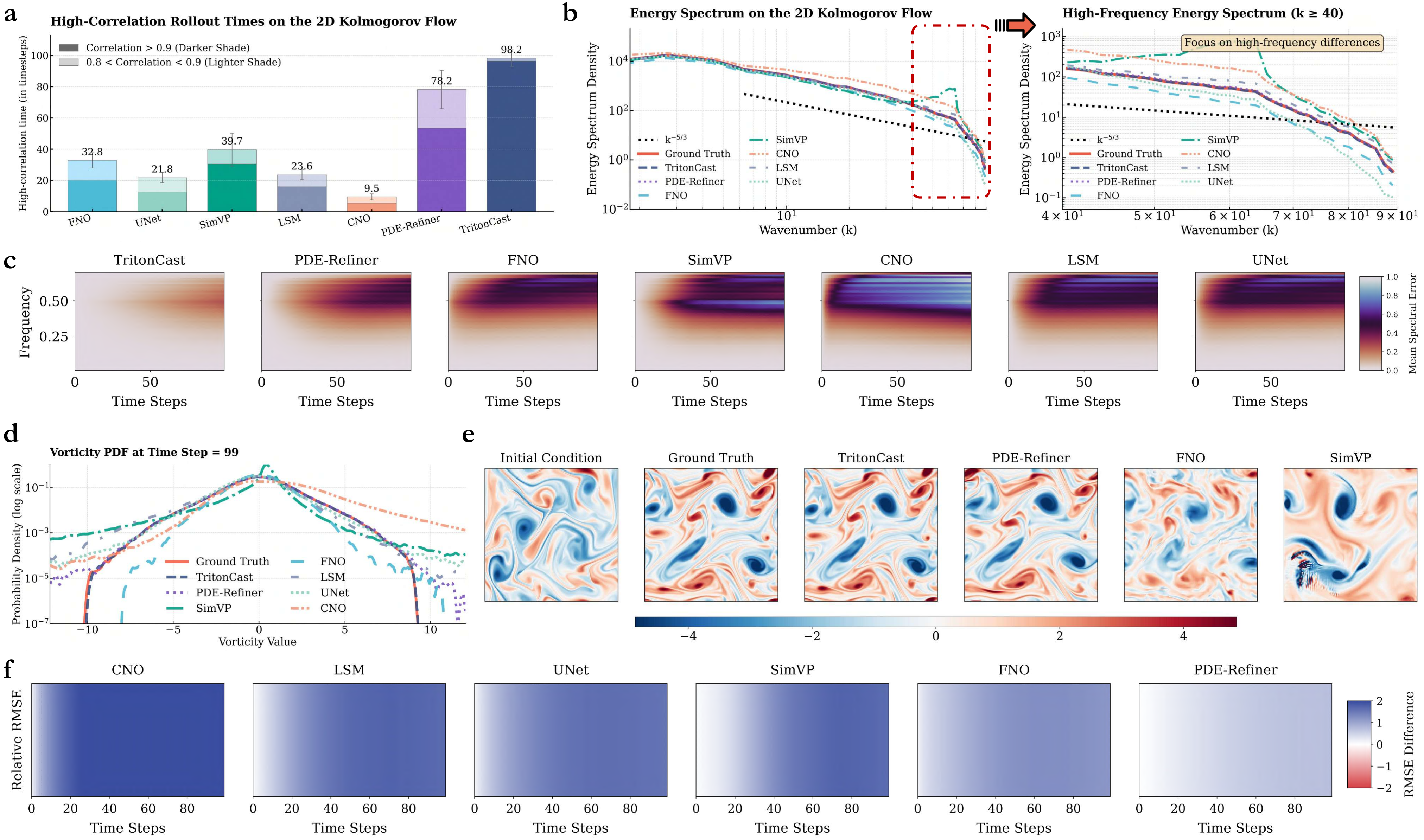}
\caption{\textbf{\method{} overcomes spectral bias for superior long-term turbulence forecasting.}
\textbf{a}, A quantitative comparison of high-correlation rollout times (correlation > $0.9$). \method{}'s effective forecast horizon ($98.2$ timesteps) significantly surpasses all baseline models, including the physics-informed PDE-Refiner ($78.2$ timesteps).
\textbf{b}, Energy spectra analysis reveals differences in physical fidelity. The spectrum of \method{} (dark blue) aligns closely with the ground truth (red) and the theoretical $k^{-5/3}$ slope, particularly in the high-frequency range (zoomed-in, right panel). In contrast, most baselines exhibit severe spectral bias, with unphysical energy pile-up or dissipation.
\textbf{c}, The evolution of spectral error over time. \method{} suppresses spectral error to extremely low levels across the entire forecast period, whereas other models, particularly at high frequencies, rapidly accumulate large errors from the initial steps.
\textbf{d}, The vorticity probability density function (PDF) at timestep $99$ (log scale). \method{} accurately reproduces the heavy-tailed distribution of the ground truth, indicating its ability to capture the correct probability of extreme vorticity values.
\textbf{e}, A visual comparison of vorticity fields. The \method{} forecast is visually consistent with the ground truth, preserving fine-scale vortex structures. In contrast, most baselines exhibit excessive smoothing, while models like SimVP produce unphysical numerical artifacts.
\textbf{f}, Relative Root Mean Square Error (RMSE) difference to \method{}. All baseline models exhibit substantially higher error than \method{} across the entire forecast horizon.}
\label{turbulence_comparison}
\end{figure}

To test TritonCast's ability to handle the multi-scale dynamics central to complex physical systems, we evaluate it on the challenging benchmark of 2D Kolmogorov turbulence. The results unequivocally demonstrate a step-change in performance over a suite of state-of-the-art benchmarks (\textbf{Fig.~\ref{turbulence_comparison}}). TritonCast achieves a high-correlation forecast horizon of $98.2$ timesteps, significantly outperforming all baselines, including the physics-informed PDE-Refiner (\textbf{Fig.~\ref{turbulence_comparison}a}). This superior long-term accuracy stems directly from its ability to overcome the spectral bias that plagues other AI architectures. Energy spectrum analysis reveals that while most models exhibit unphysical energy pile-up or dissipation at high frequencies, TritonCast's spectrum remains in excellent agreement with the ground truth, correctly reproducing the theoretical $k^{-5/3}$ energy cascade (\textbf{Fig.~\ref{turbulence_comparison}b}). This spectral fidelity is maintained across the entire simulation, with spectral error suppressed to near-zero levels while competitors accumulate large errors from the very first steps (\textbf{Fig.~\ref{turbulence_comparison}c}).

This superior spectral fidelity directly translates into physically realistic long-term simulations. At the $99$th timestep, TritonCast preserves fine-scale vortex structures that are visually indistinguishable from the ground truth, whereas other models suffer from excessive smoothing or produce unphysical numerical artifacts (\textbf{Fig.~\ref{turbulence_comparison}e}). Critically, TritonCast also accurately captures the heavy-tailed probability distribution of extreme vorticity events, a feature other models fail to reproduce (\textbf{Fig.~\ref{turbulence_comparison}d}). This ability to correctly model the statistics of extremes, rooted in its accurate representation of small-scale dynamics, provides a foundational explanation for its advanced capabilities in more complex forecasting tasks. This success in a canonical turbulent system underscores a general principle: accurately modeling small-scale dynamics is the key to improving long-term forecasts of critical Earth system phenomena, from ocean eddies to climate extremes.

\subsection*{Ablation studies}
\label{main:ablation}

\begin{figure}[h!]
\centering
\includegraphics[width=\linewidth]{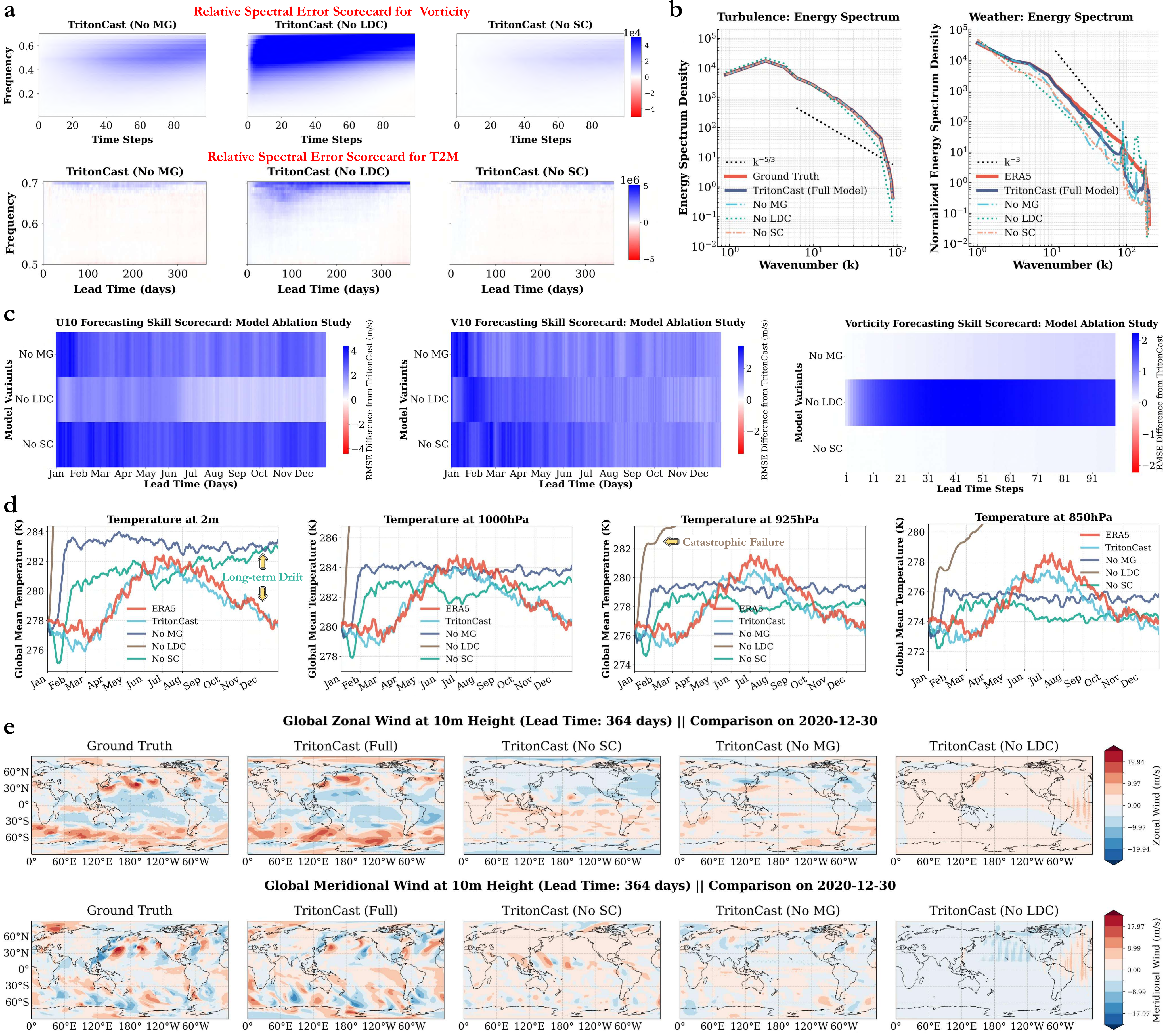}
\caption{\textbf{Ablation studies pinpoint the critical roles of \method{}'s architecture.}
\textbf{a}, Spectral error analysis shows that removing any core component Multi-Grid (\textbf{No MG}), Latent Core (\textbf{No LDC}), or Skip-Connections (\textbf{No SC}) causes rapid high-frequency error accumulation.
\textbf{b}, Energy spectra of ablated models reveal severe, unphysical energy dissipation at small scales, unlike the full model which correctly reproduces the theoretical energy cascades with characteristic slopes of $k^{-5/3}$ for turbulence and $k^{-3}$ for synoptic scale.
\textbf{c}, Skill scorecards quantify a sharp increase in RMSE for all ablated variants, confirming their degraded accuracy.
\textbf{d}, Long-term stability is critically compromised upon ablation. Removing the \textbf{No LDC} component leads to catastrophic model failure, while removing \textbf{No MG} or \textbf{No SC} induces severe drift.
\textbf{e}, After a 364-day forecast, all ablated models fail to reproduce realistic stratospheric circulation patterns, losing both large and small-scale structures. 
These findings demonstrate the synergistic interplay of the architectural components, confirming that each element is indispensable for achieving robust long-term stability and physical fidelity.}
\label{TritonCast_ablation_exp}
\end{figure}

To understand the success of \method{}, we run a series of tests where we remove its key parts one by one (\textbf{Fig.~\ref{TritonCast_ablation_exp}}). These tests show the unique and essential jobs of its three core components: the multi-grid structure (\textbf{No MG}), the latent dynamical core (\textbf{No LDC}), and the skip-connections (\textbf{No SC}). The multi-grid structure is the foundation for keeping the model's predictions physically realistic. It uses a coarse-to-fine and back-to-coarse hierarchy to compel the model to separate and learn the dynamics at each scale. This allows it to correctly reproduce the theoretical energy cascades and follow the characteristic slopes of $k^{-5/3}$ for turbulence and $k^{-3}$ for synoptic scale (\textbf{Fig.~\ref{TritonCast_ablation_exp}a, b}). Removing it (\textbf{No MG}) prevents the model from forming detailed weather patterns, and its long-term forecast becomes biased (\textbf{Fig.~\ref{TritonCast_ablation_exp}e}). The Latent Dynamical Core (\texttt{No LDC}) is the engine for long-term stability. By modeling the main dynamics only at the coarsest scale, it is shielded from high-frequency instabilities that arise from the accumulation of model errors in representing small scales. Removing it leads to rapid collapse during a year-long forecast (\textbf{Fig.~\ref{TritonCast_ablation_exp}d}). The skip-connections (\textbf{No SC}) are key for suppressing the accumulation of small-scale errors. By re-injecting precise details from the input, they directly correct the deviation of high-frequency signals, preventing the drift caused by error accumulation in long-term forecasts (\textbf{Fig.~\ref{TritonCast_ablation_exp}d}). Without them, important weather systems, like the polar vortex, move to incorrect locations (\textbf{Fig.~\ref{TritonCast_ablation_exp}e}). In short, \method{}'s success comes from how its parts work together as a well-designed team: the multi-grid structure handles multi-scale complexity, the latent core ensures long-term stability, and the skip-connections keep the details accurate. All three parts are necessary and work together to create a reliable model.

To directly test the critical role of high-frequency information for long-term stability, we conduct a targeted experiment by continuously filtering these components during the forecast. This experiment confirms that the absence of high-frequency details is a direct cause of catastrophic climate drift, providing strong physical evidence for our model's architecture. A detailed description and analysis of this experiment are available in Supplementary~\ref{appendix:ablation_Spectral}.

\section*{Discussion}

Accurate simulation and long-term forecasting of complex, multi-scale Earth system dynamics, are crucial for scientific understanding and addressing global challenges, from predicting turbulent motion on short time scales to modeling the Earth's climate response to radiative forcing. However, this goal faces significant hurdles. Traditional numerical methods contend with a trade-off between computational cost and physical accuracy. Current AI models often exhibit "spectral bias", struggling to capture high-frequency signals essential for long-term forecast, leading to error accumulation and physical inconsistencies. Therefore, overcoming spectral bias to build stable, physically consistent AI forecasting models is a key challenge in bridging Earth system science and artificial intelligence.

We have shown that \method{} address the core challenges of long-term Earth system forecasting. To validate its capability, we conducted a series of tests across atmospheric science, oceanography, and theoretical turbulence. In atmospheric science, beyond competitive medium-range predictions, the model demonstrates significant long-term stability: it not only completes a full-year forecast in a purely autoregressive mode but even successfully predicts the record-breaking $2020$ Siberian heatwave nearly six months in advance, using only initial conditions from the beginning of the year. This long-term capability is further confirmed in a 2500-day climate simulation, where after $2191$ days of integration (about $6$ years), the model's output for the global temperature field remains physically realistic, and no ensemble members show systematic drift, accurately reproducing the Earth climate's seasonal and inter-annual variations (\textbf{Fig.~\ref{TritonCast_weather}d}). This success is also seen in oceanography, where TritonCast extends the effective forecast horizon (ACC > $0.85$) for Kuroshio eddies from about $10$ days to $120$ days, a significant improvement (\textbf{Fig.~\ref{Figure_4_ocean_kurrent}a, d}). These applied achievements are rooted in the model's high physical fidelity, which is supported by its accurate reproduction of the theoretical $k^{-5/3}$ energy spectrum in turbulence tests (\textbf{Fig.~\ref{turbulence_comparison}b}) and its zero-shot cross-resolution generalization, where a model trained only on $0.25$\degree{} coarse-resolution data generates physically realistic forecasts on unseen $0.125$\degree{} fine-resolution grids (\textbf{Fig.~\ref{Figure_4_ocean_kurrent}f}). These achievements, combined with its high computational efficiency (e.g., a $365$-day global weather forecast requires only $56$ seconds on a single 40GB-A100 GPU), validate that TritonCast's architecture effectively overcomes the core bottleneck of spectral bias to provide a solid foundation for reliable and efficient, multi-scale, long-term Earth system forecasting, as also revealed by our extensive ablation studies (\textbf{Fig.~\ref{TritonCast_ablation_exp}}), .

While \method{} substantially advances in long-term Earth system forecasting, operational forecasting systems typically integrate model predictions with real-time observations to continuously correct state estimates. Therefore, effectively integrating \method{}'s powerful predictive ability with advanced Data Assimilation techniques is a crucial next step. Incorporating observational information promises to further enhance \method{}'s real-world prediction accuracy, correct potential model drift, and provide superior initial conditions for subsequent forecasts. Such synergy between models and observations is vital for transitioning AI models like \method{} to operational use and improving Earth system prediction capabilities.

\section*{Materials and Methods}\label{sec:method}

\subsection*{Dataset}
This section details the datasets employed in this study (The details of datasets can be found in Supplementary~\ref{appendix:datasets}). 

\noindent\textbf{\textit{Atmospheric Data}:} Atmospheric variables are sourced from the ECMWF Reanalysis v5 (ERA5) dataset~\cite{hersbach2020era5}. A total of 69 atmospheric variables are utilized in our analysis. These comprise five upper-air variables at 13 standard pressure levels (specifically 50 hPa, 100 hPa, 150 hPa, 200 hPa, 250 hPa, 300 hPa, 400 hPa, 500 hPa, 600 hPa, 700 hPa, 850 hPa, 925 hPa, and 1,000 hPa): geopotential (Z), temperature (T), zonal wind component (U), meridional wind component (V), and specific humidity (Q); along with four surface variables: 10 metre zonal wind component (U10M), 10 metre meridional wind component (V10M), 2 metre temperature (T2M), and mean sea level pressure (MSLP). For medium-range weather forecasting, we conduct experiments on 1.5 degree data with a 6-hour temporal resolution (00:00 UTC, 06:00 UTC, 12:00 UTC, and 18:00 UTC). We use data from 1959-2017 for training, 2018-2019 for validation, and 2020 for testing. For long-term weather forecasting, we conduct experiments on data with a 24 hour temporal resolution (12:00 UTC) and a 1 degree spatial resolution. For data partitioning, we use data from 1993-2017 for training, 2018 for validation, and 2019-2021 for testing. For multi-year climate simulation, we conduct experiments on data with a 24 hour temporal resolution and a 1.5 degree spatial resolution. Apart from the 69 ERA5 variables (12:00 UTC), Sea Surface Temperature (SST) from GLORYS12 reanalysis dataset is used to drive the atmospheric model's evolution. We use data from 1993-2017 for training and 2018-2024 for testing (See Supplementary~\ref{appendix:era5} for details on the datasets). 

%

\noindent\textbf{\textit{Ocean Data and Atmospheric Forcing}:} For oceanic data, we utilize the GLORYS12 reanalysis dataset. This dataset provides daily mean data covering latitudes from -80° to 90°, spanning the period since 1993. It features an original spatial resolution of 1/12 degree (corresponding to a 2041 × 4320 grid). In this work, we resample the GLORYS12 data to a 0.25-degree resolution (721 × 1440 grid points). To better adapt to the input of different architecture models, we use the data with size 720 × 1440. The model focuses on simulating five key ocean variables: sea salinity, sea stream zonal velocity, sea stream meridional velocity, and sea temperature across 23 vertical levels (depths: 0.49m, 2.65m, 5.08m, 7.93m, 11.41m, 15.81m, 21.60m, 29.44m, 40.34m, 55.76m, 77.85m, 92.32m, 109.73m, 130.67m, 155.85m, 186.13m, 222.48m, 266.04m, 318.13m, 380.21m, 453.94m, 541.09m and 643.57m), and sea surface height (SSH). To simulate ocean-atmosphere interactions, we incorporate four surface variables from the ERA5 reanalysis dataset as atmospheric forcing fields to drive the ocean model. These include the 10 metre zonal wind component (U10M), 10 metre meridional wind component (V10M), 2 metre temperature (T2M), and mean sea level pressure (MSLP). For data partitioning, we use data from 1993–2017 for training, 2018–2019 for validation, and 2020 for testing (See Supplementary~\ref{appendix:ocean_glorys}). 

\noindent\textbf{\textit{Global and Regional Geostrophic Velocity Data:}} This study utilizes daily sea surface geostrophic velocities (eastward ugos and northward vgos) for the period $1993$–$2024$, sourced from the Copernicus Marine Service's multi-satellite altimetry product. Our modeling strategy is twofold: first, to learn global large-scale circulation, we downsample the data to a $0.25\degree$ resolution and partition it into training ($1993$–$2018$), validation ($2019$), and test ($2020$-$2024$) sets. Second, since training in native $0.125\degree$ resolution is computationally prohibitive on a global scale, we train dedicated high-fidelity models directly on the original $0.125\degree$ data for three regions of intense mesoscale eddy activity, the Kuroshio Extension, the Gulf Stream, and the Agulhas Current, to precisely capture critical local ocean dynamics (See Supplementary~\ref{appendix:ocean_uvgos}).


\noindent\textbf{\textit{Kolmogorov Turbulence}:} We evaluate the model's handling of multi-scale dynamics using data from simulations of two-dimensional homogeneous isotropic decaying turbulence. The dataset, a canonical benchmark, is generated via direct numerical simulation of the vorticity transport equation on a $[0, 2 \pi]^2$ periodic domain. These simulations employ a $128 \times 128$ spatial resolution, a pseudo-spectral method with third-order Runge-Kutta time-stepping (CFL-constrained), and a Reynolds number of Re = $5000$. Initialization follows established procedures~\cite{mcwilliams1984emergence}, using a random vorticity field with a broad-band energy spectrum that induces a stable energy cascade (See Supplementary~\ref{appendix:dns_dataset}).

\subsection*{Problem Statement}

We aim to forecast the long-term evolution of diverse Earth system dynamics. The state of a given system at time $t$ is a multidimensional array $X^t$, whose constituent physical fields (e.g., temperature, velocity), structure, and resolution vary fundamentally across domains. Our goal is to learn a single, unified {neural solver}, denoted by $\Phi$, that implicitly learns the governing dynamics from data. Given $N$ historical states, $[X^{t-N+1}, \dots, X^t]$, $\Phi$ propagates the system forward to predict the next state, $\hat{X}^{t+1}$. For long-term forecasting, we apply this solver autoregressively, iteratively updating the input sequence with the model's own predictions:
\begin{equation}
    \hat{X}^{t+k} = \Phi(X^{t-N+K}, \dots, \hat{X}^{t+K-1})
    \label{eq:autoregressive_rollout}
\end{equation}
where predicted states $(\hat{X})$ replace observed states $(X)$ for steps $K > 1$. The central challenge is to design a solver $\Phi$ that is both versatile enough for this heterogeneity and remains numerically stable and physically consistent over long-duration rollouts ($K \gg 1$). A detailed description of the notation, including variable definitions and array dimensions, is provided in Supplementary~\ref{appendix_problem_statement}.

\subsection*{\method{} Model Architecture}

To overcome the challenge of long-term instability, we propose \method{}, a deep learning framework that instantiates the neural solver $\Phi$. Inspired by multi-grid methods, the core of \method{} lies in its hierarchical, multi-resolution architecture, explicitly designed to ``divide and conquer'' the forecasting problem. It separates the system's dynamics into a stable, low-frequency component and a detailed, high-frequency component, processes them with specialized modules, and then systematically reintegrates them. This architecture, following a V-cycle computational pattern (\textbf{Fig.~\ref{fig1}a)}, comprises three indispensable components whose synergistic roles are validated by our ablation studies (\textbf{Fig.~\ref{TritonCast_ablation_exp}}).

Let $V^{(l)}$ denote the discrete function space representing the system's state at resolution level $l$, where $l=0$ corresponds to the highest resolution $H \times W$, and $l=L$ represents the coarsest level. A state at time $t$ and level $l$ is $u_t^{(l)} \in V^{(l)}$. The state at the finest resolution, $u_t^{(0)}$, corresponds to the general state representation $X^t$ defined in the Problem Statement. The architectural design of \method{} is built upon two pillars: a {Multi-Grid Hierarchy} with {Skip-Connections} to ensure multi-scale fidelity, and a {Latent Dynamical Core} to guarantee long-term stability.

\subsubsection*{The Multi-Grid Hierarchy and Skip-Connections}
The foundation of \method{} is its hierarchical structure that processes information across multiple scales simultaneously. This is achieved through a {Restriction Path (Encoder)}, which decomposes the input, and a {Prolongation Path (Decoder)}, which reconstructs the output.

On the {Restriction Path}, a high-resolution input state $u_t^{(l)}$ is progressively downsampled. At each level $l$, a learnable operator $\mathcal{S}_{\text{enc}}^{(l)}$ extracts relevant features $f_t^{(l)}$, which are then passed through a restriction operator $\mathcal{R}_l^{(l+1)}$ to produce the state representation $u_t^{(l+1)}$ for the next coarser level. Crucially, the extracted features $f_t^{(l)}$ are stored at each level.
\begin{gather}\label{eq2}
    f_t^{(l)} = \mathcal{S}_{\text{enc}}^{(l)}(u_t^{(l)}; \theta_{\mathcal{S}_{\text{enc}}}^{(l)}) \\
    u_t^{(l+1)} = \mathcal{R}_l^{(l+1)}(f_t^{(l)}; \theta_{\mathcal{R}}^{(l)})
\end{gather}
Where $\theta$ denotes the learnable parameters of the corresponding operators. This hierarchical decomposition is essential for preserving the physical realism of the simulation. Our ablation study shows that removing this multi-grid structure ({No MG}) leads to severe, unphysical energy dissipation at small scales, causing the model to fail in reproducing the correct theoretical energy cascades for turbulence and weather ($k^{-5/3}$ and $k^{-3}$, \textbf{Fig.~\ref{TritonCast_ablation_exp}b}).

On the {Prolongation Path}, information propagates from coarse grids back to finer grids. At each level $l$, the predicted state from the coarser level, $\hat{u}_{t+1}^{(l+1)}$, is upsampled by a prolongation operator $\mathcal{P}_{l+1}^{(l)}$. This up-propagated information is then fused with the stored high-frequency features $f_t^{(l)}$ from Eq.~\ref{eq2} via {Skip-Connections}. A final refinement network $\mathcal{N}_{\text{refine}}^{(l)}$ processes this fused information to yield the final prediction $\hat{u}_{t+1}^{(l)}$ for that level.
\begin{gather}
    u_{\text{coarse} \to \text{fine}}^{(l)} = \mathcal{P}_{l+1}^{(l)}(\hat{u}_{t+1}^{(l+1)}; \theta_{\mathcal{P}}^{(l+1)}) \\
    \hat{u}_{t+1}^{(l)} = \mathcal{N}_{\text{refine}}^{(l)}\left( \mathcal{F}\text{usion}\left(u_{\text{coarse} \to \text{fine}}^{(l)}, f_t^{(l)}\right); \theta_{\mathcal{N}}^{(l)} \right)
\end{gather}
The role of Skip-Connections is to ensure long-term forecast fidelity. Without them ({No SC}), the model suffers from severe climate drift (\textbf{Fig.~\ref{TritonCast_ablation_exp}d}), causing major weather systems like the stratospheric polar vortex to be misplaced after a year-long forecast (\textbf{Fig.~\ref{TritonCast_ablation_exp}e}).

\subsubsection*{The Latent Dynamical Core}
At the coarsest level $L$ of the hierarchy, \method{} employs a {Latent Dynamical Core (LDC)}, $\mathcal{F}_{\text{latent}}$. This component's responsibility is to simulate the spatio-temporal evolution of the system's dominant, large-scale dynamics. By operating only on this coarse, spectrally simplified latent representation $Z_t^{(L)}$, the LDC is shielded from high-frequency noise and instabilities, enabling stable, long-term integration.
\begin{equation}
    \hat{z}_{t+1}^{(L)} = \mathcal{F}_{\text{latent}}(Z_t^{(L)}; \theta_{\text{latent}}), \quad \text{where } Z_t^{(L)} = [z_{t-N+1}^{(L)}, \dots, z_t^{(L)}]
\end{equation}
The LDC is the engine of \method{}'s stability. Its indispensability is starkly illustrated in our ablation study: removing the LDC ({No LDC}) leads to {catastrophic model failure} within a few forecast steps (\textbf{Fig.~\ref{TritonCast_ablation_exp}d}, "Catastrophic Failure"), with enormous spectral errors across all frequencies (\textbf{Fig.~\ref{TritonCast_ablation_exp}a}). This confirms that a dedicated, stable core for large-scale evolution is paramount for any long-term autoregressive forecast.

In concert, these three components, the Multi-Grid hierarchy for spectral fidelity, the Latent Dynamical Core for stability, and Skip-Connections for accuracy form a synergistic system. Their interplay, validated by our ablation studies, effectively mitigates spectral bias and suppresses error accumulation, enabling accurate and stable long-term Earth system forecasting.

\section*{Data Availability}

The GLORYS12 reanalysis datasets are obtained from~\url{https://data.marine.copernicus.eu/product/GLOBAL_MULTIYEAR_PHY_001_030/description}. The ERA5 reanalysis datasets are obtained from~\url{https://cds.climate.copernicus.eu}. The CMEMS datasets are obtained from~\url{https://data.marine.copernicus.eu/product/SEALEVEL_GLO_PHY_L4_MY_008_047/download}. The Turbulence datasets are obtained from~\url{https://huggingface.co/datasets/scaomath/navier-stokes-dataset}. 

\section*{Code Availability}

All the pre-trained weights, example datasets, training logs, and other detailed information for our scenarios can be found on Hugging Face~\url{https://huggingface.co/TritonCast}, Github~\url{https://github.com/Alexander-wu/TritonCast}. Our homepage is~\url{https://tritoncast4earth.netlify.app/}.

\section*{Acknowledgments}
This work was supported by the National Natural Science Foundation of China (42125503, 42430602). Niklas Boers acknowledges funding by the Volkswagen Foundation and the European Union's Horizon Europe research and innovation programme under grant agreement No. 101137601 (ClimTip).

\bibliographystyle{refstyle}
\bibliography{references}

@article{hersbach2020era5,
  title={The ERA5 global reanalysis},
  author={Hersbach, Hans and Bell, Bill and Berrisford, Paul and Hirahara, Shoji and Hor{\'a}nyi, Andr{\'a}s and Mu{\~n}oz-Sabater, Joaqu{\'\i}n and Nicolas, Julien and Peubey, Carole and Radu, Raluca and Schepers, Dinand and others},
  journal={Quarterly journal of the royal meteorological society},
  volume={146},
  number={730},
  pages={1999--2049},
  year={2020},
  publisher={Wiley Online Library}
}

@article{ling2024fengwu,
  title={FengWu-W2S: A deep learning model for seamless weather-to-subseasonal forecast of global atmosphere},
  author={Ling, Fenghua and Chen, Kang and Wu, Jiye and Han, Tao and Luo, Jing-Jia and Ouyang, Wanli and Bai, Lei},
  journal={arXiv preprint arXiv:2411.10191},
  year={2024}
}

@article{watt2025ace2,
  title={ACE2: accurately learning subseasonal to decadal atmospheric variability and forced responses},
  author={Watt-Meyer, Oliver and Henn, Brian and McGibbon, Jeremy and Clark, Spencer K and Kwa, Anna and Perkins, W Andre and Wu, Elynn and Harris, Lucas and Bretherton, Christopher S},
  journal={npj Climate and Atmospheric Science},
  volume={8},
  number={1},
  pages={205},
  year={2025},
  publisher={Nature Publishing Group UK London}
}

@article{wang2024coupled,
  title={Coupled ocean-atmosphere dynamics in a machine learning earth system model},
  author={Wang, Chenggong and Pritchard, Michael S and Brenowitz, Noah and Cohen, Yair and Bonev, Boris and Kurth, Thorsten and Durran, Dale and Pathak, Jaideep},
  journal={arXiv preprint arXiv:2406.08632},
  year={2024}
}

@article{mcwilliams1984emergence,
  title={The emergence of isolated coherent vortices in turbulent flow},
  author={McWilliams, James C},
  journal={Journal of Fluid Mechanics},
  volume={146},
  pages={21--43},
  year={1984},
  publisher={Cambridge University Press}
}

@article{liang2025foundation,
  title={Foundation Models for Spatio-Temporal Data Science: A Tutorial and Survey},
  author={Liang, Yuxuan and Wen, Haomin and Xia, Yutong and Jin, Ming and Yang, Bin and Salim, Flora and Wen, Qingsong and Pan, Shirui and Cong, Gao},
  journal={arXiv preprint arXiv:2503.13502},
  year={2025}
}

@article{rasp2024weatherbench,
  title={WeatherBench 2: A benchmark for the next generation of data-driven global weather models},
  author={Rasp, Stephan and Hoyer, Stephan and Merose, Alexander and Langmore, Ian and Battaglia, Peter and Russell, Tyler and Sanchez-Gonzalez, Alvaro and Yang, Vivian and Carver, Rob and Agrawal, Shreya and others},
  journal={Journal of Advances in Modeling Earth Systems},
  volume={16},
  number={6},
  pages={e2023MS004019},
  year={2024},
  publisher={Wiley Online Library}
}

@article{price2025probabilistic,
  title={Probabilistic weather forecasting with machine learning},
  author={Price, Ilan and Sanchez-Gonzalez, Alvaro and Alet, Ferran and Andersson, Tom R and El-Kadi, Andrew and Masters, Dominic and Ewalds, Timo and Stott, Jacklynn and Mohamed, Shakir and Battaglia, Peter and others},
  journal={Nature},
  volume={637},
  number={8044},
  pages={84--90},
  year={2025},
  publisher={Nature Publishing Group UK London}
}

@article{bordoni2025futures,
  title={The futures of climate modeling},
  author={Bordoni, S and Kang, SM and Shaw, Tiffany A and Simpson, IR and Zanna, L},
  journal={npj Climate and Atmospheric Science},
  volume={8},
  number={1},
  pages={99},
  year={2025},
  publisher={Nature Publishing Group UK London}
}

@inproceedings{
raonic2023convolutional,
title={Convolutional Neural Operators for robust and accurate learning of {PDE}s},
author={Bogdan Raonic and Roberto Molinaro and Tim De Ryck and Tobias Rohner and Francesca Bartolucci and Rima Alaifari and Siddhartha Mishra and Emmanuel de Bezenac},
booktitle={Thirty-seventh Conference on Neural Information Processing Systems},
year={2023},
url={https://openreview.net/forum?id=MtekhXRP4h}
}

@article{lam2023learning,
  title={Learning skillful medium-range global weather forecasting},
  author={Lam, Remi and Sanchez-Gonzalez, Alvaro and Willson, Matthew and Wirnsberger, Peter and Fortunato, Meire and Alet, Ferran and Ravuri, Suman and Ewalds, Timo and Eaton-Rosen, Zach and Hu, Weihua and others},
  journal={Science},
  volume={382},
  number={6677},
  pages={1416--1421},
  year={2023},
  publisher={American Association for the Advancement of Science}
}

@inproceedings{he2024mgno,
	title={Mg{NO}: Efficient Parameterization of Linear Operators via Multigrid},
	author={Juncai He and Xinliang Liu and Jinchao Xu},
	booktitle={The Twelfth International Conference on Learning Representations},
	year={2024},
	url={https://openreview.net/forum?id=8OxL034uEr}
}

@article{gettelman2022future,
  title={The future of Earth system prediction: Advances in model-data fusion},
  author={Gettelman, Andrew and Geer, Alan J and Forbes, Richard M and Carmichael, Greg R and Feingold, Graham and Posselt, Derek J and Stephens, Graeme L and Van den Heever, Susan C and Varble, Adam C and Zuidema, Paquita},
  journal={Science Advances},
  volume={8},
  number={14},
  pages={eabn3488},
  year={2022},
  publisher={American Association for the Advancement of Science}
}

@article{kochkov2024neural,
  title={Neural general circulation models for weather and climate},
  author={Kochkov, Dmitrii and Yuval, Janni and Langmore, Ian and Norgaard, Peter and Smith, Jamie and Mooers, Griffin and Kl{\"o}wer, Milan and Lottes, James and Rasp, Stephan and D{\"u}ben, Peter and others},
  journal={Nature},
  volume={632},
  number={8027},
  pages={1060--1066},
  year={2024},
  publisher={Nature Publishing Group UK London}
}

@article{li2021towards,
  title={Towards multiscale modeling of ocean surface turbulent mixing using coupled MPAS-Ocean v6. 3 and PALM v5. 0},
  author={Li, Qing and Van Roekel, Luke},
  journal={Geoscientific Model Development},
  volume={14},
  number={4},
  pages={2011--2028},
  year={2021},
  publisher={Copernicus Publications G{\"o}ttingen, Germany}
}

@article{berloff2007ocean,
  title={Ocean eddy dynamics in a coupled ocean--atmosphere model},
  author={Berloff, P and Dewar, W and Kravtsov, S and McWilliams, J},
  journal={Journal of physical oceanography},
  volume={37},
  number={5},
  pages={1103--1121},
  year={2007}
}

@article{stechmann2014multiscale,
  title={Multiscale eddy simulation for moist atmospheric convection: Preliminary investigation},
  author={Stechmann, Samuel N},
  journal={Journal of Computational Physics},
  volume={271},
  pages={99--117},
  year={2014},
  publisher={Elsevier}
}

@article{marati2004energy,
  title={Energy cascade and spatial fluxes in wall turbulence},
  author={Marati, Nicoletta and Casciola, Carlo Massimo and Piva, Renzo},
  journal={Journal of Fluid Mechanics},
  volume={521},
  pages={191--215},
  year={2004},
  publisher={Cambridge University Press}
}

@incollection{leonard1975energy,
  title={Energy cascade in large-eddy simulations of turbulent fluid flows},
  author={Leonard, Athony},
  booktitle={Advances in geophysics},
  volume={18},
  pages={237--248},
  year={1975},
  publisher={Elsevier}
}

@article{biferale2003shell,
  title={Shell models of energy cascade in turbulence},
  author={Biferale, Luca},
  journal={Annual review of fluid mechanics},
  volume={35},
  number={1},
  pages={441--468},
  year={2003},
  publisher={Annual Reviews 4139 El Camino Way, PO Box 10139, Palo Alto, CA 94303-0139, USA}
}

@article{raaisaanen2007reliable,
  title={How reliable are climate models?},
  author={Ra{\"a}isa{\"a}nen, Jouni},
  journal={Tellus A: Dynamic Meteorology and Oceanography},
  volume={59},
  number={1},
  pages={2--29},
  year={2007},
  publisher={Taylor \& Francis}
}

@book{trefethen2000spectral,
  title={Spectral methods in MATLAB},
  author={Trefethen, Lloyd N},
  year={2000},
  publisher={SIAM}
}

@article{hess2022physically,
  title={Physically constrained generative adversarial networks for improving precipitation fields from Earth system models},
  author={Hess, Philipp and Dr{\"u}ke, Markus and Petri, Stefan and Strnad, Felix M and Boers, Niklas},
  journal={Nature Machine Intelligence},
  volume={4},
  number={10},
  pages={828--839},
  year={2022},
  publisher={Nature Publishing Group UK London}
}

@article{liu2015systems,
  title={Systems integration for global sustainability},
  author={Liu, Jianguo and Mooney, Harold and Hull, Vanessa and Davis, Steven J and Gaskell, Joanne and Hertel, Thomas and Lubchenco, Jane and Seto, Karen C and Gleick, Peter and Kremen, Claire and others},
  journal={Science},
  volume={347},
  number={6225},
  pages={1258832},
  year={2015},
  publisher={American Association for the Advancement of Science}
}

@book{leveque2007finite,
  title={Finite difference methods for ordinary and partial differential equations: steady-state and time-dependent problems},
  author={LeVeque, Randall J},
  year={2007},
  publisher={SIAM}
}

@book{ferziger2019computational,
  title={Computational methods for fluid dynamics},
  author={Ferziger, Joel H and Peri{\'c}, Milovan and Street, Robert L},
  year={2019},
  publisher={springer}
}

@article{lorenz1969predictability,
  title={The predictability of a flow which possesses many scales of motion},
  author={Lorenz, Edward N},
  journal={Tellus},
  volume={21},
  number={3},
  pages={289--307},
  year={1969},
  publisher={Taylor \& Francis}
}

@book{durran2010numerical,
  title={Numerical methods for fluid dynamics: With applications to geophysics},
  author={Durran, Dale R},
  volume={32},
  year={2010},
  publisher={Springer Science \& Business Media}
}

@article{pope2001turbulent,
  title={Turbulent flows},
  author={Pope, Stephen B},
  journal={Measurement Science and Technology},
  volume={12},
  number={11},
  pages={2020--2021},
  year={2001}
}

@article{cui2025forecasting,
  title={Forecasting the eddying ocean with a deep neural network},
  author={Cui, Yingzhe and Wu, Ruohan and Zhang, Xiang and Zhu, Ziqi and Liu, Bo and Shi, Jun and Chen, Junshi and Liu, Hailong and Zhou, Shenghui and Su, Liang and others},
  journal={Nature Communications},
  volume={16},
  number={1},
  pages={2268},
  year={2025},
  publisher={Nature Publishing Group UK London}
}

@book{stensrud2007parameterization,
  title={Parameterization schemes: keys to understanding numerical weather prediction models},
  author={Stensrud, David J},
  year={2007},
  publisher={Cambridge University Press}
}

@book{pedlosky2013geophysical,
  title={Geophysical fluid dynamics},
  author={Pedlosky, Joseph},
  year={2013},
  publisher={Springer Science \& Business Media}
}

@article{palmer2001nonlinear,
  title={A nonlinear dynamical perspective on model error: A proposal for non-local stochastic-dynamic parametrization in weather and climate prediction models},
  author={Palmer, Tim N},
  journal={Quarterly Journal of the Royal Meteorological Society},
  volume={127},
  number={572},
  pages={279--304},
  year={2001},
  publisher={Wiley Online Library}
}

@incollection{lorenz2017deterministic,
  title={Deterministic Nonperiodic Flow 1},
  author={Lorenz, Edward N},
  booktitle={Universality in Chaos, 2nd edition},
  pages={367--378},
  year={2017},
  publisher={Routledge}
}

@article{meehl2000coupled,
  title={The coupled model intercomparison project (CMIP)},
  author={Meehl, Gerald A and Boer, George J and Covey, Curt and Latif, Mojib and Stouffer, Ronald J},
  journal={Bulletin of the American Meteorological Society},
  volume={81},
  number={2},
  pages={313--318},
  year={2000},
  publisher={JSTOR}
}

@article{bauer2015quiet,
  title={The quiet revolution of numerical weather prediction},
  author={Bauer, Peter and Thorpe, Alan and Brunet, Gilbert},
  journal={Nature},
  volume={525},
  number={7567},
  pages={47--55},
  year={2015},
  publisher={Nature Publishing Group UK London}
}

@article{schneider2017earth,
  title={Earth system modeling 2.0: A blueprint for models that learn from observations and targeted high-resolution simulations},
  author={Schneider, Tapio and Lan, Shiwei and Stuart, Andrew and Teixeira, Joao},
  journal={Geophysical Research Letters},
  volume={44},
  number={24},
  pages={12--396},
  year={2017},
  publisher={Wiley Online Library}
}

@article{reichstein2019deep,
  title={Deep learning and process understanding for data-driven Earth system science},
  author={Reichstein, Markus and Camps-Valls, Gustau and Stevens, Bjorn and Jung, Martin and Denzler, Joachim and Carvalhais, Nuno and Prabhat, F},
  journal={Nature},
  volume={566},
  number={7743},
  pages={195--204},
  year={2019},
  publisher={Nature Publishing Group UK London}
}

@article{irrgang2021towards,
  title={Towards neural Earth system modelling by integrating artificial intelligence in Earth system science},
  author={Irrgang, Christopher and Boers, Niklas and Sonnewald, Maike and Barnes, Elizabeth A and Kadow, Christopher and Staneva, Joanna and Saynisch-Wagner, Jan},
  journal={Nature Machine Intelligence},
  volume={3},
  number={8},
  pages={667--674},
  year={2021},
  publisher={Nature Publishing Group UK London}
}

@article{lecun2015deep,
  title={Deep learning},
  author={LeCun, Yann and Bengio, Yoshua and Hinton, Geoffrey},
  journal={nature},
  volume={521},
  number={7553},
  pages={436--444},
  year={2015},
  publisher={Nature Publishing Group UK London}
}

@article{pathak2022fourcastnet,
  title={Fourcastnet: A global data-driven high-resolution weather model using adaptive fourier neural operators},
  author={Pathak, Jaideep and Subramanian, Shashank and Harrington, Peter and Raja, Sanjeev and Chattopadhyay, Ashesh and Mardani, Morteza and Kurth, Thorsten and Hall, David and Li, Zongyi and Azizzadenesheli, Kamyar and others},
  journal={arXiv preprint arXiv:2202.11214},
  year={2022}
}

@article{xiong2023ai,
  title={Ai-goms: Large ai-driven global ocean modeling system},
  author={Xiong, Wei and Xiang, Yanfei and Wu, Hao and Zhou, Shuyi and Sun, Yuze and Ma, Muyuan and Huang, Xiaomeng},
  journal={arXiv preprint arXiv:2308.03152},
  year={2023}
}

@article{guo2025data,
  title={Data-driven global ocean modeling for seasonal to decadal prediction},
  author={Guo, Zijie and Lyu, Pumeng and Ling, Fenghua and Bai, Lei and Luo, Jing-Jia and Boers, Niklas and Yamagata, Toshio and Izumo, Takeshi and Cravatte, Sophie and Capotondi, Antonietta and others},
  journal={Science Advances},
  volume={11},
  number={33},
  pages={eadu2488},
  year={2025},
  publisher={American Association for the Advancement of Science}
}

@article{manabe1969climate,
  title={Climate calculations with a combined ocean-atmosphere model},
  author={Manabe, Syukuro and Bryan, Kirk},
  journal={Journal of Atmospheric Sciences},
  volume={26},
  number={4},
  pages={786--789},
  year={1969}
}

@article{lippe2023pde,
  title={Pde-refiner: Achieving accurate long rollouts with neural pde solvers},
  author={Lippe, Phillip and Veeling, Bas and Perdikaris, Paris and Turner, Richard and Brandstetter, Johannes},
  journal={Advances in Neural Information Processing Systems},
  volume={36},
  pages={67398--67433},
  year={2023}
}

@article{bi2023accurate,
  title={Accurate medium-range global weather forecasting with 3D neural networks},
  author={Bi, Kaifeng and Xie, Lingxi and Zhang, Hengheng and Chen, Xin and Gu, Xiaotao and Tian, Qi},
  journal={Nature},
  volume={619},
  number={7970},
  pages={533--538},
  year={2023},
  publisher={Nature Publishing Group}
}

@inproceedings{Brandt1977MultilevelAS,
  title={Multi-level adaptive solutions to boundary-value problems math comptr},
  author={Achi Brandt},
  year={1977},
  url={https://api.semanticscholar.org/CorpusID:5919303}
}

@article{wang2024xihe,
  title={Xihe: A data-driven model for global ocean eddy-resolving forecasting},
  author={Wang, Xiang and Wang, Renzhi and Hu, Ningzi and Wang, Pinqiang and Huo, Peng and Wang, Guihua and Wang, Huizan and Wang, Senzhang and Zhu, Junxing and Xu, Jianbo and others},
  journal={arXiv preprint arXiv:2402.02995},
  year={2024}
}

@article{bodnar2025foundation,
  title={A foundation model for the Earth system},
  author={Bodnar, Cristian and Bruinsma, Wessel P and Lucic, Ana and Stanley, Megan and Allen, Anna and Brandstetter, Johannes and Garvan, Patrick and Riechert, Maik and Weyn, Jonathan A and Dong, Haiyu and others},
  journal={Nature},
  pages={1--8},
  year={2025},
  publisher={Nature Publishing Group UK London}
}

@article{li2020fourier,
  title={Fourier neural operator for parametric partial differential equations},
  author={Li, Zongyi and Kovachki, Nikola and Azizzadenesheli, Kamyar and Liu, Burigede and Bhattacharya, Kaushik and Stuart, Andrew and Anandkumar, Anima},
  journal={arXiv preprint arXiv:2010.08895},
  year={2020}
}

@article{andrychowicz2023deep,
  title={Deep learning for day forecasts from sparse observations},
  author={Andrychowicz, Marcin and Espeholt, Lasse and Li, Di and Merchant, Samier and Merose, Alexander and Zyda, Fred and Agrawal, Shreya and Kalchbrenner, Nal},
  journal={arXiv preprint arXiv:2306.06079},
  year={2023}
}

@inproceedings{rahaman2019spectral,
  title={On the spectral bias of neural networks},
  author={Rahaman, Nasim and Baratin, Aristide and Arpit, Devansh and Draxler, Felix and Lin, Min and Hamprecht, Fred and Bengio, Yoshua and Courville, Aaron},
  booktitle={International conference on machine learning},
  pages={5301--5310},
  year={2019},
  organization={PMLR}
}

@article{kenneth2019international,
  title={International best track archive for climate stewardship (IBTrACS) project, version 4},
  author={Kenneth, R and Howard, J and James, P and Michael, C and Carl, J},
  journal={(No Title)},
  year={2019},
  publisher={NOAA National Centers for Environmental Information}
}

@article{knapp2010international,
  title={The international best track archive for climate stewardship (IBTrACS) unifying tropical cyclone data},
  author={Knapp, Kenneth R and Kruk, Michael C and Levinson, David H and Diamond, Howard J and Neumann, Charles J},
  journal={Bulletin of the American Meteorological Society},
  volume={91},
  number={3},
  pages={363--376},
  year={2010},
  publisher={American Meteorological Society}
}

@article{xu2019frequency,
  title={Frequency principle: Fourier analysis sheds light on deep neural networks},
  author={Xu, Zhi-Qin John and Zhang, Yaoyu and Luo, Tao and Xiao, Yanyang and Ma, Zheng},
  journal={arXiv preprint arXiv:1901.06523},
  year={2019}
}

@article{fridovich2022spectral,
  title={Spectral bias in practice: The role of function frequency in generalization},
  author={Fridovich-Keil, Sara and Gontijo Lopes, Raphael and Roelofs, Rebecca},
  journal={Advances in Neural Information Processing Systems},
  volume={35},
  pages={7368--7382},
  year={2022}
}

@article{lang2024aifs,
  title={AIFS-CRPS: ensemble forecasting using a model trained with a loss function based on the continuous ranked probability score},
  author={Lang, Simon and Alexe, Mihai and Clare, Mariana CA and Roberts, Christopher and Adewoyin, Rilwan and Bouall{\`e}gue, Zied Ben and Chantry, Matthew and Dramsch, Jesper and Dueben, Peter D and Hahner, Sara and others},
  journal={arXiv preprint arXiv:2412.15832},
  year={2024}
}

@inproceedings{pfaff2020learning,
  title={Learning mesh-based simulation with graph networks},
  author={Pfaff, Tobias and Fortunato, Meire and Sanchez-Gonzalez, Alvaro and Battaglia, Peter},
  booktitle={International conference on learning representations},
  year={2020}
}

@article{keisler2022forecasting,
  title={Forecasting global weather with graph neural networks},
  author={Keisler, Ryan},
  journal={arXiv preprint arXiv:2202.07575},
  year={2022}
}

@article{yang2025generative,
  title={Generative assimilation and prediction for weather and climate},
  author={Yang, Shangshang and Nai, Congyi and Liu, Xinyan and Li, Weidong and Chao, Jie and Wang, Jingnan and Wang, Leyi and Li, Xichen and Chen, Xi and Lu, Bo and others},
  journal={arXiv preprint arXiv:2503.03038},
  year={2025}
}

@article{selz2025effective,
  title={On the effective resolution of AI weather prediction models},
  author={Selz, Tobias and Bruinsma, Wessel and Craig, George C and Markou, Stratis and Turner, Richard and Vaughan, Anna},
  journal={Authorea Preprints},
  year={2025},
  publisher={Authorea}
}

@article{dheeshjith2025samudra,
  title={Samudra: An AI global ocean emulator for climate},
  author={Dheeshjith, Surya and Subel, Adam and Adcroft, Alistair and Busecke, Julius and Fernandez-Granda, Carlos and Gupta, Shubham and Zanna, Laure},
  journal={Geophysical Research Letters},
  volume={52},
  number={10},
  pages={e2024GL114318},
  year={2025},
  publisher={Wiley Online Library}
}

@article{alet2025skillful,
  title={Skillful joint probabilistic weather forecasting from marginals},
  author={Alet, Ferran and Price, Ilan and El-Kadi, Andrew and Masters, Dominic and Markou, Stratis and Andersson, Tom R and Stott, Jacklynn and Lam, Remi and Willson, Matthew and Sanchez-Gonzalez, Alvaro and others},
  journal={arXiv preprint arXiv:2506.10772},
  year={2025}
}

@article{gupta2022towards,
  title={Towards multi-spatiotemporal-scale generalized pde modeling},
  author={Gupta, Jayesh K and Brandstetter, Johannes},
  journal={arXiv preprint arXiv:2209.15616},
  year={2022}
}

@article{hao2025deep,
  title={Deep Learning for Ocean Forecasting: A Comprehensive Review of Methods, Applications, and Datasets},
  author={Hao, Rixu and Zhao, Yuxin and Zhang, Shaoqing and Deng, Xiong},
  journal={IEEE Transactions on Cybernetics},
  year={2025},
  publisher={IEEE}
}

@article{kido2023skillful,
  title={Skillful multiyear prediction of the Kuroshio and Gulf Stream jets and eddy activity},
  author={Kido, Shoichiro and Nonaka, Masami and Miyazawa, Yasumasa},
  journal={Geophysical Research Letters},
  volume={50},
  number={15},
  pages={e2023GL103705},
  year={2023},
  publisher={Wiley Online Library}
}

\clearpage
\begin{spacing}{1.2}
\tableofcontents
\end{spacing}

\clearpage
\appendix
\renewcommand{\appendixpagename}{\Large Supplementary}
\renewcommand{\appendixtocname}{\Large Supplementary}
\begin{appendices}
\section{Notation and problem statement}
\label{appendix_problem_statement}

\subsection{General Notations and Tensor Representations}

To ensure clarity and rigor throughout this paper, we formally define the core mathematical notations and tensor representations used herein. We consider the complete physical state of an Earth system at a discrete time step $t$ as a "snapshot" of the system. Conceptually, this state is composed of multiple physical variables distributed over a three-dimensional spatial grid. The most natural representation is therefore a 4D tensor, comprising dimensions for {physical variables ($C'$)}, {vertical levels ($\mathcal{D}$)}, {latitude ($H$)}, and {longitude ($W$)}. However, to leverage deep learning operators (e.g., 2D convolutions, 2D attention mechanisms) that are highly optimized for 2D spatial data, we perform a representational transformation at the model's input stage. Specifically, the physical variable dimension ($C'$) and the vertical level dimension ($\mathcal{D}$) are {flattened and concatenated} into a single, expanded channel dimension, denoted as $C$. This process can be illustrated with a clear example:

In the atmospheric science domain, consider $C'=5$ upper-air variables (Z, T, U, V, Q), each distributed across $\mathcal{D}=13$ standard pressure levels. After flattening, this data forms $C'_\text{upper} = 5 \times 13 = 65$ channels. Combined with $C'=4$ surface variables (for which $\mathcal{D}=1$), the total number of channels for the model becomes $C = 65 + 4 = 69$. In the ocean science domain, the same methodology is applied to ocean variables at different depth levels.

Through this transformation, the original 4D state $\mathbf{X}_t \in \mathbb{R}^{\mathcal{D} \times  C' \times H \times W}$ is effectively reshaped into a 3D tensor $\mathbf{X}_t \in \mathbb{R}^{C \times H \times W}$ that the model can process directly. Consequently, for the remainder of this paper, any reference to the state tensor $\mathbf{X}_t$ implies this 3D representation with the flattened vertical dimension, unless otherwise specified. The central task of our work is to learn a {neural solver}, uniformly denoted as $\Phi$, that approximates the dynamical evolution of the Earth system. This solver $\Phi$ takes a sequence of $N$ consecutive historical states, a "history window", $[\mathbf{X}_{t-N+1}, \dots, \mathbf{X}_t]$ as its input. The model's output is the prediction for the next time step, $t+1$, which we denote as $\hat{\mathbf{X}}_{t+1}$. We use the caret symbol ($\hat{\cdot}$) to explicitly distinguish the model's {prediction} from the {ground truth} $\mathbf{X}_{t+1}$, which is derived from observations or high-fidelity simulations. This fundamental single-step forecasting task is thus formulated as a mapping from $\mathbb{R}^{\mathcal{D} \times C' \times H \times W}$ to $\mathbb{R}^{C \times H \times W}$:
\begin{equation}
\hat{\mathbf{X}}_{t+1} = \Phi(\mathbf{X}_{t-N+1}, \dots, \mathbf{X}_t)
\label{eq:appendix_general_forecast_en_revised}
\end{equation}

\subsection{Formalism of the Multi-Scale Problem}

The single, fixed-resolution representation defined in the previous section, while sufficient for describing a single-step forecast, overlooks a critical property of Earth system dynamics: its {multi-scale nature}. The system's evolution is driven by the interaction of processes spanning from global-scale atmospheric and oceanic circulations, to mesoscale weather systems and ocean eddies, down to small-scale turbulence and convection. Mainstream deep learning models exhibit a "spectral bias", meaning they preferentially learn large-scale, low-frequency signals while struggling to capture high-frequency, small-scale details that are energetically smaller but dynamically crucial. This deficiency is detrimental for long-term forecasting, as initial small-scale errors grow non-linearly and contaminate all scales, eventually causing the forecast to diverge.

Therefore, to build a model capable of stable long-term integration, we must formalize the problem within a hierarchical, multi-scale mathematical framework.

\subsubsection{Function Spaces and Discretization Levels}

We introduce a discretization level index $l \in \{0, 1, \dots, L\}$ to denote different spatial resolutions, where $l=0$ corresponds to the highest (finest) resolution and $l=L$ to the lowest (coarsest). At each level $l$, the system's state has a corresponding discrete representation, $\mathbf{u}_t^{(l)}$. We define the set of all possible states at level $l$ as a discrete function space $\mathcal{V}^{(l)}$, such that $\mathbf{u}_t^{(l)} \in \mathcal{V}^{(l)}$.

To connect with the previous notation, we establish that the state at the finest level is equivalent to the global state tensor: $\mathbf{u}_t^{(0)} \equiv \mathbf{X}_t$. Thus, a complete system state is viewed as a hierarchical set of representations, $\{\mathbf{u}_t^{(0)}, \mathbf{u}_t^{(1)}, \dots, \mathbf{u}_t^{(L)}\}$.

Information transfer between scales is achieved through specific operators. We define a {Restriction Operator}, $R^{(l+1)}$, which maps information from a fine grid to a coarser one, i.e., $R^{(l+1)}: \mathcal{V}^{(l)} \to \mathcal{V}^{(l+1)}$. Correspondingly, a {Prolongation Operator}, $P^{(l)}$, maps information from a coarse grid back to a finer one, i.e., $P^{(l)}: \mathcal{V}^{(l+1)} \to \mathcal{V}^{(l)}$. Using these operators, we can recursively generate representations at all coarser scales starting from the highest-resolution state $\mathbf{u}_t^{(0)}$:
\begin{equation}
\mathbf{u}_t^{(l+1)} = R^{(l+1)}(\mathbf{u}_t^{(l)})
\end{equation}
This multi-scale framework forms the theoretical cornerstone of the TritonCast architecture, allowing the model to separate and process different dynamical processes at their appropriate scales.

\subsubsection{Mathematical Formulation of Autoregressive Forecasting}

The true challenge in Earth system science is not single-step prediction but rather achieving stable, long-term prediction. This requires the model to perform {autoregressive} forecasting.

In an autoregressive rollout, the model's prediction at one step becomes the input for the next, allowing it to iteratively advance the simulation. Let $\hat{\mathbf{X}}_{t+k}$ denote the prediction for step $k$ initiated from time $t$. The process is mathematically formulated as:
\begin{equation}
\hat{\mathbf{X}}_{t+K} = \Phi\left(
    \begin{cases}
        [\mathbf{X}_{t-N+K}, \dots, \mathbf{X}_{t+K-1}] & \text{if } K=1 \\
        [\hat{\mathbf{X}}_{t+K-N}, \dots, \hat{\mathbf{X}}_{t+K-1}] & \text{if } K > 1
    \end{cases}
\right)
\label{eq:autoregressive_formulation_en}
\end{equation}
where, for notational consistency, we consider the historical ground truth $\mathbf{X}$ to be part of the prediction sequence $\hat{\mathbf{X}}$ for steps $K \le 0$.

Equation \eqref{eq:autoregressive_formulation_en} reveals the core dilemma of long-term forecasting: for any $K>1$, the model's input consists entirely of its own past predictions. This means that any small prediction error, $\varepsilon_{K-1} = \hat{\mathbf{X}}_{t+K-1} - \mathbf{X}_{t+K-1}$, is fed back into the model as part of the input for step $K$. In a non-linear dynamical system, such errors are rapidly amplified over time, causing the forecast trajectory to diverge from the true one and potentially leading to numerical collapse.

Therefore, the central scientific problem this paper addresses is the design of a general and robust neural solver $\Phi$ that maintains both {numerical stability} and {physical consistency} throughout the long-term autoregressive rollouts described by Equation \eqref{eq:autoregressive_formulation_en}, ultimately enabling reliable forecasting within a multi-scale framework.

\subsection{List and Description of Key Physical Variables}

To provide a clear and centralized reference, the following table details the key physical variables used across the different physical domains (Atmospheric Science, Ocean Science, and Theoretical Physics) in this study. The table includes the full name of each variable, its mathematical symbol as used in the text, a concise physical description (e.g., its vertical distribution or role as a forcing field), and its corresponding data source. These definitions are consistent with our descriptions of the datasets in the  Supplementary~\ref{appendix:datasets} (Overview of Datasets).

\begin{table}[htbp]
\centering
\caption{\textbf{A comprehensive list of the key physical variables used in this study.}}
\label{tab:physical_variables_en}
\scriptsize 
\begin{tabular}{lllll}
\toprule
\textbf{Domain} & \textbf{Variable Name} & \textbf{Symbol} & \textbf{Description} & \textbf{Data Source} \\
\midrule
\multirow{9}{*}{Atmospheric Science} & Geopotential & Z & Upper-air variable at 13 standard pressure levels & ERA5 \\
& Specific humidity & Q & Upper-air variable at 13 standard pressure levels & ERA5 \\
 & Temperature & T & Upper-air variable at 13 standard pressure levels & ERA5 \\
 & Zonal wind component & U & Upper-air variable at 13 standard pressure levels & ERA5 \\
 & Meridional wind component & V & Upper-air variable at 13 standard pressure levels & ERA5 \\
 
\cmidrule(l){2-5}
 & 10 metre u wind component & U10M & Surface variable & ERA5 \\
 & 10 metre v wind component & V10M & Surface variable & ERA5 \\
 & 2 metre temperature  & T2M & Surface variable & ERA5 \\
 & Mean sea level pressure & MSLP & Surface variable & ERA5 \\
\midrule
\multirow{7}{*}{Ocean Science} & Sea Salinity & S & 3D variable at 23 vertical depth levels & GLORYS12 \\
 & Sea stream zonal velocity & $\mathrm{U}_{\mathrm{o}}$ & 3D variable at 23 vertical depth levels & GLORYS12 \\
 & Sea stream meridional velocity & $\mathrm{V}_{\mathrm{o}}$ & 3D variable at 23 vertical depth levels & GLORYS12 \\
 & Sea temperature & $\mathrm{T}_{\mathrm{o}}$ & 3D variable at 23 vertical depth levels & GLORYS12 \\

 & Sea surface height & SSH & 2D sea-surface variable & GLORYS12 \\
 \cmidrule(l){2-5}
 & Eastward geostrophic velocity & ugos & 2D sea surface geostrophic current component & CMEMS \\
 & Northward geostrophic velocity & vgos & 2D sea surface geostrophic current component & CMEMS\\
\midrule
\multirow{4}{*}{\begin{tabular}[c]{@{}l@{}}Ocean Forcing Fields\\ (from Atmosphere)\end{tabular}} & 10 metre u wind component & U10M & Atmospheric forcing field for the ocean model & ERA5 \\
 & 10 metre v wind component & V10M & Atmospheric forcing field for the ocean model & ERA5 \\
 & 2 metre temperature & T2M & Atmospheric forcing field for the ocean model & ERA5 \\
 & Mean sea level pressure & MSLP & Atmospheric forcing field for the ocean model & ERA5 \\
\midrule
Theoretical Physics & Vorticity & $\omega$ & Scalar field in 2D Isotropic Turbulence & DNS \\
\bottomrule
\end{tabular}
\end{table}

\clearpage
\subsection{Notations}

We summarize the key notations used throughout this paper in \textbf{Table~\ref{tab:notations_en}}.

\begin{table}[h!]
\centering
\caption{\textbf{Summary of key notations used in this work.}}
\label{tab:notations_en}
\setlength{\tabcolsep}{18pt}
\small
\begin{tabular}{ll}
\toprule
\textbf{Notation} & \textbf{Meaning in this work} \\
\midrule
\multicolumn{2}{l}{\textit{General Notations}} \\
$t$ & Discrete time step index. \\
$\mathbf{X}_t \in \mathbb{R}^{C \times H \times W}$ & State tensor of the Earth system at time $t$, defined on a spatial grid. \\
$H, W$ & Spatial dimensions (e.g., height/latitude, width/longitude). \\
$C$ & Number of input channels, with physical variables and vertical levels flattened. \\
$N$ & Number of input time steps (input sequence length). \\
$K$ & {Forecast step index} (lead time). \\
$\hat{\mathbf{X}}_{t+K}$ & Predicted state by the model at future time $t+K$. \\
$\Phi$ & The AI forecasting model (i.e., the neural solver, e.g., \method{}). \\
$\theta$ & Learnable parameters of the model $\Phi$. \\
\midrule
\multicolumn{2}{l}{\textit{Notations for the \method{} Architecture}} \\
$l$ & {Resolution level index} in \method{}'s hierarchy ($l=0$ is finest). \\
$\mathbf{Z}^{(l)}$ & Feature map representation at resolution level $l$. \\
$L$ & Index of the coarsest resolution level in \method{}. \\
$\mathcal{R}^{(l \to l+1)}$ & Restriction operator (maps features from level $l$ to $l+1$; downsampling). \\
$\mathcal{P}^{(l \to l-1)}$ & Prolongation operator (maps features from level $l$ to $l-1$; upsampling). \\
$\mathcal{L}$ & The Latent Dynamical Core (LDC) operating at the coarsest level $L$. \\
\midrule
\multicolumn{2}{l}{\textit{Notations for Analysis and Evaluation}} \\
$k$ & {Wavenumber}, used in spectral analysis. \\
RMSE & Root Mean Square Error (evaluation metric). \\
ACC & Anomaly Correlation Coefficient (evaluation metric). \\
\bottomrule
\end{tabular}
\end{table}
\clearpage

\begin{figure}[t!]
\centering
\includegraphics[width=\linewidth]{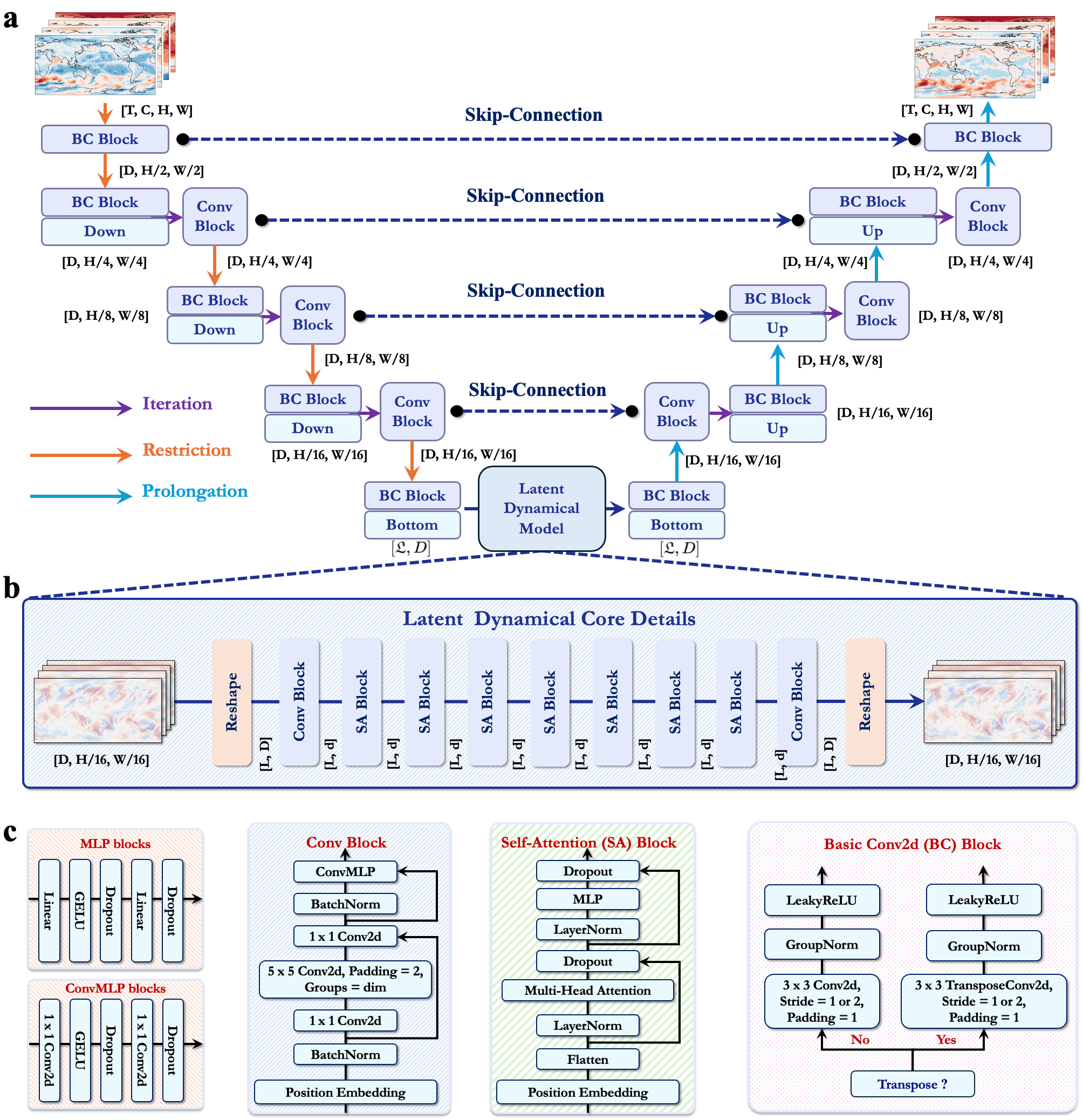}
\caption{\textbf{{Architecture of the \method{} Model.}}
{a}, Overview: The multi-grid inspired \method{} architecture hierarchically processes input data. Its core operations iterative updates within levels, downsampling via BC Blocks (orange arrows), and upsampling (cyan arrows) mimic multi-grid V-cycles. Skip-connections across resolutions (dashed blue lines) preserve fine details. A Latent Dynamical Core operates on the coarsest grid.
{b}, Latent Dynamical Core: This model reshapes the coarsest grid features and processes them through convolutional blocks and a series of Self-Attention (SA) blocks, capturing spatiotemporal dynamics in the latent space before reshaping the output back to a grid structure.
{c}, Fundamental Building Blocks: These include MLP blocks, ConvMLP blocks, the main Conv Block, the standard Self-Attention (SA) block, and the Basic Conv2d (BC) Block for resolution changes.}
\label{supply_main}
\end{figure}

\clearpage
\section{The TritonCast Model Architecture}
\label{appendix:model_details}

The core of TritonCast is a deep, hierarchical neural network structured as a symmetric {V-Cycle} framework, as depicted in \textbf{Fig.~\ref{supply_main}}. This design draws heavy inspiration from the principles of multi-grid methods in numerical computation, enabling the model to efficiently and separately process and learn the complex dynamics of the Earth system across multiple spatial scales. Our ablation studies (see the main text and Appendix~\ref{app:ablation}) experimentally validate that the three core components of this framework the multi-grid hierarchy, the Latent Dynamical Core, and the skip-connections are each indispensable pillars for achieving stable and physically consistent long-term forecasts.

\subsection{The Hierarchical V-Cycle Framework}
As illustrated in \textbf{Fig.~\ref{supply_main}a}, the overall data flow in TritonCast follows a distinct "V" shape. A high-resolution input state tensor first enters the left side of the model, the {Restriction Path} (or encoder). Along this path, the data flows through a series of downsampling modules (indicated by orange arrows), where its spatial resolution is progressively halved, as if zooming out from a macroscopic view deeper into the system's core. This hierarchical structure is the cornerstone for the model to correctly capture multi-scale physical processes; as our ablation study ({No MG}) reveals, removing this structure leads to an inability to reproduce correct energy spectra, causing erroneous energy transfers across scales and resulting in severe forecast bias.

When the data flow reaches the bottom of the V-Cycle, the coarsest resolution level, it is fed into the heart of the architecture: the {Latent Dynamical Core (LDC)}, the engine behind the model's remarkable long-term stability. The LDC (detailed in \textbf{Fig.~\ref{supply_main}b}) is responsible for the temporal iteration and evolution of the system's dominant, low-frequency dynamical modes within a highly compressed, low-dimensional latent space. Our ablation experiments ({No LDC}) dramatically illustrate its critical role: removing this stable core causes the entire model to suffer a catastrophic failure within just a few prediction steps, with forecasts rapidly diverging into physically meaningless states.

After the LDC performs its robust prediction of the future state, the updated features proceed to the right side of the model, the {Prolongation Path} (or decoder). This path symmetrically mirrors the restriction path, passing the data through a series of upsampling modules (indicated by cyan arrows) to progressively double the spatial resolution, ultimately reconstructing the detail-rich physical fields with artistic precision.

To ensure that high-frequency spatial details abstracted away during downsampling (such as small-scale eddies and turbulence) are accurately recovered during reconstruction, we establish {Skip-Connections} (represented by dashed blue lines) between corresponding levels of the encoder and decoder. These connections, acting as information highways, are the key to suppressing long-term forecast drift and ensuring the precision of physical details. Our ablation study ({No SC}) highlights their indispensability: a model without this mechanism, while stable, systematically drifts away from the true state, for instance, by misplacing critical weather systems like the polar vortex. By providing a direct path for fine-grained features from the encoder to be fused in the decoder, skip-connections act as a continuous calibrator, effectively correcting for errors that might otherwise accumulate during long-term integration.

In summary, the V-Cycle framework of TritonCast implements an elegant "divide and conquer" strategy through the synergistic interplay of its three core components. The multi-grid hierarchy manages multi-scale complexity, the Latent Dynamical Core ensures long-term stability, and the skip-connections guarantee long-term accuracy, together overcoming the central challenges of AI-based Earth system forecasting.

\subsection{Architecture Overview}
The {TritonCast} architecture is designed to effectively model the complex multi-scale dynamics inherent in the Earth system and to mitigate the spectral bias commonly found in deep learning models for long-term forecasting. At its core, it adopts an {Encoder-Latent Dynamical Core-Decoder} framework. Crucially, its design draws inspiration from classical {multi-grid methods} \cite{Brandt1977MultilevelAS}, enabling the hierarchical processing of information across different spatial scales (\textbf{Fig.~\ref{supply_main}a}).

Let $X_t \in \mathbb{R}^{B \times C \times H \times W}$ represent the input state at time $t$ (where $B$ is the batch size, $C$ is the number of input channels, and $H, W$ are the spatial dimensions). The architecture operates across $L+1$ resolution levels, indexed $l=0, \dots, L$, where $l=0$ corresponds to the original resolution $(H, W)$ and $l=L$ to the coarsest resolution $(H/2^L, W/2^L)$. In our implementation, $L=4$.

The {encoder path} progressively reduces spatial resolution in a two-step process at each level. First, features undergo {intra-level feature refinement (smoothing)} via convolutional blocks ($\mathcal{S}^{l}_{\text{enc}}$, implemented as \texttt{Conv Block}s), an operation described as:
\begin{equation}
    S^{(l)} = \mathcal{S}^{(l)}_{\text{enc}}(Z^{(l)}_{\text{enc}}; \theta^{(l)}_{\mathcal{S}_{\text{enc}}})
    \label{eq:encoder_smoothing}
\end{equation}
These refined features $S^{(l)}$ are carefully preserved to play a crucial role in the decoding stage via skip connections. Following this, a downsampling operation ($\mathcal{R}^{(l \to l+1)}$, implemented as a \texttt{BC Block Down}) performs {resolution reduction (restriction)}, mapping the features to the next coarser level:
\begin{equation}
    Z^{(l+1)}_{\text{enc}} = \mathcal{R}^{(l \to l+1)}(S^{(l)}; \theta^{(l)}_{\mathcal{R}})
    \label{eq:encoder_restriction}
\end{equation}
This process continues from $l=0$ to $L-1$, culminating in a final smoothing operation at the coarsest level $l=L$ to produce $Z_{\text{enc}}^{(L)\prime}$.

At the architecture's bottleneck, the coarsest resolution level, the features $Z_{\text{enc}}^{(L)\prime}$ are processed by the dedicated {Latent Dynamical Core (LDC)}, denoted by $\mathcal{L}$:
\begin{equation}
    Z_{\text{ldc}}^{(L)} = \mathcal{L}(Z_{\text{enc}}^{(L)\prime}; \theta_{\mathcal{L}})
    \label{eq:ldc_op}
\end{equation}
The LDC (detailed in Section~\ref{section:sub_ldc}) is specifically designed to capture and propagate the large-scale, slowly evolving dynamics crucial for long-term prediction. Its output, $Z^{(L)}_{\text{ldc}}$, serves as the initial state for the decoder at the coarsest level.

Symmetrically, the {decoder path} progressively reconstructs the high-resolution state. At each level, it begins with {resolution increase (prolongation)} via an upsampling operation ($\mathcal{P}^{(l \to l-1)}$, implemented as a \texttt{BC Block Up}):
\begin{equation}
    Z^{(l-1)}_{\text{up}} = \mathcal{P}^{(l \to l-1)}(Z^{(l)}_{\text{dec}}; \theta^{(l)}_{\mathcal{P}})
    \label{eq:decoder_prolongation}
\end{equation}
The critical step of detail restoration is {skip connection concatenation}, which fuses the upsampled features with the corresponding preserved features $S^{(l-1)}$ from the encoder path:
\begin{equation}
    Z^{(l-1)}_{\text{cat}} = \text{Concat}(Z^{(l-1)}_{\text{up}}, S^{(l-1)})
    \label{eq:decoder_concat}
\end{equation}
These fused features then undergo a final {intra-level feature refinement} by a convolutional block ($\mathcal{S}^{(l-1)}_{\text{dec}}$) to produce the level's output:
\begin{equation}
    Z^{(l-1)}_{\text{dec}} = \mathcal{S}^{(l-1)}_{\text{dec}}(Z^{(l-1)}_{\text{cat}}; \theta^{(l-1)}_{\mathcal{S}_{\text{dec}}})
    \label{eq:decoder_smoothing}
\end{equation}
Ultimately, the decoder's output at the original resolution, $Z^{(0)}_{\text{dec}}$, represents the predicted state for the next time step, $X_{t+1}$. Through this hierarchical approach, {TritonCast} can simultaneously represent and interact with dynamics at different spatial frequencies, fostering a more faithful representation of cross-scale energy transfers and suppressing the uncontrolled error growth associated with spectral bias in long-term autoregressive predictions.

\subsection{Latent Dynamical Core}\label{section:sub_ldc}
The Latent Dynamical Core (LDC) operates at the bottleneck ($l=L$) of the hierarchical architecture. It processes the coarsest-scale features $Z_{\text{enc}}^{(L)\prime} \in \mathbb{R}^{B \times D \times H' \times W'}$ from the final encoder smoothing step. The LDC's primary function is to evolve this latent state over a time step, capturing the essential dynamics of the large-scale, low-frequency modes. To do this, the LDC first reshapes the spatial grid features $Z_{\text{enc}}^{(L)\prime}$ into a sequence format, $X_{\text{seq}} = \text{Reshape}(Z_{\text{enc}}^{(L)\prime}) \in \mathbb{R}^{B \times \mathfrak{L} \times D}$. This sequence then undergoes iterative refinement through a series of $N$ transformation blocks, denoted as $\mathrm{LDM\_Block}$. Let $X^{(0)} = X_{\text{seq}}$, the evolution within the LDC is described as:
\begin{equation}
    X^{(i)} = \text{LDM\_Block}(X^{(i-1)}; \theta^{(i)}_{\mathcal{L}}) \quad \text{for } i = 1, \dots, N,
    \label{eq:ldm_block_iter_revised}
\end{equation}
Each $\text{LDM\_Block}$ strategically combines the strengths of self-attention and convolutional operations. While self-attention excels at capturing long-term dependencies and low-frequency global patterns, convolutional layers are effective at extracting local features and higher-frequency spatial patterns. A typical block incorporates a Multi-Head Self-Attention (MHSA) layer and a feed-forward network (MLP), both using residual connections and layer normalization, characteristic of Transformer encoders:
\begin{align}
    Y &= X^{(i-1)} + \text{PositionalEncoding} \\
    Y_{\text{attn}} &= Y + \text{Dropout}(\text{MHSA}(\text{LayerNorm}(Y); \theta^{(i)}_{\text{attn}})) \label{eq:mhsa_res_revised}\\
    X'_{\text{sa}} &= Y_{\text{attn}} + \text{Dropout}(\text{MLP}(\text{LayerNorm}(Y_{\text{attn}}); \theta^{(i)}_{\text{mlp}})) \label{eq:mlp_res_revised}
\end{align}
Following or interwoven with self-attention, convolutional components (denoted $\mathcal{F}_{\text{conv}}$) refine the features, enhancing local spatial structures or residual high-frequency information \textit{at this coarse scale}:
\begin{equation}
    X^{(i)} = \mathcal{F}_{\text{conv}}(X'_{\text{sa}}; \theta^{(i)}_{\text{conv}})
    \label{eq:ldm_conv_refine_final_revised}
\end{equation}
After passing through $N$ such blocks, the final refined sequence $X^{(N)}$ is reshaped back into the spatial grid format to produce the LDC's output:
\begin{equation}
    Z^{(K)}_{\text{ldm}} = \text{Reshape}^{-1}(X^{(N)}) \in \mathbb{R}^{B \times D \times H' \times W'}
    \label{eq:ldm_reshape_out_revised}
\end{equation}
Thus, the overall operation is $Z_{\text{ldm}}^{(K)} = \mathcal{L}\left( Z_{\text{enc}}^{(K)\prime}; \theta_{\mathcal{L}} \right).$ This synergistic design within the LDC is pivotal for suppressing spectral bias and enabling stable, physically consistent long-term autoregressive forecasting.

\subsection{Fundamental Building Blocks}
\label{sec:app:blocks}

The {TritonCast} architecture is constructed from several meticulously designed fundamental building blocks, as detailed in \textbf{Fig.~\ref{supply_main}c}, which are foundational to its multi-scale modeling and complex dynamics capturing capabilities.

\subsubsection{MLP and ConvMLP Blocks}
The Multi-Layer Perceptron (MLP) block serves as a standard unit for non-linear feature transformation, typically operating on vectorized inputs. It comprises a sequence of linear layers, GELU activation functions, and Dropout layers. For an input vector $x$, the transformation is as follows:
\begin{equation}
\begin{aligned}
h_1 & =\operatorname{GELU}\left(\operatorname{Linear}_1(x)\right) \\
d_1 & \left.=\operatorname{Dropout}( h_1\right) \\
h_2 & =\operatorname{Linear}_2\left(d_1\right) \\
y_{\mathrm{mlp}} & =\operatorname{Dropout}\left(h_2\right)
\end{aligned}
\end{equation}
Its spatial domain equivalent, the Convolutional MLP (ConvMLP) block, employs 1x1 convolutions to achieve a similar function while preserving spatial dimensions. For an input feature map $X$, the operation is:
\begin{equation}
\begin{aligned}
H_1 & =\operatorname{GELU}\left(\operatorname{Conv} 2 \mathrm{~d}_{1 \times 1}^{(1)}(X)\right) \\
D_1 & =\operatorname{Dropout}\left(H_1\right) \\
H_2 & =\operatorname{Conv} 2 \mathrm{~d}_{1 \times 1}^{(2)}\left(D_1\right) \\
Y_{\text {convmlp }} & =\operatorname{Dropout}\left(H_2\right)\label{eq:app_convmlp_output}
\end{aligned}
\end{equation}

\subsubsection{Conv Block}
The Conv Block is central to intra-level feature refinement, and its intricate internal structure showcases a clever parallel design. The process begins by adding a learnable {Position Embedding} to the input feature map $X$. The feature map is then fed into a dual-path structure: in the {Main Spatial Path}, spatial patterns are extracted through a series of normalization and convolution operations (notably a $5 \times 5$ depthwise-separable-like convolution); simultaneously, in the {Channel-wise Path}, a parallel {ConvMLP Block} focuses on enhancing channel-wise representations. The outputs of these two specialized paths are fused via element-wise addition, and finally, a {Residual Connection} adds the original input $X$ back to the fused features, ensuring lossless information flow. This operation can be abstractly denoted as:
\begin{equation}
Y_{\text{conv}} = \mathcal{S}(X + \text{PositionEmbedding}; \theta_{\mathcal{S}})
\label{eq:app_conv_block}
\end{equation}

\subsubsection{Self-Attention (SA) Block}
The Self-Attention (SA) Block implements a standard Transformer encoder layer and is the workhorse for capturing global information within the Latent Dynamical Core. For an input sequence $X_{\text{seq}}$, the block first incorporates positional information. Its core consists of two canonical sub-layers, both employing residual connections and layer normalization: first, a {Multi-Head Self-Attention} (MHSA) mechanism computes the dependencies among all pairs of elements in the sequence; this is followed by a position-wise feed-forward network, implemented by an {MLP block}, for further non-linear transformation. For an input sequence $Y$ ($X_{\text{seq}}$ plus positional encoding), the process is summarized as:
\begin{equation}
\begin{aligned}
A &= \text{MHSA}(\text{LayerNorm}(Y)) \\
Y' &= Y + \text{Dropout}(A) \\
F &= \text{MLP}(\text{LayerNorm}(Y')) \\
Z_{\text{sa}} &= Y' + \text{Dropout}(F) \label{eq:app_sa_output}
\end{aligned}
\end{equation}

\subsubsection{Basic Conv2d (BC) Block}
The Basic Conv2d (BC) Block is the agent of scale transformation within the hierarchical architecture. Its operation is conditioned on a \texttt{Transpose?} flag. For {downsampling} (\texttt{Transpose? = No}), it applies a standard $3 \times 3$ convolution with a stride of 2:
\begin{equation}
Y_{\text{down}} = \text{LeakyReLU}(\text{GroupNorm}(\text{Conv2d}_{3\times3}(\text{Stride}=2, \text{Padding}=1)(X))) \label{eq:app_bc_down}
\end{equation}
Conversely, for {upsampling} (\texttt{Transpose? = Yes}), it utilizes a $3 \times 3$ transpose convolution with a stride of 2 to double the spatial dimensions:
\begin{equation}
Y_{\text{up}} = \text{LeakyReLU}(\text{GroupNorm}(\text{TransposeConv2d}_{3\times3}(\text{Stride}=2, \text{Padding}=1, ...)(X))) \label{eq:app_bc_up}
\end{equation}
In both cases, the convolution is followed by Group Normalization and a LeakyReLU activation function, ensuring stability and non-linearity during the scale transformation process.

\subsection{TritonCast Model Family Configurations}
To address a diverse range of Earth system forecasting tasks, from global atmospheric circulation to regional ocean eddies, TritonCast is designed not as a single, monolithic model but as a highly scalable architecture. By adjusting the depth, width, and overall scale of its core components, we construct a family of models, each optimized for a specific application scenario. This philosophy enables an optimal balance between computational resources, forecast accuracy, and task complexity.
Tab.~\ref{tab:model_family_summary_en} systematically summarizes the full suite of TritonCast models developed in this work. The table categorizes each model by its application domain and parameter scale, providing a clear overview of the TritonCast family's composition and capability spectrum.

\begin{table}[h!]
\centering
\footnotesize
\caption{\textbf{Summary of the complete TritonCast trained model family.}}
\label{tab:model_family_summary_en}
\begin{tabularx}{\textwidth}{ l | l | c | >{\raggedright\arraybackslash}X | >{\raggedright\arraybackslash}X }
\toprule
\textbf{Application Domain} & \textbf{Specific Task} & \textbf{Params} & \textbf{Core Training Data} & \textbf{Notes / Key Features} \\
\midrule
\multirow{3}{*}{\textbf{Atmospheric Science}} & Medium-range Weather & 1B & ERA5 @ 1.5°, 6-hr steps & \textbf{Heavyweight accuracy model}, for SOTA benchmark \\
\cmidrule{2-5}
& Long-term Stability Test & 0.02B & ERA5 @ 1.0°, 24-hr steps & \textbf{Lightweight stability model}, completes 365-day forecast \\
\cmidrule{2-5}
& Multi-year Climate Sim. & 0.1B & ERA5 + GLORYS12 @ 1.5° & \textbf{Climate simulation model}, for long-term physical response \\
\midrule
\multirow{3}{*}{\textbf{Oceanography}} & Global Ocean Simulation & 0.02B & GLORYS12 + ERA5 @ 0.25° & \textbf{Global ocean model}, validates coupled forecast robustness \\
\cmidrule{2-5}
& High-Fidelity Eddy Forecast & 0.028B & CMEMS @ 0.125° (3 regions) & \textbf{High-res regional model}, for eddy-resolving forecasts \\
\cmidrule{2-5}
& Zero-shot Generalization & 0.002B & Global CMEMS @ 0.25° coarse-res & \textbf{Ultra-lightweight model}, for cross-resolution capability \\
\midrule
\textbf{Theoretical Physics} & 2D Isotropic Turbulence & 0.02B & 128x128 DNS Data & \textbf{Physics-benchmark model}, for turbulence dynamics test \\
\bottomrule
\end{tabularx}
\end{table}

From the heavyweight 1-billion-parameter model, aiming for state-of-the-art accuracy in medium-range weather forecasting, to the ultra-lightweight 2-million-parameter model, designed to explore the theoretical limits of generalization, the TritonCast family showcases the versatility and power of its architectural design. This suite of meticulously designed models allows us to conduct a comprehensive and in-depth evaluation of its performance across several key scientific problems. For the detailed training configurations of each model, please refer to Supplementary~\ref{appendix:EXPSTEUP}.

\clearpage

\section{Datasets and Evaluation Methods}

\subsection{Overview of Datasets}
\label{appendix:datasets}
\subsubsection{Atmospheric Science Dataset (ERA5)}
\label{appendix:era5}
\textbf{\ding{182} ERA5-6h for medium-range weather forecasting dataset.} For medium-range weather forecasting, similar to previous works, such as GraphCast \cite{lam2023learning}, we use 6h time interval ERA5 to train \method{} for fair comparison. A total of 69 atmospheric variables are utilized in our analysis, as shown in \textbf{Tab.~\ref{tb_atmo_6h}}. These comprise five upper-air variable fields at 13 standard pressure levels (specifically 50, 100, 150, 200, 250, 300, 400, 500, 600, 700, 850, 925, and 1,000 hPa): geopotential (Z), temperature (T), zonal wind component (U), meridional wind component (V), and specific humidity (Q); along with four surface variables: 10 metre u wind component (U10M), 10 metre v wind component (V10M), 2 metre temperature (T2M), and mean sea level pressure (MSLP). All data for this task are processed to a 1.5° spatial resolution and a 6-hour temporal resolution, corresponding to 00:00 UTC, 06:00 UTC, 12:00 UTC, and 18:00 UTC. We use the data with a size 120 × 240 (excluding Antarctica). The model is trained using data spanning 59 years from 1959 to 2017, while data from 2018 to 2019 are used for validation, and data from 2020 are used for testing.

\begin{table}[h]
\centering
\small
\caption{\textbf{Atmospheric variables used by \method{} for medium-range weather forecasting.} This table details the 69 atmospheric variables sourced from the ERA5-6h reanalysis dataset~\cite{hersbach2020era5} and processed for training and evaluating the \method{} model. It lists five upper-air variables specified at 13 standard pressure levels (resulting in 65 upper-air fields) and four single-level surface variables. For each variable type, the full name, abbreviation, number of vertical layers, and the uniform temporal (6h) and spatial (1.5°) resolutions used in this study are provided.}
\label{tb_atmo_6h}
\setlength{\tabcolsep}{3.5pt}
\begin{tabular}{cccccc}
\toprule
\textbf{Type}            & \textbf{Full name}                          & \textbf{Abbreviation}& \textbf{Layers}     & \textbf{Time Resolution} & \textbf{Spatial Resolution} \\ \midrule
\multirow{5}{*}{Upper-air variables} & Geopotential                        & Z    & 13      & 6h             & 1.5°                 \\ 
                                & Specific humidity                        & Q    & 13      & 6h             & 1.5°                 \\ 
                                & Temperature                              & T    & 13      & 6h             & 1.5°                 \\ 
                                & Zonal wind component              & U    & 13      & 6h             & 1.5°                 \\ 
                                & Meridional wind component              & V    & 13      & 6h             & 1.5°                 \\ 
\midrule
\multirow{4}{*}{Surface variables} & Mean sea level pressure               & MSLP & 1      & 6h             & 1.5°                  \\ 
                                & 2 metre temperature                           & T2M  & 1      & 6h             & 1.5°                  \\ 
                                & 10 metre u wind component          & U10M  & 1      & 6h             & 1.5°                  \\ 
                                & 10 metre v wind component          & V10M  & 1      & 6h             & 1.5°                  \\ \bottomrule
\end{tabular}
\end{table}

\clearpage
\textbf{\ding{183} Long-term weather forecasting dataset.} Atmospheric variable data for training and evaluating the \method{} global weather forecasting model are sourced from the ERA5-24h reanalysis dataset~\cite{hersbach2020era5}. A central challenge in long-term atmospheric forecasting is the effective management of error accumulation over extended autoregressive rollouts. Inspired by how state-of-the-art models like NeuralGCM recommend lower temporal resolutions to ensure stability in long-term climate simulations, we find that adopting a {daily (24-hour) temporal resolution} is a deliberate and optimal choice for our long-term stability tests. This coarser time step helps to suppress the rapid accumulation of errors introduced by high-frequency transient processes, which is crucial for maintaining physical consistency and numerical stability throughout year-long forecast integrations.

A total of 69 atmospheric variables are utilized in our analysis, as detailed in \textbf{Tab.~\ref{tb_atmo_daily}}. These comprise five upper-air variable fields at 13 standard pressure levels (specifically 50, 100, 150, 200, 250, 300, 400, 500, 600, 700, 850, 925, and 1,000 hPa): geopotential (Z), specific humidity (Q), temperature (T), zonal wind component (U), and meridional wind component (V); along with four surface variables: 10-meter u-wind component (U10M), 10-meter v-wind component (V10M), 2-meter temperature (T2M), and mean sea level pressure (MSLP). All data for this study are processed to a 1.0° spatial resolution and a 24-hour temporal resolution, corresponding to 12:00 UTC. And to better adapt to the input of different architecture models, we use the data with a size 180 × 360 (excluding Antarctica). For data partitioning, the model is trained using data spanning 25 years from 1993 to 2017. Subsequently, data from 2018 are used for validation, and data from 2019, 2020, and 2021 are used for testing.

\begin{table}[h]
\centering
\small
\caption{\textbf{Atmospheric variables from ERA5 used for \method{} long-term weather forecasting.} This table details the 69 atmospheric variables sourced from the ERA5 reanalysis dataset~\cite{hersbach2020era5} and processed for training and evaluating the \method{} model. It lists five upper-air variables specified at 13 standard pressure levels (resulting in 65 upper-air fields) and four single-level surface variables. For each variable type, the full name, abbreviation, number of vertical layers, and the uniform temporal (24h) and spatial (1.0°) resolutions used in this study are provided.}
\label{tb_atmo_daily}
\setlength{\tabcolsep}{3.5pt}
\begin{tabular}{cccccc}
\toprule
\textbf{Type}            & \textbf{Full name}                          & \textbf{Abbreviation}& \textbf{Layers}     & \textbf{Time Resolution} & \textbf{Spatial Resolution} \\ \midrule
\multirow{5}{*}{Upper-air variables} & Geopotential                        & Z    & 13      & 24h             & 1.0°                 \\ 
                                & Specific humidity                        & Q    & 13      & 24h             & 1.0°                 \\ 
                                & Temperature                              & T    & 13      & 24h             & 1.0°                 \\ 
                                & Zonal wind component              & U    & 13      & 24h             & 1.0°                 \\ 
                                & Meridional wind component              & V    & 13      & 24h             & 1.0°                 \\ 
\midrule
\multirow{4}{*}{Surface variables} & Mean sea level pressure               & MSLP & 1      & 24h             & 1.0°                  \\ 
                                & 2 metre temperature                           & T2M  & 1      & 24h             & 1.0°                  \\ 
                                & 10 metre u wind component          & U10M  & 1      & 24h             & 1.0°                  \\ 
                                & 10 metre v wind component          & V10M  & 1      & 24h             & 1.0°                  \\ \bottomrule
\end{tabular}
\end{table}

\clearpage
\textbf{\ding{184} Multi-year climate simulation dataset.} Atmospheric variable data for training and evaluating the \method{} multi-year climate simulation model are sourced from the ERA5-24h reanalysis dataset~\cite{hersbach2020era5} and GLORYS12 reanalysis dataset. To investigate the potential of \method{} in climate simulation, following the approach of NeuralGCM \cite{kochkov2024neural}, we use sea surface temperature as a driver to simulate the variability of temperature. All data for this task are processed to a 1.5° spatial resolution and a 24-hour temporal resolution. And to better adapt to the input of different architecture models, we use the data with a size 120 × 240 (excluding Antarctica). For model development, data from 1993–2017 are used for training and data from 2018-2024 are used for testing. The data details can be found in \textbf{Tab.~\ref{tb_climate_simulation}}. 

\begin{table}[h!]
\centering
\small
\caption{\textbf{Atmospheric variables and corresponding forcing (Sea surface temperature) used by \method{} for interannual simulation.} This table details the 70 variables sourced from the ERA5-24h reanalysis dataset~\cite{hersbach2020era5} and GLORYS12 reanalysis dataset used in multi-year climate simulation task. It lists five upper-air variables specified at 13 standard pressure levels (resulting in 65 upper-air fields) and four single-level surface variables. And 1 forcing variable is also included. For each variable type, the full name, abbreviation, number of vertical layers, and the uniform temporal (24h) and spatial (1.5°) resolutions used in this task are provided.}
\label{tb_climate_simulation}
\setlength{\tabcolsep}{3.5pt}
\begin{tabular}{cccccc}
\toprule
\textbf{Type}            & \textbf{Full name}                          & \textbf{Abbreviation}& \textbf{Layers}     & \textbf{Time Resolution} & \textbf{Spatial Resolution} \\ \midrule
\multirow{5}{*}{Upper-air variables} & Geopotential                        & Z    & 13      & 24h             & 1.5°                 \\ 
                                & Specific humidity                        & Q    & 13      & 24h             & 1.5°                 \\ 
                                & Temperature                              & T    & 13      & 24h             & 1.5°                 \\ 
                                & Zonal wind component              & U    & 13      & 24h             & 1.5°                 \\ 
                                & Meridional wind component              & V    & 13      & 24h             & 1.5°                 \\ 
\midrule
\multirow{4}{*}{Surface variables} & Mean sea level pressure               & MSLP & 1      & 24h             & 1.5°                  \\ 
                                & 2 metre temperature                           & T2M  & 1      & 24h             & 1.5°                  \\ 
                                & 10 metre u wind component          & U10M  & 1      & 24h             & 1.5°                  \\ 
                                & 10 metre v wind component          & V10M  & 1      & 24h             & 1.5° \\
\midrule
\multirow{1}{*}{Oceanic variables} & Sea surface temperature               & SST & 1      & 24h             & 1.5°
                            \\ \bottomrule
\end{tabular}
\end{table}

\clearpage
\subsubsection{Ocean Science Datasets (GLORYS12)}
\label{appendix:ocean_glorys}
For the global ocean simulation task, oceanic data are sourced from the GLORYS12 reanalysis dataset. This dataset provides daily mean data, originally at a 1/12° spatial resolution. We resample these data to a 0.25° resolution (721 × 1440 grid points) to match the atmospheric data resolution and facilitate coupled simulations. And to better adapt to the input of different architecture models, we use the data with a size 720 × 1440 (excluding Antarctica). And atmospheric variable data used as forcings are sourced from the ERA5-24h reanalysis dataset~\cite{hersbach2020era5}. As shown in \textbf{Tab.~\ref{tb_supp_data_ocean}}, the model simulates five key ocean variables: sea salinity ($\mathrm{S}$), sea stream zonal velocity ($\mathrm{U}_{\mathrm{o}}$), sea stream meridional velocity ($\mathrm{V}_{\mathrm{o}}$), sea temperature ($\mathrm{T}_{\mathrm{o}}$) across 23 specified vertical levels (depths: 0.49m, 2.65m, 5.08m, 7.93m, 11.41m, 15.81m, 21.60m, 29.44m, 40.34m, 55.76m, 77.85m, 92.32m, 109.73m, 130.67m, 155.85m, 186.13m, 222.48m, 266.04m, 318.13m, 380.21m, 453.94m, 541.09m and 643.57m), and sea surface height (SSH). To account for ocean-atmosphere interactions, four surface variables from the ERA5 reanalysis dataset serve as atmospheric forcing fields: 10 metre zonal wind component (U10M), 10 metre meridional wind component (V10M), 2 metre temperature (T2M), and mean sea level pressure (MSLP). A static land-sea mask (LSM) is also included. All utilized data possess a final spatial resolution of 0.25° and a temporal resolution of 24 hours. For model development, data from 1993–2017 are used for training, data from 2018–2019 for validation, and data from 2020 for testing.

\begin{table*}[h!]
\centering
\small
\caption{\textbf{Datasets for the \method{} global ocean simulation task.} This table details the variables used for training, validating, and testing the \method{} ocean model. It includes four atmospheric forcing variables sourced from ERA5~\cite{hersbach2020era5} (U10M, V10M, T2M, MSLP) and five oceanic state variables derived from the GLORYS12 reanalysis dataset (S, $\mathrm{U}_{\mathrm{o}}$, $\mathrm{V}_{\mathrm{o}}$, $\mathrm{T}_{\mathrm{o}}$, SSH), along with a static land-sea mask (LSM). For each variable, the table specifies its type, full name, abbreviation, number of vertical layers (23 levels for S, $\mathrm{U}_{\mathrm{o}}$, $\mathrm{V}_{\mathrm{o}}$, $\mathrm{T}_{\mathrm{o}}$), the total time span covered (1993-2020), and the uniform temporal (24h) and spatial (0.25°) resolution used in this task after resampling oceanic data.}
\label{tb_supp_data_ocean}
\setlength{\tabcolsep}{1pt}
\begin{tabular}{ccccccc}
\toprule
\textbf{Type}        & \textbf{Full name}                 & \textbf{Abbreviation} & \textbf{Layers} & \textbf{Time}      & \textbf{Time Resolution} & \textbf{Spatial Resolution}\\ \midrule
Atmospheric & 10 metre u wind component     & U10M         & 1      & 1993-2020 & 24h             & 0.25°               \\
Atmospheric & 10 metre v wind component     & V10M         & 1      & 1993-2020 & 24h             & 0.25°               \\
Atmospheric & 2 metre temperature           & T2M          & 1      & 1993-2020 & 24h             & 0.25°               \\
Atmospheric & Mean sea level pressure       & MSLP         & 1      & 1993-2020 & 24h             & 0.25°               \\ \midrule
Oceanic     & Sea salinity                  & S            & 23     & 1993-2020 & 24h             & 0.25°               \\
Oceanic     & Sea stream zonal velocity     & $\mathrm{U}_{\mathrm{o}}$         & 23     & 1993-2020 & 24h             & 0.25°               \\
Oceanic     & Sea stream meridional velocity & $\mathrm{V}_{\mathrm{o}}$         & 23     & 1993-2020 & 24h             & 0.25°               \\
Oceanic     & Sea temperature               & $\mathrm{T}_{\mathrm{o}}$         & 23     & 1993-2020 & 24h             & 0.25°               \\
Oceanic     & Sea surface height            & SSH          & 1      & 1993-2020 & 24h             & 0.25°               \\ \midrule
Static      & Land-sea mask                 & LSM          & ---    & ---       & ---             & 0.25°               \\ \bottomrule
\end{tabular}
\end{table*}

\clearpage
\subsubsection{CMEMS Surface Geostrophic Velocity Datasets}
\label{appendix:ocean_uvgos}
For the tasks of high-fidelity ocean eddy forecasting and cross-resolution generalization, we utilize the multi-satellite altimetry product from the {Copernicus Marine Service}. This dataset provides daily global sea surface geostrophic velocities since 1993 and serves as a benchmark in ocean dynamics research. We primarily use two key variables: the zonal (\texttt{UGOS}) and meridional (\texttt{VGOS}) sea surface geostrophic velocities. Notably, given that the numerical range of this geostrophic velocity data is small (typically between -$2$ and $2$ m/s), we do not apply any normalization or standardization; our only preprocessing step consists of handling potential missing values (NaNs).

\begin{figure}[h!]
\centering
\includegraphics[width=0.9\linewidth]{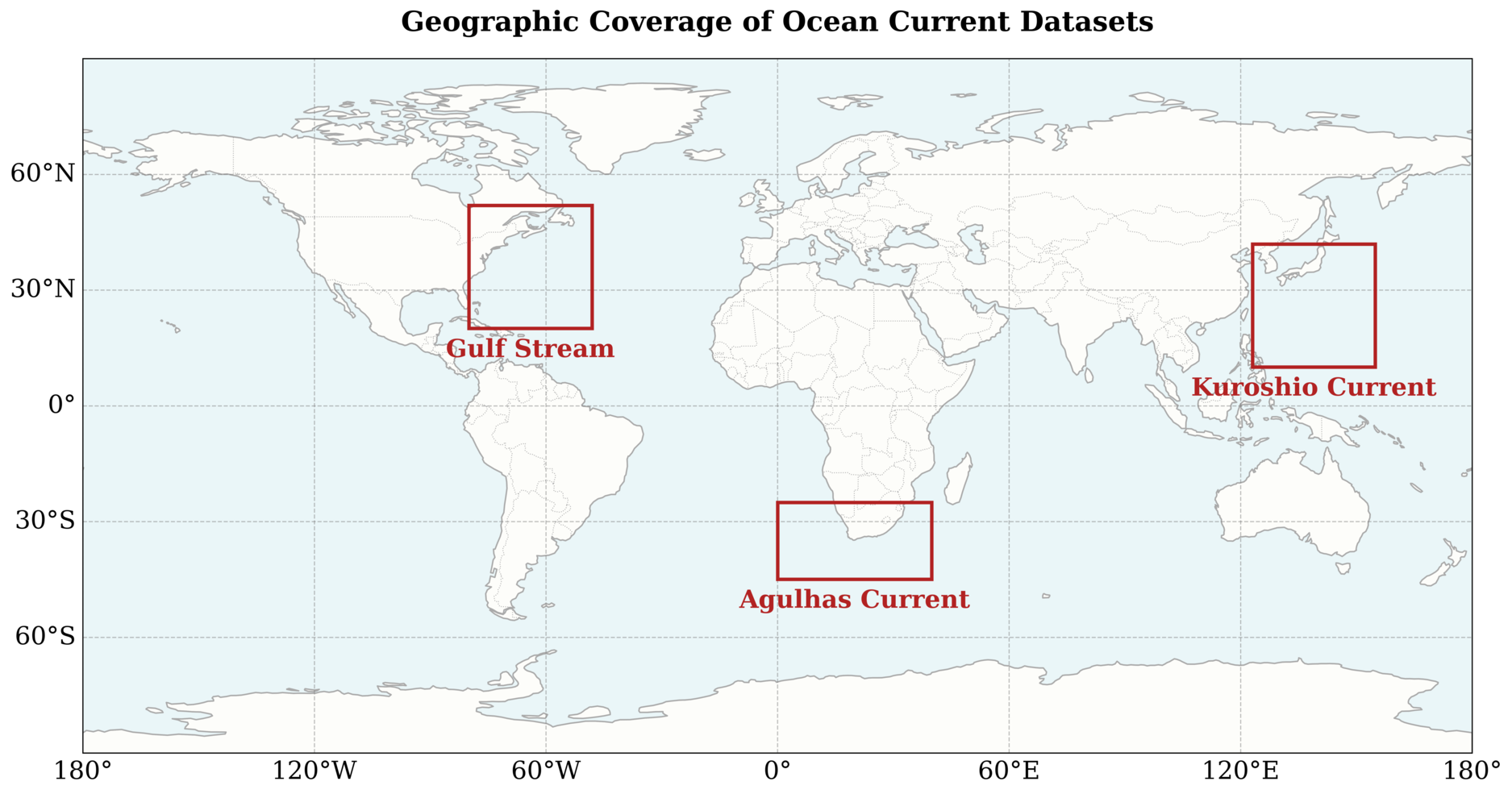}
\caption{\textbf{Geographic coverage of the four ocean current datasets used for model training and evaluation.} The global dataset (not boxed) spans the entire world map, while the three high-resolution regional datasets (red boxes) focus on the Gulf Stream, Kuroshio Current, and Agulhas Current systems, three major western boundary currents.}
\label{fig:ocean_current_regions}
\end{figure}

Our experimental strategy is twofold to comprehensively evaluate the model's capabilities. First, to train a foundational model that learns global large-scale circulation patterns, we downsample the native $0.125\degree$ resolution global data from 1993 to 2024 to a coarser $0.25\degree$ resolution using bilinear interpolation. Second, to rigorously test the model's ability to capture and forecast fine-scale dynamical processes such as mesoscale eddies, we retain the native $0.125\degree$ high resolution for three key regions known for their intense eddy activity. These regions are the Gulf Stream, the Kuroshio Current Extension, and the Agulhas Current, as illustrated in \textbf{Fig.~\ref{fig:ocean_current_regions}}.

This dual-track approach allows us to perform both long-term, stable global forecasts and high-precision, eddy-resolving regional forecasts within a unified framework. The key characteristics of the four derived experimental datasets are summarized in Tab.~\ref{tab:ocean_datasets_en}.

\begin{table}[h!]
\centering
\small
\caption{\textbf{Summary of ocean current dataset characteristics.}}
\label{tab:ocean_datasets_en}
\begin{tabular}{lcccc}
\toprule
\textbf{Dataset Name} & \textbf{Geographic Area (Approx.)} & \textbf{Spatial Resolution} & \textbf{Temporal Coverage} & \textbf{Key Variables} \\
\midrule
Global & Global Coverage & 0.25\degree & 1993-2024 & UGOS, VGOS \\
Gulf Stream & 25\degree N-50\degree N, 80\degree W-50\degree W & 0.125\degree & 1993-2024 & UGOS, VGOS \\
Kuroshio Current & 20\degree N-45\degree N, 120\degree E-150\degree E & 0.125\degree & 1993-2024 & UGOS, VGOS \\
Agulhas Current & 45\degree S-25\degree S, 10\degree E-50\degree E & 0.125\degree & 1993-2024 & UGOS, VGOS \\
\bottomrule
\end{tabular}
\end{table}

\subsubsection{2D Kolmogorov Turbulence DNS}
\label{appendix:dns_dataset}
Accurately predicting the dynamics of turbulent flows governed by the Navier–Stokes equations (NSE) represents a fundamental challenge in science and engineering. Data-driven neural operator learning offers a promising avenue, but its rigorous validation hinges on high-fidelity benchmark datasets capable of capturing complex physical phenomena. The dataset utilized in this study is specifically designed for this purpose, derived from high-fidelity numerical simulations of the two-dimensional incompressible NSE on a periodic domain, typically the torus $\mathbb{T}^2 = [0, 2\pi]^2$ with periodic boundary conditions.

The core dynamics describe the evolution of the fluid velocity field $\mathbf{u}(\mathbf{x}, t)$ and pressure $p(\mathbf{x}, t)$. In the velocity-pressure formulation, the governing equations are:
\begin{equation}
\partial_t \mathbf{u} + (\mathbf{u} \cdot \nabla) \mathbf{u} + \nabla p = \nu \Delta \mathbf{u} + \mathbf{f}, \quad \nabla \cdot \mathbf{u} = 0
\end{equation}
where $\nu$ is the kinematic viscosity and $\mathbf{f}$ represents any external forcing. The Reynolds number (Re), a key parameter characterizing the flow regime (with higher Re indicating stronger turbulence), is inversely proportional to the viscosity $\nu$.

Alternatively, for 2D flows, the dynamics can be conveniently expressed in the vorticity-streamfunction $(\omega, \psi)$ formulation. The vorticity is defined as $\omega = \nabla \times \mathbf{u} = \partial_x u_y - \partial_y u_x$, and the streamfunction $\psi$ relates to the velocity via $\mathbf{u} = (\partial_y \psi, -\partial_x \psi)$. The governing equations become:
\begin{equation}
\partial_t \omega + (\mathbf{u} \cdot \nabla) \omega = \nu \Delta \omega + \nabla \times \mathbf{f} - \Delta \psi = \omega
\end{equation}
where $\mathbf{u} \cdot \nabla \omega$ represents the advection of vorticity by the velocity field.

The dataset comprises detailed spatio-temporal trajectories, $\omega(t, \mathbf{x})$ or $\mathbf{u}(t, \mathbf{x})$, generated using advanced numerical methods. Typically, pseudo-spectral methods are employed for spatial discretization due to their high accuracy on periodic domains, coupled with high-order time integration schemes (e.g., Runge-Kutta methods like RK3 or IMEX RK4) to ensure temporal accuracy and stability. These methods accurately capture essential physical properties, such as the conservation laws and the characteristic energy cascade in turbulent flows. The simulations encompass canonical turbulence scenarios relevant for benchmarking. This includes decaying homogeneous isotropic turbulence, often initialized using a random field with a specific energy spectrum like the McWilliams initial conditions~\cite{mcwilliams1984emergence}, which evolves intricate vortical structures. Forced turbulence scenarios, where energy is continuously injected by the forcing term $\mathbf{f}$, are also common. The datasets typically cover a range of turbulent intensities, corresponding to Reynolds numbers such as Re=1000 ($\nu=10^{-3}$) and Re=5000. A critical aspect of comprehensive NSE benchmark suites is often the availability of simulations across multiple spatial resolutions (e.g., $64^2$, $128^2$, $256^2$, up to $1024^2$). This multi-resolution structure provides a stringent testbed for evaluating the generalization capabilities and resolution-invariance of modern neural operators. For the specific experiments on 2D decaying turbulence presented in this paper (\textbf{Fig.~\ref{turbulence_comparison}}), we utilize a dataset version generated at a spatial resolution of $128 \times 128$.

\clearpage
\subsection{Data Preprocessing and Normalization Strategy}

\subsubsection{Atmospheric Science Dataset (ERA5) }
\textbf{\ding{182} ERA5-6h for medium-range weather forecasting.} This task involves 69 variables from ERA5. Different variables exhibit substantial differences in magnitude. To allow the model focusing on predictions rather than learning the differences between variables, we normalize the data before putting the data into the model. We calculated the mean and standard deviation of all variables using data from 1959 to 2017 (training set). Each variable has a corresponding mean and standard deviation. Before feeding the data into the model, we first subtract the respective mean and divided it by the standard deviation.

\textbf{\ding{183} Long-term weather forecasting.} Similar to medium-range weather forecasting, this task involves 69 variables from ERA5. We normalize the data before putting the data into the model. We calculate the mean and standard deviation of all variables using data from 1993 to 2017 (training set). Each variable has a corresponding mean and standard deviation. Before feeding the data into the model, we first subtract the respective mean and divided it by the standard deviation.

\textbf{\ding{184} Multi-year climate simulation.} This task involves 69 variables from ERA5 and 1 variable (SST) from GLORYS12. We normalize the data before putting the data into the model. For 69 atmospheric variables, the mean and standard deviation from medium-range weather forecasting are directly used in this task. For forcing variable SST, the mean and standard deviation are calculated using the 1.5 degree GLORYS12 data from 1993 to 2017 (the first 365 days of each year in the training dataset). Before feeding the data into the model, we first subtract the respective mean and divided it by the standard deviation. For the ‘NaNs’ values of land, we fill them with zero before inputting the data into the model.

\subsubsection{Ocean Science Datasets (GLORYS12)} This task involves 4 atmospheric variables from ERA5 and 93 oceanic variables from GLORYS12. For oceanic variables, we first remove the climatological mean from variables exhibiting strong periodicity (Sea salinity, Sea temperature, and Sea surface height), thereby directing the model’s attention toward small anomalies. We calculate the climatological mean of periodic variables using data from 1993 to 2017 (the first 365 day of each year). Therefore, Sea salinity, Sea temperature, and Sea surface height are preprocessing to Sea salinity anomaly, Sea temperature anomaly, and Sea surface height anomaly. We then calculate the mean and standard deviation of all variables (for periodic variables, they corrspond to the anomaly) using data from 1993 to 2017 (the first 365 days of each year in the training dataset). Before feeding the data into the model, we first subtract the respective mean and divided it by the standard deviation. For the ‘NaNs’ values of land, we fill them with zero before inputting the data into the model.

\subsubsection{CMEMS Surface Geostrophic Velocity Datasets}

The CMEMS dataset comprises two key variables: the zonal (\texttt{UGOS}) and meridional (\texttt{VGOS}) sea surface geostrophic velocities. As noted in the dataset description, the numerical range of this data is inherently small, typically falling within -2 to 2 m/s. Consequently, standard data normalization or standardization procedures are deemed unnecessary, as they would not significantly impact model training while preserving the original physical meaning of the velocity values. The sole preprocessing step applied to this dataset is the handling of potential missing values (NaNs) to ensure the integrity of the data fed into the model. Apart from this, the data is utilized in its raw numerical form.

\subsubsection{2D Kolmogorov Turbulence DNS Dataset}

The dataset for the 2D Kolmogorov turbulence simulation consists of a single variable: the vorticity field, generated via Direct Numerical Simulation (DNS). The data is derived from a high-fidelity numerical simulation under controlled conditions, resulting in a well-behaved and consistent numerical range for the vorticity values. Therefore, no normalization, standardization, or any other form of data preprocessing is applied to this dataset. The raw vorticity data from the simulation is used directly as input for the model, providing an unadulterated environment to test the model's ability to learn the underlying physical dynamics without data augmentation or transformation.

\subsection{Spatiotemporal Data Handling and Prediction Modalities}
This section details the technical aspects of training the \method{} model. A key characteristic of the \method{} architecture, which enables its flexibility in both training and inference, is its native handling of spatiotemporal data using tensors with the shape \texttt{[T, C, H, W]}. Here, \texttt{T} represents the temporal dimension (number of time steps), \texttt{C} denotes the number of channels or variables, and \texttt{H} and \texttt{W} are the spatial height and width, respectively. This stands in contrast to several conventional data-driven forecast models (such as Pangu-Weather) that operate primarily on a single-step prediction paradigm, processing inputs of shape \texttt{[C, H, W]} to predict the state at the next immediate step.

\begin{figure}[h!]
\centering
\includegraphics[width=0.95\linewidth]{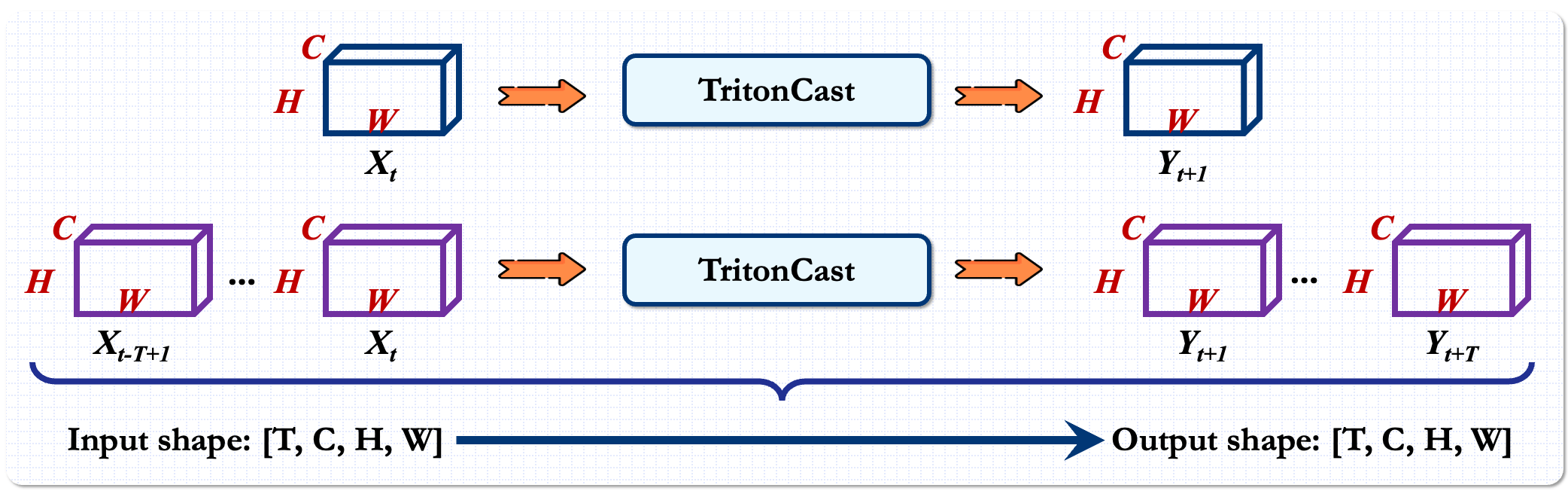}
\caption{\textbf{Flexible input-output processing in \method{}.} Schematic illustrating the model's capability to handle different temporal input structures. Top: Single-step prediction mode, where the state at time $t$, $X_t$, predicts the state at $t+1$, $Y_{t+1}$. Bottom: Multi-step prediction mode, where a block of $T$ consecutive input states $\{X_{t-T+1}, \dots, X_t\}$, represented as a tensor of shape $[T, C, H, W]$, is used to predict a block of $T$ future states $\{Y_{t+1}, \dots, Y_{t+T}\}$, also represented as a tensor of shape $[T, C, H, W]$.}
\label{Training_method}
\end{figure}

As illustrated schematically in \textbf{Fig.~\ref{Training_method}}, this \texttt{[T, C, H, W]} structure endows \method{} with two distinct prediction modalities. First, it can operate in a conventional {single-step prediction mode}, mapping the state at time $t$, $X_t$, to the predicted state at $t+1$, $Y_{t+1}$. Second, and more powerfully, it can operate in a {multi-step, sequence-to-sequence mode}, mapping a temporal block of $T$ input states $\{X_{t-T+1}, \dots, X_t\}$ to a corresponding block of $T$ predicted future states $\{Y_{t+1}, \dots, Y_{t+T}\}$ in a single, parallel forward pass.

This parallel, sequence-to-sequence prediction capability is particularly advantageous in scenarios involving slowly evolving dynamics, such as the forecasting of large-scale ocean currents. By processing and predicting temporal blocks as a whole, the model can potentially capture longer-range dependencies and model a smoother, more physically consistent temporal evolution. This approach may lead to improved performance in certain long-term forecasting applications compared to purely sequential, single-step autoregression. The specific values of $T$ for the input and output sequences used in our experiments are detailed in the respective parts.

\subsubsection{Loss Function Definition}
Across all our training experiments, we employ a single, consistent loss function: the {Mean Squared Error (MSE)} or its variant: the {Relative Mean Squared Error}. As a standard and powerful workhorse for physical system modeling and a wide range of regression tasks, MSE directly quantifies the discrepancy between the model's predictions and the ground truth. By minimizing the MSE, we are effectively driving the model's parameters to learn a mapping that minimizes the L2 norm of the error between the predicted and target fields.

A key aspect of our methodology is the universal application of this same, unmodified MSE loss function across all our diverse experimental domains from global atmospheric forecasting and ocean circulation modeling to the theoretical evolution of 2D turbulence. This domain-agnostic approach provides strong evidence for the robustness and generality of the \method{} architecture itself. It demonstrates that the model's powerful learning capabilities stem primarily from its intrinsic architectural design, rather than from a reliance on complex, task-specific loss functions tailored to particular physics.

For a given training sample with input $\mathbf{X}_t$ and ground truth target $\mathbf{Y}_t$ (e.g., $\mathbf{Y}_t = \mathbf{X}_{t+1}$ in single-step prediction), the model produces a prediction $\hat{\mathbf{Y}}_t$. The loss, $\mathcal{L}_{\text{MSE}}$, is defined as the average of the squared errors over all channels ($C$), height ($H$), and width ($W$) dimensions:
\begin{equation}
    \mathcal{L}_{\text{MSE}} = \frac{1}{C \times H \times W} \sum_{c=1}^{C} \sum_{h=1}^{H} \sum_{w=1}^{W} \left( (\hat{\mathbf{Y}}_t)_{c,h,w} - (\mathbf{Y}_t)_{c,h,w} \right)^2
\end{equation}
And the loss, $\mathcal{L}_{\text{MSE}_{relative}}$, is defined as the average of the squared errors over all channels ($C$), height ($H$), and width ($W$) dimensions:
\begin{equation}
    \mathcal{L}_{\text{MSE}_{relative}} = \frac{1}{C \times H \times W} \sum_{c=1}^{C} \sum_{h=1}^{H} \sum_{w=1}^{W} \left( \frac{(\hat{\mathbf{Y}}_t)_{c,h,w} - (\mathbf{Y}_t)_{c,h,w}}{(\mathbf{Y}_t)_{c,h,w}} \right)^2
\end{equation}

It is important to emphasize that we use an {unweighted MSE} during training. This means that every physical variable and every spatial grid point contributes equally to the total loss. This contrasts with our evaluation metrics like the latitude-weighted RMSE and ACC, which account for the Earth's geometry, and further highlights the simplicity and effectiveness of our training strategy. This straightforward objective function is sufficient to guide the optimizer in adjusting the model's billions of parameters to autonomously learn the complex physical dynamics governing each of the different Earth system domains from data alone.
\subsection{Detailed Description of Evaluation Metrics}

\subsubsection{Root Mean Square Error (RMSE) and Anomaly Correlation Coefficient (ACC)}
We utilize two metrics, RMSE (Root Mean Square Error) and ACC (Anomalous Correlation Coefficient), to evaluate the forecasting performance, which can be defined as:
\begin{equation}
    {RMSE}(c, t) = \sqrt{\frac{\sum\limits_{i=1}^{N_{\text{lat}}} \sum\limits_{j=1}^{N_{\text{lon}}} L(i) \left( \hat{\mathbf{A}}_{ij,t}^c - \mathbf{A}_{ij,t}^c \right)^2}{N_{\text{lat}} \times N_{\text{lon}}}}
\end{equation}
\begin{equation}
    \operatorname{ACC}(c, t) = \frac{\sum\limits_{i=1}^{N_{\text{lat}}} \sum\limits_{j=1}^{N_{\text{lon}}} L(i) \hat{\mathbf{A'}}_{ij,t}^c \mathbf{A'}_{ij,t}^c}{\sqrt{\sum\limits_{i=1}^{N_{\text{lat}}} \sum\limits_{j=1}^{N_{\text{lon}}} L(i) \left( \hat{\mathbf{A'}}_{ij,t}^c \right)^2 \times \sum\limits_{i=1}^{N_{\text{lat}}} \sum\limits_{j=1}^{N_{\text{lon}}} L(i) \left( \mathbf{A'}_{ij,t}^c \right)^2}}
\end{equation}
where $\mathbf{A}_{i, j, t}^c$ represents the value of variable $c$ at horizontal coordinate $(i, j)$ and time t. Latitude-dependent weights are defined as $L(i)=N_{\text {lat }} \times \frac{\cos \phi_i}{\sum_{i’=1}^{N_{\text {lat}}} \cos \phi_{i’}}$, where $\phi_i$ is the latitude at index i. The anomaly of $A$, denoted as $A'$, is computed as the deviation from its climatology, for example, it corresponds to the long-term mean of the meteorological state estimated from multiple years of training data. To evaluate model performance, RMSE and ACC are averaged across all time steps and spatial coordinates, providing summary statistics for each variable $c$ at a given lead time $\Delta t$.

\subsection{Visualization of Model Performance: The Skill Score Matrix}

To facilitate a comprehensive and concise comparison of multiple models across various physical variables within the medium-range weather forecasting window (e.g., 1-10 days), we devise a visualization method termed the {Skill Score Matrix}. \textbf{Fig.~\ref{TritonCast_weather}a} in the main text serves as a prime example, providing a unified view that clearly illustrates the performance disparities of different baseline models (e.g., IFS-HRES, Pangu) relative to our proposed {TritonCast} model across different lead times and variables. This section details, with this figure as a reference, the fundamental concept and implementation pipeline for its construction.

The first step in this methodology is to calculate the day-by-day {spatially-averaged Root Mean Square Error (RMSE)} for all models under comparison (including {TritonCast}) and for all target variables (such as U10M, V10M, U850 in \textbf{Fig.~\ref{TritonCast_weather}a}) over the 1- to 10-day forecast range. This process yields a comprehensive dataset of RMSE values, containing the specific error for each model, each variable, and at each lead time.

The second and central step is the calculation of the {Relative Skill Score}. We establish the RMSE of {TritonCast} as the benchmark. The skill score for any given baseline model on a specific variable and at a particular forecast day is defined as the percentage relative difference between the baseline's RMSE and {TritonCast}'s RMSE under the same conditions. The mathematical definition is as follows:
\begin{equation}
    \text{Skill Score}_{\text{baseline}}(v, d) = \frac{\text{RMSE}_{\text{baseline}}(v, d) - \text{RMSE}_{\text{TritonCast}}(v, d)}{\text{RMSE}_{\text{TritonCast}}(v, d)} \times 100\%
\end{equation}
where $v$ represents the physical variable and $d$ is the lead time in days. According to this definition, the meaning of our red-blue diverging colormap becomes highly intuitive:
\begin{itemize}
    \item \textbf{Positive values (colored blue)}: These indicate that the baseline model has a higher RMSE than {TritonCast}. This means the baseline model's error is larger; therefore, {blue areas signify that {TritonCast} performs better}.
    \item \textbf{Negative values (colored red)}: These indicate that the baseline model has a lower RMSE. This means the baseline model's error is smaller; therefore, {red areas signify that {TritonCast} performs relatively worse}.
    \item \textbf{A value of zero (colored white)}: This indicates that the two models have comparable performance.
\end{itemize}

The final step is the {matrix visualization}. We organize all calculated skill scores into a 2D matrix where the {rows represent the different baseline models} (e.g., IFS-HRES, Pangu in \textbf{Fig.~\ref{TritonCast_weather}a}) and the {columns represent the different physical variables}. Each cell within this matrix is itself a one-dimensional heatmap, where its horizontal axis represents the lead time from 1 to 10 days, and the color maps to the daily skill score of that model for that variable. 

This construction results in a highly information-dense visualization matrix. An observer can not only quickly perform inter-model (vertical) and inter-variable (horizontal) performance comparisons but can also discern the trend of a model's performance relative to {TritonCast} as a function of lead time by observing the color progression within each cell.

\subsection{Power Spectral Density (PSD) and Energy Spectrum Analysis}

To conduct an in-depth evaluation of the model's fidelity in representing physical processes across different spatial scales from a frequency-domain perspective, we employ {Power Spectral Density (PSD)} analysis. This method reveals how the energy of a physical field, such as sea surface temperature, is distributed across various spatial frequencies or wavenumbers. A physically realistic model should accurately reproduce the energy distribution of the ground truth field across all scales. Any significant deviation may indicate the presence of "spectral bias" in the model either under-representing certain scales (excessive energy dissipation) or over-representing them (spurious energy accumulation). \textbf{Fig.~\ref{Figure2_Ocean}d} in the main text is a typical application of our PSD analysis.

Our PSD calculation pipeline is based on the two-dimensional Discrete Fourier Transform (DFT). For a given 2D physical field $\mathbf{A}(x, y)$ (e.g., a predicted temperature field at a specific time), the process is as follows:

First, to mitigate spectral leakage caused by discontinuities at the data boundaries, we preprocess the data by applying a 2D window function (e.g., a Hanning window).

Second, we compute the 2D Fast Fourier Transform (FFT) of the windowed data field:
\begin{equation}
    \hat{\mathbf{A}}(k_x, k_y) = \mathcal{F}\{\mathbf{A}(x, y)\} = \iint \mathbf{A}(x, y) e^{-i(k_x x + k_y y)} dx dy
\end{equation}
where $k_x$ and $k_y$ are the zonal and meridional wavenumbers, respectively, which are proportional to the spatial frequencies.

The Power Spectral Density is then defined as the squared magnitude of the Fourier transform result. This represents the energy density at each wavenumber pair $(k_x, k_y)$:
\begin{equation}
    \text{PSD}(k_x, k_y) = |\hat{\mathbf{A}}(k_x, k_y)|^2
\end{equation}

For visualization and analysis, we typically perform two post-processing steps on the PSD result. First, we use an `fftshift` operation to move the zero-frequency component (corresponding to the largest scale or the mean value) to the center of the spectrum. Second, we take the base-10 logarithm ($\log_{10}$) of the result. This logarithmic scaling allows for better visualization of energy differences across orders of magnitude, which is especially useful in the high-wavenumber (small-scale) regions where energy is typically much lower than in the low-wavenumber regions.

In our experiments, as exemplified by \textbf{Fig.~\ref{Figure2_Ocean}d}, we perform an ensemble average over the results from multiple initial conditions (ICs). Specifically, for each IC, we compute the PSD of its forecast at a specific lead time (e.g., day 60). We then average the PSDs from all ICs in logarithmic space. This ensemble-averaging approach effectively smooths out noise arising from the stochasticity of individual initial conditions, revealing a more robust statistical signature of the model's spectral properties. Finally, we present a side-by-side visual comparison of the ensemble-averaged PSD of the ground truth against that of {TritonCast} and other baseline models. The PSD plot of an ideal model should closely match that of the ground truth across all wavenumbers. Any pronounced decay of energy at high wavenumbers signals a model's deficiency in capturing fine-scale details.

\subsection{Probability Density Function (PDF) Analysis}
In addition to evaluating the energy distribution in the frequency domain, we also examine the model's ability to reproduce the statistical properties of the physical fields through {Probability Density Function (PDF)} analysis. We focus particularly on the PDF of the vorticity field, as shown in \textbf{Fig.~\ref{turbulence_comparison}d} of the main text. In nonlinear dynamical systems such as 2D turbulence, extreme events i.e., the occurrence of very large or small vorticity values are rare but critically important for the overall evolution of the system. A physically faithful model must not only predict the mean state accurately but also capture the correct statistical distribution of these extreme events.

The PDF of a vorticity field typically exhibits a {heavy-tailed distribution}. This means that compared to a standard Gaussian distribution, its tails (corresponding to extreme values) decay much more slowly, indicating that extreme events are far more probable than a Gaussian assumption would suggest. A failure to accurately reproduce this heavy-tailed characteristic often implies that the model is excessively smoothing or suppressing small-scale, high-intensity vortical structures, which is a direct statistical manifestation of "spectral bias."

Our PDF analysis pipeline is as follows:
First, for a specific time step (e.g., t=99), we extract the vorticity values from all spatial grid points of the simulation output, flattening the 2D (or 3D) vorticity field into a one-dimensional array.

Second, as vorticity is a continuous variable, we employ the {Kernel Density Estimation (KDE)} method to estimate its continuous probability density function from the discrete sample points. KDE generates a smooth density curve by placing a kernel (e.g., a Gaussian kernel) on each data point and then summing all kernels. This is a more accurate and smoother density estimation method than traditional histograms. For a given set of vorticity samples $\{ \omega_i \}_{i=1}^n$, the PDF estimate $\hat{f}(\omega)$ can be expressed as:
\begin{equation}
    \hat{f}(\omega) = \frac{1}{nh} \sum_{i=1}^{n} K\left(\frac{\omega - \omega_i}{h}\right)
\end{equation}
where $n$ is the total number of samples, $h$ is the bandwidth, and $K$ is the kernel function.

Finally, to clearly observe and compare the tail distributions, we plot the calculated PDFs on a {semi-logarithmic scale} (with the Y-axis in log scale). As shown in \textbf{Fig.~\ref{turbulence_comparison}d}, we overlay the PDF of the Ground Truth with the PDFs from {TritonCast} and other baseline models for direct comparison. The PDF curve of an ideal model should closely match the ground truth curve over the entire range of values, especially in the heavy-tailed regions that are crucial for the system's dynamics.

\clearpage

\section{Experimental Setup}
\label{appendix:EXPSTEUP}
This section provides the detailed experimental step, including data proprocessing details, model architecture configurations, training hyperparameters, and inference procedure for the various experimental domains addressed in this study, covering atmospheric science, ocean science, and theoretical physics (turbulence). The primary objective of providing this information is to ensure the full reproducibility of our experimental results, thereby offering a solid foundation for other researchers to validate and extend our work. To maintain clarity, we follow the paper's experimental divisions and present an independent, standardized configuration table for each distinct model setup (e.g., the large-scale model for medium-range weather forecasting and the lightweight model for long-term stability tests).
\subsection{Experimental Setup for Atmospheric Science}
\label{appendix:weather_appendix_exp}

\subsubsection{Medium-range Weather Forecasting (WeatherBench 2 Benchmark)}

This section provides detailed information regarding the training configuration of the \method{} model for the global medium-range weather forecasting experiments, facilitating the reproducibility of the results presented.

For this task, we design and train a 1B parameter version of TritonCast from scratch using $1.5 \degree$ resolution ERA5 data. Input data were spatially processed to a regular 1.5° × 1.5° latitude-longitude grid (121 × 240 points). We use the data with size 120 × 240. The model takes a single time step as input to predict the atmospheric state at the next time step, corresponding to a 6-hour lead time in this study. Both input and output states consist of 69 atmospheric variables. The dataset is partitioned chronologically: data from 1959 to 2017 served as the training set, 2018-2019 data were used for validation, and 2020 for testing. We use training data to calculate mean and standard deviation. Before inputting the data to the neural network, we first subtract the respective mean and divided it by the standard deviation. The training data loader incorporated shuffling, managed via a distributed sampler for multi-GPU training.

\paragraph{Training and Optimization}: The model is trained on 48 GPUs using Distributed Data Parallel (DDP) with a per-GPU batch size of $1$, resulting in an effective global batch size of $48$. We employ a standard relative $L_2$ as the loss function, without any specific variable or latitude-based weighting. Following the previous works, we apply a pre-training and fine-tuning strategy. In the pre-training stage, the data range from 1959 to 2017 are used for training and 2018 to 2019 for validating. The Adam optimizer is used with an initial learning rate of $1e-3$. We adopt a simple learning rate schedule, `CosineAnnealingLR` with no warm-up phase. The learning rate gradually decreases to 0. The model is trained for a maximum of $200$ epochs, and the weights corresponding to the lowest validation loss are saved as the best model. In the fine-tuning phase, we begin with the best-performing model from the pre-training stage. The training data spans from 2013 to 2017, while the validation data covers the period from 2018 to 2019. The Adam optimizer is employed with a fixed learning rate of $1e-6$. The fine-tuning process is conducted in multiple stages. Initially, we start from the best model in pre-training stage and fine-tuning the model gradually from 2-step to 12-step supervision within 1 epoch, with a total of 10 epochs. In the subsequent stage, we start from the weight of the last epoch and fine-tuning the model from 13-step to 24-step supervision over 1 epoch, again with a total of 10 epochs. The next phase involves fine-tuning the model from 25-step to 48-step supervision over 1 epoch, start from the last model in previous fine-tuning stage, maintaining a total of 10 epochs. Finally, we fine-tuning the model from 49-step to 60-step supervision in 1 epoch, with the total epochs set to 10. The weight of the last epoch is used for evaluating \method{}. Detailed gradient descent updates can be found in \textbf{Tab.~\ref{tab:detailed_config_atmos_medium_finetune}}.

\begin{table}[h!]
\centering
\footnotesize
\caption{\textbf{Detailed gradient descent updates for each epoch during the medium-range weather forecasting fine-tuning phase of TritonCast.}}
\label{tab:detailed_config_atmos_medium_finetune}
\setlength{\tabcolsep}{10pt}
\begin{tabular}{l|cc|cc|cc|cc}
\toprule
\multirow{2}{*}{Fine-tuning Stage} & \multicolumn{2}{c|}{2-step to 12-step} & \multicolumn{2}{c|}{13-step to 24-step} & \multicolumn{2}{c|}{25-step to 48-step} & \multicolumn{2}{c}{49-step to 60-step} \\ \cmidrule{2-9} 
                                   & 2 to 11              & 12              & 13 to 23              & 24              & 25 to 47              & 48              & 49 to 59              & 60             \\ \midrule
Gradient Descent Updates           & 10                   & 50              & 10                    & 49              & 5                     & 32              & 5                     & 35             \\ \bottomrule
\end{tabular}
\end{table}

\begin{table}[h!]
\centering
\footnotesize
\caption{\textbf{Detailed model configuration and hyperparameter for TritonCast in medium-range weather forecasting.}}
\label{tab:detailed_config_atmos_medium}
\setlength{\tabcolsep}{15pt}
\begin{tabular}{lll}
\toprule
{Parameter} & {Code Variable} & {Value} \\
\midrule
\multicolumn{3}{l}{\textit{{1. Overall Architecture}}} \\
Input Channels & \texttt{input\_channel} & 69 \\
Output Channels & \texttt{output\_channels} & 69 \\
V-Cycle Depth / Number of Levels & \texttt{num\_spatial\_layers} & 4 \\
Spatial Hidden Dimension & \texttt{spatial\_hidden\_dim} & 1536\\
Gradient Checkpointing & \texttt{gradient\_checkpointing} & False\\
\midrule
\multicolumn{3}{l}{\textit{{2. Latent Dynamical Core (LDC)}}} \\
Number of LDC Blocks & \texttt{num\_temporal\_layers} & 8\\
LDC Temporal Hidden Dimension & \texttt{temporal\_hidden\_dim} (\texttt{channel\_hid}) & 3072\\
LDC MLP Expansion Ratio & \texttt{mlp\_ratio} & 8 \\
\midrule
\multicolumn{3}{l}{\textit{{3. Core Module: Self-Attention}}} \\
Number of Attention Heads & \texttt{num\_heads} & 4\\
QKV Bias & \texttt{qkv\_bias} & False \\
Attention Dropout Rate & \texttt{attn\_drop} &  0\\
Projection Dropout Rate & \texttt{proj\_drop} (in Attention) &0 \\
LayerScale Init Value & \texttt{init\_value} (for gamma) &1e-6 \\
\midrule
\multicolumn{3}{l}{\textit{{4. Core Module: Convolution}}} \\
BC Block Activation Function & \texttt{nn.LeakyReLU} & None \\
BC Block Normalization Type & \texttt{nn.GroupNorm} & 2\\
ConvBlock MLP Expansion Ratio & \texttt{mlp\_ratio} & 4\\
ConvBlock/SA Block Activation Fn & \texttt{act\_layer=nn.GELU} & None\\
ConvBlock/SA Block Norm Type & \texttt{norm\_layer=nn.LayerNorm} & None\\
\midrule
\multicolumn{3}{l}{\textit{{5. Training Hyperparameters}}} \\
Optimizer & \texttt{optimizer = optim.Adam} & \\
Learning Rate Scheduler & \texttt{torch.optim.lr\_scheduler.CosineAnnealingLR} &  \\
Max Learning Rate & \texttt{lr=1e-3} &1e-3 \\
Min Learning Rate & \texttt{Min\_LR} & 0\\
Batch Size & \texttt{batch\_size=1} &  1 per GPU \\
Num Gpus & \texttt{num\_gpus} & 48 \\
Training Duration / Epochs & \texttt{num\_epochs = 200} & 200 \\
\midrule
\multicolumn{3}{l}{\textit{{6. Regularization}}} \\
General Dropout Rate & \texttt{drop} & None \\
Drop Path Rate (LDC) & \texttt{drop\_path} & None \\
\midrule
\multicolumn{3}{l}{\textit{{7. Model Scale}}} \\
Total Parameters & Total Parameters & 1B \\
Total Training Time & Total Training Time  & 400 hours\\
\bottomrule
\end{tabular}
\end{table}

\paragraph{Evaluation Protocol}: During evaluation, we conduct a rigorous, 10-day autoregressive forecast. The process is initialized with a single input: the true atmospheric state from the 0:00 UTC, Jan. 1, 2020. Subsequently, the model performs 40 consecutive forecast steps with a $6$-hour step size. In each step, the output from the previous step serves as the complete input for the current step. No corrections from observations or reanalysis (i.e., no data assimilation) are introduced during the entire rollout. For baseline moedels, the forecast results are sourced for WeatherBench 2. And following NeuralGCM, we compare the quantitative results of all methods by bilinear interpolation to 1.5 degree. The interval between adjacent initial conditions (ICs) is 12 hours, and we report the average results of the first 700 ICs.

The following \textbf{Tab.~\ref{tab:detailed_config_atmos_medium}} provides a comprehensive breakdown of all key parameters.

\clearpage
\subsubsection{Long-term Autoregressive Stability Tests}

To ensure the reproducibility of our experiments and to provide a clear reference for the long-term atmospheric stability tests discussed in the main text, this section details the specific model configuration and hyperparameters for the TritonCast model used. This test represents one of the most demanding benchmarks in AI weather forecasting, with the objective of completing a full year-long, purely autoregressive global forecast without any external data forcing or intervention. The success of the model hinges on its ability to maintain numerical stability throughout the entire integration period, avoiding divergence or significant climate drift.

For this task, we design and train a lightweight, 0.02B parameter version of TritonCast from scratch using $1 \degree$ resolution ERA5 data, without any fine-tuning. Notably, this exceptional long-term stability is achieved with minimal use of conventional regularization techniques such as dropout and weight decay. This strongly suggests that the model's stability is primarily derived from its novel, multi-grid inspired architectural design, rather than from explicit regularization tricks.

\paragraph{Training and Optimization}: The model is trained on 8 GPUs using Distributed Data Parallel (DDP) with a per-GPU batch size of $1$, resulting in an effective global batch size of $8$. We employ a standard {Mean Squared Error (MSE)} as the loss function, without any specific variable or latitude-based weighting. The Adam optimizer is used with an initial learning rate of $1e-3$. We adopt a simple learning rate schedule, `StepLR`, which decays the learning rate by a factor of $0.2$ (gamma=$0.2$) every $50$ epochs, with no warm-up phase. The model is trained for a maximum of $1000$ epochs, and the weights corresponding to the lowest validation loss are saved as the best model.

\paragraph{Evaluation Protocol}: During evaluation, we conduct a rigorous, year-long autoregressive forecast. The process is initialized with a single input: the true atmospheric state from the first day of the test year (e.g., January 1, 00:00 UTC). Subsequently, the model performs 364 (or 365) consecutive forecast steps with a $24$-hour step size. In each step, the output from the previous step serves as the complete input for the current step. No corrections from observations or reanalysis (i.e., no data assimilation) are introduced during the entire rollout.

The following \textbf{Tab.~\ref{tab:year_forcast_detailed_config_en}} provides a comprehensive breakdown of all key parameters.

\begin{table}[h!]
\centering
\footnotesize
\caption{\textbf{Detailed model configuration and hyperparameter for TritonCast model Used in long-term autoregressive stability tests.}}
\label{tab:year_forcast_detailed_config_en}
\setlength{\tabcolsep}{18pt}
\begin{tabular}{lll}
\toprule
{Parameter} & {Code Variable} & {Value} \\
\midrule
\multicolumn{3}{l}{\textit{{1. Overall Architecture}}} \\
Input Channels & \texttt{input\_channel} & 69 \\
Output Channels & \texttt{output\_channels} & 69 \\
V-Cycle Depth / Number of Levels & \texttt{num\_spatial\_layers} & 4 \\
Spatial Hidden Dimension & \texttt{spatial\_hidden\_dim} & 256\\
Gradient Checkpointing & \texttt{gradient\_checkpointing} & True\\
\midrule
\multicolumn{3}{l}{\textit{{2. Latent Dynamical Core (LDC)}}} \\
Number of LDC Blocks & \texttt{num\_temporal\_layers} & 8\\
LDC Temporal Hidden Dimension & \texttt{temporal\_hidden\_dim} (\texttt{channel\_hid}) & 512\\
LDC MLP Expansion Ratio & \texttt{mlp\_ratio} & 8 \\
\midrule
\multicolumn{3}{l}{\textit{{3. Core Module: Self-Attention}}} \\
Number of Attention Heads & \texttt{num\_heads} & 4\\
QKV Bias & \texttt{qkv\_bias} & False \\
Attention Dropout Rate & \texttt{attn\_drop} &  0\\
Projection Dropout Rate & \texttt{proj\_drop} (in Attention) &0 \\
LayerScale Init Value & \texttt{init\_value} (for gamma) &1e-6 \\
\midrule
\multicolumn{3}{l}{\textit{{4. Core Module: Convolution}}} \\
BC Block Activation Function & \texttt{nn.LeakyReLU} & None \\
BC Block Normalization Type & \texttt{nn.GroupNorm} & 2\\
ConvBlock MLP Expansion Ratio & \texttt{mlp\_ratio} & 4\\
ConvBlock/SA Block Activation Fn & \texttt{act\_layer=nn.GELU} & None\\
ConvBlock/SA Block Norm Type & \texttt{norm\_layer=nn.LayerNorm} & None\\
\midrule
\multicolumn{3}{l}{\textit{{5. Training Hyperparameters}}} \\
Optimizer & \texttt{optimizer = optim.Adam} & \\
Learning Rate Scheduler & \texttt{torch.optim.lr\_scheduler.StepLR} &  \\
Max Learning Rate & \texttt{lr=1e-3} &1e-3 \\
Min Learning Rate & \texttt{Min\_LR} & 0\\
Batch Size & \texttt{batch\_size} &  1 per GPU \\
Num Gpus & \texttt{num\_gpus} & 8 \\
Training Duration / Epochs & \texttt{num\_epochs = 1000} & 1000 \\
\midrule
\multicolumn{3}{l}{\textit{{6. Regularization}}} \\
General Dropout Rate & \texttt{drop} & None \\
Drop Path Rate (LDC) & \texttt{drop\_path} & None \\
\midrule
\multicolumn{3}{l}{\textit{{7. Model Scale}}} \\
Total Parameters & Total Parameters & 0.02B \\
Total Training Time & Total Training Time  & 32.9 hours\\
\bottomrule
\end{tabular}
\end{table}

\subsubsection{Multi-year Climate Simulations}
\label{appendix:Climate_Simulations}

This section provides detailed information regarding the training configuration of the \method{} model for the multi-year climate simulations experiments, facilitating the reproducibility of the results presented.

\begin{table}[h!]
\centering
\footnotesize
\caption{\textbf{Detailed model configuration and hyperparameter for TritonCast in multi-year climate simulations.}}
\label{tab:detailed_config_climate_simulations}
\setlength{\tabcolsep}{15pt}
\begin{tabular}{lll}
\toprule
{Parameter} & {Code Variable} & {Value} \\
\midrule
\multicolumn{3}{l}{\textit{{1. Overall Architecture}}} \\
Input Channels & \texttt{input\_channel} & 70 \\
Output Channels & \texttt{output\_channels} & 69 \\
V-Cycle Depth / Number of Levels & \texttt{num\_spatial\_layers} & 4 \\
Spatial Hidden Dimension & \texttt{spatial\_hidden\_dim} & 512\\
Gradient Checkpointing & \texttt{gradient\_checkpointing} & False\\
\midrule
\multicolumn{3}{l}{\textit{{2. Latent Dynamical Core (LDC)}}} \\
Number of LDC Blocks & \texttt{num\_temporal\_layers} & 8\\
LDC Temporal Hidden Dimension & \texttt{temporal\_hidden\_dim} (\texttt{channel\_hid}) & 1024\\
LDC MLP Expansion Ratio & \texttt{mlp\_ratio} & 8 \\
\midrule
\multicolumn{3}{l}{\textit{{3. Core Module: Self-Attention}}} \\
Number of Attention Heads & \texttt{num\_heads} & 4\\
QKV Bias & \texttt{qkv\_bias} & False \\
Attention Dropout Rate & \texttt{attn\_drop} &  0\\
Projection Dropout Rate & \texttt{proj\_drop} (in Attention) &0 \\
LayerScale Init Value & \texttt{init\_value} (for gamma) &1e-6 \\
\midrule
\multicolumn{3}{l}{\textit{{4. Core Module: Convolution}}} \\
BC Block Activation Function & \texttt{nn.LeakyReLU} & None \\
BC Block Normalization Type & \texttt{nn.GroupNorm} & 2\\
ConvBlock MLP Expansion Ratio & \texttt{mlp\_ratio} & 4\\
ConvBlock/SA Block Activation Fn & \texttt{act\_layer=nn.GELU} & None\\
ConvBlock/SA Block Norm Type & \texttt{norm\_layer=nn.LayerNorm} & None\\
\midrule
\multicolumn{3}{l}{\textit{{5. Training Hyperparameters}}} \\
Optimizer & \texttt{optimizer = optim.Adam} & \\
Learning Rate Scheduler & \texttt{torch.optim.lr\_scheduler.CosineAnnealingLR} &  \\
Max Learning Rate & \texttt{lr=1e-3} &1e-3 \\
Min Learning Rate & \texttt{Min\_LR} & 0\\
Batch Size & \texttt{batch\_size=1} &  1 per GPU \\
Num Gpus & \texttt{num\_gpus} & 16 \\
Training Duration / Epochs & \texttt{num\_epochs = 200} & 200 \\
\midrule
\multicolumn{3}{l}{\textit{{6. Regularization}}} \\
General Dropout Rate & \texttt{drop} & None \\
Drop Path Rate (LDC) & \texttt{drop\_path} & None \\
\midrule
\multicolumn{3}{l}{\textit{{7. Model Scale}}} \\
Total Parameters & Total Parameters & 0.1B \\
Total Training Time & Total Training Time  & 7 hours\\
\bottomrule
\end{tabular}
\end{table}

For this task, we design and train a 0.1B parameter version of TritonCast from scratch using $1.5 \degree$ resolution ERA5 data and $1.5 \degree$ resolution GLORYS12 data. The simulation is conducted based on the experimental design of the Atmospheric Model Intercomparison Project (AMIP), where prescribed sea surface temperature (SST) is used as the boundary condition to drive the evolution of the atmospheric model. Input data were spatially processed to a regular 1.5° × 1.5° latitude-longitude grid (121 × 240 points). We use the data with size 120 × 240. For the atmospheric data from ERA5, we use the data at 12:00 UTC of each day. And for the forcing data (oceanic variable) SST from GLORYS12 data, it corresponding to the daily mean state. The model takes a single time step as input (composed of atmospheric state and oceanic forcing at time $t$) to simulate the atmospheric state at the next time step, corresponding to a 24-hour lead time in this study. The input state consists of 70 variables and the output states consists of 69 variables. The dataset is partitioned chronologically: data from 1993 to 2017 served as the training set and data from 2018 to 2020 served as testing data. We use training data to calculate mean and standard deviation. Before inputting the data to the neural network, we first subtract the respective mean and divided it by the standard deviation. The training data loader incorporated shuffling, managed via a distributed sampler for multi-GPU training.

\paragraph{Training and Optimization}: The model is trained on 16 GPUs using Distributed Data Parallel (DDP) with a per-GPU batch size of $1$, resulting in an effective global batch size of $16$. We employ a standard relative $L_2$ as the loss function, without any specific variable or latitude-based weighting. The Adam optimizer is used with an initial learning rate of $1e-3$. We adopt a simple learning rate schedule, `CosineAnnealingLR` with no warm-up phase. The learning rate gradually decreases to 0. The model is trained for a maximum of $200$ epochs, and the weight corresponding to the last epoch is used for analysis.

\paragraph{Evaluation Protocol}: During evaluation, we conduct a 2500-day climate simulation. The process is initialized with a single input: the true atmospheric state and SST forcing. Subsequently, the model performs 2500 consecutive forecast steps with a $24$-hour step size. In each step, the atmospheric output from the previous step and the true SST forcing serve as the complete input for the current step. We run 2500-day simulations with 40 different initial conditions spaced at 1 day for the year 2018. Out of these 40 initial conditions, 40 successfully completed full 2500-day without encountering model instability. And following NeuralGCM, the comparison is conducted over the time period available for all initial conditions (similar to ensemble mean).

The following \textbf{Tab.~\ref{tab:detailed_config_climate_simulations}} provides a comprehensive breakdown of all key parameters.

\begin{figure}[h!]
\centering
\includegraphics[width=1\linewidth]{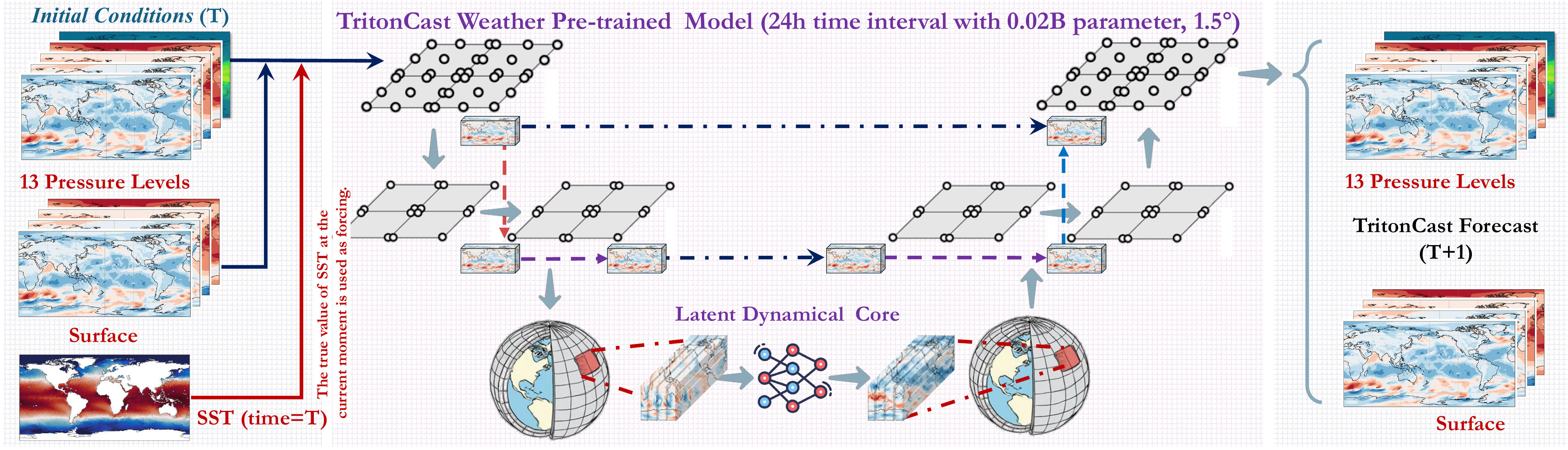}
\caption{
    \textbf{Workflow of the Multi-year Climate Simulation (AMIP-style).}
    This figure illustrates the experimental setup for long-term climate simulation. 
    At each time step (T), the TritonCast model is provided with two inputs: 
    (1) the atmospheric initial conditions (across 13 pressure levels and the surface), which are the model’s own forecast from the previous step, and 
    (2) the true, observed Sea Surface Temperature (SST) at time T, which acts as a prescribed external forcing. 
    The model then integrates this combined state forward for 24 hours to produce the atmospheric forecast for time T+1. 
    This process is iterated autoregressively over thousands of days, enabling a rigorous test of the model's long-term stability and its physical response to realistic boundary forcing.
}
\label{fig:climate_simluation}
\end{figure}

\paragraph{Simulation Workflow.}
To visually illustrate the experimental setup, the end-to-end workflow for the multi-year climate simulation is schematically detailed in \textbf{Fig.~\ref{fig:climate_simluation}}. This procedure employs a one-way coupling strategy where the atmospheric model operates autoregressively, driven by prescribed real-world boundary conditions. Specifically, for each 24-hour integration step, the model's input is twofold: (1) the complete atmospheric state (69 variables) as predicted by the model from the previous time step, and (2) the true, observed sea surface temperature (SST) for the current time step. This SST field serves as the external forcing. The model then outputs the atmospheric state for the next time step, which is subsequently used as the atmospheric input for the next iteration. This iterative cycle is repeated for the entire 2500-day simulation period, providing a rigorous test of the model's intrinsic stability and its ability to maintain a physically consistent climate trajectory under realistic forcing.

\subsection{Experimental Setup for Oceanography}
\label{appendix:ocean_appendix_exp}

\subsubsection{Global Ocean Long-term Simulations}
\label{section:global_ocean_simulation}
This section provides detailed information regarding the training configuration of the \method{} model for the global ocean long-term simulations experiments, facilitating the reproducibility of the results presented.

\begin{table}[h!]
\centering
\footnotesize
\caption{\textbf{Detailed model configuration and hyperparameter for TritonCast in global ocean long-term simulations.}}
\label{tab:detailed_config_ocean_simulations}
\setlength{\tabcolsep}{15pt}
\begin{tabular}{lll}
\toprule
{Parameter} & {Code Variable} & {Value} \\
\midrule
\multicolumn{3}{l}{\textit{{1. Overall Architecture}}} \\
Input Channels & \texttt{input\_channel} & 97 \\
Output Channels & \texttt{output\_channels} & 93 \\
V-Cycle Depth / Number of Levels & \texttt{num\_spatial\_layers} & 4 \\
Spatial Hidden Dimension & \texttt{spatial\_hidden\_dim} & 256\\
Gradient Checkpointing & \texttt{gradient\_checkpointing} & False\\
\midrule
\multicolumn{3}{l}{\textit{{2. Latent Dynamical Core (LDC)}}} \\
Number of LDC Blocks & \texttt{num\_temporal\_layers} & 8\\
LDC Temporal Hidden Dimension & \texttt{temporal\_hidden\_dim} (\texttt{channel\_hid}) & 512\\
LDC MLP Expansion Ratio & \texttt{mlp\_ratio} & 8 \\
\midrule
\multicolumn{3}{l}{\textit{{3. Core Module: Self-Attention}}} \\
Number of Attention Heads & \texttt{num\_heads} & 4\\
QKV Bias & \texttt{qkv\_bias} & False \\
Attention Dropout Rate & \texttt{attn\_drop} &  0\\
Projection Dropout Rate & \texttt{proj\_drop} (in Attention) &0 \\
LayerScale Init Value & \texttt{init\_value} (for gamma) &1e-6 \\
\midrule
\multicolumn{3}{l}{\textit{{4. Core Module: Convolution}}} \\
BC Block Activation Function & \texttt{nn.LeakyReLU} & None \\
BC Block Normalization Type & \texttt{nn.GroupNorm} & 2\\
ConvBlock MLP Expansion Ratio & \texttt{mlp\_ratio} & 4\\
ConvBlock/SA Block Activation Fn & \texttt{act\_layer=nn.GELU} & None\\
ConvBlock/SA Block Norm Type & \texttt{norm\_layer=nn.LayerNorm} & None\\
\midrule
\multicolumn{3}{l}{\textit{{5. Training Hyperparameters}}} \\
Optimizer & \texttt{optimizer = optim.Adam} & \\
Learning Rate Scheduler & \texttt{torch.optim.lr\_scheduler.CosineAnnealingLR} &  \\
Max Learning Rate & \texttt{lr=1e-3} &1e-3 \\
Min Learning Rate & \texttt{Min\_LR} & 0\\
Batch Size & \texttt{batch\_size=1} &  1 per GPU \\
Num Gpus & \texttt{num\_gpus} & 48 \\
Training Duration / Epochs & \texttt{num\_epochs = 200} & 200 \\
\midrule
\multicolumn{3}{l}{\textit{{6. Regularization}}} \\
General Dropout Rate & \texttt{drop} & None \\
Drop Path Rate (LDC) & \texttt{drop\_path} & None \\
\midrule
\multicolumn{3}{l}{\textit{{7. Model Scale}}} \\
Total Parameters & Total Parameters & 0.02B \\
Total Training Time & Total Training Time  & 400 hours\\
\bottomrule
\end{tabular}
\end{table}

For this task, we design and train a 0.02B parameter version of TritonCast from scratch using $0.25 \degree$ resolution GLORYS12 data. Input data were spatially processed to a regular 0.25° × 0.25° latitude-longitude grid (721 × 1440 points). We use the data with size 720 × 1440. For the forcing atmospheric data from ERA5, we use the data at 12:00 UTC of each day. And for the oceanic variables from GLORYS12 data, they correspond to the daily mean state. We first remove the climatological mean from variables exhibiting strong periodicity, thereby directing the model’s attention toward small anomalies, which includes Sea Salinity, Sea temperature, and Sea surface height. The model takes a single time step as input (composed of ocean state at time $t$ and atmospheric forcing at time $t+1$) to simulate the ocean state at the next time step, corresponding to a 24-hour lead time in this study. The input state consists of 97 variables and the output states consists of 93 variables. The dataset is partitioned chronologically: data from 1993 to 2017 served as the training set, data from 2018 to 2019 served as validation data, and 2020 for testing. We use training data to calculate mean and standard deviation. Before inputting the data to the neural network, we first subtract the respective mean and divided it by the standard deviation. The training data loader incorporated shuffling, managed via a distributed sampler for multi-GPU training.

\paragraph{Training and Optimization}: Baseline models except WenHai and our model are trained using the same experimental setup with pre-training and fine-tuning stages. They are trained on 48 GPUs using Distributed Data Parallel (DDP) with a per-GPU batch size of $1$, resulting in an effective global batch size of $48$. Prior to calculating the loss, we first apply a mask with a value of zero to the land areas, thereby guiding the model to focus on simulating the ocean state. We employ a standard relative $L_2$ as the loss function, without any specific variable or latitude-based weighting. Following the previous works, we apply a pre-training and fine-tuning strategy. In the pre-training stage, the Adam optimizer is used with an initial learning rate of $1e-3$. We adopt a simple learning rate schedule, `CosineAnnealingLR` with no warm-up phase. The learning rate gradually decreases to 0. The model is trained for a maximum of $200$ epochs, and the weights corresponding to the lowest validation loss are saved as the best model. In the fine-tuning phase, we start from the best-performing model obtained during the pre-training stage. The Adam optimizer is employed with a fixed learning rate of $1\times10^{-6}$. The model is progressively fine-tuned from 2-step to 7-step supervision. For $m$-step supervision ($m \geq 2$), the training begins with the best model from the $(m-1)$-step supervision model and is trained for 10 epochs. For each $m$-step supervision stage, the best model is defined as the one achieving the lowest validation loss during $m$-step inference.

\paragraph{Evaluation Protocol}: During evaluation, we conduct a 120-day global ocean simulation. The process is initialized with a single input: the true ocean state at time $t$ and true atmospheric forcing at time $t+1$. Subsequently, the model performs 120 consecutive simulation steps with a $24$-hour step size. In each step, the ocean output from the previous step and the true atmospheric forcing serve as the complete input for the current step. Similar to the training stage, we mask the ocean state before next step simulation. We run 120-day simulations with 240 different initial conditions spaced at 1 day for the year 2020. Due to WenHai model is close-sourced, following the common practice, we use its official released onnx file to inference under the same ICs. And we use bilinea interpolation to scale the result of WenHai to 0.25 degrees. We report the average results of 240 ICs.

The following \textbf{Tab.~\ref{tab:detailed_config_ocean_simulations}} provides a comprehensive breakdown of all key parameters.

\subsubsection{Coupled Ocean-Atmosphere Long-term Forecasts}
\label{sec:appendix_coupled_forecast_en}
This section provides detailed information regarding the configuration of the \method{} model for the global coupled ocean-atmosphere long-term forecasts experiments, facilitating the reproducibility of the results presented.

For this task, we use the TritonCast's oceanic model and atmospheric model to produce global ocean forecasts. For the predicted forcing atmospheric data from TritonCast's atmospheric, we use the data at 12:00 UTC of each day and bilinear interpolation is then applied to resample the data to a resolution of 0.25 degrees, aligning it with the resolution of the ocean model. And for the oceanic variables from GLORYS12 data, they correspond to the daily mean state. We first remove the climatological mean from variables exhibiting strong periodicity, thereby directing the model’s attention toward small anomalies, which includes Sea Salinity, Sea temperature, and Sea surface height. Similar to global ocean long-term simulations, before inputting the data to the neural network, we first subtract the respective mean and divided it by the standard deviation.

\paragraph{Evaluation Protocol}: To assess practical applicability of different models, we drive the ocean models with atmospheric forecasts from TritonCast's atmospheric model (6h time interval with 1B parameter), which introduces real-world uncertainty. As common atmospheric forecast models including TritonCast don’t provide all forcings required by WenHai at 60-day forecasting scale, we can’t assess its results. During evaluation, we conduct a 60-day global ocean forecasts. The process is initialized with a single input: the true ocean state at time $t$ and the predicted atmospheric forcing at time $t+1$. Subsequently, the model performs 60 consecutive simulation steps with a $24$-hour step size. In each step, the ocean output from the previous step and the predicted atmospheric forcing serve as the complete input for the current step. Similar to the training stage, we mask the ocean state before next step simulation. We run 60-day forecasts with 240 different initial conditions spaced at 1 day for the year 2020. We report the average results of 240 ICs.

\begin{figure}[h!]
\centering
\includegraphics[width=1\linewidth]{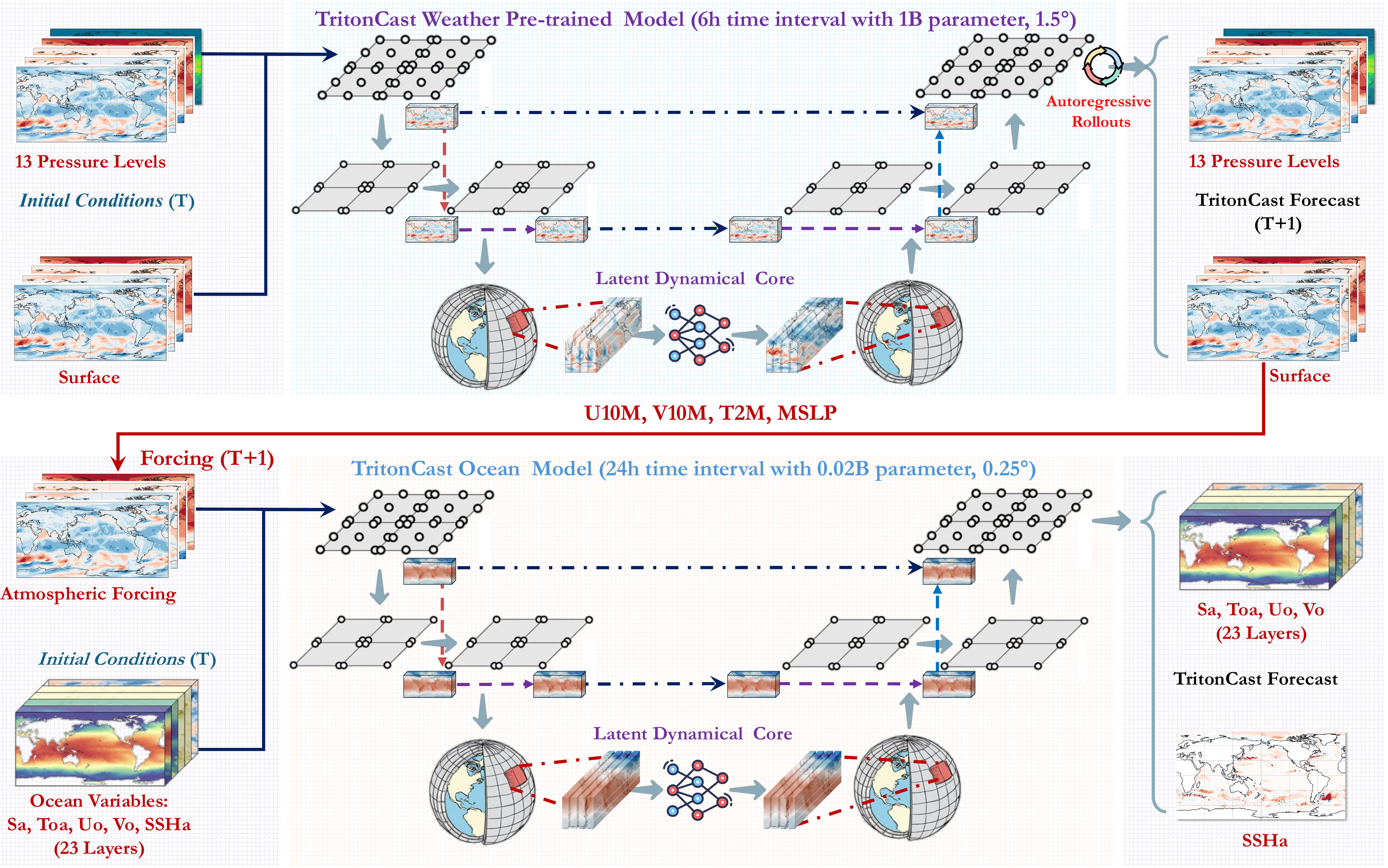}
\caption{\textbf{Schematic of the TritonCast coupled ocean-atmosphere forecast workflow.} This figure illustrates a one-way coupled forecasting workflow designed to evaluate the ocean model's performance under realistic uncertainty. \textbf{Top panel:} A pre-trained, 1-billion (1B) parameter TritonCast weather model generates forecasts at a 6-hour interval from initial atmospheric conditions (13 pressure levels and surface variables). For time alignment, the atmosphere model runs 4 times and the ocean model runs 1 time. Key surface variables predicted by this model (U10M, V10M, T2M, MSLP) serve as the atmospheric forcing. \textbf{Bottom panel:} This atmospheric forcing, combined with initial ocean conditions (including salinity \textit{S}, temperature \textit{To}, velocity \textit{Uo}/\textit{Vo}, and sea surface height anomaly SSHa), is input into a 20-million (0.02B) parameter TritonCast ocean model. The ocean model, operating at a 24-hour time interval, forecasts the future ocean state, which comprises variables across 23 layers and the sea surface height anomaly. This setup mimics an operational forecasting scenario where the ocean model is driven by predicted, rather than observed, atmospheric data.}
\label{fig:ao_couple_workflow}
\end{figure}

To visually illustrate the experimental setup for the coupled ocean-atmosphere long-term forecasts described in Supplementary~\ref{sec:appendix_coupled_forecast_en}, \textbf{Fig.~\ref{fig:ao_couple_workflow}} provides a detailed schematic of the end-to-end workflow. This procedure employs a one-way coupling strategy, where the predictive output from a standalone TritonCast atmospheric model is used to drive the ocean model, rather than relying on idealized observational or reanalysis data. The core purpose of this design is to simulate a scenario that closely resembles operational forecasting, thereby rigorously evaluating the robustness and practical applicability of the ocean model when handling upstream inputs that contain inherent uncertainties and errors.

As depicted in \textbf{Fig.~\ref{fig:ao_couple_workflow}}, the specific workflow is divided into two main stages:

\begin{enumerate}
    \item \textbf{Generation of Atmospheric Forcing (Top Panel of \textbf{Fig.~\ref{fig:ao_couple_workflow}}):} The process begins with a large-scale, pre-trained TritonCast weather model (1B parameters, 1.5$^{\circ}$ resolution). Taking the initial atmospheric state (including 13 pressure levels and surface variables) as input, this model generates an autoregressive forecast with a 6-hour step size. At each forecast step, a specific subset of its output the surface variables of 10-meter winds (U10M, V10M), 2-meter temperature (T2M), and mean sea level pressure (MSLP) is extracted. These predicted variables constitute the atmospheric forcing required to drive the ocean model.

    \item \textbf{Prediction of Ocean State (Bottom Panel of \textbf{Fig.~\ref{fig:ao_couple_workflow}}):} Subsequently, the atmospheric forcing generated in the previous stage is combined with the ocean's own initial conditions (including salinity \textit{S}, temperature \textit{To}, velocity \textit{Uo}/\textit{Vo}, and sea surface height SSH). These are then fed into a separate, higher-resolution TritonCast ocean model (0.02B parameters, 0.25$^{\circ}$ resolution). This ocean model produces a forecast with a 24-hour step size, ultimately generating a detailed prediction of the future ocean state (comprising variables across 23 vertical layers and the sea surface height).
\end{enumerate}

Through this methodology, any inevitable errors from the atmospheric forecast are propagated to the ocean model, forming a complete AI predictive chain. This provides a more stringent and realistic testbed for assessing the practical performance of the entire system.

\clearpage
\subsection{Experimental Setup for Ocean Current (CMEMS)}
\label{appendix:CMEMS_appendix_exp}
This section provides the detailed experimental setup for high-fidelity ocean eddy forecasting and cross-resolution generalization tests. We use the multi-satellite altimetry product data from the Copernicus Marine Service (CMEMS), which serves as a benchmark for evaluating the model's performance in complex ocean dynamics scenarios.

\subsubsection{High-Fidelity Eddy and Current Forecasts}

This experiment aims to rigorously evaluate the TritonCast model's ability to capture and forecast fine-scale phenomena in the most energetic and dynamically complex western boundary current regions, namely the Gulf Stream, the Kuroshio Extension, and the Agulhas Current. To achieve the highest fidelity, we train three specialized models, one for each region, using their native 0.125° high-resolution data.

We adopt a Sequence-to-Sequence forecasting paradigm, where the model takes 10 consecutive days of sea surface geostrophic velocity fields (two variables: UGOS and VGOS) as input to predict the corresponding fields for the next 10 days. The dataset is partitioned chronologically: data from 1993 to 2018 are used for training, 2019 for validation, and 2020 to 2024 for testing. As described in the dataset section, due to the small numerical range of the original data, we do not apply normalization and only handle potential missing values (NaNs).

\paragraph{Training and Optimization}: Each regional model is trained on 8 GPUs using the Distributed Data Parallel (DDP) strategy. We set the batch size to 4 per GPU, resulting in an effective global batch size of 32. The optimization objective is to minimize the standard Mean Squared Error (MSE) loss function. We use the Adam optimizer with an initial learning rate of 1e-3, paired with a CosineAnnealingLR learning rate scheduler that smoothly decays the learning rate to 1e-6 over 500 epochs. The model is trained for a total of 1000 epochs, and the model weights corresponding to the lowest validation loss are saved as the final best model.

The following \textbf{Tab.~\ref{tab:detailed_config_ocean_regional}} provides a comprehensive breakdown of the configuration and hyperparameters for the TritonCast model used in high-fidelity regional forecasting.

\begin{table}[h!]
\centering
\footnotesize
\caption{\textbf{Detailed configuration and hyperparameters for the specialized TritonCast models used in the high-fidelity eddy and current forecasting task.} These models are trained on the 0.125° high-resolution datasets for specific regions (i.e., the Gulf Stream, Kuroshio Extension, and Agulhas Current).}
\label{tab:detailed_config_ocean_regional}
\begin{tabular}{lll}
\toprule
{Parameter} & {Code Variable} & {Value} \\
\midrule
\multicolumn{3}{l}{\textit{{1. Overall Architecture}}} \\
Input Channels & \texttt{input\_channels} & 2 (UGOS, VGOS) \\
Output Channels & \texttt{output\_channels} & 2 (UGOS, VGOS) \\
Input/Output Sequence Length & \texttt{input/output\_steps} & 10 days \\
V-Cycle Depth / Number of Levels & \texttt{num\_spatial\_layers} & 4 \\
Spatial Hidden Dimension & \texttt{spatial\_hidden\_dim} & 256\\
Gradient Checkpointing & \texttt{gradient\_checkpointing} & False\\
\midrule
\multicolumn{3}{l}{\textit{{2. Latent Dynamical Core (LDC)}}} \\
Number of LDC Blocks & \texttt{num\_temporal\_layers} & 8\\
LDC Temporal Hidden Dimension & \texttt{temporal\_hidden\_dim} (\texttt{channel\_hid}) & 512\\
LDC MLP Expansion Ratio & \texttt{mlp\_ratio} & 8 \\
\midrule
\multicolumn{3}{l}{\textit{{3. Core Module: Self-Attention}}} \\
Number of Attention Heads & \texttt{num\_heads} & 8\\
QKV Bias & \texttt{qkv\_bias} & False \\
Attention Dropout Rate & \texttt{attn\_drop} &  0\\
Projection Dropout Rate & \texttt{proj\_drop} (in Attention) & 0 \\
LayerScale Init Value & \texttt{init\_value} (for gamma) & 1e-6 \\
\midrule
\multicolumn{3}{l}{\textit{{4. Core Module: Convolution}}} \\
BC Block Activation Function & \texttt{nn.LeakyReLU} & None \\
BC Block Normalization Type & \texttt{nn.GroupNorm} & 2\\
ConvBlock MLP Expansion Ratio & \texttt{mlp\_ratio} & 4\\
ConvBlock/SA Block Activation Fn & \texttt{act\_layer=nn.GELU} & None\\
ConvBlock/SA Block Norm Type & \texttt{norm\_layer=nn.LayerNorm} & None\\
\midrule
\multicolumn{3}{l}{\textit{{5. Training Hyperparameters}}} \\
Optimizer & \texttt{optimizer = optim.Adam} & \\
Learning Rate Scheduler & \texttt{torch.optim.lr\_scheduler.CosineAnnealingLR} &  \\
Max Learning Rate & \texttt{lr=1e-3} & 1e-3 \\
Min Learning Rate & \texttt{eta\_min} & 1e-6\\
Batch Size & \texttt{batch\_size} &  4 per GPU \\
Num Gpus & \texttt{num\_gpus} & 8 \\
Training Duration / Epochs & \texttt{num\_epochs = 1000} & 1000 \\
\midrule
\multicolumn{3}{l}{\textit{{6. Regularization}}} \\
General Dropout Rate & \texttt{drop} & None \\
Drop Path Rate (LDC) & \texttt{drop\_path} & None \\
\midrule
\multicolumn{3}{l}{\textit{{7. Model Scale}}} \\
Total Parameters & Total Parameters & 0.028B \\
Total Training Time (Gulf Stream) & Total Training Time  & 32.2 hours\\
Total Training Time (Agulhas Current) & Total Training Time  & 25.4 hours\\
Total Training Time (Kuroshio Extension) & Total Training Time  & 32.1 hours\\

\bottomrule
\end{tabular}
\end{table}

\begin{table}[h!]
\centering
\footnotesize
\caption{\textbf{Detailed configuration and hyperparameters for the lightweight global TritonCast model used in the zero-shot cross-resolution generalization experiments.} This model is trained on the 0.25° coarse-resolution global ocean current data and serves as the foundational model for subsequent high-resolution inference tests.}
\label{tab:lightweighted_ocean}
\setlength{\tabcolsep}{10pt}
\begin{tabular}{lll}
\toprule
{Parameter} & {Code Variable} & {Value} \\
\midrule
\multicolumn{3}{l}{\textit{{1. Overall Architecture}}} \\
Input Channels & \texttt{input\_channels} & 2 (UGOS, VGOS) \\
Output Channels & \texttt{output\_channels} & 2 (UGOS, VGOS) \\
Input/Output Sequence Length & \texttt{input/output\_steps} & 10 days \\
V-Cycle Depth / Number of Levels & \texttt{num\_spatial\_layers} & 4 \\
Spatial Hidden Dimension & \texttt{spatial\_hidden\_dim} & 64\\
Gradient Checkpointing & \texttt{gradient\_checkpointing} & False\\
\midrule
\multicolumn{3}{l}{\textit{{2. Latent Dynamical Core (LDC)}}} \\
Number of LDC Blocks & \texttt{num\_temporal\_layers} & 4\\
LDC Temporal Hidden Dimension & \texttt{temporal\_hidden\_dim} (\texttt{channel\_hid}) & 256\\
LDC MLP Expansion Ratio & \texttt{mlp\_ratio} & 8 \\
\midrule
\multicolumn{3}{l}{\textit{{3. Core Module: Self-Attention}}} \\
Number of Attention Heads & \texttt{num\_heads} & 4\\
QKV Bias & \texttt{qkv\_bias} & False \\
Attention Dropout Rate & \texttt{attn\_drop} &  0\\
Projection Dropout Rate & \texttt{proj\_drop} (in Attention) & 0 \\
LayerScale Init Value & \texttt{init\_value} (for gamma) & 1e-6 \\
\midrule
\multicolumn{3}{l}{\textit{{4. Core Module: Convolution}}} \\
BC Block Activation Function & \texttt{nn.LeakyReLU} & None \\
BC Block Normalization Type & \texttt{nn.GroupNorm} & 2\\
ConvBlock MLP Expansion Ratio & \texttt{mlp\_ratio} & 4\\
ConvBlock/SA Block Activation Fn & \texttt{act\_layer=nn.GELU} & None\\
ConvBlock/SA Block Norm Type & \texttt{norm\_layer=nn.LayerNorm} & None\\
\midrule
\multicolumn{3}{l}{\textit{{5. Training Hyperparameters}}} \\
Optimizer & \texttt{optimizer = optim.Adam} & \\
Learning Rate Scheduler & \texttt{torch.optim.lr\_scheduler.CosineAnnealingLR} &  \\
Max Learning Rate & \texttt{lr=1e-3} & 1e-3 \\
Min Learning Rate & \texttt{eta\_min} & 1e-6\\
Batch Size & \texttt{batch\_size} &  1 per GPU \\
Num Gpus & \texttt{num\_gpus} & 8 \\
Training Duration / Epochs & \texttt{num\_epochs = 200} & 200 \\
\midrule
\multicolumn{3}{l}{\textit{{6. Regularization}}} \\
General Dropout Rate & \texttt{drop} & None \\
Drop Path Rate (LDC) & \texttt{drop\_path} & None \\
\midrule
\multicolumn{3}{l}{\textit{{7. Model Scale}}} \\
Total Parameters & Total Parameters & 0.002B \\
Total Training Time & Total Training Time  & 36.1 hours\\
\bottomrule
\end{tabular}
\end{table}

\clearpage
\subsubsection{Zero-shot Cross-Resolution Generalization Test}

This experiment aims to investigate whether the TritonCast model learns the scale-invariant underlying physical laws that govern ocean dynamics, rather than merely fitting statistical patterns at a specific resolution. To this end, we design a rigorous "zero-shot" cross-resolution generalization test.

We first train a lightweight, 0.02B parameter TritonCast global model using the global sea surface geostrophic velocity data, which is downsampled to a 0.25° resolution via bilinear interpolation. The training data for this model also spans from 1993 to 2024 and follows the same training, validation, and testing split strategy.

\paragraph{Training and Optimization}: The model is trained on 8 GPUs with a batch size of 1 per GPU, resulting in an effective global batch size of 8. We use the Adam optimizer and a StepLR learning rate scheduler with an initial learning rate of $1e-3$, which is decayed by a factor of $0.2$ every $50$ epochs. The training, aimed at minimizing MSE loss, runs for a total of 1000 epochs.

\paragraph{Evaluation Protocol}: The key to this test lies in the evaluation phase. We take the best model, trained on 0.25° global coarse data, and apply it directly to forecast on the unseen 0.125° high-resolution regional data. No fine-tuning or modifications are made to the model during this inference process. We assess its zero-shot generalization performance by quantitatively and qualitatively evaluating its ability to generate physically realistic and detailed eddy structures at the higher resolution.

The detailed configuration and hyperparameters for the lightweight global model used in this experiment are largely consistent with those in \textbf{Tab.~\ref{tab:lightweighted_ocean}}.

\subsection{Experimental Setup for Turbulence}
\label{appendix:turbulence}

To fundamentally test TritonCast's ability to handle multi-scale dynamics and suppress the spectral bias that plagues many AI models, we evaluate it on the challenging benchmark of two-dimensional Kolmogorov turbulence. This experiment serves as a canonical test case in physics to assess whether the model can accurately capture the energy cascade and preserve the physical fidelity of fine-scale vortex structures in long-term autoregressive rollouts.

The dataset used for this task is a high-fidelity numerical simulation of decaying turbulence, initialized with McWilliams conditions at a spatial resolution of $128 \times 128$ and a Reynolds number of Re=5000. The model is trained on a single-step prediction task, where it takes the vorticity field at one timestep as input to predict the field at the subsequent timestep.

\paragraph{Training and Optimization}: A TritonCast model with approximately 0.02B parameters is configured for this task. The model is trained using the Distributed Data Parallel (DDP) strategy. We use a batch size of 20 and the Adam optimizer with a fixed learning rate of 1e-3. The model is trained for 500 epochs to minimize the Mean Squared Error (MSE) between the predicted and ground truth vorticity fields. All experiments are conducted with a fixed random seed of 42 to ensure reproducibility.

The following \textbf{Tab.~\ref{tab:detailed_config_turbulence}} provides a comprehensive breakdown of the specific model configuration and hyperparameters used for the turbulence forecasting experiment.

\begin{table}[h!]
\centering
\footnotesize
\caption{\textbf{Detailed model configuration and hyperparameters for TritonCast in the 2D turbulence forecasting task.}}
\label{tab:detailed_config_turbulence}
\setlength{\tabcolsep}{20pt}
\begin{tabular}{lll}
\toprule
{Parameter} & {Code Variable} & {Value} \\
\midrule
\multicolumn{3}{l}{\textit{{1. Overall Architecture}}} \\
Input Channels (Vorticity) & \texttt{input\_channel} & 1 \\
Output Channels (Vorticity) & \texttt{output\_channels} & 1 \\
Input/Output Sequence Length & \texttt{input/output\_length} & 1 \\
V-Cycle Depth / Number of Levels & \texttt{num\_spatial\_layers} & 4 \\
Spatial Hidden Dimension & \texttt{spatial\_hidden\_dim} & 256\\
\midrule
\multicolumn{3}{l}{\textit{{2. Latent Dynamical Core (LDC)}}} \\
Number of LDC Blocks & \texttt{num\_temporal\_layers} & 8\\
LDC Temporal Hidden Dimension & \texttt{temporal\_hidden\_dim} & 512\\
LDC MLP Expansion Ratio & \texttt{mlp\_ratio} & 8 \\
\midrule
\multicolumn{3}{l}{\textit{{3. Core Module: Self-Attention}}} \\
Number of Attention Heads & \texttt{num\_heads} & 4\\
QKV Bias & \texttt{qkv\_bias} & False \\
Attention Dropout Rate & \texttt{attn\_drop} &  0\\
Projection Dropout Rate & \texttt{proj\_drop} (in Attention) & 0 \\
LayerScale Init Value & \texttt{init\_value} (for gamma) & 1e-6 \\
\midrule
\multicolumn{3}{l}{\textit{{4. Core Module: Convolution}}} \\
BC Block Activation Function & \texttt{nn.LeakyReLU} & None \\
BC Block Normalization Type & \texttt{nn.GroupNorm} & 2\\
ConvBlock MLP Expansion Ratio & \texttt{mlp\_ratio} & 4\\
ConvBlock/SA Block Activation Fn & \texttt{act\_layer=nn.GELU} & None\\
ConvBlock/SA Block Norm Type & \texttt{norm\_layer=nn.LayerNorm} & None\\
\midrule
\multicolumn{3}{l}{\textit{{5. Training Hyperparameters}}} \\
Optimizer & \texttt{optimizer = optim.Adam} & \\
Learning Rate Scheduler & \texttt{torch.optim.lr\_scheduler.StepLR} &  None \\
Learning Rate & \texttt{learning\_rate} & 1e-3 \\
Batch Size & \texttt{batch\_size} &  20 \\
Training Duration / Epochs & \texttt{num\_epochs} & 500 \\
\midrule
\multicolumn{3}{l}{\textit{{6. Regularization}}} \\
General Dropout Rate & \texttt{drop} & None \\
Drop Path Rate (LDC) & \texttt{drop\_path} & None \\
\midrule
\multicolumn{3}{l}{\textit{{7. Model Scale}}} \\
Total Parameters & Total Parameters &  0.02B \\
Total Training Time & Total Training Time  & 21.7 hours\\
\bottomrule
\end{tabular}
\end{table}

\subsection{Baseline Models: Implementation and Evaluation Details}
To ensure that our evaluation of TritonCast's performance is comprehensive, fair, and transparent, we systematically compare it against a suite of state-of-the-art (SOTA) baseline models spanning various domains and tasks. The selected baselines represent the gold standards and technological frontiers in the field and can be broadly categorized as follows:

\begin{itemize}
    \item \textbf{Traditional Physical Numerical Models}: Represented by the \textbf{IFS-HRES}~\cite{rasp2024weatherbench} from the European Centre for Medium-Range Weather Forecasts (ECMWF). This operational system is the global gold standard for weather forecasting and provides a critical benchmark for the performance of our AI model.

    \item \textbf{Hybrid Physics-ML Models}: Represented by \texttt{NeuralGCM}~\cite{kochkov2024neural}. This class of models attempts to integrate machine learning components with traditional physical cores or principles, representing an important modeling paradigm distinct from purely data-driven approaches. We evaluate both its deterministic and stochastic versions in our long-term stability tests.

    \item \textbf{Purely Data-Driven AI Models}: This category includes a diverse range of deep learning architectures that have recently achieved breakthroughs in Earth science. The goal is to compare TritonCast with its direct AI competitors, including:
    \begin{enumerate}
        \item \textit{Graph Neural Networks}: Such as Google DeepMind's \texttt{GraphCast}~\cite{lam2023learning}, which demonstrates exceptional performance in weather forecasting.
        \item \textit{Transformer Architectures}: Such as Huawei Cloud's \texttt{Pangu-Weather} and models specialized for ocean forecasting like \texttt{WenHai}~\cite{cui2025forecasting} and \texttt{ORCA\_DL}~\cite{guo2025data}, representing SOTA attention-based models.
        \item \textit{Neural Operators}: Such as \texttt{FNO}~\cite{li2020fourier}, \texttt{CNO}~\cite{raonic2023convolutional}, and \texttt{MGNO}~\cite{he2024mgno}. These models excel at learning solutions to Partial Differential Equations (PDEs) and are particularly well-suited for problems in theoretical physics like turbulence.
        \item \textit{Diffusion Models}: Such as \texttt{GenCast}~\cite{price2025probabilistic} and \texttt{PDE-Refiner}~\cite{lippe2023pde}, which represent the forefront of generative modeling applications in Earth science.
    \end{enumerate}
\end{itemize}

To guarantee the rigor and fairness of all comparisons, we follow a strict implementation and evaluation protocol. Our core principle is to adhere as closely as possible to the original research while ensuring, where necessary, that all models are evaluated under unified and equitable conditions. Specifically, we adopt a dual-track strategy, which is detailed in the following subsections: \ding{182} first, we prioritize the use of officially released, open-source pre-trained model weights; \ding{183} second, for models where official weights are not available, we meticulously reproduce them according to the original papers and open-source code.

\subsubsection{Use of Official Open-source Weights}
\label{appendix:Open_source_Weights}
To ensure a fair and direct comparison with the original research, we prioritize the use of officially released and open-source pre-trained model weights. This approach not only ensures that we evaluate the optimal model performance presented by the original authors but also lends the highest credibility to our results. In this section, we provide a detailed list of the official weights and their corresponding configurations used in the different experimental scenarios.

\paragraph{\ding{182} Medium-range Atmospheric Forecasting.}
To comprehensively evaluate the forecast accuracy of TritonCast in the standard medium-range weather forecasting task (1-to-10 days), we employ the internationally recognized \href{https://weatherbench2.readthedocs.io/en/latest/data-guide.html#forecast-datasets}{WeatherBench 2 benchmark}. This platform provides a rigorous, standardized evaluation protocol and a comprehensive set of metrics for data-driven global weather models. Following this protocol, we perform a direct comparison of TritonCast against a suite of state-of-the-art models. These baselines include not only leading AI systems such as Pangu-Weather and GraphCast but also the operational gold standard, IFS-HRES from ECMWF.

\paragraph{\ding{183} Long-term Atmospheric Forecasting.}
In the demanding year-long autoregressive forecasting task, we use the official weights for several baseline models. All inference is performed on NVIDIA H200 GPUs. To ensure the accuracy of our evaluation, we conduct a detailed verification of the data structures by analyzing the NetCDF/HDF5 output files generated by each model, using standard Python libraries such as \texttt{xarray} and \texttt{h5py}.

\begin{itemize}
    \item \textbf{GenCast} We use the officially released \href{https://huggingface.co/kashif/gencast/blob/main/params/GenCast%201p0deg%20Mini%20%3C2019.npz}{\texttt{GenCast\_1p0deg\_<2019.npz}} weights. Our data structure analysis confirms that the model operates on a 1.0° resolution (181×360 grid) with 13 standard pressure levels. It uses a \textbf{12-hour} forecast step, thus requiring 730 inference steps to complete a one-year forecast. The presence of a \texttt{sample} dimension (size 50) also reflects its nature as a probabilistic (ensemble) forecast model.

\begin{tcolorbox}[
    colback=black!5!white, 
    colframe=black!75!black, 
    sharp corners, 
    boxsep=2mm,
    fontupper=\footnotesize]
\begin{verbatim}
--- Gencast Inference Results Info ---
<xarray.Dataset>
Dimensions:   (batch: 1, time: 730, lat: 181, lon: 360, level: 13)
Coordinates:
  * lon       (lon) float64 0.0 1.0 2.0 ... 357.0 358.0 359.0
  * lat       (lat) float64 -90.0 -89.0 -88.0 ... 88.0 89.0 90.0
  * level     (level) int32 50 100 150 200 ... 700 850 925 1000
  * sample    (sample) int64 0 1
  * time      (time) datetime64[ns] 2020-01-02T12:00:00 ...
Data variables:
    10m_u_component_of_wind   (sample, batch, time, lat, lon) float32 ...
    10m_v_component_of_wind   (sample, batch, time, lat, lon) float32 ...
    2m_temperature            (sample, batch, time, lat, lon) float32 ...
    geopotential              (sample, batch, time, level, lat, lon) float32 ...
    mean_sea_level_pressure   (sample, batch, time, lat, lon) float32 ...
    sea_surface_temperature   (sample, batch, time, lat, lon) float32 ...
    specific_humidity         (sample, batch, time, level, lat, lon) float32 ...
    temperature               (sample, batch, time, level, lat, lon) float32 ...
    total_precipitation_12hr  (sample, batch, time, lat, lon) float32 ...
    u_component_of_wind       (sample, batch, time, level, lat, lon) float32 ...
    v_component_of_wind       (sample, batch, time, level, lat, lon) float32 ...
    vertical_velocity         (sample, batch, time, level, lat, lon) float32 ...
    ...
\end{verbatim}
\end{tcolorbox}

\clearpage
\item \textbf{GraphCast}

    We employ the official \href{https://huggingface.co/shermansiu/dm_graphcast_small/blob/main/GraphCast_small%20-%20ERA5%201979-2015%20-%20resolution%201.0%20-%20pressure%20levels%2013%20-%20mesh%202to5%20-%20precipitation%20input%20and%20output.npz}{\texttt{GraphCast\_small}} weights. The model also operates at 1.0° resolution (181×360) and with 13 pressure levels. However, it uses a finer \textbf{6-hour} forecast step, requiring a total of 1460 inference steps for a full year-long autoregressive forecast.

\begin{tcolorbox}[
    colback=black!5!white, 
    colframe=black!75!black, 
    sharp corners, 
    boxsep=2mm,
    fontupper=\footnotesize]
\begin{verbatim}
--- Graphcast Inference Results Info ---
<xarray.Dataset>
Dimensions:                  (time: 1460, batch: 1, lat: 181, lon: 360,
                              level: 13)
Coordinates:
  * lon                      (lon) float32 0.0 1.0 2.0 3.0 ... 357.0 358.0 359.0
  * lat                      (lat) float32 -90.0 -89.0 -88.0 ... 88.0 89.0 90.0
  * batch                    (batch) int64 0
  * level                    (level) int32 50 100 150 200 ... 700 850 925 1000
  * time                     (time) datetime64[ns] 2020-01-02T06:00:00 ... 20...
Data variables:
    10m_u_component_of_wind  (time, batch, lat, lon) float32 ...
    10m_v_component_of_wind  (time, batch, lat, lon) float32 ...
    2m_temperature           (time, batch, lat, lon) float32 ...
    geopotential             (time, batch, level, lat, lon) float32 ...
    mean_sea_level_pressure  (time, batch, lat, lon) float32 ...
    specific_humidity        (time, batch, level, lat, lon) float32 ...
    temperature              (time, batch, level, lat, lon) float32 ...
    total_precipitation_6hr  (time, batch, lat, lon) float32 ...
    u_component_of_wind      (time, batch, level, lat, lon) float32 ...
    v_component_of_wind      (time, batch, level, lat, lon) float32 ...
    vertical_velocity        (time, batch, level, lat, lon) float32 ...
\end{verbatim}
\end{tcolorbox}

    \clearpage
    \item \textbf{NeuralGCM (Deterministic \& Stochastic Versions)}

    We evaluate both official versions of NeuralGCM. The \href{https://neuralgcm.readthedocs.io/en/latest/checkpoints.html}{weight files} are \texttt{deterministic\_1\_4\_deg.pkl} and \texttt{stochastic\_1\_4\_deg.pkl}, respectively. Both models operate on a distinct, lower-resolution ~1.4° Gaussian grid (128×256) but feature a much richer vertical structure with \textbf{37 levels}. Following the standard setup, we use a \textbf{24-hour} forecast step and provide the sea surface temperature (SST) and sea ice cover from the initial time step as fixed boundary forcings throughout the simulation.

\begin{tcolorbox}[
    colback=black!5!white, 
    colframe=black!75!black, 
    sharp corners, 
    boxsep=2mm,
    fontupper=\footnotesize]
\begin{verbatim}
--- NeuralGCM-Stochastic Inference Results Info ---
<xarray.Dataset>
Dimensions:                             (time: 365, level: 37, longitude: 256,
                                          latitude: 128)
Coordinates:
  * longitude                           (longitude) float64 0.0 1.406 ... 358.6
  * latitude                            (latitude) float64 -88.93 ... 88.93
  * level                               (level) int64 1 2 3 5 ... 950 975 1000
  * time                                (time) datetime64[ns] 2020-01-01T12:...
Data variables:
 specific_cloud_ice_water_content    (time, level, longitude, latitude) float32 ...
 v_component_of_wind                 (time, level, longitude, latitude) float32 ...
 sim_time                            (time) float32 ...
 geopotential                        (time, level, longitude, latitude) float32 ...
 specific_humidity                   (time, level, longitude, latitude) float32 ...
 u_component_of_wind                 (time, level, longitude, latitude) float32 ...
 specific_cloud_liquid_water_content (time, level, longitude, latitude) float32 ...
 temperature                         (time, level, longitude, latitude) float32 ...
Attributes:
    longitude_wavenumbers:     128
    total_wavenumbers:         129
    longitude_nodes:           256
    latitude_nodes:            128
    latitude_spacing:          gauss
    longitude_offset:          0.0
    radius:                    1.0
    spherical_harmonics_impl:  RealSphericalHarmonicsWithZeroImag
    spmd_mesh:                 
    centers:                   [   1    2    3    5    7   10   20   30   50 ...
    horizontal_grid_type:      Grid
    vertical_grid_type:        PressureCoordinates
\end{verbatim}
\end{tcolorbox}

\clearpage
\begin{tcolorbox}[
    colback=black!5!white, 
    colframe=black!75!black, 
    sharp corners, 
    boxsep=2mm,
    fontupper=\footnotesize]
\begin{verbatim}
--- NeuralGCM-Deterministic Inference Results Info ---
<xarray.Dataset>
Dimensions:                              (time: 365, level: 37, longitude: 256,
                                          latitude: 128)
Coordinates:
  * longitude                            (longitude) float64 0.0 1.406 ... 358.6
  * latitude                             (latitude) float64 -88.93 ... 88.93
  * level                                (level) int64 1 2 3 5 ... 950 975 1000
  * time                                 (time) datetime64[ns] 2020-01-01T12:...
Data variables:
 geopotential                        (time, level, longitude, latitude) float32 ...
 v_component_of_wind                 (time, level, longitude, latitude) float32 ...
 specific_humidity                   (time, level, longitude, latitude) float32 ...
 temperature                         (time, level, longitude, latitude) float32 ...
 sim_time                            (time) float32 ...
 u_component_of_wind                 (time, level, longitude, latitude) float32 ...
 specific_cloud_ice_water_content    (time, level, longitude, latitude) float32 ...
 specific_cloud_liquid_water_content (time, level, longitude, latitude) float32 ...
Attributes:
    longitude_wavenumbers:     128
    total_wavenumbers:         129
    longitude_nodes:           256
    latitude_nodes:            128
    latitude_spacing:          gauss
    longitude_offset:          0.0
    radius:                    1.0
    spherical_harmonics_impl:  RealSphericalHarmonicsWithZeroImag
    spmd_mesh:                 
    centers:                   [   1    2    3    5    7   10   20   30   50 ...
    horizontal_grid_type:      Grid
    vertical_grid_type:        PressureCoordinates
\end{verbatim}
\end{tcolorbox}

\clearpage
\item \textbf{TritonCast \& Ground Truth}
For comparison, the output format for our TritonCast model and the ground truth data is standardized to 1.0° resolution (180×360), a \textbf{24-hour} time step, and 13 standard pressure levels. This ensures that our model's output is directly comparable with major baselines on key dimensions.
    
\begin{tcolorbox}[
    colback=black!5!white, 
    colframe=black!75!black, 
    sharp corners, 
    boxsep=2mm,
    fontupper=\footnotesize]
\begin{verbatim}
<xarray.Dataset>
Dimensions:      (time: 365, level: 13, latitude: 180, longitude: 360)
Coordinates:
  * time         (time) datetime64[ns] 2020-01-02T12:00:00 ... 2020-12-30T12:...
  * level        (level) int64 50 100 150 200 250 300 ... 600 700 850 925 1000
  * latitude     (latitude) float64 89.5 88.5 87.5 86.5 ... -87.5 -88.5 -89.5
  * longitude    (longitude) float64 0.0 1.0 2.0 3.0 ... 356.0 357.0 358.0 359.0
Data variables:
    temperature  (time, level, latitude, longitude) float32 201.0 ... 238.0

--- Ground Truth Dataset Info ---
<xarray.Dataset>
Dimensions:      (time: 365, level: 13, latitude: 180, longitude: 360)
Coordinates:
  * time         (time) datetime64[ns] 2020-01-02T12:00:00 ... 2020-12-30T12:...
  * level        (level) int64 50 100 150 200 250 300 ... 600 700 850 925 1000
  * latitude     (latitude) float64 89.5 88.5 87.5 86.5 ... -87.5 -88.5 -89.5
  * longitude    (longitude) float64 0.0 1.0 2.0 3.0 ... 356.0 357.0 358.0 359.0
Data variables:
    temperature  (time, level, latitude, longitude) float32 193.5 ... 261.1
==================================================
\end{verbatim}
\end{tcolorbox}
    
\end{itemize}

\paragraph{Configuration Summary}
\textbf{Tab.~\ref{tab:baseline_configs_long_range}} provides a concise summary and direct comparison of the key configurations for the baseline models used.

\begin{table}[h!]
\centering
\caption{\textbf{Key configurations for official open-source models in long-term atmospheric forecasting.}}
\label{tab:baseline_configs_long_range}
\footnotesize
\begin{tabularx}{\textwidth}{lllll}
\toprule
\textbf{Model} & \textbf{Official Weight File (Partial)} & \textbf{Spatial Res.} & \textbf{Time Step} & \textbf{Levels} \\
\midrule
GenCast & \texttt{GenCast\_1p0deg\_<2019.npz} & \SI{1.0}{\degree} (181×360) & 12 hours & 13 \\
\addlinespace
GraphCast (Small) & \texttt{GraphCast\_small...1.0...npz} & \SI{1.0}{\degree} (181×360) & 6 hours & 13 \\
\addlinespace
NeuralGCM (Stochastic) & \texttt{stochastic\_1\_4\_deg.pkl} & $\sim$\SI{1.4}{\degree} (128×256) & 24 hours & 37 \\
\addlinespace
NeuralGCM (Deterministic) & \texttt{deterministic\_1\_4\_deg.pkl} & $\sim$\SI{1.4}{\degree} (128×256) & 24 hours & 37 \\
\addlinespace
\textbf{TritonCast (Lightweighted)} & \texttt{triton\_weather\_1.0\_model.pth} & \SI{1.0}{\degree} (180×360) & 24 hours & 13 \\
\bottomrule
\end{tabularx}
\end{table}

\paragraph{\ding{184} Wenhai Model For Global Ocean Simulation.} Due to the model of WenHai is close-sourced, we employ the official \href{https://drive.google.com/file/d/1BMfoS4KJVXr9isth9PQgenroCeRqSa2l/view?usp=drive_link}{\texttt{WenHai.onnx}} weights to produce ocean simulation under the same ICs with our \method{}. The inference is performed on a NVIDIA A100 40GB GPU. Different from other baseline models and our \method{}, WenHai requires more forcing variables, which limits its real-world application in ocean forecasting task. The model operates at 1/12° resolution (2041×4320) and with 23 depth levels. Then, we use bilinear interpolation to interpolate the results of WenHai to 0.25°, ensuring consistency with the spatial resolution of other methods.

\subsubsection{Protocol for Model Reproducibility and Training}

\paragraph{\ding{182} Baseline Models For Global Ocean Simulation.}
For the global ocean simulation task at $0.25\degree$ resolution, we reproduce and retrain three representative baseline models, \textbf{FourCastNet}, \textbf{AI-GOMS}, and \textbf{ORCA\_DL}, based on their official open-source code to ensure a direct and fair performance comparison with TritonCast. This retraining step is crucial for maintaining the integrity of our comparison, as the official pre-trained weights for these models are not directly applicable to this specific experimental setup.

We follow a rigorous reproduction protocol centered on the principle of a \textbf{"head-to-head competition,"} ensuring all models are trained and evaluated under identical conditions All training is performed on \textbf{48 NVIDIA A800 GPUs} using the Distributed Data Parallel (DDP) strategy, and all models use the same dataset splits for training (1993-2017), validation (2018-2019), and testing (2020). All reproduced baseline models and our model are trained using the same experimental setup with pre-training and fine-tuning stages. More training details refer to Supplementary~\ref{section:global_ocean_simulation}.

\begin{itemize}
    \item \textbf{FourCastNet}: We reproduce \href{https://github.com/NVlabs/FourCastNet}{its Adaptive Fourier Neural Operator-based Transformer architecture}, resulting in a model with approximately \textbf{0.03B} 
    parameters.

    \item \textbf{AI-GOMS}: We reproduce \href{https://github.com/xiangyanfei212/Ocean_AI_model}{its Adaptive Fourier Neural Operator-based (AFNO) Masked Autoencoders architecture (MAE)}, resulting in a model with approximately \textbf{0.02B} parameters.
    
    \item \textbf{ORCA\_DL}: We reproduce \href{https://github.com/OpenEarthLab/ORCA-DL}{its Swin Transformer-based U-Net architecture}, resulting in a model with approximately \textbf{0.03B} parameters. A key detail is that this model has an \textbf{inherent architectural limitation that supports only single-step prediction}. Therefore, we train it to predict the next single time step from a single input time step.

\end{itemize}

These approaches establishe a solid foundation for a direct comparison against \method{}. The configurations for these reproduced models are detailed in \textbf{Tab.~\ref{tab:repro_fourcastnet_config_global_ocean_simulation}}, \textbf{Tab.~\ref{tab:repro_aigoms_config_global_ocean_simulation}}, and \textbf{Tab.~\ref{tab:repro_orca_dl_config_global_ocean_simulation}}.

To visually illustrate the trade-offs among architectural capability, parameter efficiency, and training cost, \textbf{Tab.~\ref{tab:baseline_summary_ocean_simulation}} summarizes the key metrics for TritonCast against the reproduced baselines.

\begin{table}[h!]
\centering
\caption{\textbf{Configuration details for the reproduced FourCastNet model in global ocean simulation task.}}
\label{tab:repro_fourcastnet_config_global_ocean_simulation}
\setlength{\tabcolsep}{8pt}
\footnotesize
\begin{tabular}{lll}
\toprule
\textbf{Category} & \textbf{Parameter} & \textbf{Value} \\
\midrule
\multirow{8}{*}{\textbf{1. Architecture}} & Core Architecture & Adaptive Fourier Neural Operator-based Transformer \\
& Task Paradigm & \textbf{Single-step prediction} \\
& Input/Output Sequence Length & 1 day / 1 day \\
& Input/Output Channels & 97 (1 step × 93 vars) \\
& Embedding Dimension & 832 \\
& Patch Size & 8×8\\
& Depths / Num Blocks & 4 / 16 \\
\midrule
\multirow{7}{*}{\textbf{2. Training Hyperparameters}} & Optimizer & Adam \\
& Max Learning Rate & 1e-3 \\
& Min Learning Rate & 0 \\
& Learning Rate Scheduler & CosineAnnealingLR (T\_max=200) \\
& Batch Size & 1 per GPU \\
& GPU Configuration & 48 × NVIDIA A800 \\
& Training Epochs & 200 \\
\midrule
\multirow{2}{*}{\textbf{3. Data \& Task}} & Dataset & 0.25° Global CMEMS \\
& Data Timestep Sampling & 1 sample every 1 days \\
\midrule
\multirow{2}{*}{\textbf{4. Model Scale \& Cost}} & Total Parameters & ~0.03B (approx. 32M) \\
& Training Time & ~77 hours \\
\bottomrule
\end{tabular}
\end{table}

\begin{table}[h!]
\centering
\caption{\textbf{Configuration details for the reproduced AI-GOMS model in global ocean simulation task.}}
\label{tab:repro_aigoms_config_global_ocean_simulation}
\setlength{\tabcolsep}{6pt}
\footnotesize
\begin{tabular}{lll}
\toprule
\textbf{Category} & \textbf{Parameter} & \textbf{Value} \\
\midrule
\multirow{8}{*}{\textbf{1. Architecture}} & Core Architecture & AFNO-based MAE \\
& Task Paradigm & \textbf{Single-step prediction} \\
& Input/Output Sequence Length & 1 day / 1 day \\
& Input/Output Channels & 97 (1 step × 93 vars) \\
& Embedding Dimension / Decoder Embed Dimension & 256 / 512 \\
& Patch Size & 16×16\\
& Depths / Num Blocks & 4 / 8 \\
\midrule
\multirow{7}{*}{\textbf{2. Training Hyperparameters}} & Optimizer & Adam \\
& Max Learning Rate & 1e-3 \\
& Min Learning Rate & 0 \\
& Learning Rate Scheduler & CosineAnnealingLR (T\_max=200) \\
& Batch Size & 1 per GPU \\
& GPU Configuration & 48 × NVIDIA A800 \\
& Training Epochs & 200 \\
\midrule
\multirow{2}{*}{\textbf{3. Data \& Task}} & Dataset & 0.25° Global CMEMS \\
& Data Timestep Sampling & 1 sample every 1 days \\
\midrule
\multirow{2}{*}{\textbf{4. Model Scale \& Cost}} & Total Parameters & ~0.02B (approx. 29M) \\
& Training Time & ~75 hours \\
\bottomrule
\end{tabular}
\end{table}

\begin{table}[h!]
\centering
\caption{\textbf{Configuration details for the reproduced ORCA\_DL model in global ocean simulation task.}}
\label{tab:repro_orca_dl_config_global_ocean_simulation}
\setlength{\tabcolsep}{13pt}
\footnotesize
\begin{tabular}{lll}
\toprule
\textbf{Category} & \textbf{Parameter} & \textbf{Value} \\
\midrule
\multirow{8}{*}{\textbf{1. Architecture}} & Core Architecture & Swin Transformer U-Net \\
& Task Paradigm & \textbf{Single-step prediction (model limitation)} \\
& Input/Output Sequence Length & 1 day / 1 day \\
& Input/Output Channels & 97 (1 step × 93 vars) \\
& Embedding Dimension & 96 \\
& Patch / Window Size & 12×18 / 8×15 \\
& Encoder Depths / Heads & [3, 3, 3] / [3, 6, 12] \\
& Mixture-of-Experts (MoE) & True \\
\midrule
\multirow{7}{*}{\textbf{2. Training Hyperparameters}} & Optimizer & Adam \\
& Max Learning Rate & 1e-3 \\
& Min Learning Rate & 0 \\
& Learning Rate Scheduler & CosineAnnealingLR (T\_max=200) \\
& Batch Size & 1 per GPU \\
& GPU Configuration & 48 × NVIDIA A800 \\
& Training Epochs & 200 \\
\midrule
\multirow{2}{*}{\textbf{3. Data \& Task}} & Dataset & 0.25° Global CMEMS \\
& Data Timestep Sampling & 1 sample every 1 days \\
\midrule
\multirow{2}{*}{\textbf{4. Model Scale \& Cost}} & Total Parameters & ~0.03B (approx. 34M) \\
& Training Time & ~58 hours \\
\bottomrule
\end{tabular}
\end{table}

\begin{table}[h!]
\centering
\footnotesize
\caption{\textbf{Summary comparison of models for Global Ocean Simulation ($0.25 \degree$).}}
\label{tab:baseline_summary_ocean_simulation}
\setlength{\tabcolsep}{2pt}
\begin{tabular}{lcccc}
\toprule
\textbf{Model} & \textbf{Parameters} & \textbf{Task Paradigm} & \textbf{Training Time (Pre-training and Fine-tuning)} & \textbf{Core Architecture} \\
\midrule
\textbf{TritonCast (Ours)} & \textbf{~0.02B (~28M)} & \textbf{Single-step} & \textbf{400 hrs} & \textbf{Hierarchical V-Cycle} \\
FourCastNet & ~0.03B (~32M) & {Single-step} & ~77 hrs & AFNO-based Transformer \\
AI-GOMS & ~0.02B (~29M) & {Single-step} & ~75 hrs & AFNO-based MAE \\
ORCA\_DL & ~0.03B (~34M) & {Single-step} & ~58 hrs & Swin Transformer U-Net \\
\bottomrule
\end{tabular}
\end{table}

\paragraph{\ding{183} Baseline Models For Global Ocean Current Forecasting ($0.25 \degree$).}

For the global ocean current forecasting task at $0.25\degree$ resolution, we reproduce and retrain two representative baseline models, \textbf{PDE-Refiner} and \textbf{ORCA\_DL}, based on their official open-source code to ensure a direct and fair performance comparison with TritonCast. It is important to emphasize that this is purely a long-term \textbf{forecasting} task: the models perform autoregressive rollouts based only on the \texttt{UGOS} and \texttt{VGOS} variables, \textbf{without reliance on any external variable forcing} (e.g., atmospheric winds). This retraining step is crucial for maintaining the integrity of our comparison, as the official pre-trained weights for these models are not directly applicable to this specific experimental setup.

We follow a rigorous reproduction protocol centered on the principle of a \textbf{"head-to-head competition,"} ensuring all models are trained and evaluated under identical conditions. All training is performed on \textbf{8 NVIDIA A100 GPUs} using the Distributed Data Parallel (DDP) strategy, and all models use the same dataset splits for training (1993-2018), validation (2019), and testing (2020-2024).

\begin{itemize}
    \item \textbf{PDE-Refiner}: We reproduce the \href{https://github.com/pdearena/pdearena}{FourierUnet} backbone used in its \texttt{PDE-Arena} benchmark. The model has approximately \textbf{0.04B} parameters. Following its design, we train it as a \textbf{sequence-to-sequence} model, which takes 10 time steps as input to predict the next 10 time steps. The model is trained with the Adam optimizer and a cosine-annealed learning rate starting from \texttt{1e-3}. The training for the first 200 epochs takes approximately 47 hours.

    \item \textbf{ORCA\_DL}: We reproduce \href{https://github.com/OpenEarthLab/ORCA-DL}{its Swin Transformer-based U-Net architecture}, resulting in a model with approximately \textbf{0.16B} parameters. A key detail is that this model has an \textbf{inherent architectural limitation that supports only single-step prediction}. Therefore, we train it to predict the next single time step from a single input time step. It is also trained with the Adam optimizer and an initial learning rate of \texttt{1e-3} with a cosine annealing schedule. Despite its larger size, training the first 200 epochs takes only about 3.1 hours due to its single-step nature.
\end{itemize}

These approaches establishe a solid foundation for a direct comparison against \method{}. The configurations for these reproduced models are detailed in \textbf{Tab.~\ref{tab:repro_pde_refiner_config_en}} and \textbf{Tab.~\ref{tab:repro_orca_dl_config_en}}.

\begin{table}[h!]
\centering
\caption{\textbf{Configuration details for the reproduced PDE-Refiner (FourierUnet backbone).}}
\label{tab:repro_pde_refiner_config_en}
\setlength{\tabcolsep}{15pt}
\footnotesize
\begin{tabular}{lll}
\toprule
\textbf{Category} & \textbf{Parameter} & \textbf{Value} \\
\midrule
\multirow{6}{*}{\textbf{1. Architecture}} & Core Architecture & FourierUnet \\
& Input/Output Sequence Length & 10 days / 10 days \\
& Input/Output Channels & 20 (10 steps × 2 vars) \\
& Hidden Channels & 32 \\
& Fourier Modes (modes1, modes2) & 8, 8 \\
& Fourier Layers Depth & 1 \\
\midrule
\multirow{7}{*}{\textbf{2. Training Hyperparameters}} & Optimizer & Adam \\
& Max Learning Rate & 1e-3 \\
& Min Learning Rate & 0 \\
& Learning Rate Scheduler & CosineAnnealingLR (T\_max=200) \\
& Batch Size & 1 per GPU \\
& GPU Configuration & 8 × NVIDIA A100 \\
& Training Epochs & 200 \\
\midrule
\multirow{2}{*}{\textbf{3. Data \& Task}} & Dataset & 0.25° Global CMEMS (ugos, vgos) \\
& Data Timestep Sampling & 1 sample every 3 days \\
\midrule
\multirow{2}{*}{\textbf{4. Model Scale \& Cost}} & Total Parameters & ~0.04B (approx. 40M) \\
& Training Time & ~47 hours \\
\bottomrule
\end{tabular}
\end{table}

\begin{table}[h!]
\centering
\caption{\textbf{Configuration details for the reproduced ORCA\_DL model.}}
\label{tab:repro_orca_dl_config_en}
\setlength{\tabcolsep}{13pt}
\footnotesize
\begin{tabular}{lll}
\toprule
\textbf{Category} & \textbf{Parameter} & \textbf{Value} \\
\midrule
\multirow{8}{*}{\textbf{1. Architecture}} & Core Architecture & Swin Transformer U-Net \\
& Task Paradigm & \textbf{Single-step prediction (model limitation)} \\
& Input/Output Sequence Length & 1 day / 1 day \\
& Input/Output Channels & 2 (1 step × 2 vars) \\
& Embedding Dimension & 96 \\
& Patch / Window Size & 4×4 / 6×12 \\
& Encoder Depths / Heads & [2, 2, 2] / [3, 6, 12] \\
& Mixture-of-Experts (MoE) & True \\
\midrule
\multirow{7}{*}{\textbf{2. Training Hyperparameters}} & Optimizer & Adam \\
& Max Learning Rate & 1e-3 \\
& Min Learning Rate & 1e-6 \\
& Learning Rate Scheduler & CosineAnnealingLR (T\_max=500) \\
& Batch Size & 1 per GPU \\
& GPU Configuration & 8 × NVIDIA A100 \\
& Training Epochs & 200 \\
\midrule
\multirow{2}{*}{\textbf{3. Data \& Task}} & Dataset & 0.25° Global CMEMS (ugos, vgos) \\
& Data Timestep Sampling & 1 sample every 3 days \\
\midrule
\multirow{2}{*}{\textbf{4. Model Scale \& Cost}} & Total Parameters & ~0.16B (approx. 160M) \\
& Training Time & ~3.1 hours \\
\bottomrule
\end{tabular}
\end{table}

To visually illustrate the trade-offs among architectural capability, parameter efficiency, and training cost, \textbf{Tab.~\ref{tab:baseline_summary_en}} summarizes the key metrics for TritonCast against the reproduced baselines.

\begin{table}[h!]
\centering
\footnotesize
\caption{\textbf{Summary comparison of models for Global Ocean Current Forecasting ($0.25 \degree$).}}
\label{tab:baseline_summary_en}
\setlength{\tabcolsep}{8pt}
\begin{tabular}{lcccc}
\toprule
\textbf{Model} & \textbf{Parameters} & \textbf{Task Paradigm} & \textbf{Training Time (200 Epochs)} & \textbf{Core Architecture} \\
\midrule
\textbf{TritonCast (Ours)} & \textbf{~0.003B (~3M)} & \textbf{Seq-to-Seq} & \textbf{36.1 hrs} & \textbf{Hierarchical V-Cycle} \\
PDE-Refiner & ~0.04B (~40M) & Seq-to-Seq & ~47 hrs & FourierUnet \\
ORCA\_DL & ~0.16B (~160M) & {Single-step} & ~3.1 hrs & Swin Transformer U-Net \\
\bottomrule
\end{tabular}
\end{table}

\paragraph{\ding{184} High-resolution Regional Ocean Current Forecasting ($0.125 \degree$).}

To rigorously evaluate TritonCast's capability in capturing fine-scale ocean dynamics, we select the highly challenging Kuroshio Extension region at $0.125\degree$ resolution as a testbed. For this scenario, we focus on reproducing and evaluating three representative state-of-the-art models: \textbf{SimVP}, \textbf{U-Net}, and \textbf{DiT}.

To ensure an absolutely fair comparison, all reproduction efforts follow a strict \textbf{"head-to-head competition"} principle. All models, including our TritonCast, use the identical Kuroshio region dataset, adhere to a unified \textbf{sequence-to-sequence (10 days in, 10 days out)} task paradigm, and are subjected to a consistent training procedure, including the use of the Adam optimizer, a Cosine Annealing learning rate scheduler, and a total training duration of 1000 epochs.

It is important to note a key adaptation for the DiT (Diffusion Transformer) model. The original DiT architecture is designed primarily for image generation and cannot directly perform spatio-temporal sequence forecasting. To enable it for our sequence-to-sequence task, we perform a necessary modification: \textbf{we retain its core Transformer blocks and integrate them with the same convolutional encoder and decoder used in TritonCast}. This hybrid architecture allows the adapted DiT model to process spatio-temporal data and generate sequence predictions. \textbf{Tab.~\ref{tab:kuroshio_baselines_summary_en}} summarizes and compares the key configurations for TritonCast and the reproduced baselines.

\begin{table}[h!]
\centering
\caption{\textbf{Comparison of reproduced baseline models for the high-resolution Kuroshio region ($0.125\degree$) forecasting task.}}
\label{tab:kuroshio_baselines_summary_en}
\setlength{\tabcolsep}{10pt}
\small
\begin{tabular}{lcccc}
\toprule
\textbf{Model} & \textbf{Task Paradigm} & \textbf{Batch Size (per GPU)} & \textbf{Initial LR} & \textbf{Parameters} \\
\midrule
\textbf{TritonCast (Ours)}& Seq-to-Seq (10→10 days) & 4 & 1e-3 &   0.028B\\
\addlinespace
SimVP  & Seq-to-Seq (10→10 days) & 4 & 1e-3&  0.019B \\
\addlinespace
U-Net  & Seq-to-Seq (10→10 days)\textsuperscript{1} & 4 & 1e-3 &0.032B \\
\addlinespace
DiT  & Seq-to-Seq (10→10 days) & 4 & 1e-3 & 0.015B \\
\bottomrule
\multicolumn{5}{l}{\textsuperscript{1}\footnotesize{U-Net processes the spatio-temporal sequence by merging the time and channel dimensions.}}
\end{tabular}
\end{table}

\paragraph{\ding{185} Baselines For Turbulence Forecasting.}

In the classic physics benchmark of 2D Kolmogorov turbulence, we conduct a comprehensive comparison between TritonCast and an extensive suite of baseline models that span a wide range of mainstream AI architectures. To ensure that all models are evaluated under strictly fair conditions, we reproduce and retrain all baselines within a highly standardized and automated experimental framework. This framework is driven by a unified YAML configuration file, which guarantees that all training and evaluation conditions are identical, except for the model architecture itself.

Our reproduction protocol adheres to the core principle of a \textbf{"head-to-head competition"} as follows:

\begin{enumerate}
    \item \textbf{Unified Dataset and Task}: All models, including TritonCast, use the exact same 2D turbulence DNS dataset (128×128 resolution, Re=5000). The dataset is consistently split into 80\% for training, 10\% for validation, and 10\% for testing. The task paradigm is unified as \textbf{single-step prediction}, where the model predicts the vorticity field at the next time step given the field at the current time step.

    \item \textbf{Unified Training Paradigm}: With minor exceptions for a few models requiring slight adjustments to batch size or learning rate due to memory constraints or convergence properties, the vast majority of models are trained with identical hyperparameters. This includes using the Adam optimizer, an initial learning rate of \texttt{1e-3} with a cosine annealing scheduler, and a total training duration of 500 epochs.

    \item \textbf{Unified Evaluation Protocol}: All trained models are subjected to a long-term \textbf{autoregressive rollout for 99 time steps} on the same test set. This means that, after the first time step, the input to the model exclusively depends on its own prediction from the previous step, posing a significant challenge to the model's long-term stability.
\end{enumerate}

The baselines we reproduce cover a diverse set of architectures, primarily including:
\begin{itemize}
    \item \textbf{Neural Operators}: Such as \textbf{FNO}, \textbf{CNO}, and \textbf{MGNO}, which are representative models designed for learning PDE dynamical systems.
    \item \textbf{U-Net Like Architectures}: Including the classic \textbf{U-Net} and the \textbf{FourierUnet} that serves as the backbone for PDE-Refiner.
    \item \textbf{Video Prediction Models}: Such as \textbf{SimVP}, which have inherent advantages in handling spatio-temporal sequences.
    \item \textbf{Other Mainstream Architectures}: Including \textbf{LSM (Latent Spectral Model)} and others.
\end{itemize}

This rigorously unified reproduction and evaluation process ensures that performance differences can be attributed primarily to the intrinsic design of the models and their ability to capture the underlying physical dynamics. A summary of the key configurations for the main baselines in this experiment is provided in \textbf{Tab.~\ref{tab:turbulence_baselines_summary}}.

\begin{table}[h!]
\centering
\caption{\textbf{Summary of reproduction configurations for the main baseline models in the 2D turbulence forecasting task.} All models perform a single-step prediction task.}
\label{tab:turbulence_baselines_summary}
\small
\begin{tabular}{lcccc}
\toprule
\textbf{Model} & \textbf{Core Architecture} & \textbf{Batch Size} & \textbf{Learning Rate} & \textbf{Approx. Params} \\
\midrule
\textbf{TritonCast (Ours)} & \textbf{Hierarchical V-Cycle} & \textbf{20} & \textbf{1e-3} & \textbf{0.01B} \\
\midrule
FNO & Neural Operator & 20 & 1e-3 & 0.01B \\
CNO\textsuperscript{1} & Convolutional Neural Operator & 5 & 1e-3 & 0.09B\\
PDE-Refiner& Diffusion Model & 20 & 1e-3 & 0.02B \\
SimVP & Convolutional RNN & 20 & 1e-3 & 0.06B \\
LSM & Latent Spectral Model & 20 & 1e-3 & 0.02B \\
U-Net & Conv. Encoder-Decoder & 20 & 1e-3 & 0.02B \\
\bottomrule
\multicolumn{5}{l}{\textsuperscript{1}\footnotesize{When using an NVIDIA A100 GPU with 40GB of memory, an out-of-memory error occurs if the batch size exceeds 5.}}
\end{tabular}
\end{table}
\clearpage

\section{Extended Results in Atmospheric Dynamics Forecasting and Simulation}

\subsection{Comprehensive Evaluation of Medium-range Weather Forecasting Performance}
\label{appendix:wb2_full_results}

\begin{figure}[h!]
\centering
\includegraphics[width=1\linewidth]{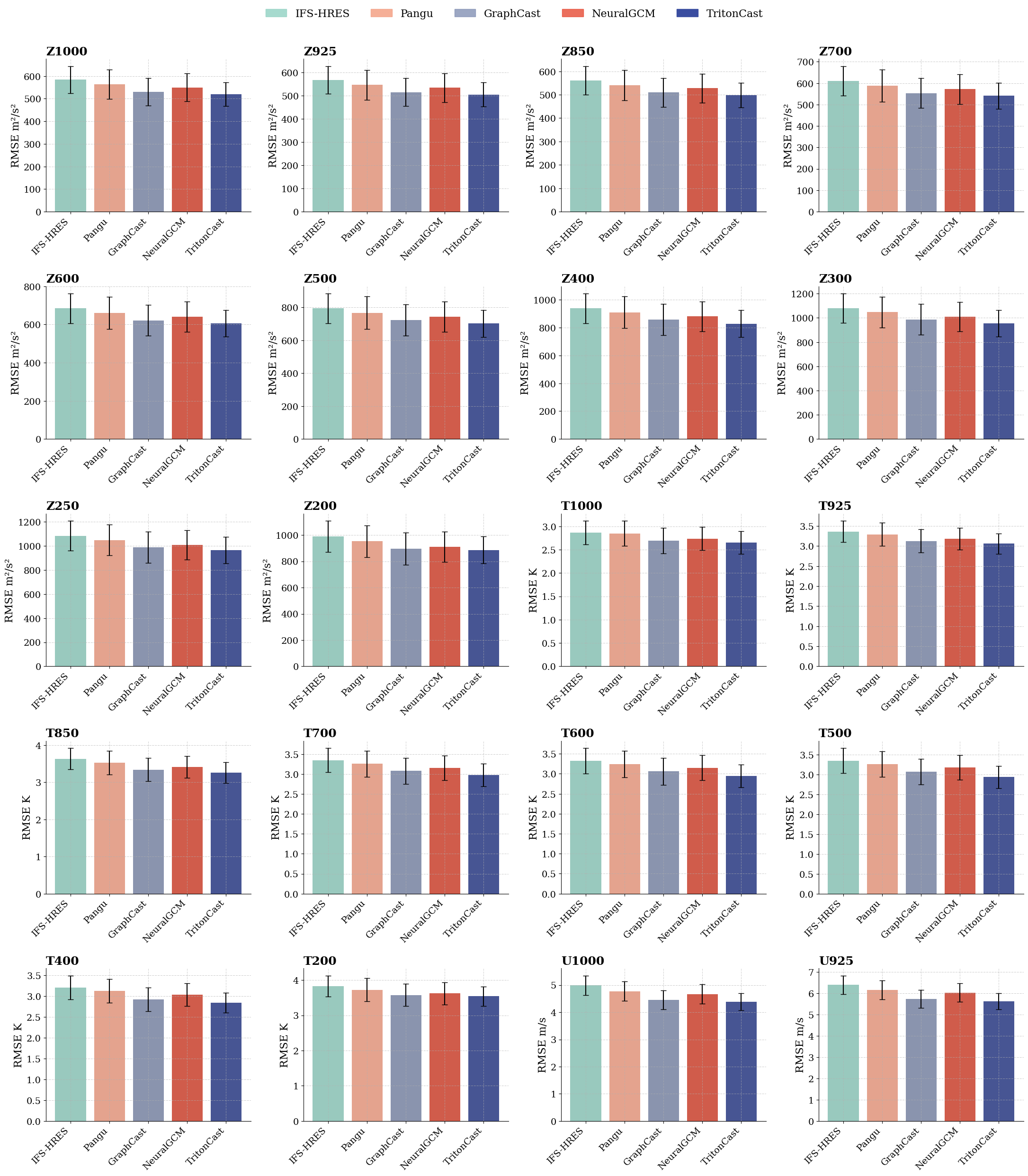}
\caption{\textbf{Performance advantage of TritonCast in 10-day medium-range forecasts.} The figure shows the results of key variables. All results are the average of predictions based on 700 initial conditions (ICs), which are selected consecutively at 12-hour intervals starting from 00:00 UTC on January 1, 2020.}
\label{fig:appendix_wb2}
\end{figure}

\begin{figure}[h!]
\centering
\includegraphics[width=1\linewidth]{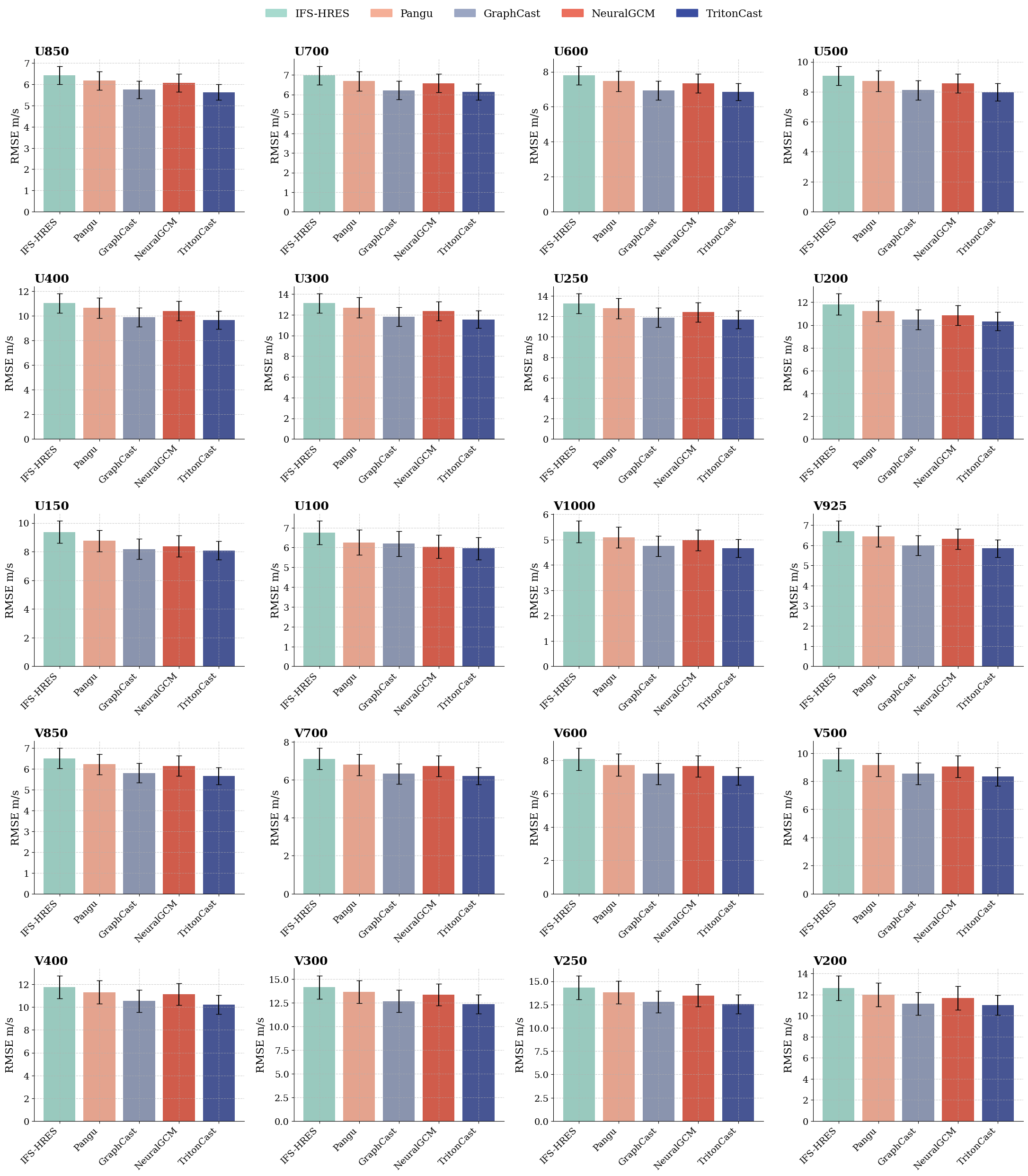}
\caption{\textbf{Performance advantage of TritonCast in 10-day medium-range forecasts.} The figure shows the results of key variables. All results are the average of predictions based on 700 initial conditions (ICs), which are selected consecutively at 12-hour intervals starting from 00:00 UTC on January 1, 2020.}
\label{fig:Appendix_wb2_v2}
\end{figure}

To provide a more comprehensive evaluation of the TritonCast model's medium-range forecasting performance, we systematically analyze key variables at a 10-day lead time. The results of baselines are sourced from WeatherBench 2 bechmark. This evaluation is based on the average results from 700 initial conditions (ICs), selected consecutively at 12-hour intervals starting from 00:00 UTC on January 1, 2020, ensuring the statistical robustness of the conclusions. Since TritonCast demonstrates a leading advantage across numerous variables, we present these results in two figures. \textbf{Fig.~\ref{fig:appendix_wb2}} shows the top 20 superior variables, ranked by meteorological importance, while \textbf{Fig.~\ref{fig:Appendix_wb2_v2}} displays a subsequent set of variables. Taken together, these two figures clearly demonstrate that TritonCast's strong performance is not limited to a few variables but extends across multiple physical quantities and pressure levels, including geopotential (Z), temperature (T), zonal wind component (U), and meridional wind component (V), where it achieves the best 10-day forecast accuracy. This result strongly highlights the TritonCast model's robust competitiveness and broad applicability for medium-range weather forecasting task.

\subsection{In-depth Analysis of Long-term Autoregressive Stability}
\label{appendix:year_long}

To conduct a comprehensive evaluation of model performance over year-long forecast horizons, we perform both a quantitative error accumulation analysis and a qualitative visualization of physical field evolution. All models are initialized with data from January 1, 2020, and operate in a fully autoregressive mode for the entire year, without any intervention from reanalysis forcing. This section combines these two evaluation methods to provide a deep dive into the intrinsic stability and physical consistency of each model, from the perspective of both global mean statistics and instantaneous weather patterns.

\subsubsection{Quantitative Evaluation: Performance Hierarchy Revealed by Error Accumulation Curves}
\begin{figure}[h!]
\centering
\includegraphics[width=1\linewidth]{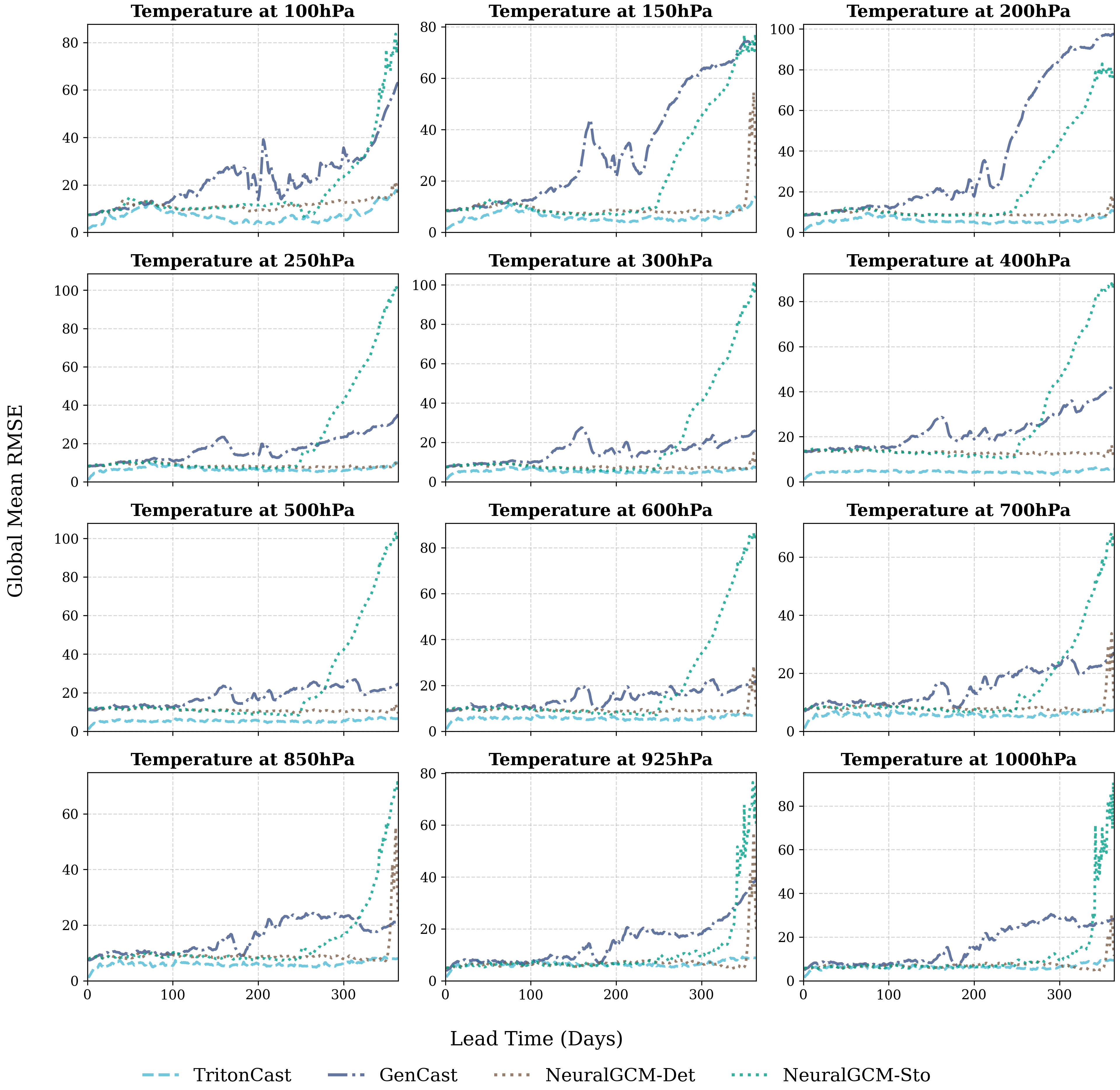}
\caption{\textbf{Day-by-day Error Accumulation Curves for Temperature.}}
\label{fig:error_accumulation_overview_plot_T}
\end{figure}

First, we quantitatively assess performance degradation by computing the day-by-day global mean Root Mean Squared Error (RMSE). \textbf{Figs.~\ref{fig:error_accumulation_overview_plot_T}} through\textbf{~\ref{fig:error_accumulation_overview_plot_Z}} present the RMSE accumulation curves for TritonCast against GenCast and the NeuralGCM models for five key variables. These curves clearly reveal a distinct performance hierarchy:

\begin{enumerate}
    \item \textbf{Sustained Lead of TritonCast:} Across all variables and atmospheric levels, the RMSE curve for TritonCast (dashed light-blue line) exhibits both the lowest error and the gentlest slope. This quantitatively proves its unparalleled accuracy and stability throughout the entire one-year forecast period.
    
    \item \textbf{Rapid Error Growth in Baseline Models:} The error accumulation for GenCast (dash-dotted dark-blue line) and the NeuralGCM models (dotted brown and green lines) is substantially faster, indicating a rapid decay of their forecast skill over long integration times.
    
    \item \textbf{The Unique Signal in Geopotential:} The geopotential RMSE curve (\textbf{Fig.~\ref{fig:error_accumulation_overview_plot_Z}}) is particularly noteworthy. TritonCast displays a unique phenomenon where its error \textit{decreases} during the initial forecast stages. This strongly suggests that the model is not only free from "model drift" but also that its internally learned climate state is more self-consistent and stable than the ERA5 analysis fields used for initialization.
\end{enumerate}

While RMSE provides a precise global metric, it cannot reveal the spatial characteristics of errors or the specific successes and failures in simulating physical processes. To address this, we introduce a qualitative analysis of the physical fields, which also includes the GraphCast model for a broader visual comparison.

\begin{figure}[h!]
\centering
\includegraphics[width=1\linewidth]{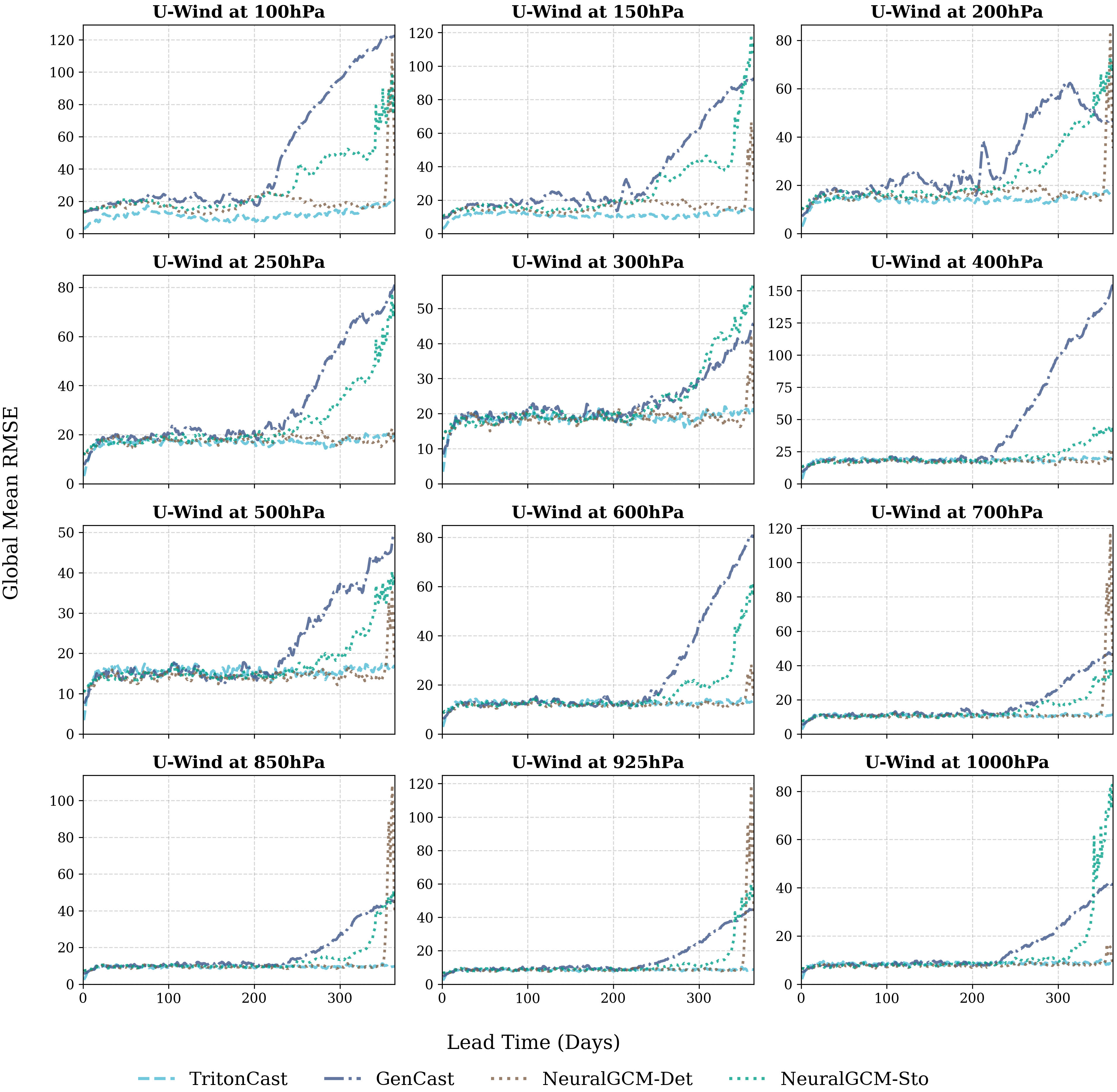}
\caption{\textbf{Day-by-day Error Accumulation Curves for U-Wind.}}
\label{fig:error_accumulation_overview_plot_U}
\end{figure}

\begin{figure}[h!]
\centering
\includegraphics[width=1\linewidth]{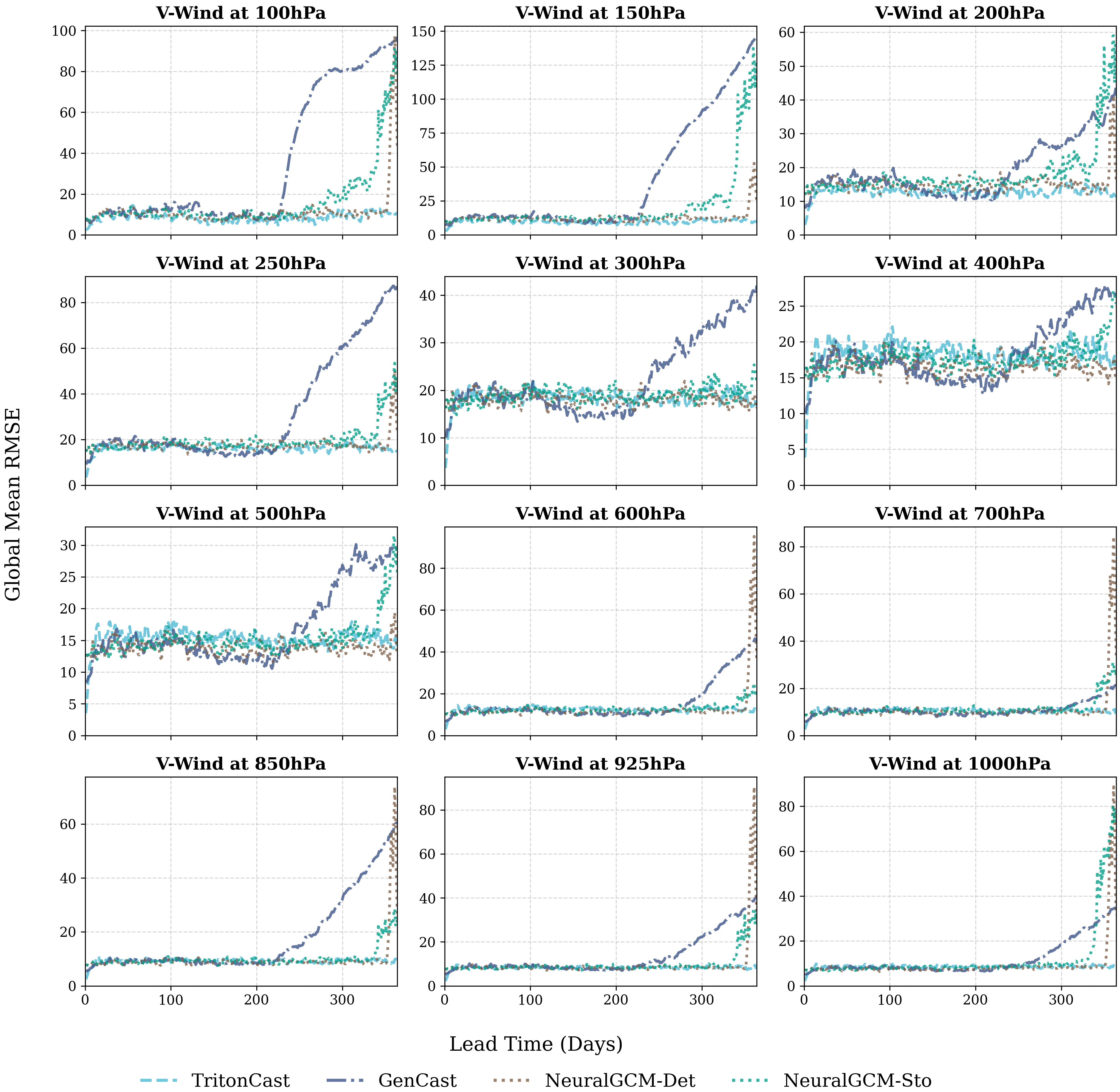}
\caption{\textbf{Day-by-day Error Accumulation Curves for V-Wind.}}
\label{fig:error_accumulation_overview_plot_V}
\end{figure}

\begin{figure}[h!]
\centering
\includegraphics[width=1\linewidth]{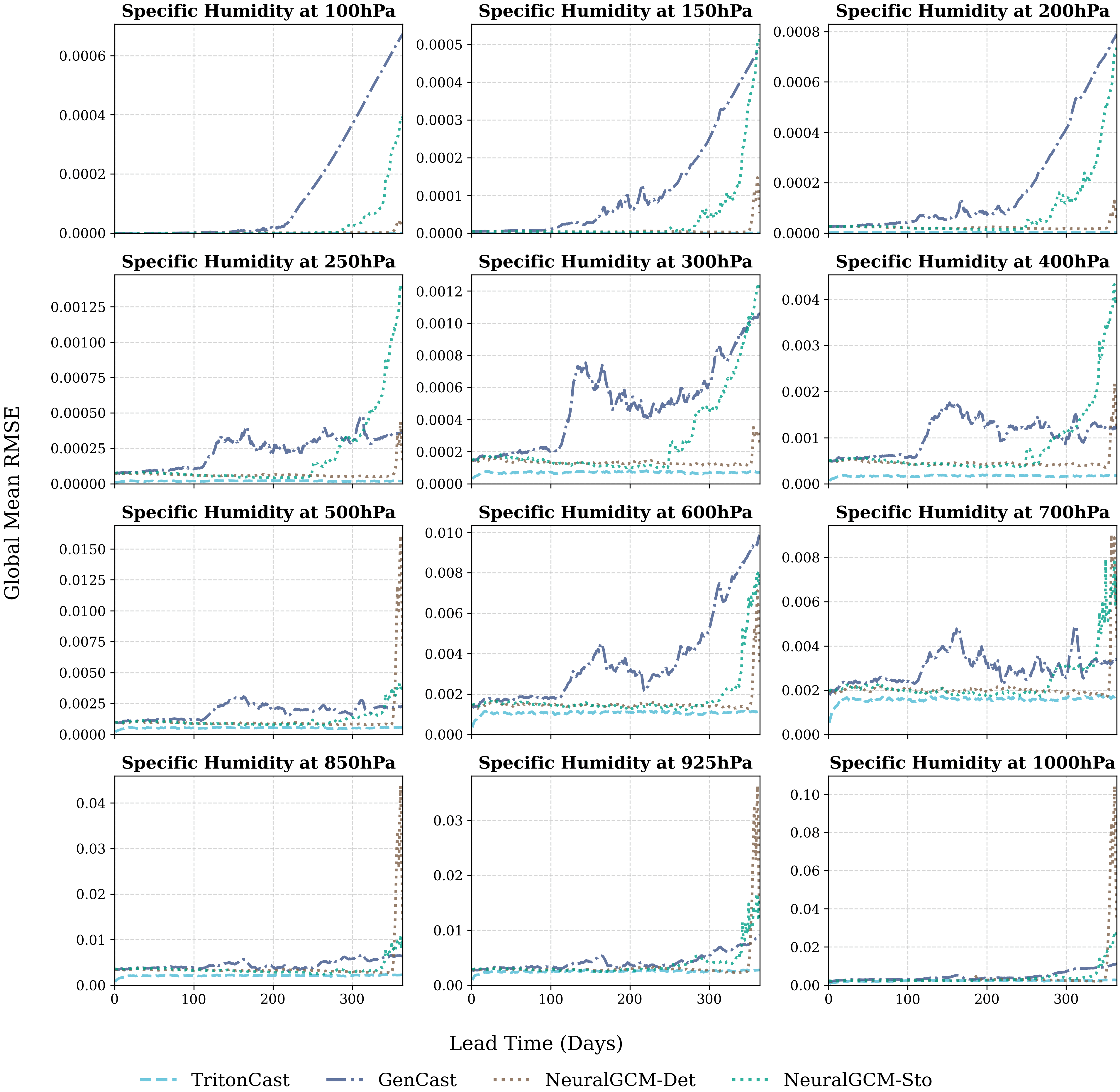}
\caption{\textbf{Day-by-day Error Accumulation Curves for Specific Humidity.}}
\label{fig:error_accumulation_overview_plot_Q}
\end{figure}

\begin{figure}[h!]
\centering
\includegraphics[width=1\linewidth]{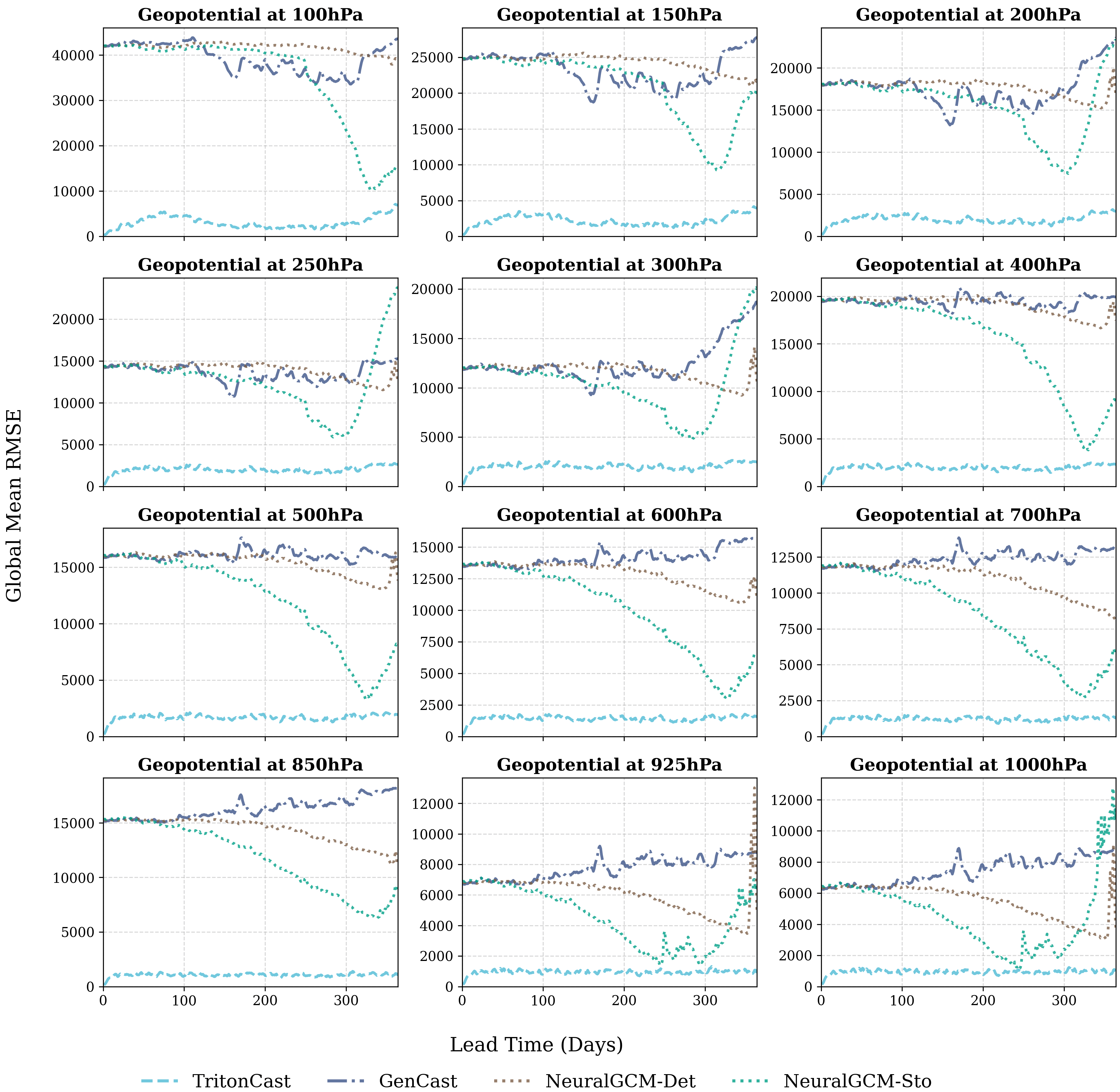}
\caption{\textbf{Day-by-day Error Accumulation Curves for Geopotential.}}
\label{fig:error_accumulation_overview_plot_Z}
\end{figure}

\subsubsection{Qualitative Evaluation: Diagnosing Model Stability Through Physical Field Evolution}

\begin{figure}[h!]
\centering
\includegraphics[width=1\linewidth]{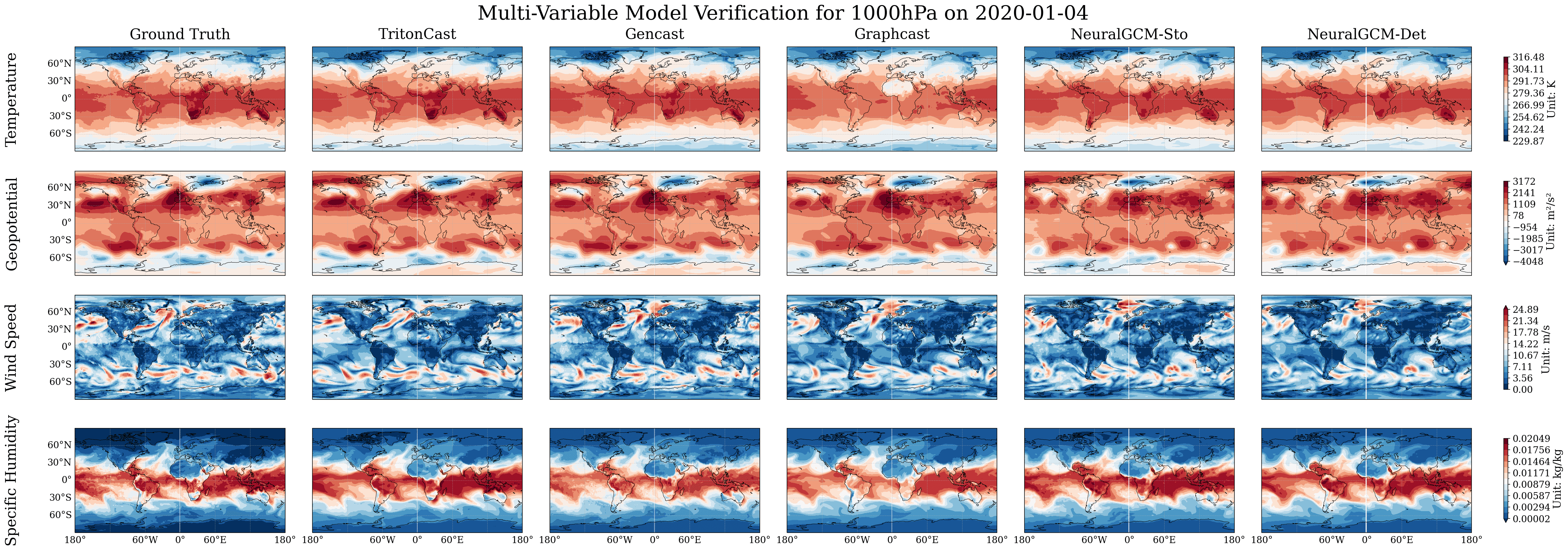}
\caption{\textbf{Multi-variable comparison for 1000hPa on 2020-01-04.}}
\label{fig:v1}
\end{figure}

We selected eight snapshots in time distributed across the seasons of 2020 (from January 4 to December 26) to visualize the key 1000hPa variables for all models, as shown in \textbf{Figs.~\ref{fig:v1}} through \textbf{\ref{fig:v8}}. These "weather snapshots" offer a direct and insightful view into the long-term behavior of the models.

\paragraph{Short-Term Forecasts (January - March): Early Signs of Performance Gaps}

In the initial phase of the forecast (\textbf{Figs.~\ref{fig:v1}} to\textbf{~\ref{fig:v4}}), all models can capture the large-scale circulation patterns. However, subtle yet telling differences in detail are already apparent:
\begin{itemize}
    \item \textbf{TritonCast:} Its generated fields are visually almost indistinguishable from the Ground Truth (ERA5). It clearly resolves the fine structures of small- to medium-scale weather systems, such as frontal systems, oceanic vortices, and tropical atmospheric rivers in the wind speed and specific humidity fields.
    \item \textbf{GraphCast:} As the next-best model, its large-scale structures are accurate, but its fields are visibly "smoother" than TritonCast's, losing some small-scale detail and energy.
    \item \textbf{GenCast \& NeuralGCM:} Even in the short term, their outputs begin to show signs of physical inconsistency. For instance, in \textbf{Fig.~\ref{fig:v3}} (March 1), the geopotential field from GenCast already exhibits significant amplitude decay and a smoothing effect: although the low-pressure system over the North Atlantic is still visible, its central intensity is markedly weaker, and its shape is far less defined than in the Ground Truth or TritonCast forecasts. This early dissipation of weather system energy serves as a precursor to the model's eventual collapse in long-range forecasts.
\end{itemize}

\paragraph{Long-Term Forecasts (June - December): The Ultimate Test of Stability and Model Collapse}

As the forecast lead time extends into the later months (\textbf{Figs.~\ref{fig:v5}} to\textbf{~\ref{fig:v8}}), the differences between models become stark and dramatic, shifting from a matter of precision to a question of fundamental stability.

 \textbf{\textit{The Astonishing Stability of TritonCast}:} Even on day 360 of the forecast (\textbf{Fig.~\ref{fig:v8}}, December 26), TritonCast's output remains a physically realistic and detailed weather map. From the wintertime mid-latitude waves in the Northern Hemisphere to the summertime subtropical anticyclones in the Southern Hemisphere, the forecast maintains a high degree of fidelity to reality.

\textbf{\textit{Catastrophic Failure of Baseline Models}:} In stark contrast to TritonCast's stability, the other models experience catastrophic failure, each in a distinct mode:
    \begin{itemize}
        \item \textbf{GenCast's "Model Collapse":} Starting from \textbf{Fig.~\ref{fig:v5}} (June 29), the output of GenCast degenerates into a lifeless, smooth, zonally-averaged state. The wind field becomes nearly static, and the geopotential and humidity fields lose all gradients, indicating a complete cessation of internal model physics.
        \item \textbf{NeuralGCM's "Numerical Divergence":} The NeuralGCM models experience numerical "blow-up." In \textbf{Fig.~\ref{fig:v5}}, the geopotential field has already become a solid color, and by \textbf{Fig.~\ref{fig:v8}}, the temperature field has disintegrated into a "mosaic" of extreme values and random noise, devoid of any physical meaning. This is a classic case of error amplification leading to computational overflow.
        \item \textbf{GraphCast's "Model Drift":} While GraphCast does not fail as catastrophically as GenCast or NeuralGCM, its long-range forecast quality degrades significantly. In \textbf{Fig.~\ref{fig:v8}}, its wind and humidity fields have become excessively smooth, losing most of their energy and detail. This is a classic visual representation of "model drift."
    \end{itemize}

\begin{figure}[h!]
\centering
\includegraphics[width=1\linewidth]{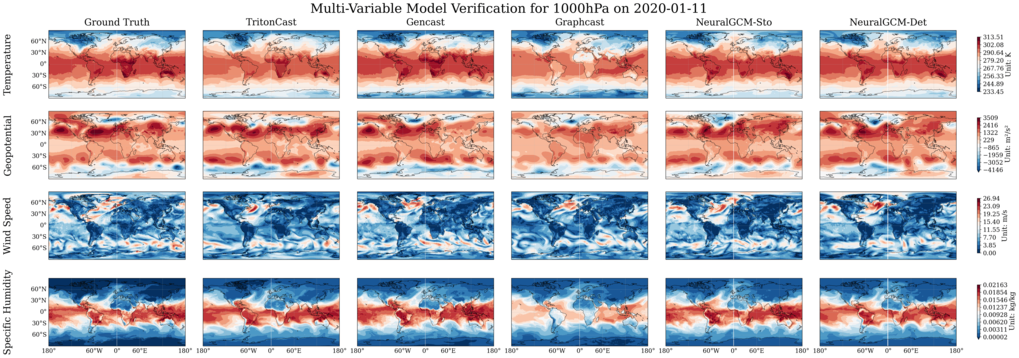}
\caption{\textbf{Multi-variable comparison for 1000hPa on 2020-01-11.}}
\label{fig:v2}
\end{figure}

\begin{figure}[h!]
\centering
\includegraphics[width=1\linewidth]{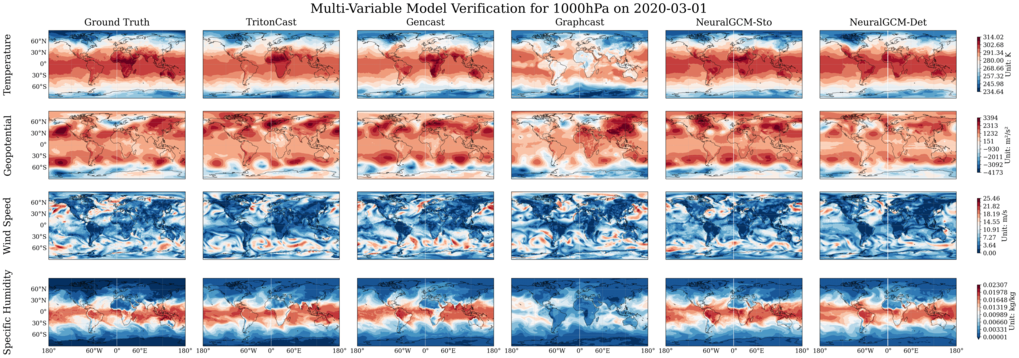}
\caption{\textbf{Multi-variable comparison for 1000hPa on 2020-03-01.}}
\label{fig:v3}
\end{figure}

\begin{figure}[h!]
\centering
\includegraphics[width=1\linewidth]{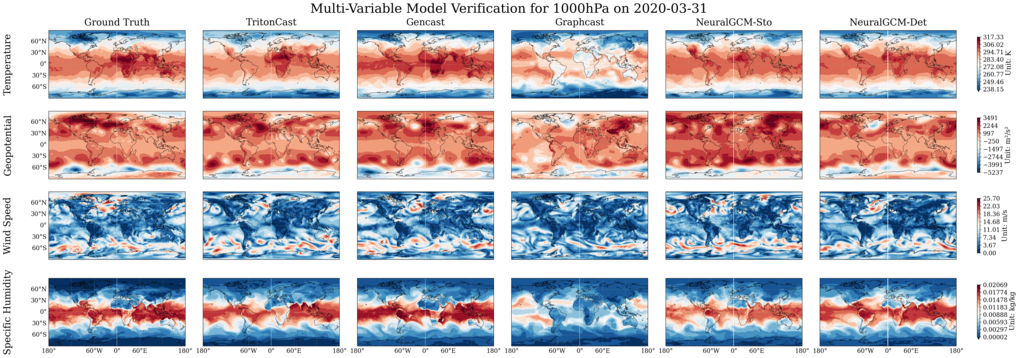}
\caption{\textbf{Multi-variable comparison for 1000hPa on 2020-03-31.}}
\label{fig:v4}
\end{figure}

\begin{figure}[h!]
\centering
\includegraphics[width=1\linewidth]{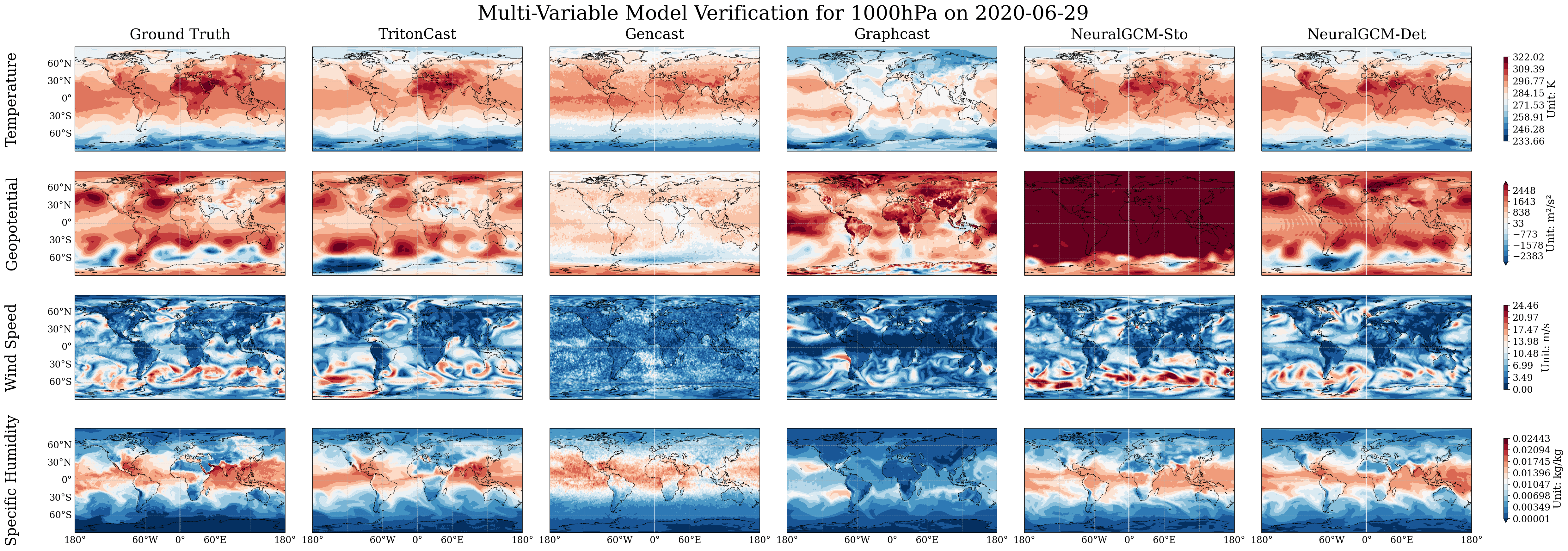}
\caption{\textbf{Multi-variable comparison for 1000hPa on 2020-06-29.}}
\label{fig:v5}
\end{figure}

\begin{figure}[h!]
\centering
\includegraphics[width=1\linewidth]{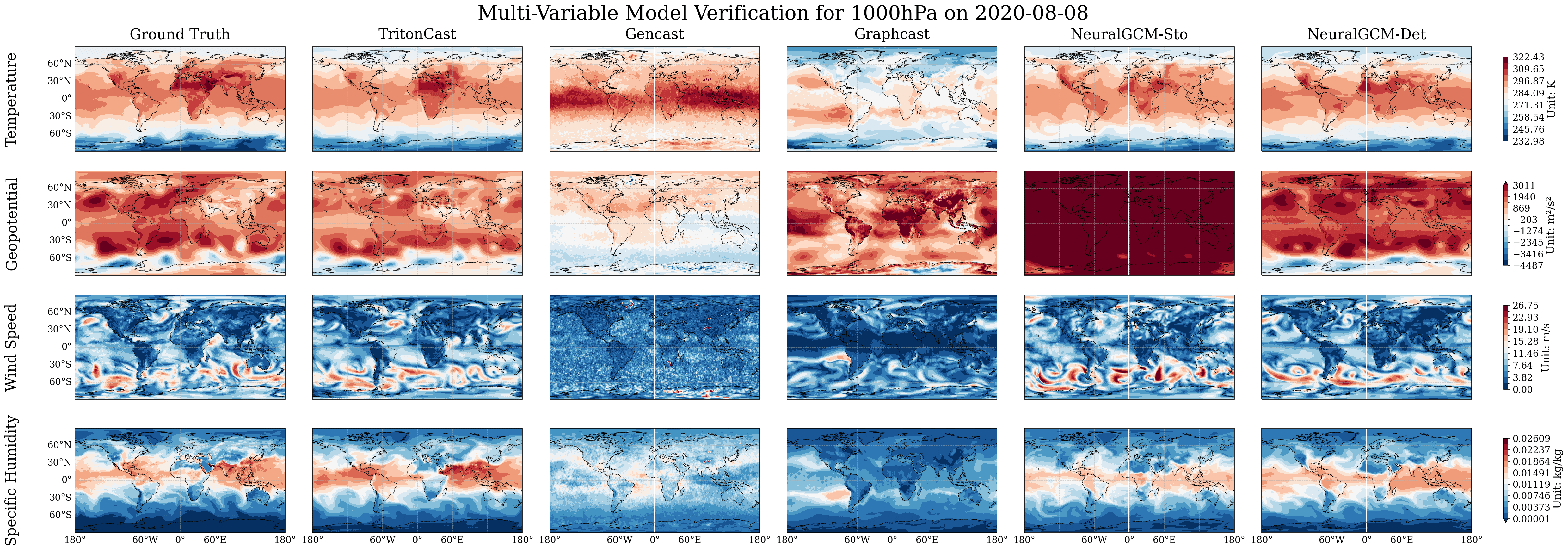}
\caption{\textbf{Multi-variable comparison for 1000hPa on 2020-08-08.}}
\label{fig:v6}
\end{figure}

\begin{figure}[h!]
\centering
\includegraphics[width=1\linewidth]{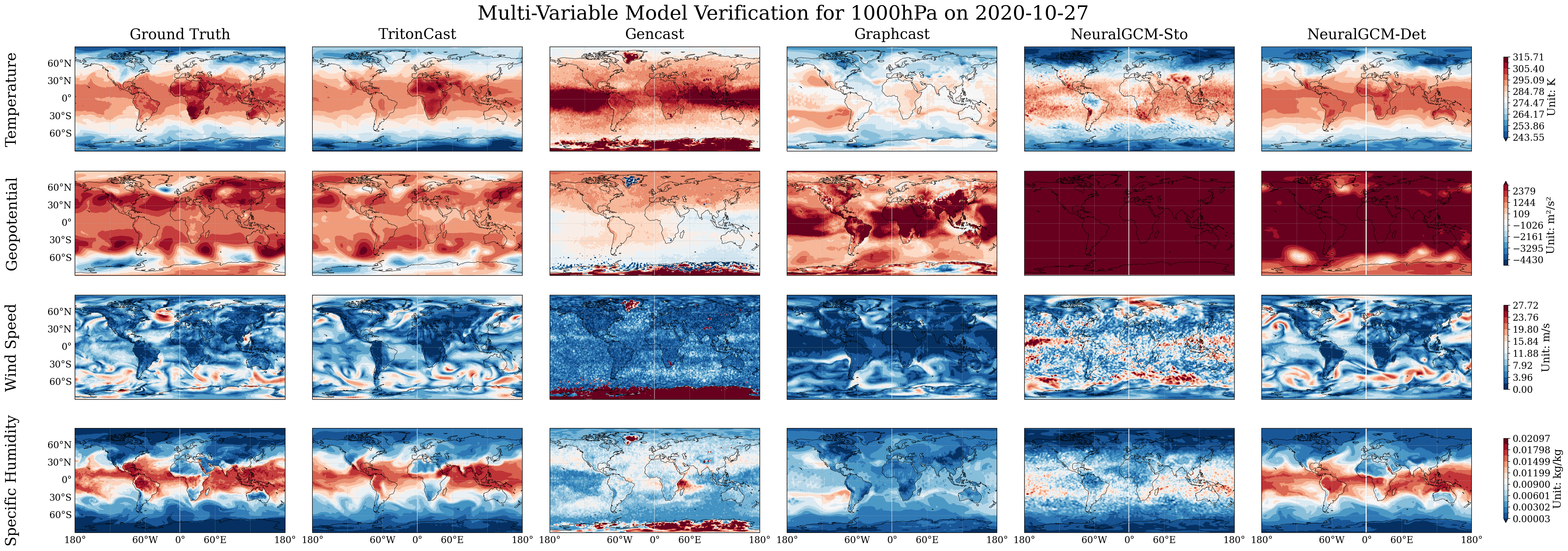}
\caption{\textbf{Multi-variable comparison for 1000hPa on 2020-10-27.}}
\label{fig:v7}
\end{figure}

\begin{figure}[h!]
\centering
\includegraphics[width=1\linewidth]{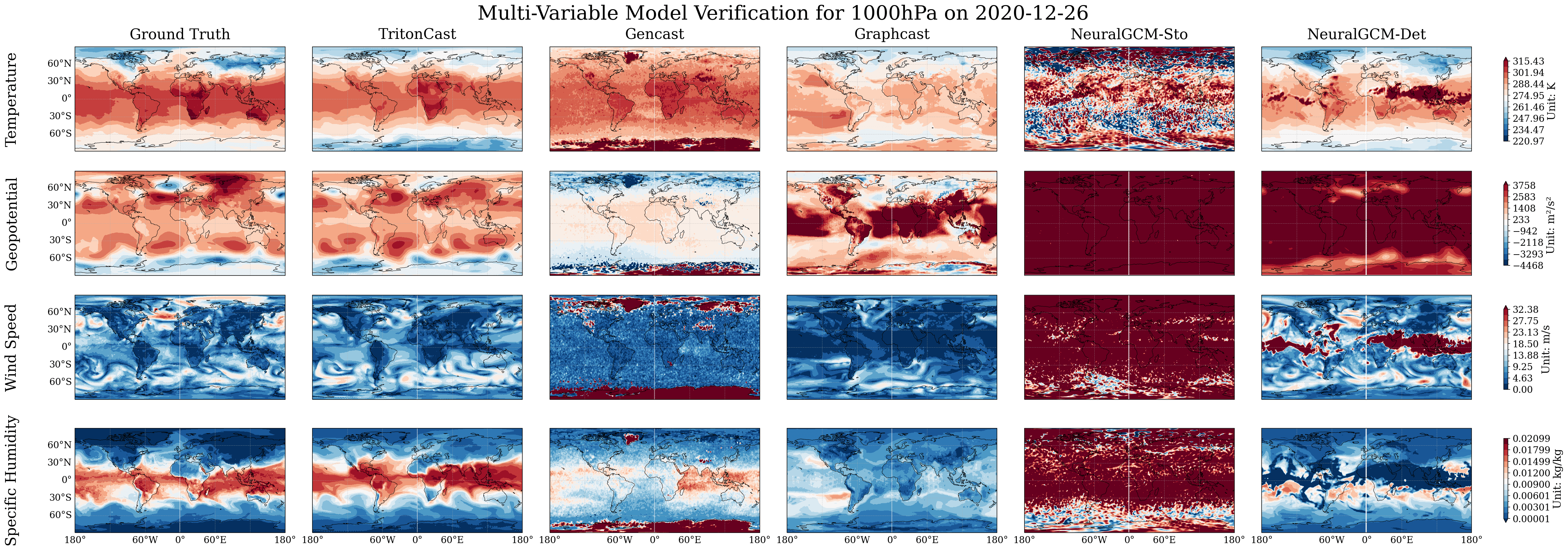}
\caption{\textbf{Multi-variable comparison for 1000hPa on 2020-12-26.}}
\label{fig:v8}
\end{figure}

\subsubsection{Integrated Analysis: Small-scale Learning as the Foundation for Long-term Stability}

By integrating the quantitative RMSE curves with the qualitative visualizations, we can form a complete and compelling argument. The RMSE curves tell us \textit{that} TritonCast's error grows slowly; the visualizations, in comparison with the baseline models, tell us \textit{why}.

Our central thesis is that {TritonCast achieves stable, year-long integration because it has successfully learned to accurately represent small-scale physical processes from data.}

\textbf{\textit{Maintenance of Detail is a Prerequisite for Stability}:} The wind and humidity fields are particularly sensitive to small-scale processes. TritonCast's ability to consistently generate fields with fine-grained vortices and moisture filaments for over a year (as seen in the wind and humidity rows of \textbf{Figs.~\ref{fig:v1}} to\textbf{~\ref{fig:v8}}) proves its capacity to stably handle key processes like convection, turbulence, and frontogenesis.

\textbf{\textit{Suppression of Error Originates from Physical Realism}:} In conventional modeling, inaccurate approximations of these small-scale processes (parameterizations) are a primary source of error. These errors, born at small scales, then contaminate the large-scale circulation via nonlinear upscale cascades. The failure modes seen in the visualizations collapse, divergence, and drift are all different manifestations of this uncontrolled error growth.
    
\textbf{\textit{The TritonCast Advantage}:} TritonCast, through its powerful learning capacity, has implicitly mastered these cross-scale interactions. It accurately models the energy and moisture budgets at small scales, effectively cutting off the primary pathway for error growth at its source. This ensures that small initial errors remain bounded and do not corrupt the large-scale circulation, which is ultimately reflected in the slow-growing RMSE curves seen in \textbf{Figs.~\ref{fig:error_accumulation_overview_plot_T}} through\textbf{~\ref{fig:error_accumulation_overview_plot_Z}}.

\paragraph{Conclusion} The success of TritonCast is therefore not coincidental. Its slow error growth in quantitative metrics and its long-term stability in qualitative visualizations are direct manifestations of its profound physics-learning capabilities. It demonstrates that an AI model that can accurately capture and stably iterate small-scale processes is the key to overcoming the primary bottleneck in long-range weather forecasting and achieving effective predictions from days to a full year.

\clearpage
\subsubsection{Time-to-Divergence Analysis for Baseline Models}
\label{appendix:time_Stability}
To provide a deeper, qualitative understanding of when and how baseline models fail during the year-long autoregressive forecast, this section presents a chronological analysis of their physical field evolution. These visualizations complement the quantitative RMSE scores by revealing the distinct physical mechanisms behind model divergence. Our analysis identifies three primary failure modes: total model collapse, excessive energy damping, and mode drift leading to numerical blow-up.

\begin{figure}[h!]
\centering
\includegraphics[width=0.95\linewidth]{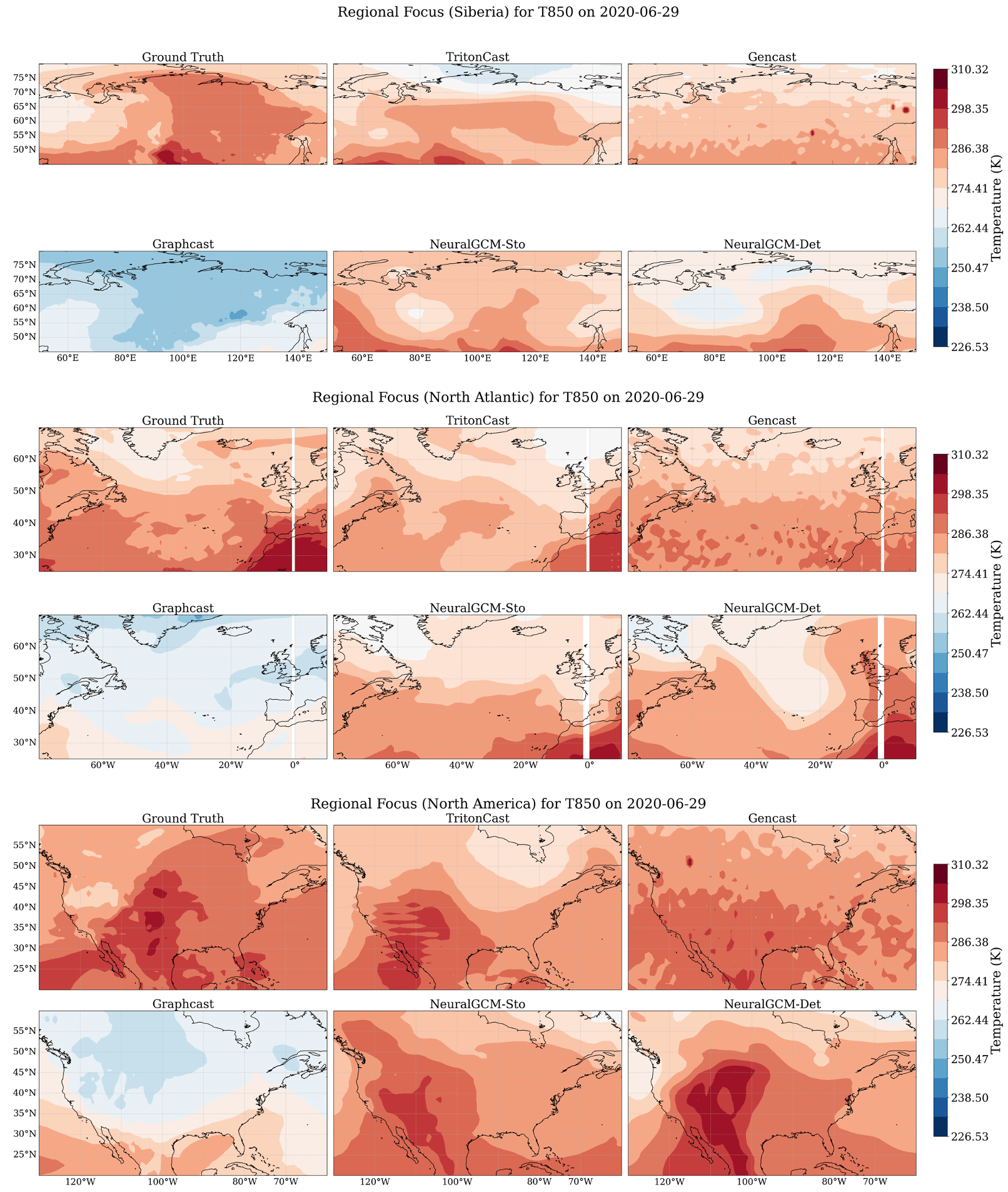}
\caption{\textbf{Regional Comparison of T850 at Day 180: Emergence of Mid-term Divergence.} This figure shows the 850hPa temperature field on June 29, 2020 (Day 180) across three key regions. TritonCast successfully predicts the extreme Siberian heatwave. Meanwhile, GenCast has completely collapsed, Graphcast is overly smoothed, and NeuralGCM exhibits significant mode drift.}
\label{fig:Regional_t850}
\end{figure}

\paragraph{Mid-term Divergence: The First Point of Catastrophic Failure.}
By the middle of the forecast period (approximately Day 180), a clear performance hierarchy emerges, with most baseline models already exhibiting signs of catastrophic failure. As shown in the regional analyses for June 29, 2020 (\textbf{Fig.~\ref{fig:Regional_t850}}), while TritonCast accurately captures the exceptional Siberian heatwave, the baseline models diverge significantly. \textbf{GenCast} undergoes a complete model collapse; its output degenerates into a latitudinally uniform, physically unrealistic state devoid of any weather patterns. \textbf{Graphcast} suffers from excessive energy damping, resulting in an overly smoothed field that completely misses the regional heatwave anomaly. The physics-informed \textbf{NeuralGCM} models, while maintaining a more structured field, demonstrate significant mode drift; they produce plausible but incorrect weather patterns that are decoupled from the ground truth.

\begin{figure}[h!]
\centering
\includegraphics[width=1\linewidth]{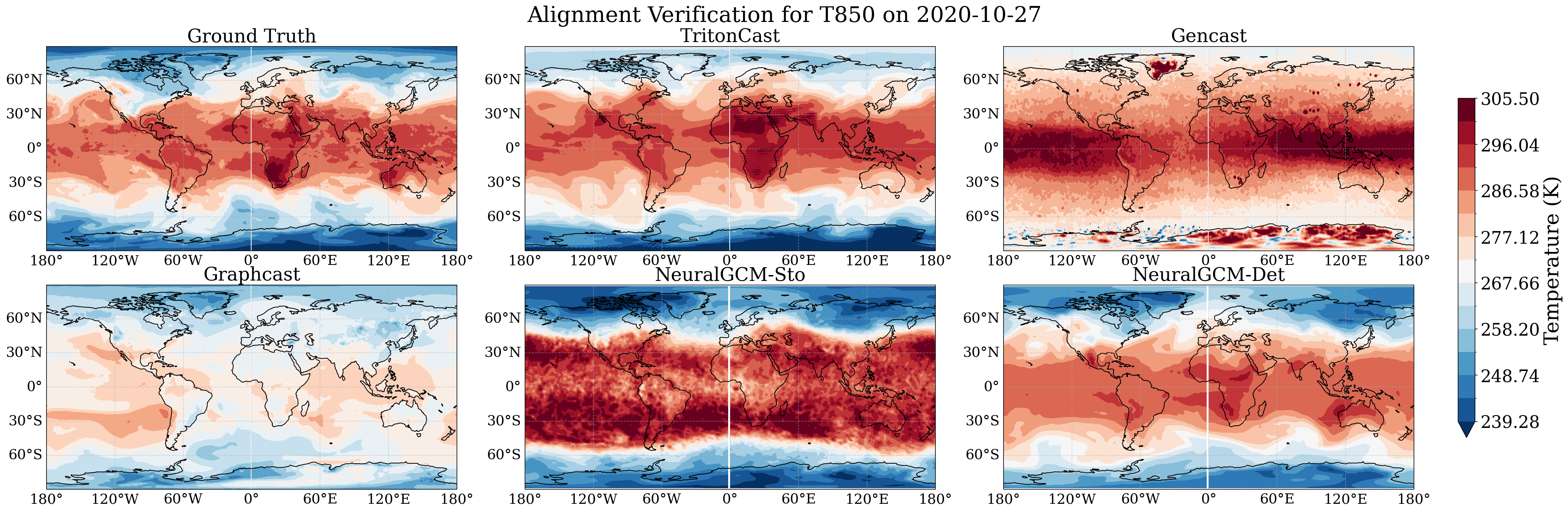}
\caption{\textbf{Global Comparison of T850 at Day 300.} This figure shows the global 850hPa temperature field on October 27, 2020 (Day 300). At this stage, GenCast's forecast field is now populated with severe numerical artifacts, while the failure modes of Graphcast (smoothing) and NeuralGCM (drift) persist.}
\label{fig:alignment_verification_T850_day299}
\end{figure}

\begin{figure}[h!]
\centering
\includegraphics[width=0.95\linewidth]{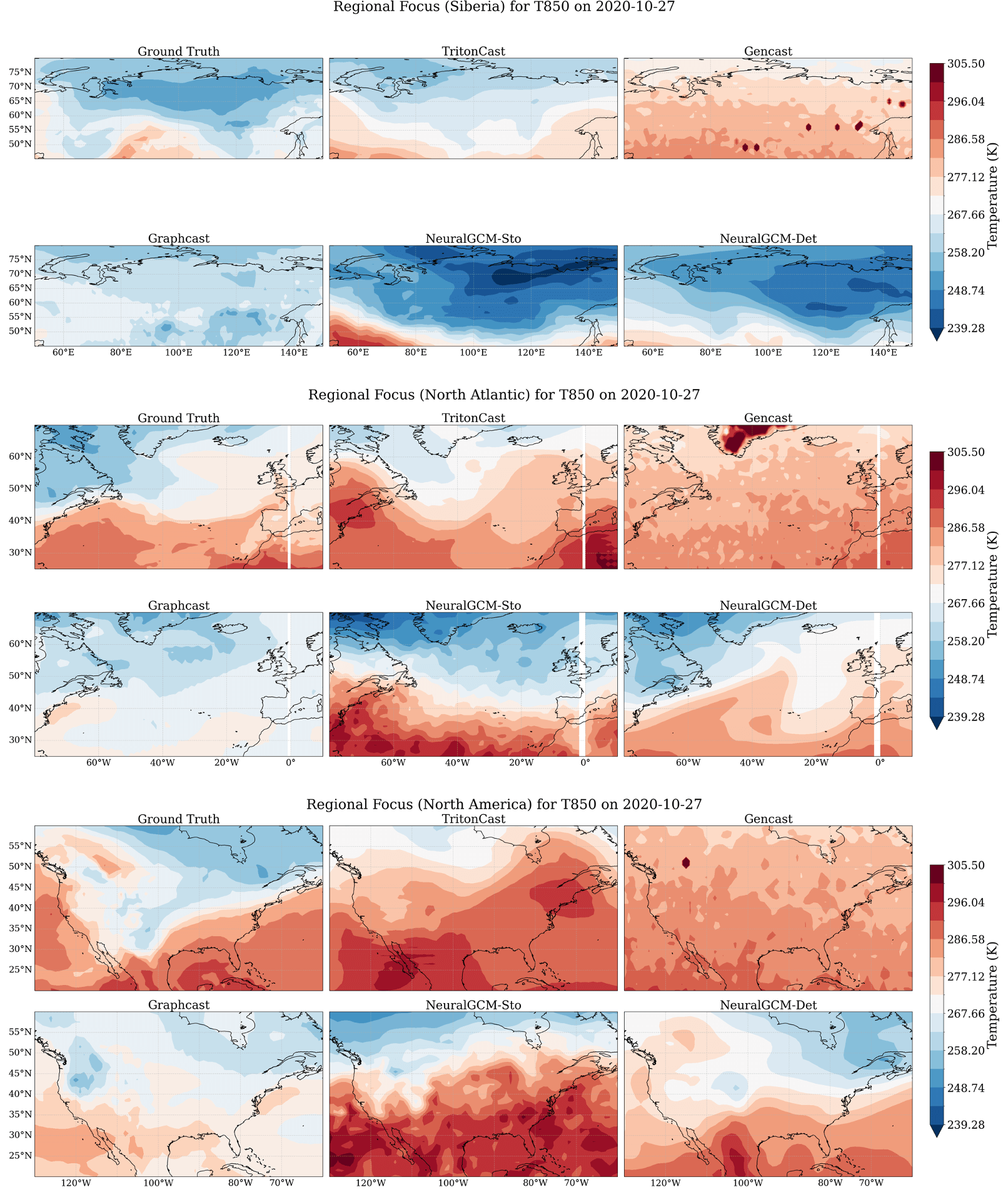}
\caption{\textbf{Regional Comparison of T850 at Day 300: Intensification of Failure Modes.} This figure focuses on regional details at Day 300. The numerical instability of GenCast, the smoothing of Graphcast, and the mode drift of NeuralGCM have all become substantially more pronounced.}
\label{fig:Regional_t850_300}
\end{figure}

\begin{figure}[h!]
\centering
\includegraphics[width=0.95\linewidth]{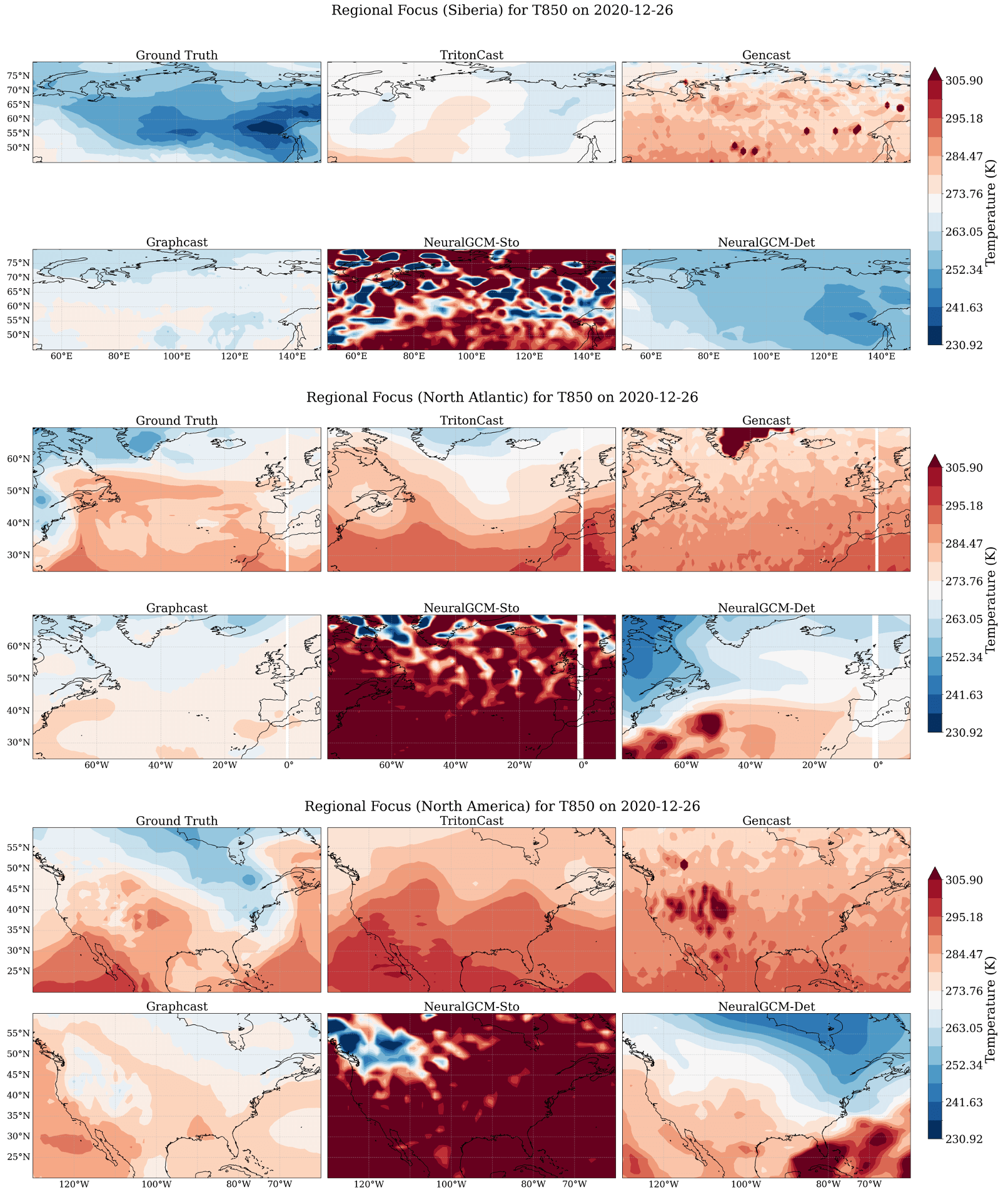}
\caption{\textbf{Regional Comparison of T850 at Day 360: Final Model Collapse.} On the final day of the forecast (December 26, 2020), all baseline models have unequivocally failed. Notably, the NeuralGCM-Stochastic model experiences a numerical blow-up at this stage, producing a field of unphysical noise.}
\label{fig:Regional_t850_360}
\end{figure}

\begin{figure}[h!]
\centering
\includegraphics[width=0.95\linewidth]{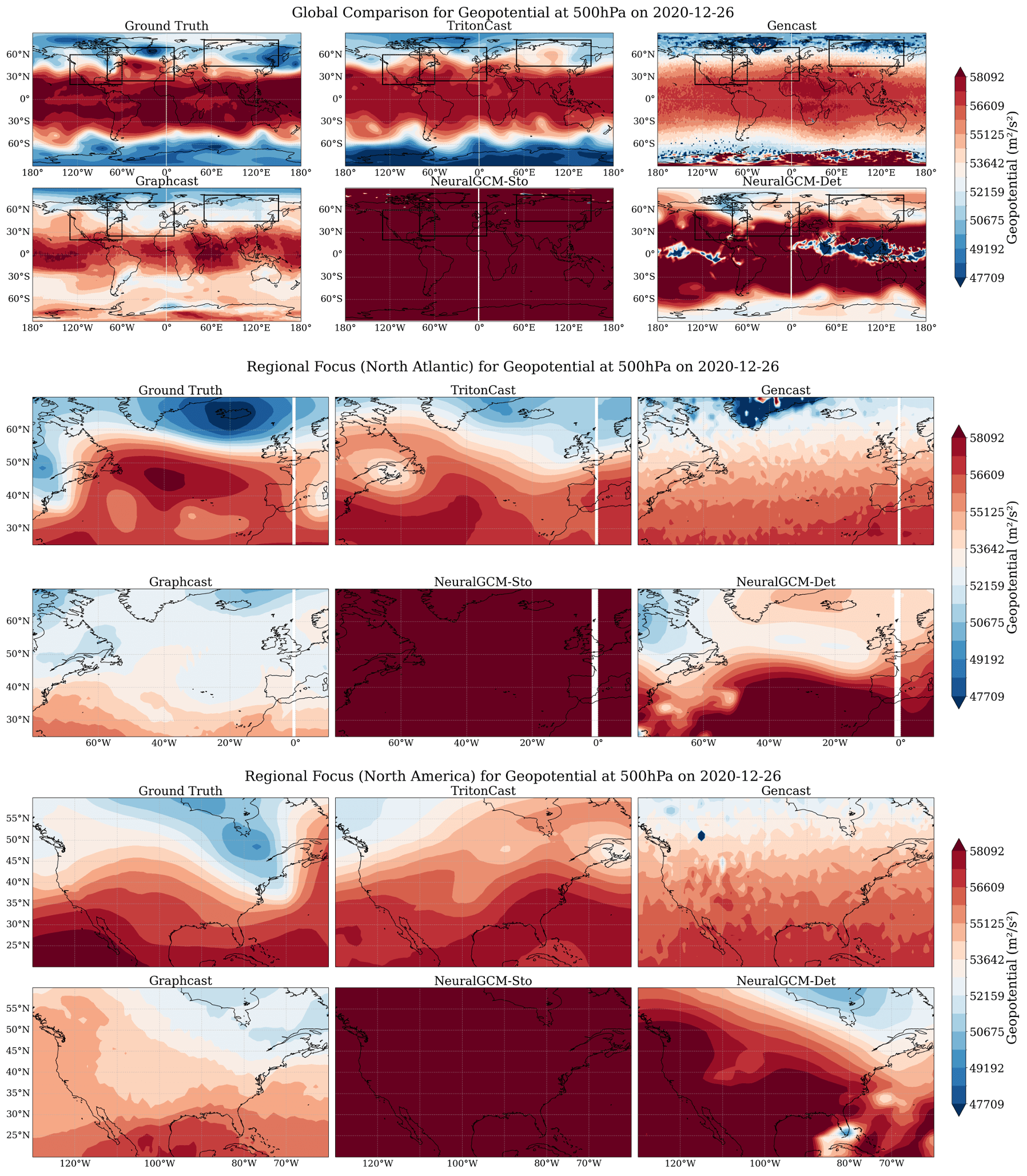}
\caption{\textbf{Global and Regional Comparison of Geopotential at 500hPa at Day 360.} This figure shows the upper-air circulation (500hPa geopotential height) on Day 360. It confirms the model failures from a dynamics perspective: TritonCast maintains realistic mid-latitude wave structures, whereas the circulation fields of all baseline models have either collapsed (GenCast, NeuralGCM-Sto), drifted (NeuralGCM-Det), or become overly smoothed (Graphcast), indicating a complete breakdown of their simulated atmospheric dynamics.}
\label{fig:Q850_360}
\end{figure}

\paragraph{Late-term Evolution of Failure Modes.}
As the forecast progresses into the later months, these failure modes intensify and evolve. The global and regional snapshots from October 27 (Day 300, \textbf{Figs.~\ref{fig:alignment_verification_T850_day299}} and~{\ref{fig:Regional_t850_300}}) show that GenCast's collapse transitions from smoothing to the generation of severe numerical artifacts. By the end of the 360-day forecast (\textbf{Figs.~\ref{fig:Regional_t850_360}} and~\textbf{\ref{fig:Q850_360}}), all baseline models have unequivocally failed. Notably, the NeuralGCM-Stochastic model, which maintains stability for a long period, ultimately succumbs to error accumulation and experiences a numerical blow-up, resulting in a field dominated by high-frequency, unphysical noise. In stark contrast, TritonCast continues to produce physically coherent and accurate fields for all variables, from the 850hPa temperature to the 500hPa geopotential height, maintaining realistic large-scale circulation patterns such as mid-latitude wave trains.

\begin{figure}[h!]
\centering
\includegraphics[width=0.95\linewidth]{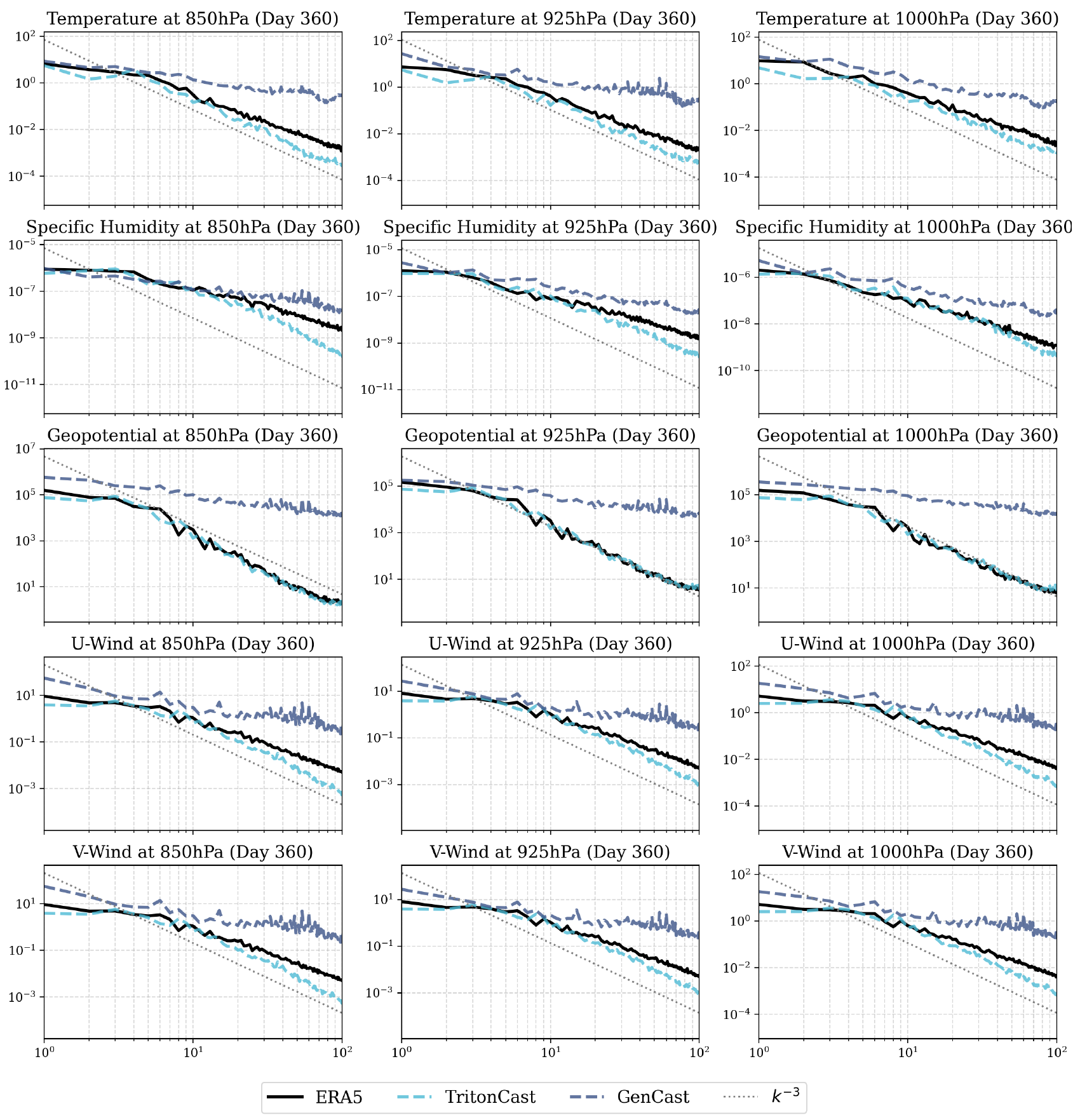}
\caption{\textbf{Power Spectra of Key Variables at Day 360.} This figure compares the zonal mean power spectra for temperature, specific humidity, geopotential, and wind components at the 850hPa, 925hPa, and 1000hPa levels after a 360-day forecast. The TritonCast forecast (light blue) successfully maintains a realistic energy distribution that closely follows the ERA5 ground truth (black) across all wavenumbers. In contrast, GenCast (dark blue) shows a characteristic failure mode with excessive energy damping at small scales (high wavenumbers) and spurious energy accumulation at large scales (low wavenumbers). The dotted gray line indicates a reference slope of $k^{-3}$.}
\label{fig:spectrum_overview_plot}
\end{figure}

\paragraph{The Underlying Cause: A Breakdown in Spectral Fidelity.}
The physical root of these failures is quantitatively revealed by the zonal mean power spectra analysis at Day 360 (\textbf{Fig.~\ref{fig:spectrum_overview_plot}}). TritonCast's spectrum (light blue) remains in excellent agreement with the ERA5 ground truth (black) across all wavenumbers, demonstrating its ability to preserve a realistic energy distribution across scales. Conversely, GenCast's spectrum (dark blue) exemplifies a classic failure mechanism in numerical modeling: a severe, unphysical damping of energy at small scales (high wavenumbers) coupled with a spurious accumulation of energy at large scales (low wavenumbers). This mishandling of the cross-scale energy cascade is the direct cause of the excessive smoothing, loss of detail, and eventual instability observed in the physical field visualizations. TritonCast's architectural design, by fundamentally addressing this spectral bias, ensures its long-term stability and physical fidelity.

\clearpage
\subsection{Extended Validation of Multi-year Climate Simulations}
\label{appendix:Climate}
To comprehensively assess the performance of the TritonCast model in long-term climate simulations, this supplementary presents a suite of extended validation results. These analyses are designed to rigorously examine, from multiple perspectives, the model's long-term stability, physical consistency, and fidelity in reproducing key climatological features throughout a nearly seven-year (2465-day) purely autoregressive simulation. The analysis herein is structured into two core sections: first, an evaluation of the fundamental stability and physical properties during the integration; and second, an assessment of the model's fidelity in simulating key climate modes that emerge from complex physical interactions.

\subsubsection{Assessment of Long-term Integration Stability and Physical Consistency}
For data-driven AI models, maintaining stability and physical realism in long-term autoregressive simulations is a central challenge. This section systematically evaluates TritonCast's performance against this challenge through energy spectrum analysis, global mean variable evolution, and spatial field comparisons at specific time steps.

\begin{figure}[h!]
\centering
\includegraphics[width=1\linewidth]{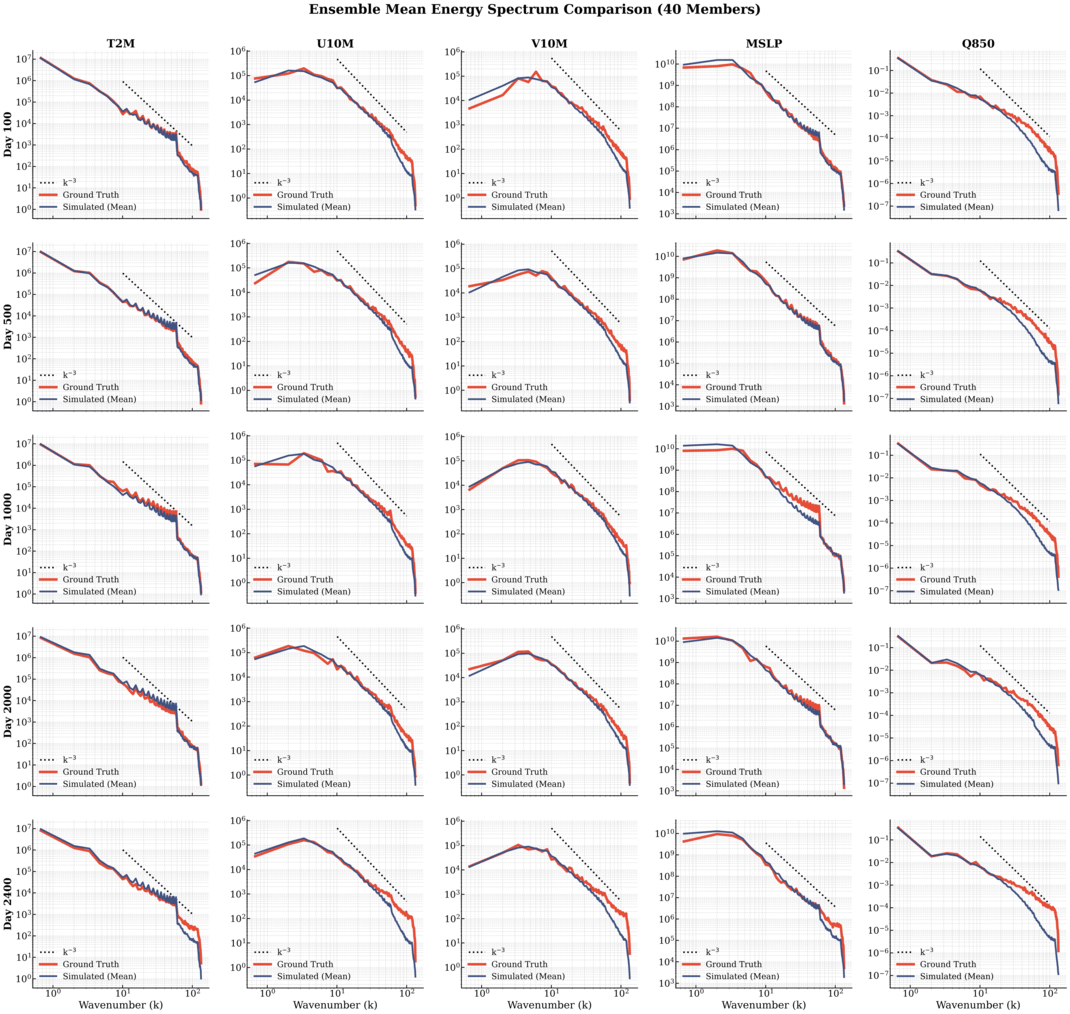}
\caption{Ensemble mean energy spectra for \textbf{surface and near-surface variables} in a 2400-day climate simulation. The figure compares the TritonCast 40-member ensemble mean (red line) to the ERA5 ground truth (blue line). The variables are 2-meter temperature (T2M), 10-meter zonal wind (U10M), 10-meter meridional wind (V10M), mean sea level pressure (MSLP), and specific humidity at 850hPa (Q850).}
\label{fig:energy_spectrum_surface_en}
\end{figure}

\begin{figure}[h!]
\centering
\includegraphics[width=1\linewidth]{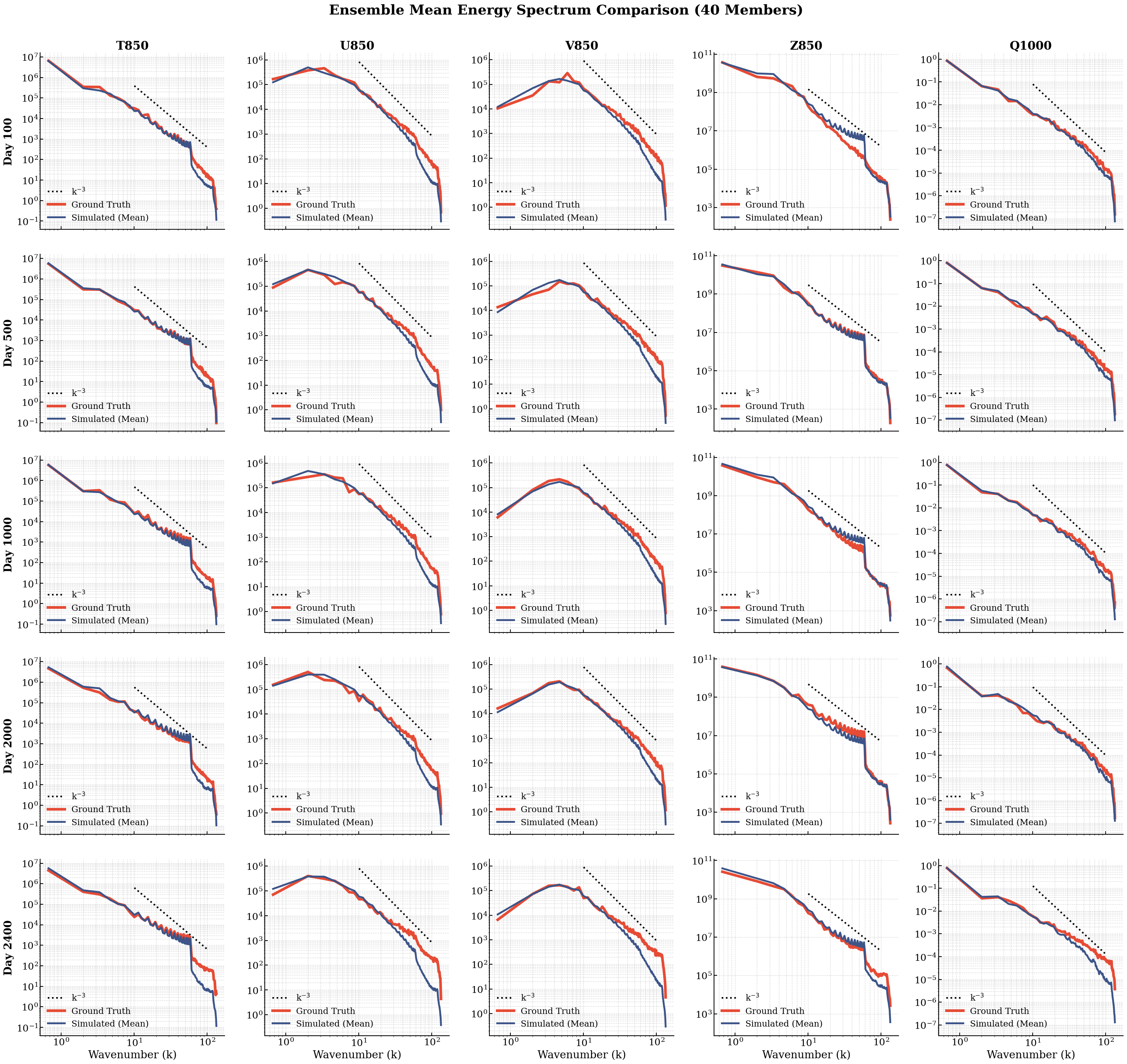}
\caption{Ensemble mean energy spectra for \textbf{upper-air variables} in a 2400-day climate simulation. The plots compare the TritonCast simulation (red line) with the ERA5 ground truth (blue line). The variables shown are temperature at 850hPa (T850), zonal wind at 850hPa (U850), meridional wind at 850hPa (V850), geopotential at 850hPa (Z850), and specific humidity at 1000hPa (Q1000). The consistent alignment with both the ground truth and the theoretical $k^{-3}$ slope highlights the model's exceptional long-term stability and physical realism.}
\label{fig:energy_spectrum_upper_air_en}
\end{figure}

\paragraph{Energy Spectrum Analysis:}
The energy spectrum is a fundamental tool for examining how energy is distributed and transferred across different spatial scales in a fluid simulation. A physically consistent model must correctly reproduce the theoretical $k^{-3}$ power law of the atmospheric energy spectrum. Significant deviations can indicate an anomalous accumulation or dissipation of energy at certain scales, a phenomenon known as "spectral bias", which often leads to the failure of long-term simulations. \textbf{Figs.~\ref{fig:energy_spectrum_surface_en}} and\textbf{~\ref{fig:energy_spectrum_upper_air_en}} present the evolution of the mean energy spectra of the ensemble for the variables near the surface and in the upper air, respectively. The results clearly demonstrate that, over the entire 2400-day simulation period, the TritonCast simulation (red line) exhibits remarkable consistency with the ERA5 ground truth (blue line) across the full range of wavenumbers and perfectly adheres to the theoretical slope. This provides strong evidence that the TritonCast architecture successfully overcomes spectral bias and maintains a physically correct energy cascade process during extremely long integrations.

\begin{figure}[h!]
\centering
\includegraphics[width=1\linewidth]{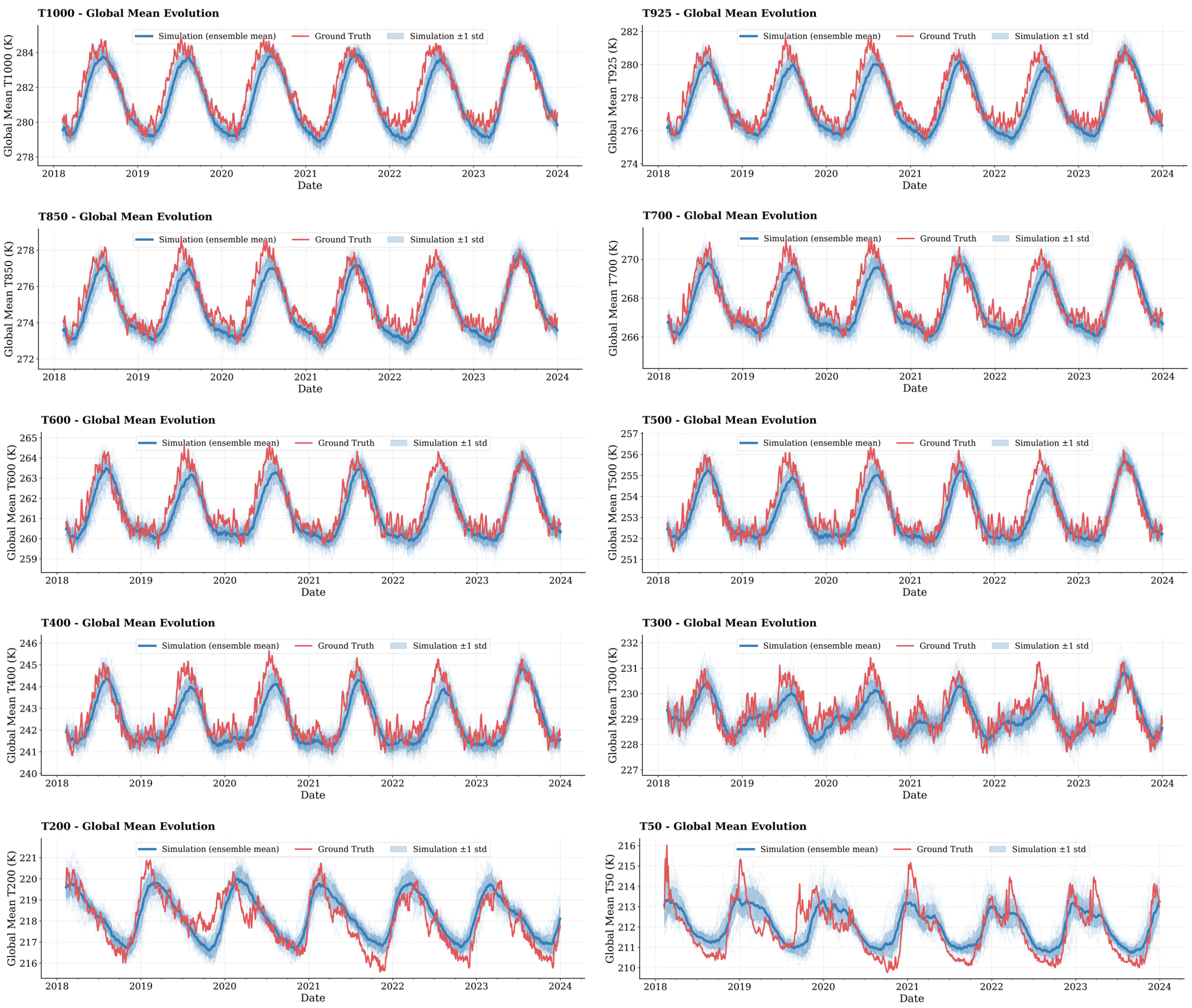}
\caption{Time series of \textbf{global mean temperature} evolution at various pressure levels from 2018 to 2024. The plots compare the TritonCast ensemble mean simulation (dark blue line) against the ERA5 ground truth (red line). The light blue shading indicates the ensemble spread ($\pm 1$ standard deviation). The model accurately reproduces the seasonal cycle and exhibits no signs of climate drift, demonstrating its exceptional long-term stability across the troposphere and stratosphere.}
\label{fig:multi_year_climate_simulation_en}
\end{figure}

\begin{figure}[h!]
\centering
\includegraphics[width=0.8\linewidth]{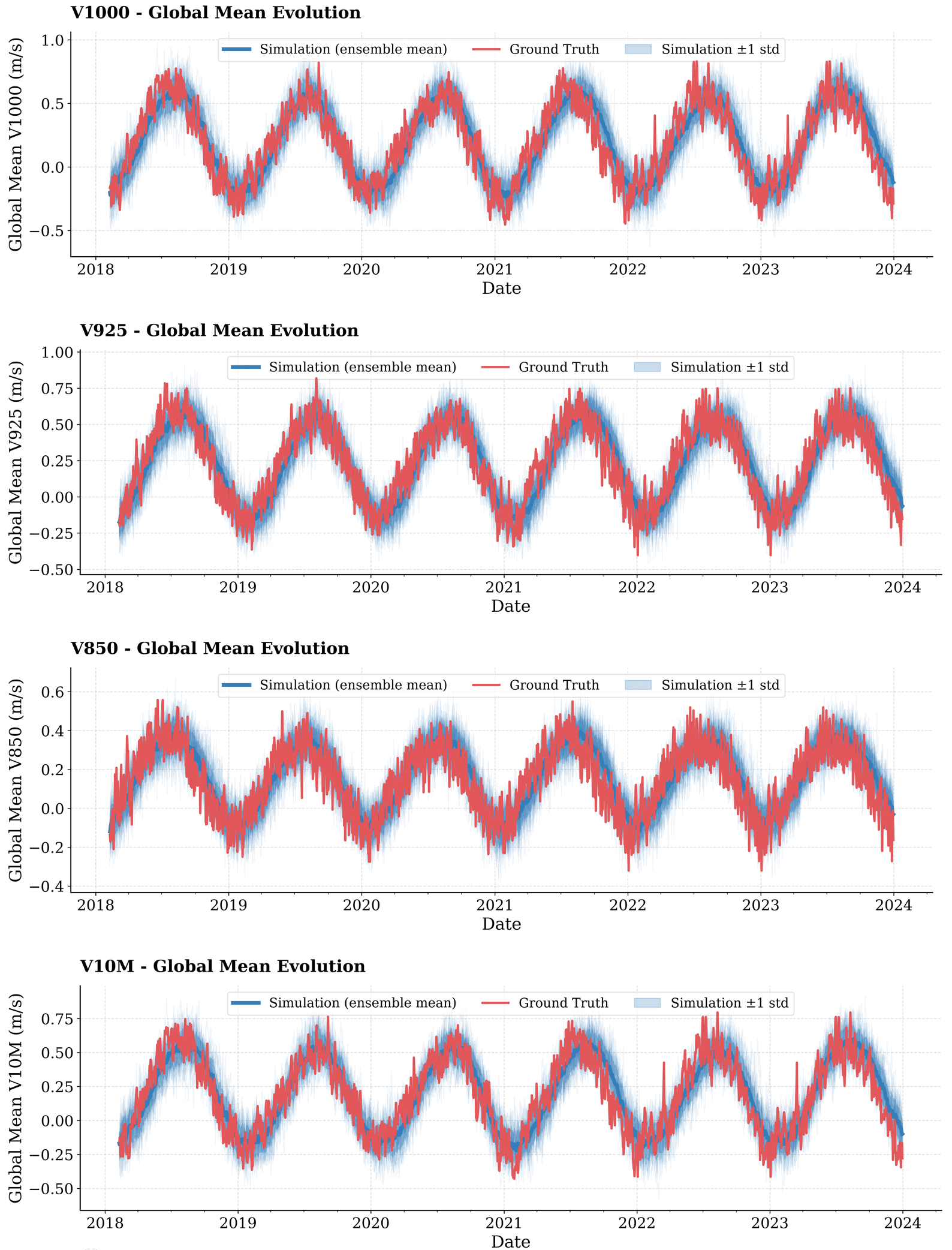}
\caption{Time series of \textbf{global mean meridional wind (V-component)} evolution at different atmospheric levels from 2018 to 2024. Following the same format, the figure shows that TritonCast's ensemble mean (dark blue line) closely tracks the ERA5 ground truth (red line). The model successfully captures seasonal oscillations and maintains long-term stability, further validating its physical realism in simulating atmospheric circulation.}
\label{fig:climate_v_en}
\end{figure}

\paragraph{Analysis of Global Mean State Evolution:}
Examining the time series of global mean variables is a direct method for diagnosing "climate drift" in a model. \textbf{Figs.~\ref{fig:multi_year_climate_simulation_en}} and\textbf{~\ref{fig:climate_v_en}} illustrate the evolution of global mean temperature and meridional wind at various atmospheric levels from 2018 to 2024. The results show that TritonCast's ensemble mean simulation (dark blue line) not only accurately reproduces the seasonal cycle of the real climate (red line), including its amplitude and phase, but also exhibits no systematic deviation over the six-year integration. Furthermore, the ensemble spread (light blue shading) remains stable and within a reasonable range, indicating that the ensemble simulation is healthy and non-divergent. These findings offer conclusive evidence of TritonCast's drift-free characteristic.

\begin{figure}[h!]
\centering
\includegraphics[width=0.85\linewidth]{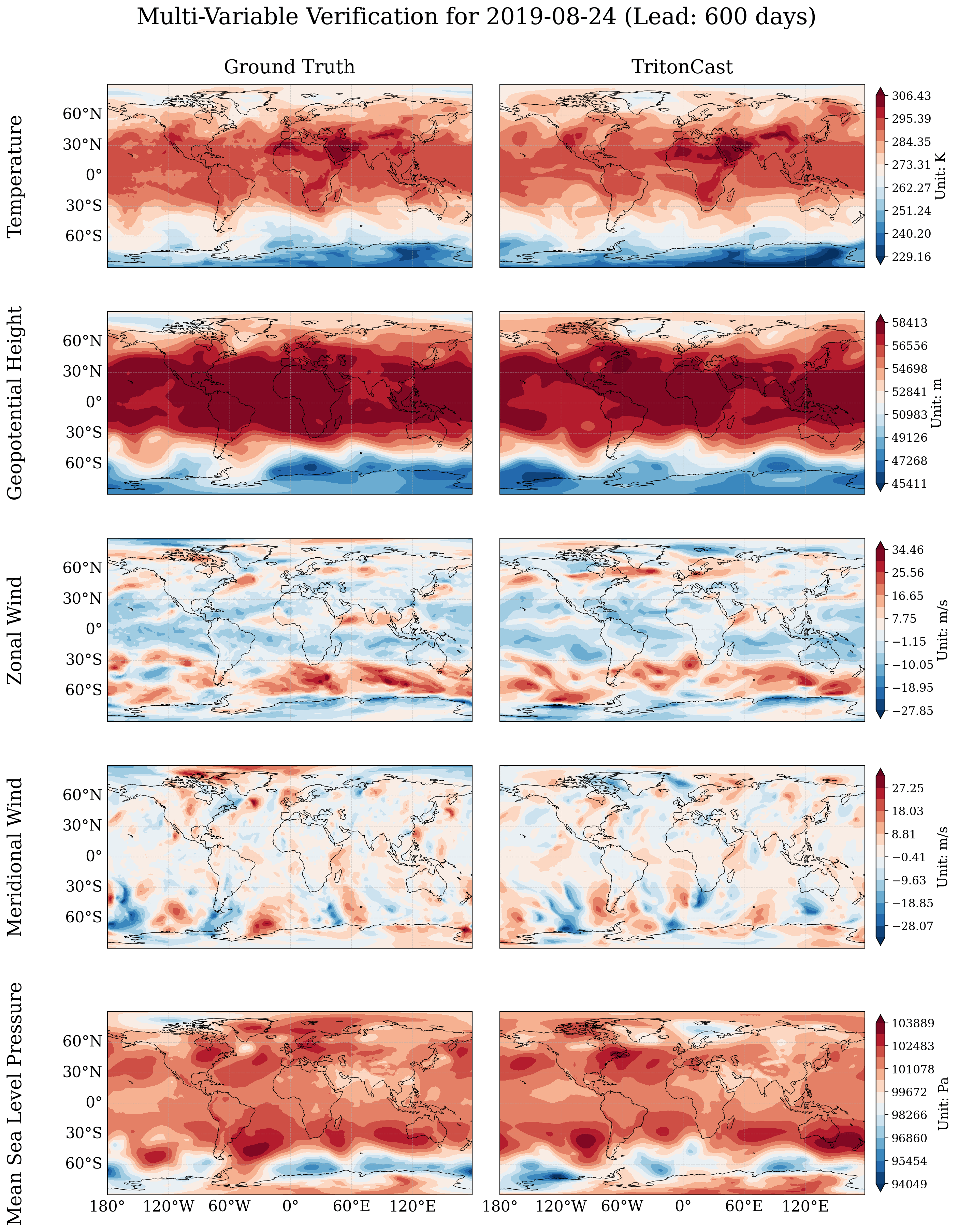}
\caption{Multi-variable verification of a single ensemble member forecast at a lead time of \textbf{600 days}. The figure compares the spatial fields of key atmospheric variables from the TritonCast simulation (right column) with the ERA5 ground truth (left column). The high degree of visual similarity in large-scale patterns, such as jet streams and pressure systems, demonstrates the model's high fidelity in long-range forecasting.}
\label{fig:verification_IC0_20190824_en}
\end{figure}

\begin{figure}[h!]
\centering
\includegraphics[width=0.85\linewidth]{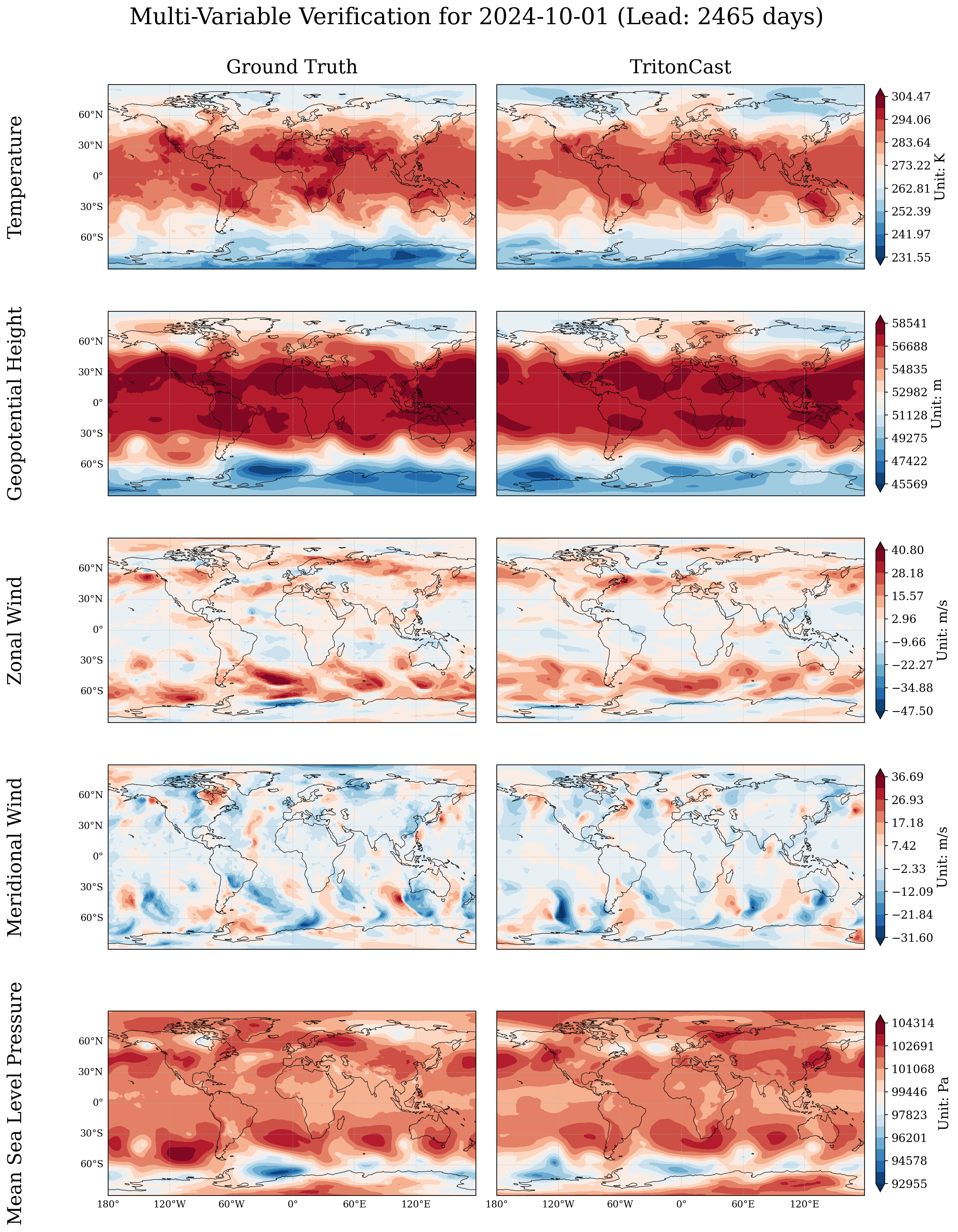}
\caption{Multi-variable verification at an extreme lead time of \textbf{2465 days} (approx. 6.75 years). This comparison showcases the model's exceptional long-term stability. Despite the extended integration period, the TritonCast simulation (right column) maintains a physically realistic climate state that is structurally consistent with the ground truth (left column), indicating a complete absence of catastrophic climate drift.}
\label{fig:verification_IC0_20241001_en}
\end{figure}

\paragraph{Verification of Spatial Field Realism:}
To visually assess the physical realism of the fields produced after long-term integration, \textbf{Figs.~\ref{fig:verification_IC0_20190824_en}} and\textbf{~\ref{fig:verification_IC0_20241001_en}} compare the multi-variable spatial distributions of a single ensemble member at lead times of 600 days and an extremely long 2465 days. Remarkably, even after nearly seven years of continuous autoregressive forecasting, the weather systems simulated by TritonCast from the large-scale gradients in the temperature field and the long-wave troughs and ridges in the geopotential height field to the jet streams in the wind field remain highly consistent with the real world in terms of their spatial location, morphology, and intensity. This demonstrates that the model does not degenerate over time into a smoothed-out climatology or a noisy, non-physical state, but rather continues to generate a dynamically evolving and physically plausible virtual Earth climate.

\subsubsection{Assessment of Fidelity in Reproducing Key Climatological Modes and Processes}
Having established the model's fundamental stability and physical consistency, this section further assesses TritonCast's ability to simulate more complex, emergent phenomena that are core components of the climate system. These diagnostics are "gold standards" for evaluating the performance and scientific value of a climate model.

\begin{figure}[h!]
\centering
\includegraphics[width=0.75\linewidth]{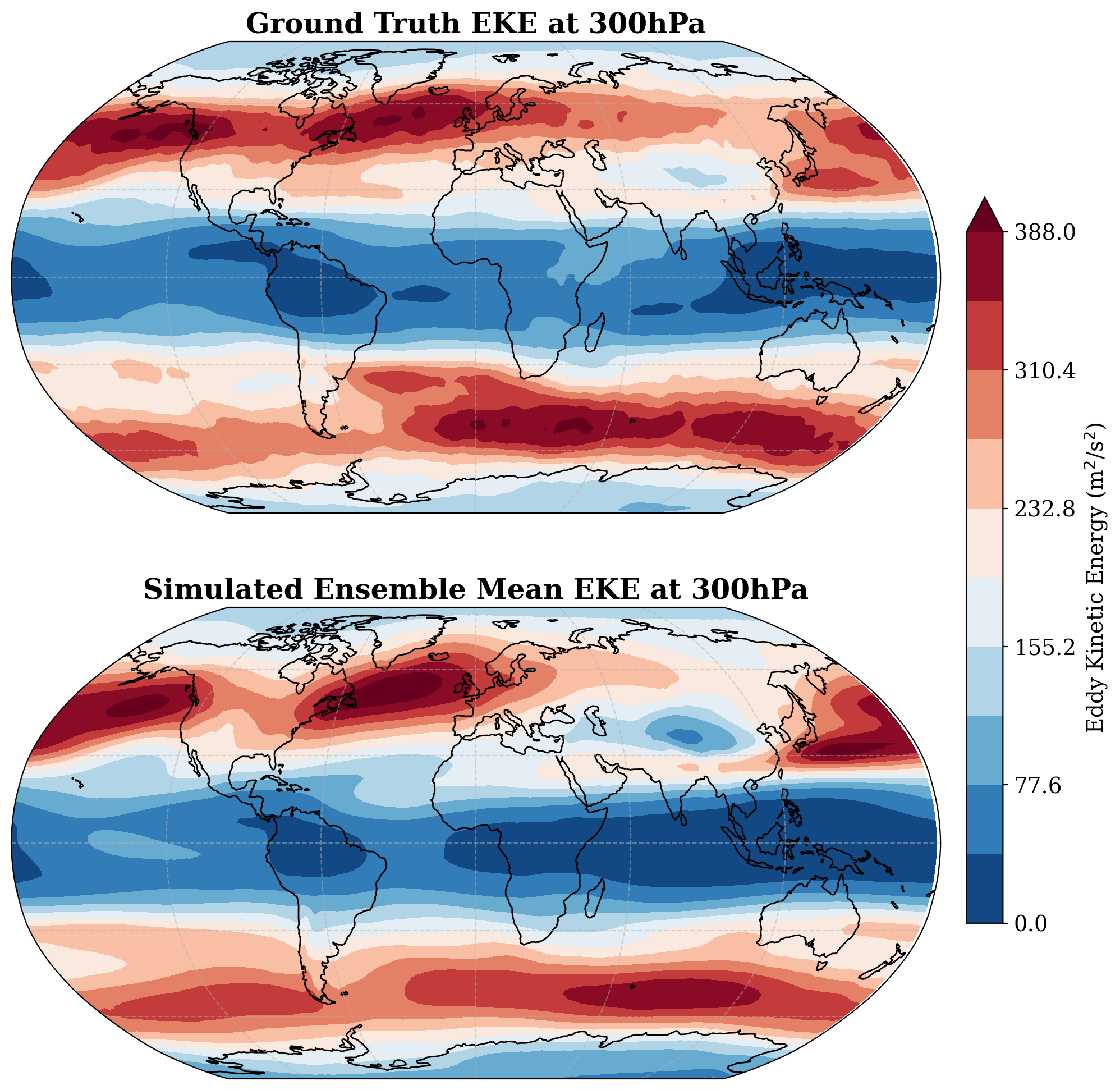}
\caption{Comparison of the climatological mean \textbf{Eddy Kinetic Energy (EKE)} at 300hPa. The top panel shows the Ground Truth, and the bottom panel shows the TritonCast ensemble mean simulation. EKE represents the activity of transient weather systems, with high-value regions (red) corresponding to the major storm tracks. The model successfully captures the geographical location and large-scale structure of the Pacific, Atlantic, and Southern Hemispheric storm tracks.}
\vspace{-15pt}
\label{fig:eke_analysis_comparison_en}
\end{figure}

\paragraph{Simulation of Eddy Kinetic Energy and Storm Tracks:}
Eddy Kinetic Energy (EKE) is a key metric that quantifies the activity of transient mid-latitude weather systems, such as extratropical cyclones. Regions of high EKE correspond to the world's major "storm tracks". \textbf{Fig.~\ref{fig:eke_analysis_comparison_en}} compares the climatological EKE at 300hPa between the ground truth and the model simulation. The result shows that TritonCast successfully reproduces the three major storm tracks: the North Pacific, the North Atlantic, and the Southern Hemispheric annular storm track. The simulated storm tracks are highly consistent with reality in terms of their geographical location, spatial structure, and intensity distribution, indicating that the model has implicitly learned the core dynamical mechanisms, such as baroclinic instability, that drive the life cycles of weather systems.

\begin{figure}[h!]
\centering
\includegraphics[width=0.75\linewidth]{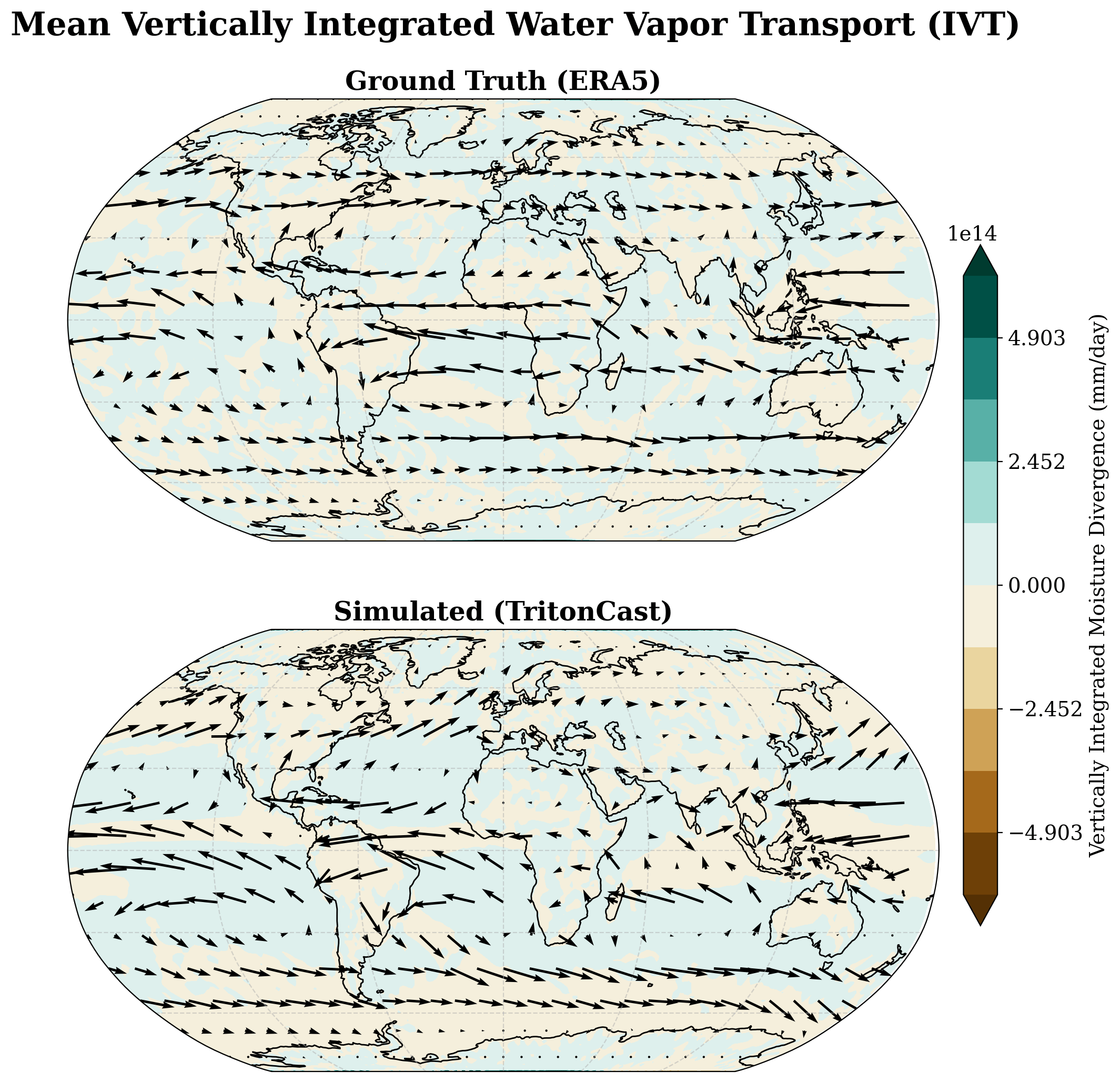}
\caption{Comparison of the climatological mean \textbf{Vertically Integrated Water Vapor Transport (IVT)}. The top panel shows the Ground Truth (ERA5), and the bottom panel shows the TritonCast simulation. Arrows represent the direction and magnitude of vapor transport, while shading indicates moisture divergence (teal, source regions) and convergence (brown, sink regions). The model demonstrates high fidelity in reproducing the key features of the global water cycle.}
\label{fig:ivt_analysis_comparison_en}
\end{figure}

\paragraph{Simulation of Global Water Vapor Transport:}
Integrated Water Vapor Transport (IVT) is a key physical field that depicts the global water cycle. \textbf{Fig.~\ref{fig:ivt_analysis_comparison_en}} compares the climatological mean IVT. The TritonCast simulation (bottom panel) clearly reproduces the main features of the global hydrological cycle, including the flow of moisture from subtropical oceanic source regions (teal background), its convergence into the Intertropical Convergence Zone (ITCZ), and its transport pathways toward the mid-to-high latitudes (arrows). The model also accurately captures the primary regions of precipitation (convergence, brown background). This demonstrates the model's high fidelity in simulating the global water cycle, highlighting its potential for studying extreme weather phenomena like atmospheric rivers.

\begin{figure}[h!]
\centering
\includegraphics[width=0.85\linewidth]{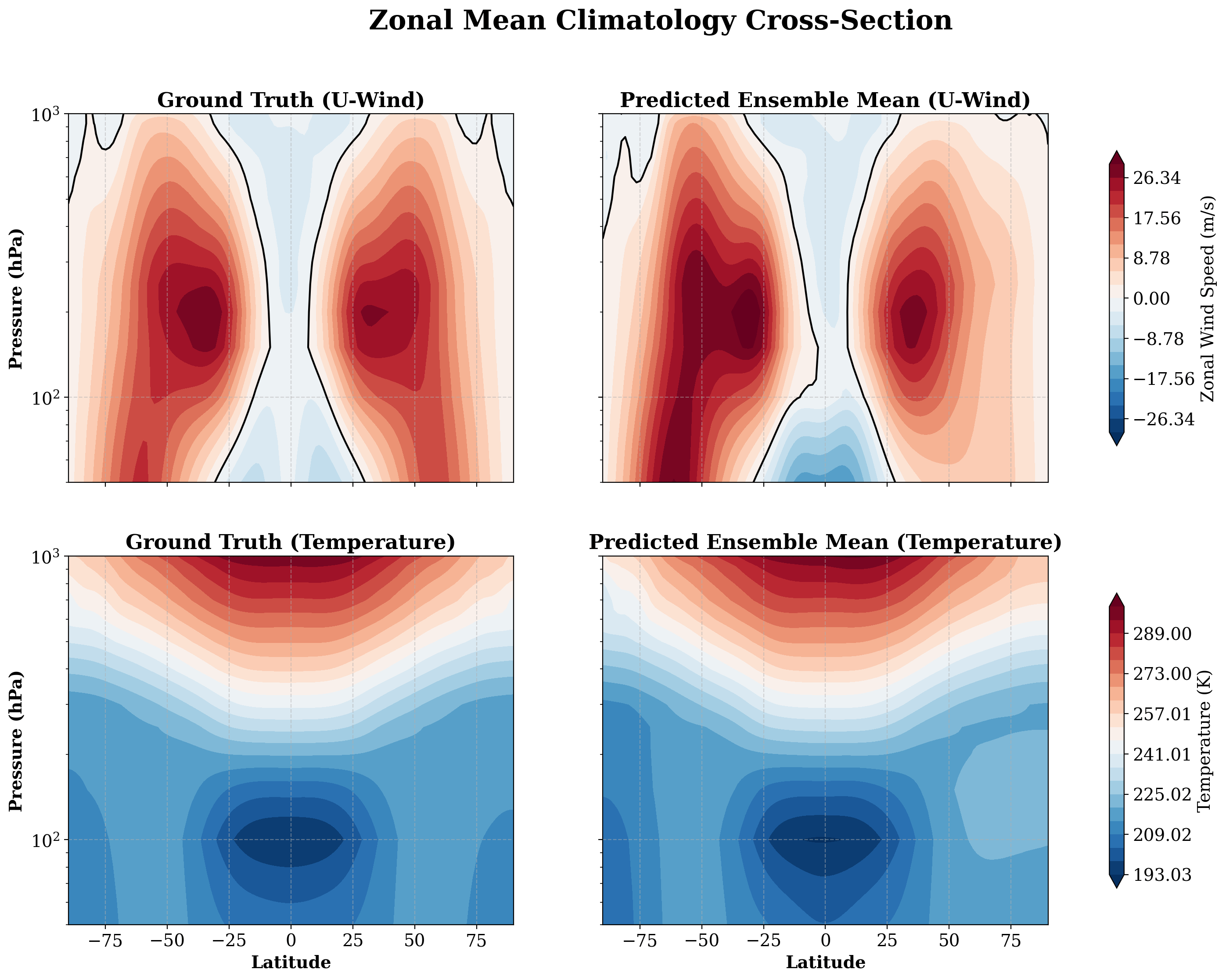}
\caption{Comparison of the \textbf{zonal mean climatological cross-section}. The top row shows Zonal Wind (U-Wind) and the bottom row shows Temperature. The left column is the Ground Truth, and the right column is the 40-member ensemble mean prediction. The model accurately reproduces the core structure of the subtropical jets and the atmospheric thermal gradient. A minor bias is observed in the predicted jet stream intensity, which appears slightly weaker than the ground truth.}
\label{fig:zonal_mean_cross_section_comparison_en}
\end{figure}

\paragraph{Structure of the Zonal Mean General Circulation:}
The zonal mean cross-section is a classic diagnostic tool for illustrating the fundamental structure of the atmospheric general circulation. \textbf{Fig.~\ref{fig:zonal_mean_cross_section_comparison_en}} compares the climatological vertical structure of the zonal mean zonal wind (U-Wind) and temperature. TritonCast accurately reproduces the core features of the atmospheric circulation, including the subtropical jet stream cores near the tropopause in both hemispheres, the polar night jet in the stratosphere, and both the latitudinal temperature gradient from the equator to the poles and the vertical temperature profile through the troposphere and stratosphere. Objectively, the simulated jet stream intensity is slightly weaker than the ground truth, which points to a direction for future model improvements. Overall, however, the model's depiction of the fundamental climate state of the atmosphere is of exceptionally high quality.

\clearpage

\section{Extended Results in Ocean Simulation and Forecasting}
\label{appendix:ocean_sim_fore}
\subsection{Evaluation of Mid-to-Short-Term Ocean Simulation Accuracy}

Although the core contribution of this study is to address the stability challenge in long-term integrations of AI models, a superior Earth system model must exhibit high fidelity across all time scales. To comprehensively evaluate TritonCast's intrinsic capability to reproduce ocean dynamics, this section presents a comparison of its accuracy against several mainstream baseline models (FourCastNet, AI-GOMS, ORCA\_DL, and WenHai) in 30-day atmospherically forced global ocean simulation experiments. These experiments, driven by ERA5 reanalysis data as the atmospheric forcing, are designed to isolate the performance from errors in atmospheric forecasts and to purely evaluate the ocean model's intrinsic performance.

\begin{figure}[h!]
\centering
\includegraphics[width=1\linewidth]{Appendix_figures/visual_rmse_10days.jpg}
\caption{\textbf{Comparison of Root Mean Square Error (RMSE) for key sea surface variables in short-term (10-day) ocean simulations.} From left to right: Sea Surface Salinity (SSS), Sea Surface Zonal Velocity ($\mathrm{U}_{\mathrm{o}}0$), Sea Surface Meridional Velocity ($\mathrm{V}_{\mathrm{o}}0$), Sea Surface Temperature (SST), and Sea Surface Height (SSH). The solid lines represent the mean over 240 simulations, and the shaded areas indicate the standard deviation range.}
\label{fig:visual_rmse_10days}
\end{figure}

First, we focus on the sea surface state during the initial phase of the simulation. \textbf{Fig.~\ref{fig:visual_rmse_10days}} shows the evolution of the Root Mean Square Error (RMSE) for Sea Surface Salinity (SSS), Zonal Velocity (U0), Meridional Velocity (V0), Sea Surface Temperature (SST), and Sea Surface Height (SSH) over the first 10 days. The results clearly indicate that for all five variables, TritonCast (dark blue solid line) exhibits the slowest RMSE growth, consistently remaining at the lowest level among all baseline models. This demonstrates that TritonCast reproduces the atmospherically driven sea surface dynamics with the highest accuracy, showcasing its strong competitive edge in short-term simulations.

\begin{figure}[h!]
\centering
\includegraphics[width=1\linewidth]{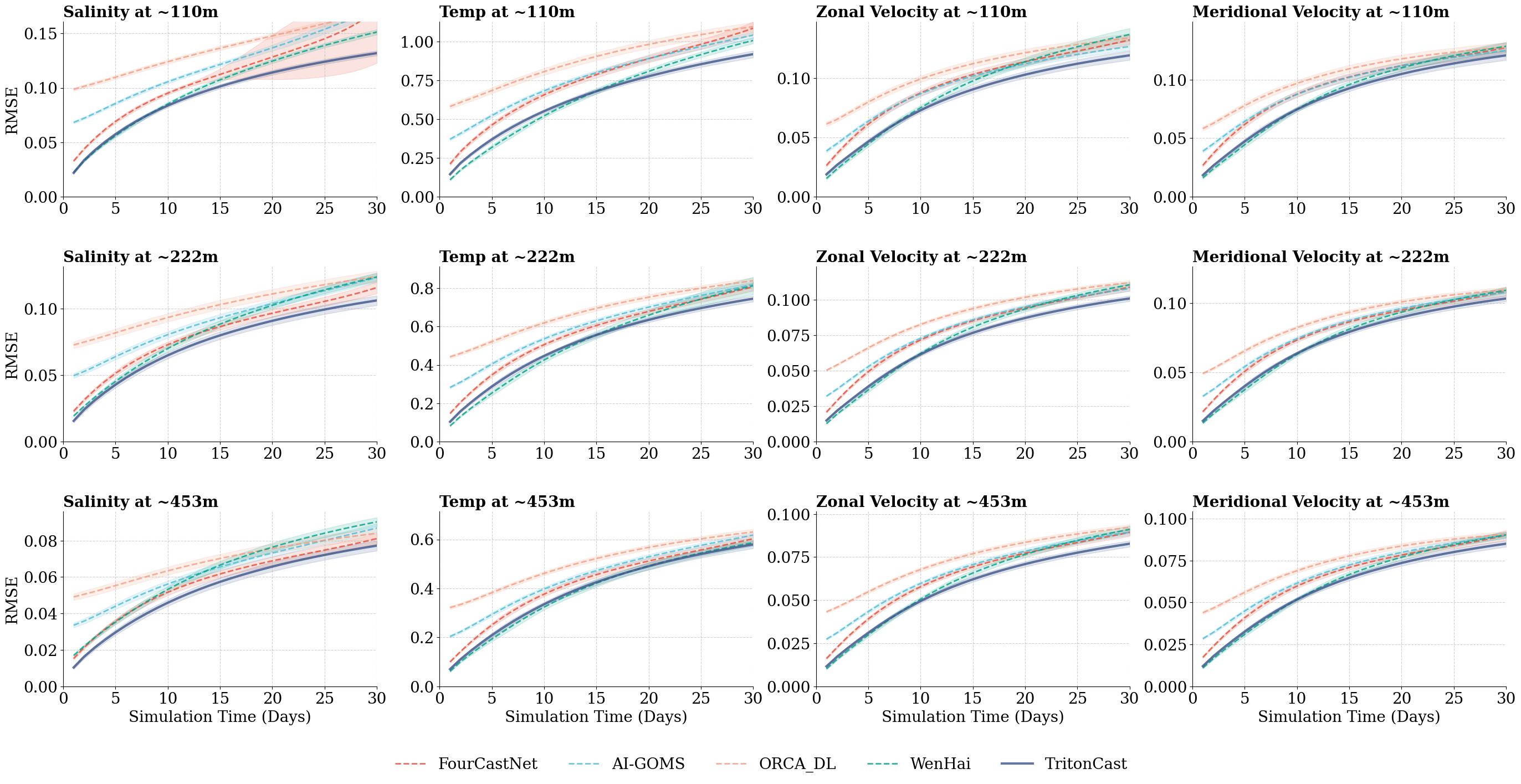}
\caption{\textbf{Comparison of Root Mean Square Error (RMSE) for deep-ocean variables in mid-term (30-day) ocean simulations.} The figure illustrates the temporal evolution of RMSE for salinity, temperature, zonal velocity, and meridional velocity at three different depth levels: approximately 110m, 222m, and 453m.}
\label{fig:rmse_all_variables_30days}
\end{figure}

A model's performance depends not only on the sea surface but also on its ability to capture the three-dimensional ocean structure. Therefore, we extend the simulation horizon to 30 days and evaluate the performance on deep-ocean variables. \textbf{Fig.~\ref{fig:rmse_all_variables_30days}} presents the simulation RMSE for key physical quantities at three depth levels: approximately 110m, 222m, and 453m. As shown, TritonCast's advantage extends from the short term into the deep ocean. Throughout the 30-day simulation period, its error accumulation in reproducing both the thermohaline structure and the three-dimensional velocity fields is significantly lower than that of other baseline models, demonstrating superior stability and fidelity.

In summary, these comprehensive simulation results provide strong evidence for the superiority of the TritonCast architecture. It is not only designed to solve the core challenge of long-term stability but also demonstrates state-of-the-art accuracy in mid-to-short-term simulation tasks. This high-fidelity simulation capability, demonstrated under an ideal forcing scenario, serves as a solid foundation for TritonCast's success in more challenging fully coupled forecasting tasks and its ultimate achievement of stable, long-term integrations.

\subsection{Physical Fidelity in Long-term Coupled Forecasting: A Multi-variable Spectral Analysis}

To provide a deeper and more comprehensive validation of TritonCast's physical fidelity in long-term coupled forecasting, this section presents a Power Spectral Density (PSD) analysis for multiple key ocean variables at various depths. PSD analysis is the gold standard for diagnosing spectral bias, as it reveals a model's ability to maintain the energy distribution across different spatial scales. The experimental setup here is a 60-day, fully AI-driven coupled forecast—where the ocean model is driven by forecasts from the TritonCast atmospheric model. The results are averaged over an ensemble of 240 different initial conditions, representing a highly stringent test of the models' performance.

\begin{figure}[h!]
\centering
\includegraphics[width=1\linewidth]{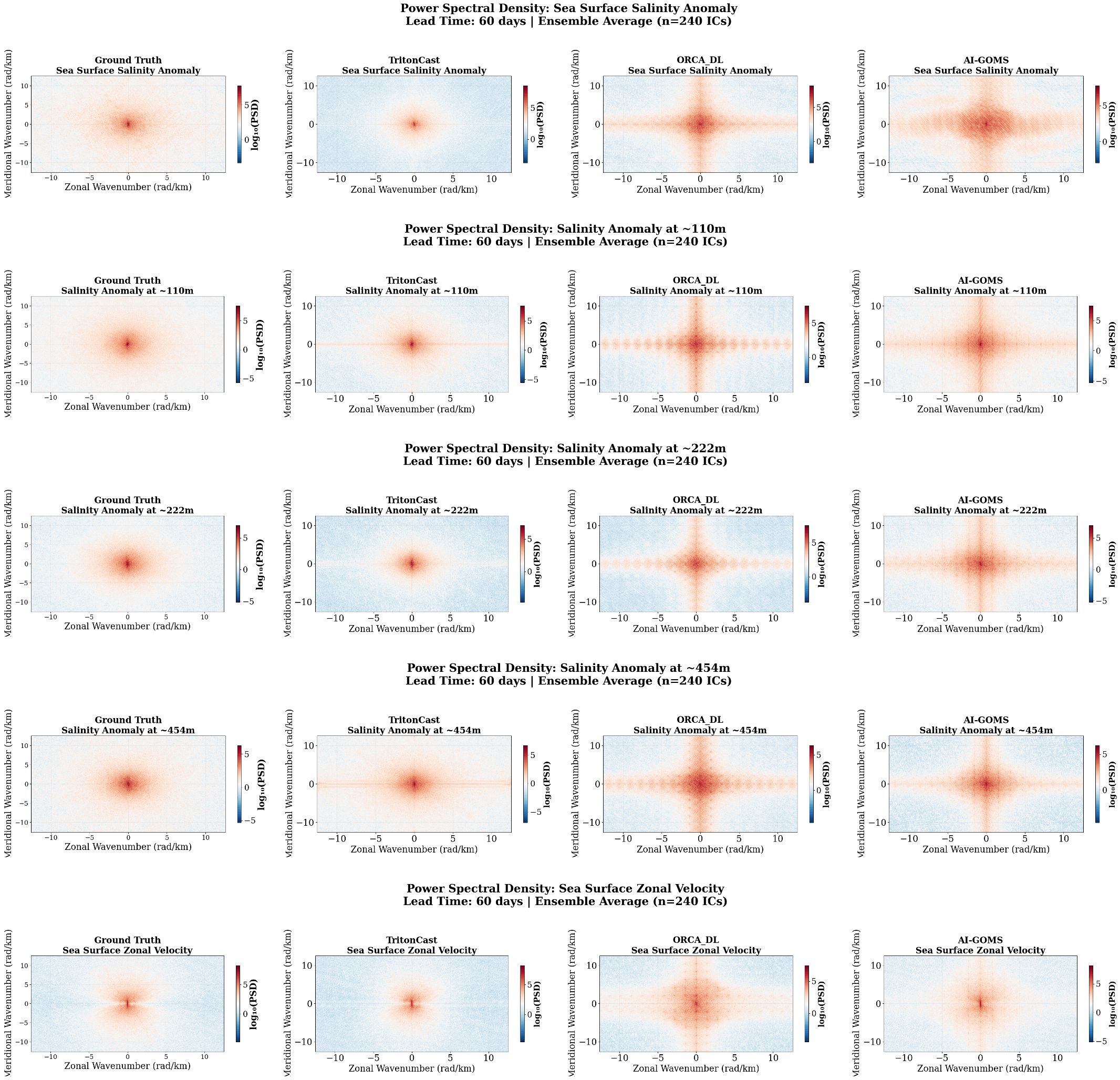}
\caption{\textbf{Power Spectral Density comparison for Salinity Anomaly and Zonal Velocity at the sea surface and various depths after a 60-day coupled forecast.}}
\label{fig:ocean_power_s1}
\end{figure}

\begin{figure}[h!]
\centering
\includegraphics[width=1\linewidth]{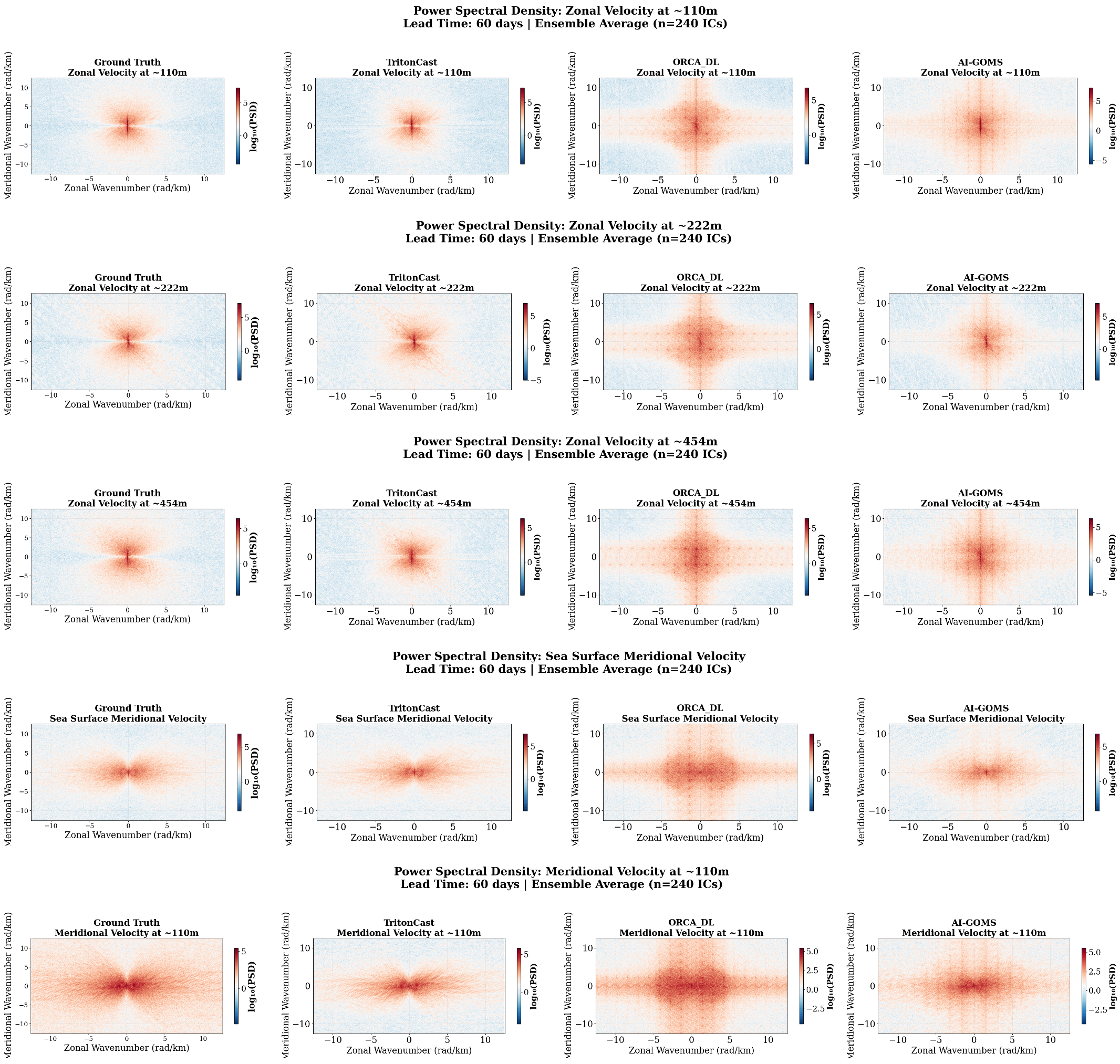}
\caption{\textbf{Power Spectral Density comparison for Zonal Velocity at deeper levels and Meridional Velocity at the sea surface and deeper levels after a 60-day coupled forecast.}}
\label{fig:ocean_power_s2}
\end{figure}

\begin{figure}[h!]
\centering
\includegraphics[width=1\linewidth]{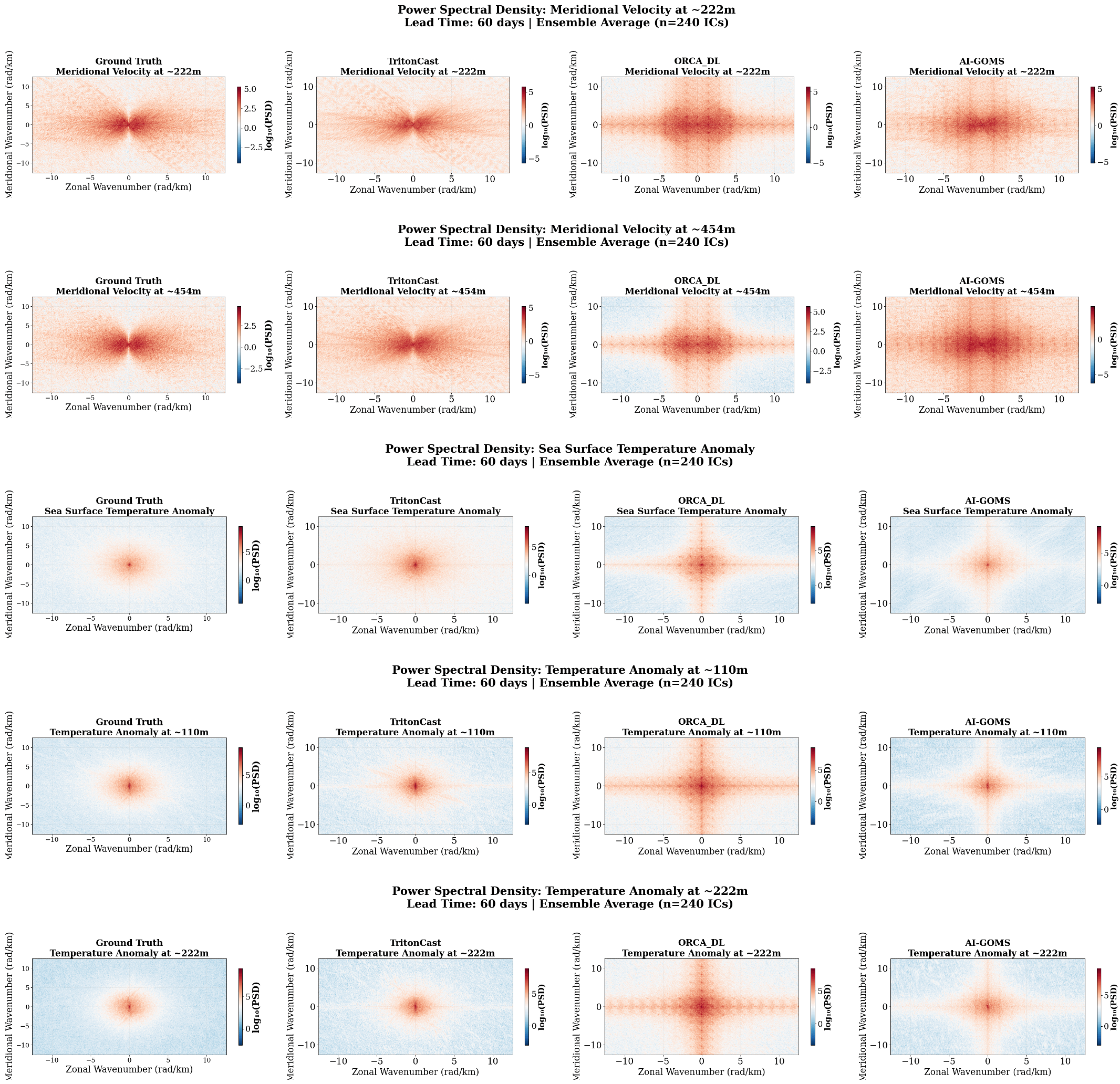}
\caption{\textbf{Power Spectral Density comparison for Meridional Velocity at deeper levels and Temperature Anomaly at the sea surface and deeper levels after a 60-day coupled forecast.}}
\label{fig:ocean_power_s3}
\end{figure}

\begin{figure}[h!]
\centering
\includegraphics[width=1\linewidth]{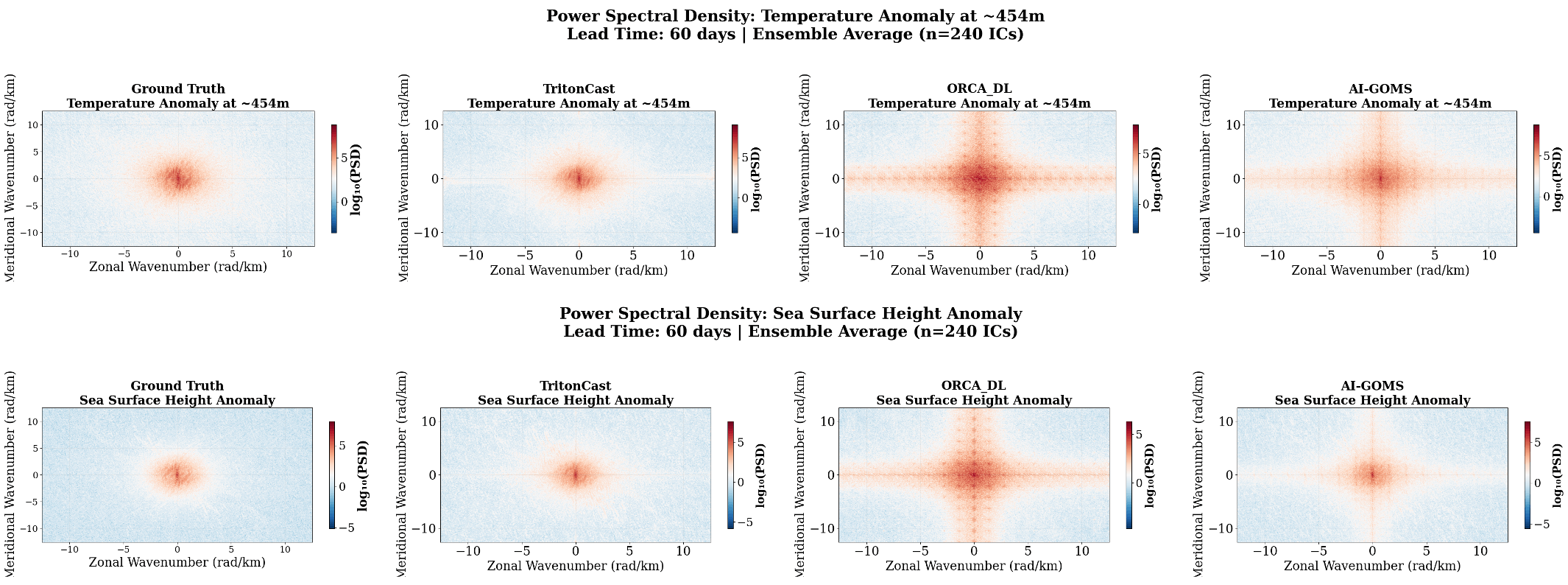}
\caption{\textbf{Power Spectral Density comparison for Temperature Anomaly at deeper levels and Sea Surface Height Anomaly after a 60-day coupled forecast.}}
\label{fig:ocean_power_s4}
\end{figure}

\textbf{Fig.~\ref{fig:ocean_power_s1}}, \textbf{Fig.~\ref{fig:ocean_power_s2}}, \textbf{Fig.~\ref{fig:ocean_power_s3}}, and \textbf{Fig.~\ref{fig:ocean_power_s4}} systematically present the PSD results for salinity anomaly, zonal/meridional velocity, temperature anomaly, and sea surface height anomaly at the surface and at depths of approximately 110m, 222m, and 453m. Analyzing these spectra leads to the following key conclusions:

\textbf{1. TritonCast maintains the full-spectrum energy with high fidelity.} Across all analyzed variables and depths, the power spectra of TritonCast's forecasts are visually almost identical to the Ground Truth. TritonCast accurately reproduces both the energy distribution in the low-wavenumber (large-scale) region and the smooth energy decay characteristics towards high wavenumbers (small-to-meso scales). This indicates that even after 60 days of autoregressive coupled forecasting, TritonCast continues to generate physically realistic ocean states that contain rich multi-scale structures, showing no signs of non-physical energy accumulation or dissipation.

\textbf{2. Baseline models exhibit severe spectral dissipation and numerical artifacts.} In stark contrast to TritonCast's excellent performance, the baseline models ORCA\_DL and AI-GOMS both show significant spectral bias.
\begin{itemize}
    \item \textbf{Spectral Dissipation}: The power spectra of AI-GOMS and ORCA\_DL exhibit a sharp decay of energy in the high-wavenumber region (far from the center), where their spectral density values are much lower than the ground truth. This is a typical over-smoothing phenomenon, signifying that the models lose a substantial amount of critical small-scale dynamical information, such as the fine structures of ocean eddies and fronts, during the iterative process. This finding corresponds directly to the blurry fields observed in \textbf{Fig.~\ref{Figure2_Ocean}c} of the main text.
    \item \textbf{Numerical Artifacts}: Particularly in the deep-variable spectra of ORCA\_DL (e.g., salinity at 454m in \textbf{Fig.~\ref{fig:ocean_power_s1}}), distinct grid-like or cross-shaped artificial patterns are evident. These structures, which are absent in real physical processes, are computational biases introduced by the model architecture itself and severely compromise the physical realism of the forecast results.
\end{itemize}

\textbf{3. Spectral fidelity is fundamental to achieving long-term stable forecasts.} This detailed spectral analysis provides solid evidence for the core argument of our study. The fundamental reason why baseline models suffer from rapid error accumulation and eventual collapse in long-term forecasts is their inherent spectral bias. This bias prevents them from accurately capturing small-scale processes and their energy exchange with large-scale motions, causing errors to be amplified non-linearly during iteration. TritonCast, through its innovative multi-grid hierarchical architecture, fundamentally mitigates the spectral bias problem, ensuring that energy is realistically maintained and transferred across all scales. This superior spectral fidelity constitutes the physical foundation for TritonCast's ability to perform long-term, stable, and reliable Earth system forecasting.

\subsection{Spatio-temporal Evolution of Physical Fields in Long-term Coupled Forecasting}

In addition to quantitative error and spectral analyses, a visual comparison of the spatio-temporal evolution of physical fields offers the most intuitive method for evaluating a model's long-term forecasting capabilities. This section presents a series of forecast results from 3 to 60 days for key sea surface variables (SSTa, SSSa, SSHa, $\mathrm{U}_{\mathrm{o}}0$, and $\mathrm{V}_{\mathrm{o}}0$) under the highly challenging AI-coupled forecasting scenario.

\begin{figure}[h!]
\centering
\includegraphics[width=1\linewidth]{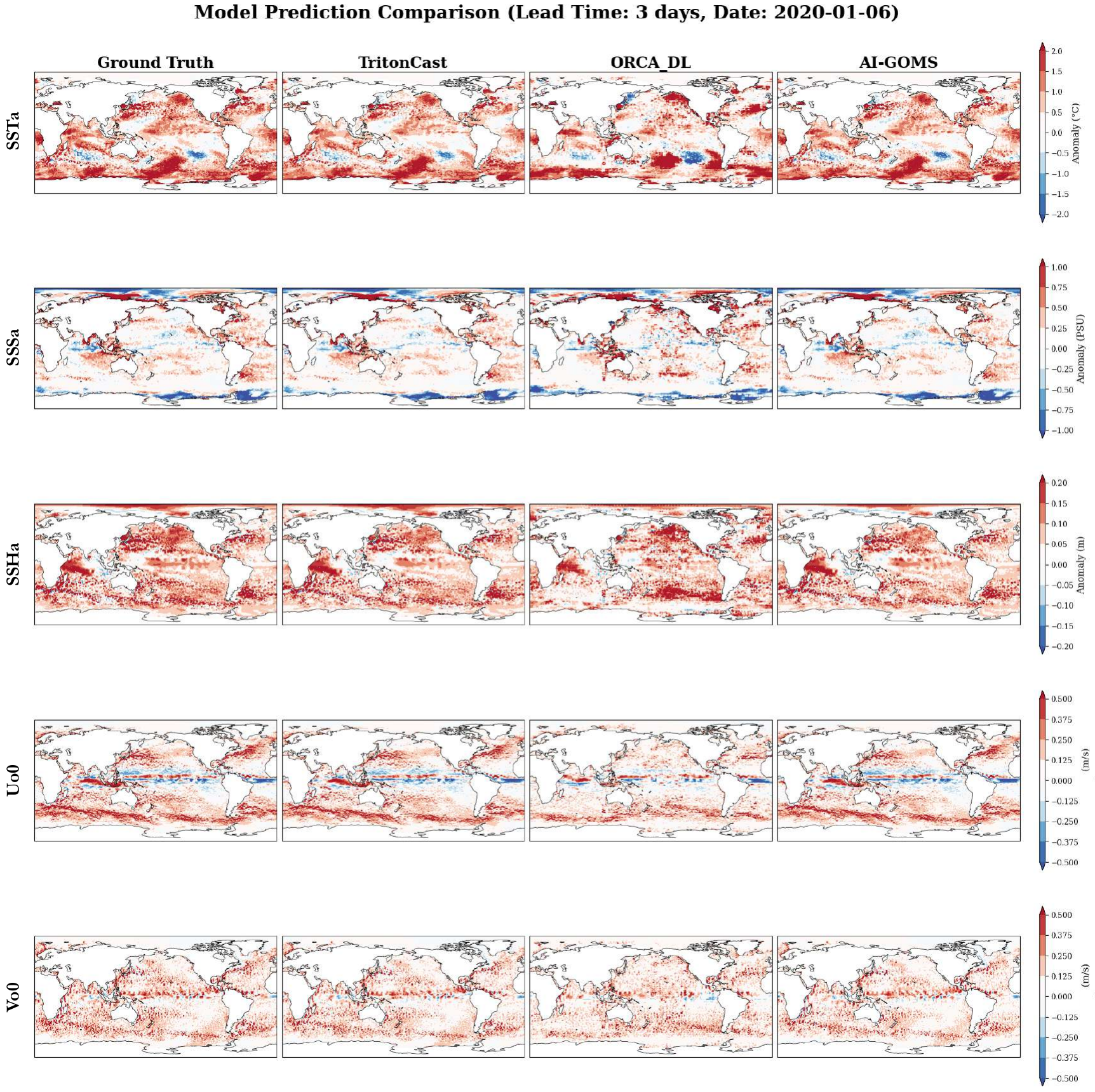}
\caption{\textbf{Comparison of physical fields for a 3-day forecast.}}
\label{fig:ocean_v1}
\end{figure}

\begin{figure}[h!]
\centering
\includegraphics[width=1\linewidth]{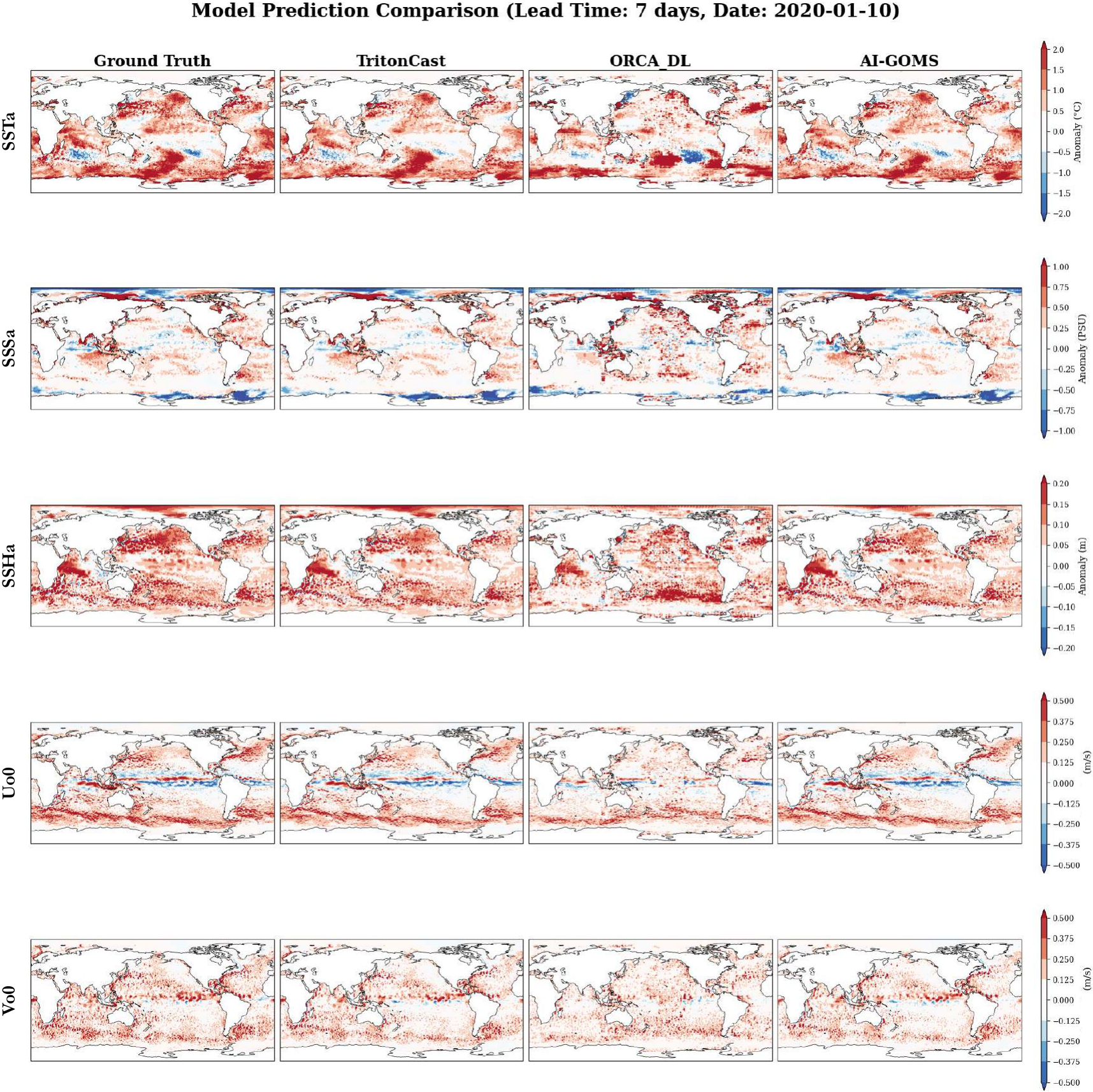}
\caption{\textbf{Comparison of physical fields for a 7-day forecast.}}
\label{fig:ocean_v2}
\end{figure}

\begin{figure}[h!]
\centering
\includegraphics[width=1\linewidth]{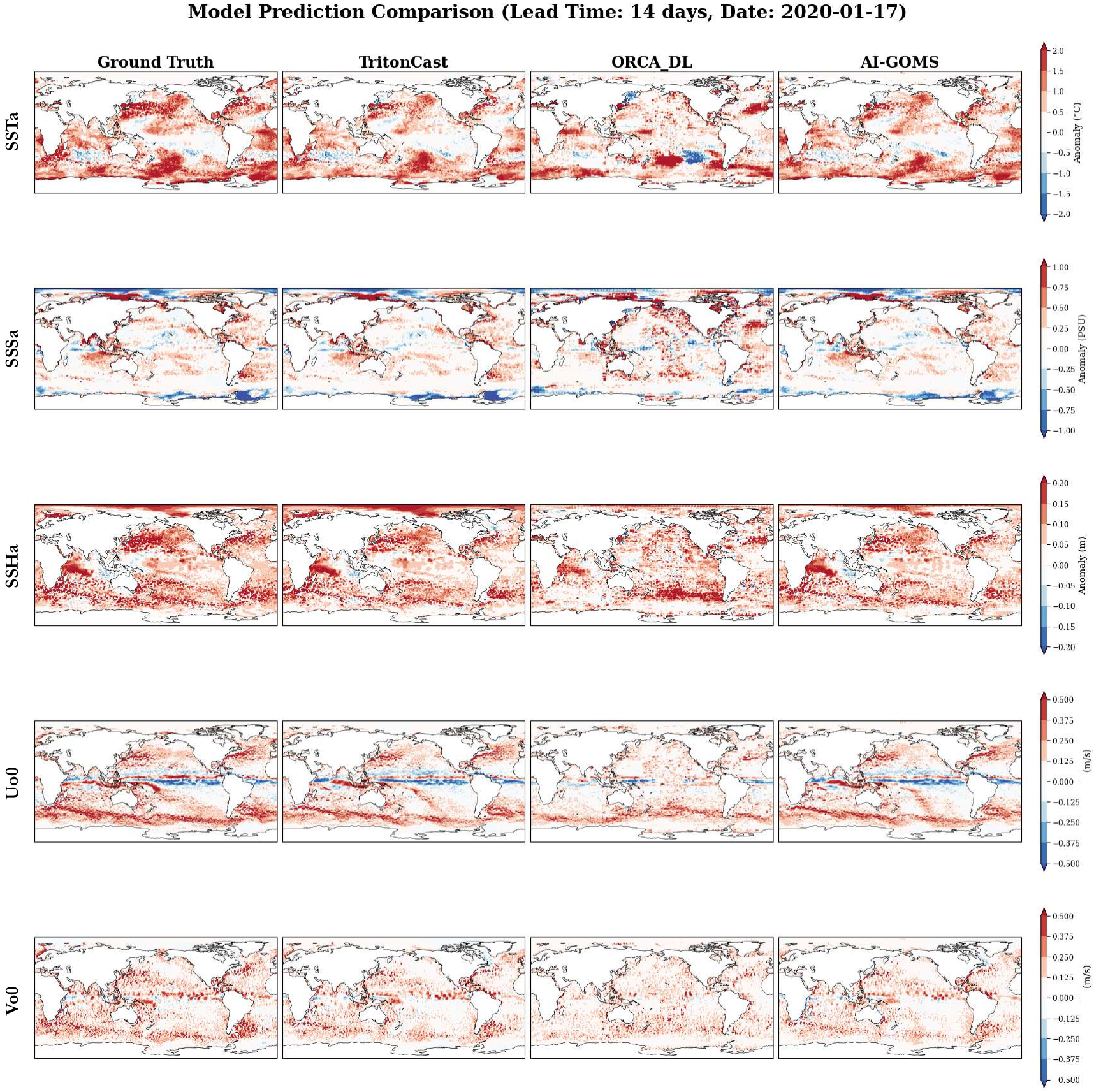}
\caption{\textbf{Comparison of physical fields for a 14-day forecast.}}
\label{fig:ocean_v3}
\end{figure}

\begin{figure}[h!]
\centering
\includegraphics[width=1\linewidth]{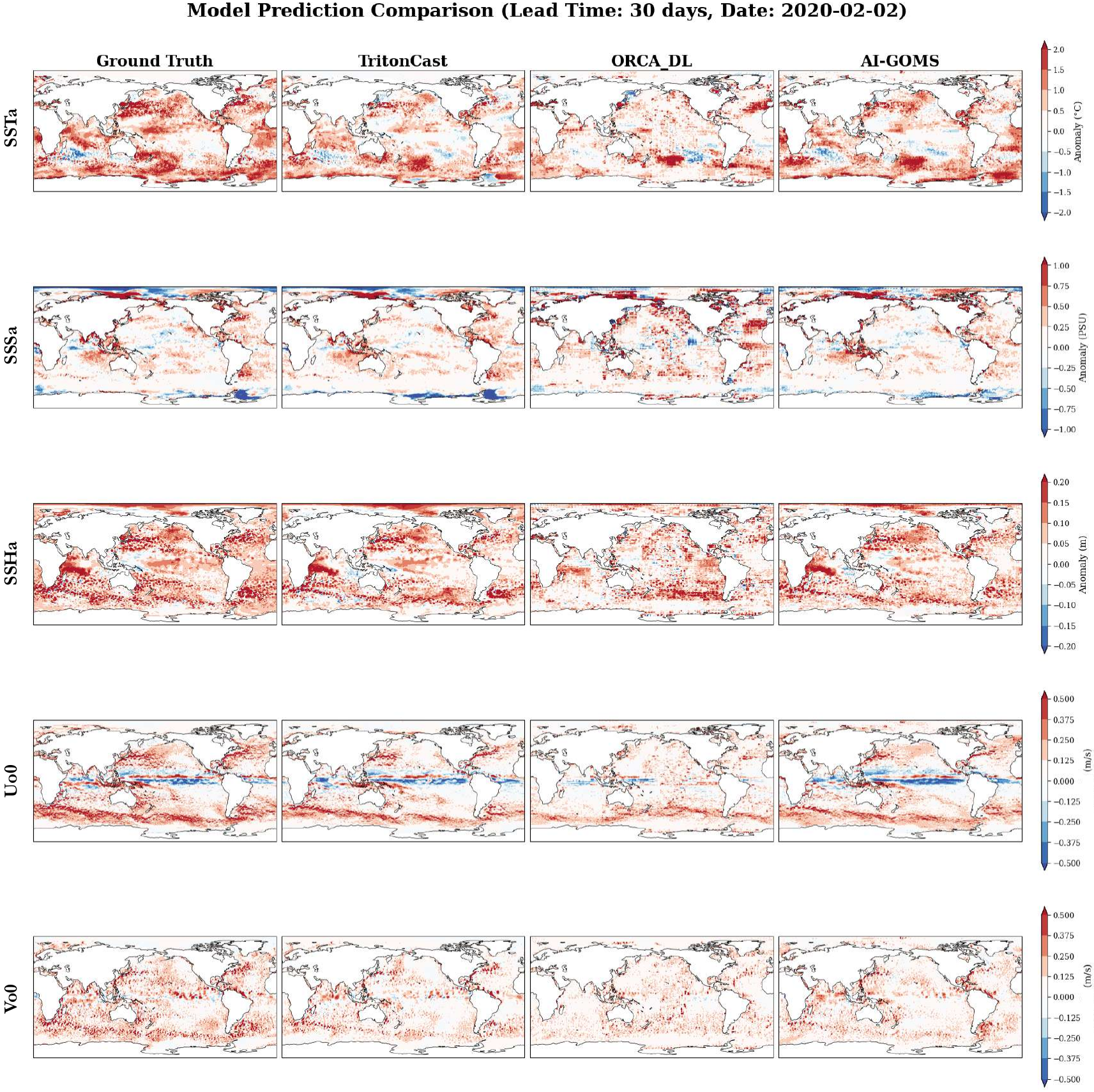}
\caption{\textbf{Comparison of physical fields for a 30-day forecast.}}
\label{fig:ocean_v4}
\end{figure}

\begin{figure}[h!]
\centering
\includegraphics[width=1\linewidth]{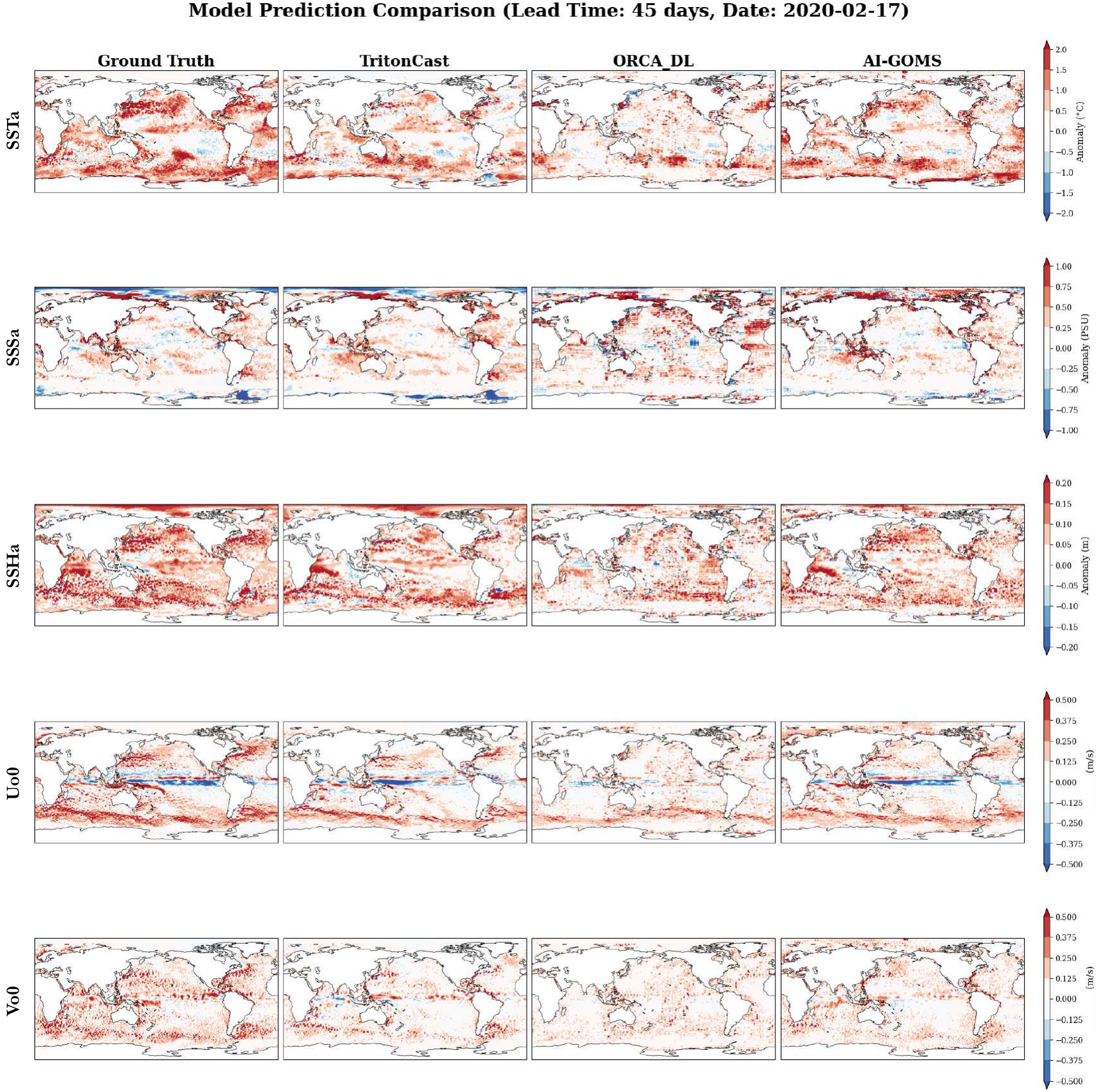}
\caption{\textbf{Comparison of physical fields for a 45-day forecast.}}
\label{fig:ocean_v5}
\end{figure}

\begin{figure}[h!]
\centering
\includegraphics[width=1\linewidth]{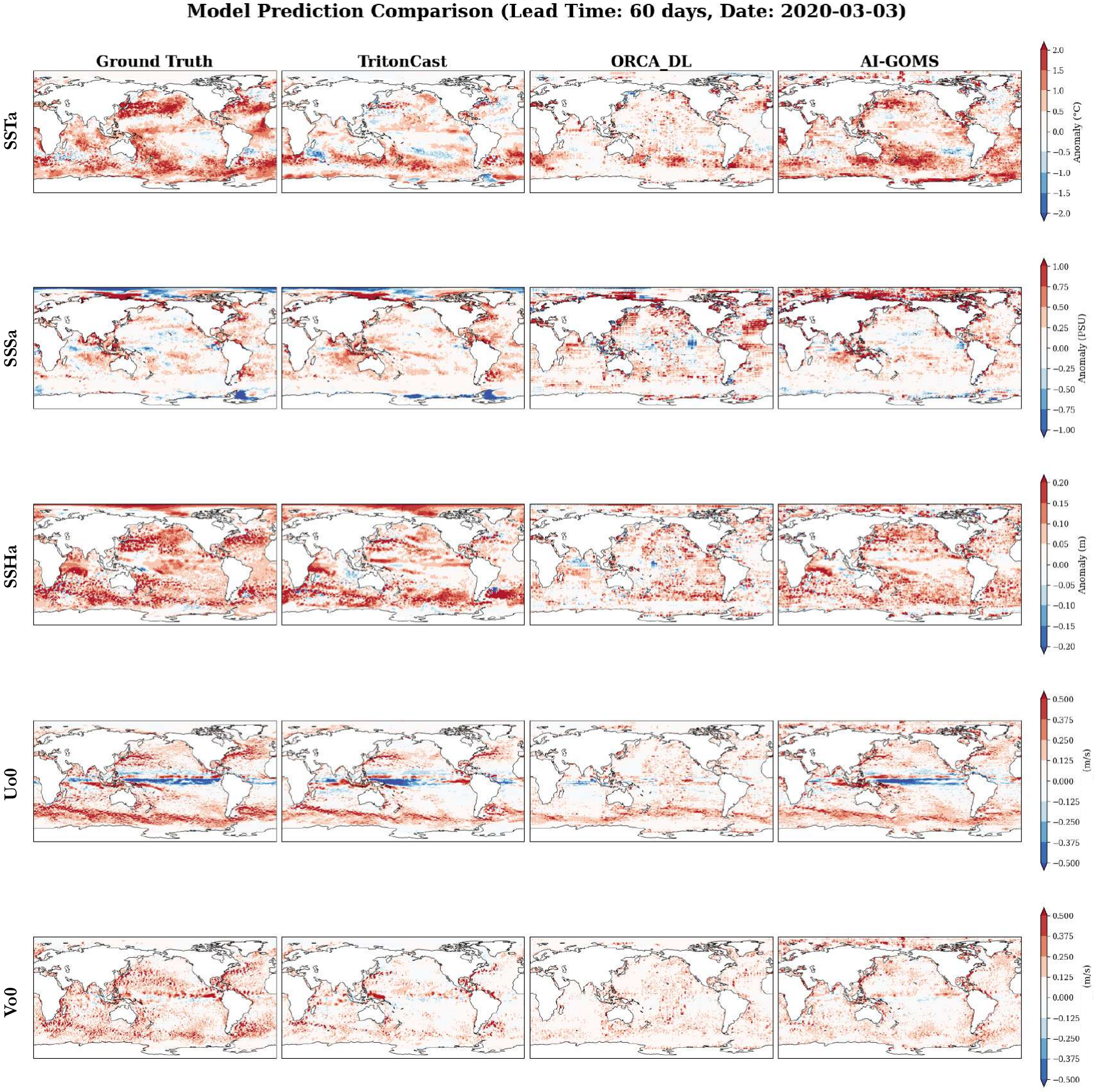}
\caption{\textbf{Comparison of physical fields for a 60-day forecast.}}
\label{fig:ocean_v6}
\end{figure}

By comparing the evolution across different lead times from \textbf{Fig.~\ref{fig:ocean_v1}} to \textbf{Fig.~\ref{fig:ocean_v6}}, we observe a significant divergence in the performance of the models:

\textbf{1. Short-term Forecasts (3-14 days):} In the initial forecast period, all models perform reasonably well, with their main modes aligning with the Ground Truth. However, TritonCast's predictions are visually the closest to the ground truth, almost indistinguishable in detail. Meanwhile, the baseline models already begin to show minor deviations: the fields from AI-GOMS are slightly smoother, while those from ORCA\_DL start to exhibit subtle noise in certain regions.

\textbf{2. Mid-to-Long-term Forecasts (30-60 days):} As the forecast lead time increases, the differences between the models become stark, intuitively revealing their fundamental disparities in long-term integration capability.
\begin{itemize}
    \item \textbf{TritonCast:} Even after 60 days, TritonCast's forecast fields maintain a high degree of physical realism. It accurately captures and sustains both the temperature and salinity anomalies in the equatorial Pacific (related to ENSO events) and the eddy activities in the mid-to-high latitudes (visible as fine-scale structures in the SSHa and velocity fields). This strongly demonstrates that its architecture effectively suppresses error accumulation and maintains the long-term stability of multi-scale dynamical structures.
    \item \textbf{AI-GOMS:} This model suffers from severe \textbf{spectral dissipation}. After 30 days, its forecast fields become excessively smooth, with nearly all small-to-meso-scale information lost. For example, in the SSHa and velocity fields, signals of ocean eddies are completely erased, leaving only large-scale, slowly varying structures. This observation perfectly aligns with the PSD analysis in the previous section and is a direct manifestation of the model's inability to retain high-frequency information.
    \item \textbf{ORCA\_DL:} This model veers to the other extreme,\textbf{numerical instability and model drift}. Starting from day 30, its forecast fields are populated with numerous unphysical artifacts and noise, particularly the large, spurious cold/warm patches in the SSTa field. By day 60, its forecast has completely diverged and shows no correlation with the ground truth, indicating a total collapse of the model integration.
\end{itemize}

In conclusion, this series of visualizations provides the most direct evidence supporting the core thesis of our study. TritonCast, through its unique architectural design, successfully overcomes the spectral bias and error amplification problems prevalent in long-term autoregressive forecasting with AI models, enabling it to generate long-range, stable, and physically realistic forecasts. In contrast, mainstream baseline models, due to their architectural limitations, either tend towards excessive smoothing or numerical collapse in long-term forecasting, rendering them unsuitable for reliable, long-range Earth system prediction tasks.

\subsection{Assessment of Extreme SST Event Prediction Skill}

\begin{figure}[h!]
\centering
\includegraphics[width=0.85\linewidth]{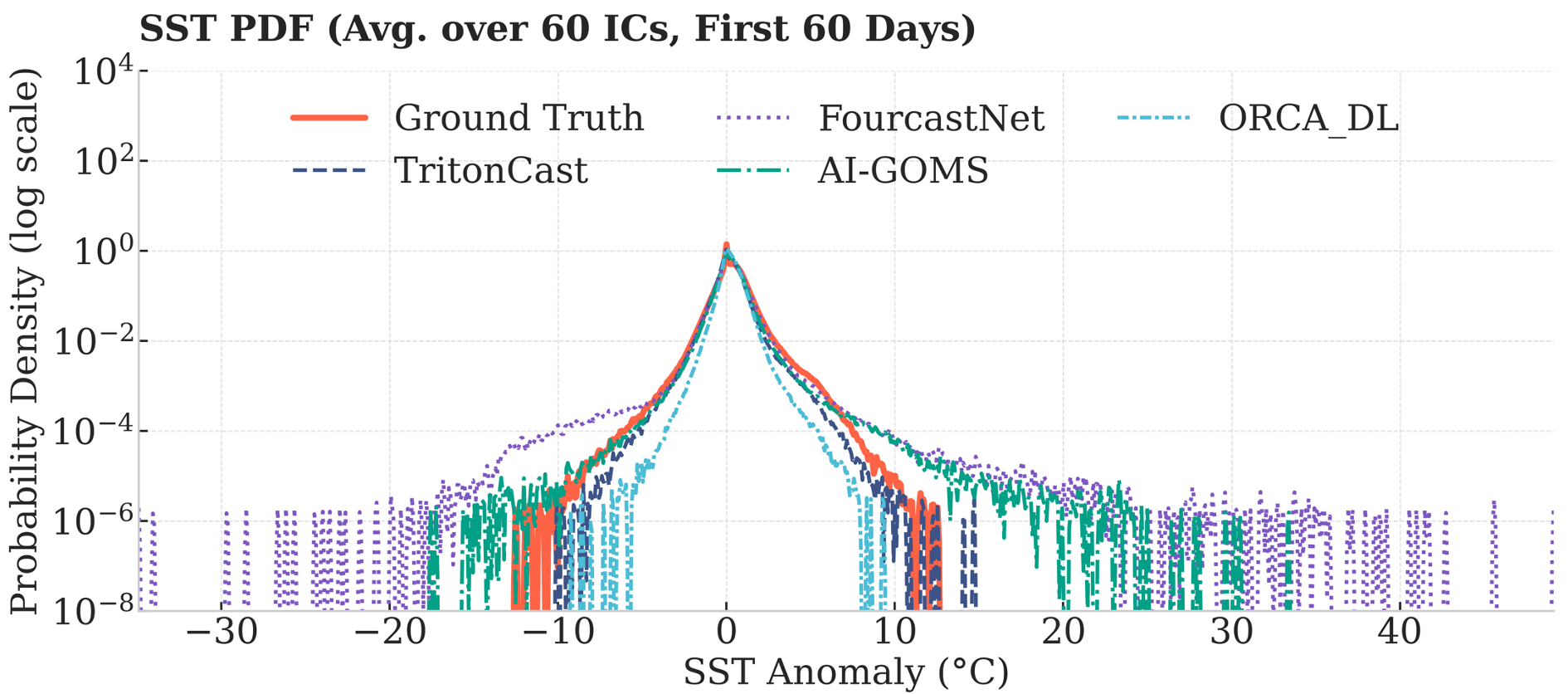}
\caption{\textbf{Comparison of Probability Density Functions (PDFs) for Sea Surface Temperature (SST) anomalies.} This figure shows the statistical distributions of SST anomalies from four AI models (TritonCast, AI-GOMS, FourcastNet, and ORCA\_DL) compared to the ground truth. The data are aggregated from the first 60 days of forecast results across 60 different initial conditions. The y-axis uses a logarithmic scale to clearly display the distribution tails, which represent extreme events such as marine heatwaves and cold spells. The results indicate that the distribution curve of TritonCast (dark blue dashed line) shows excellent agreement with the ground truth (red solid line) in both the central and tail regions, demonstrating its superior capability of capturing extreme events. In contrast, AI-GOMS and FourcastNet tend to overestimate the frequency of extreme events (a heavy-tailed distribution), whereas ORCA\_DL severely underestimates their frequency (a thin-tailed distribution).}
\label{fig:sst_pdf}
\end{figure}

To quantitatively assess the statistical prediction skill of various AI models for Sea Surface Temperature (SST) anomalies, particularly for extreme events, we aggregate forecast results from the first 60 days across 60 different initial conditions, as shown in \textbf{Fig.~\ref{fig:sst_pdf}}. After applying a land mask to ensure the analysis is confined to ocean regions, we randomly sample five million points from the hundreds of millions of data points generated by each model. We then employ the Kernel Density Estimation (KDE) method to compute the Probability Density Function (PDF). Finally, we plot all PDF curves on a single figure with a logarithmic y-axis, evaluating model performance by comparing the agreement of the distribution tails, which represent extreme events, with the ground truth.

\clearpage

\begin{figure}[h!]
\centering
\includegraphics[width=0.85\linewidth]{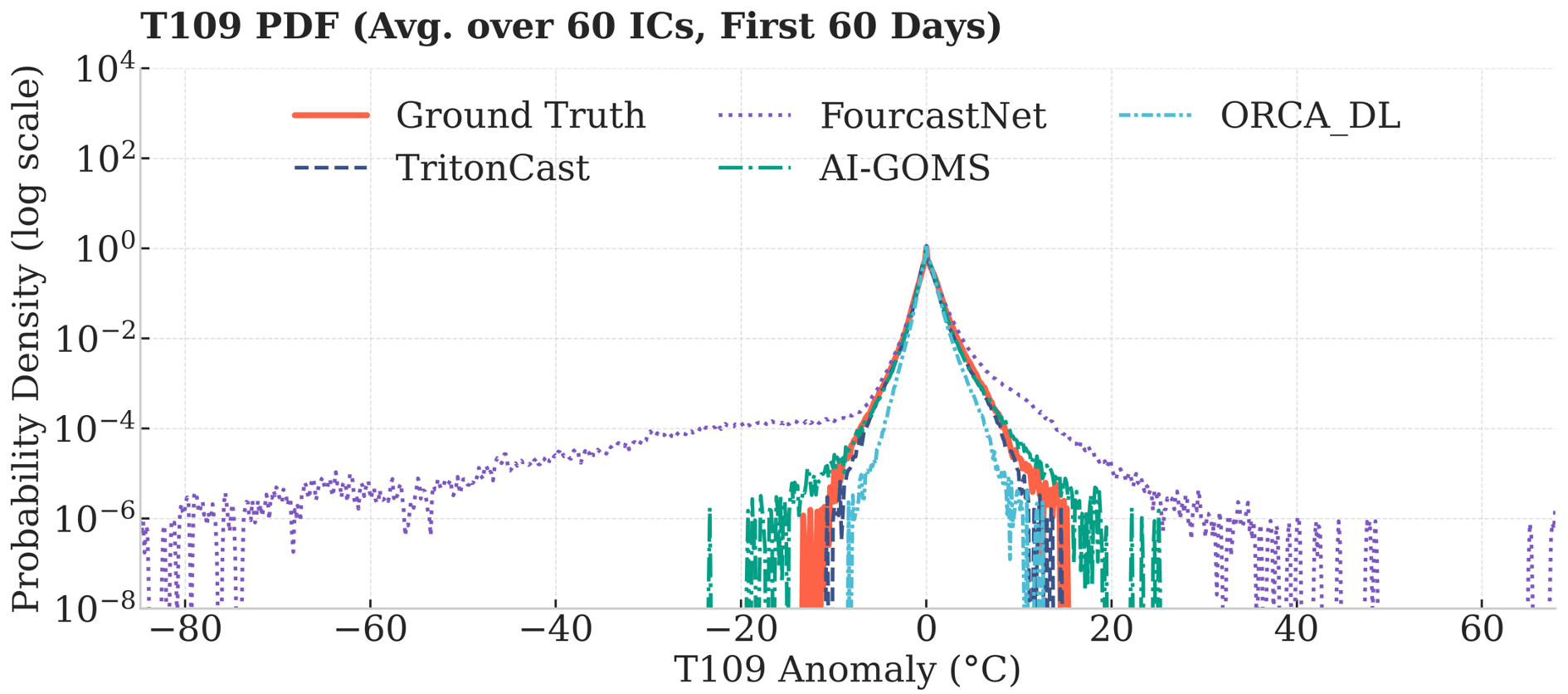}
\caption{\textbf{Comparison of Probability Density Functions (PDFs) for temperature anomalies at 100m depth (T109).} This figure presents the statistical distributions of T109 anomalies from four AI models against the ground truth. The data are similarly aggregated from the first 60 days of forecasts across 60 different initial conditions. The y-axis uses a logarithmic scale to highlight the distribution tails, which represent extreme deep-ocean temperature events. The results clearly show that TritonCast (dark blue dashed line) is the only model that maintains a high degree of agreement with the ground truth (red solid line) across the entire distribution. In comparison, the distribution of FourcastNet is severely distorted and vastly overestimates the frequency of extreme events, while both AI-GOMS and ORCA\_DL exhibit typical thin-tailed characteristics, significantly underestimating the frequency of extremes.}
\label{fig:t_pdf}
\end{figure}

As shown in \textbf{Fig.~\ref{fig:t_pdf}}, for subsurface temperature anomalies at a depth of 100 meters (T109), the performance differences among the models are significantly amplified. TritonCast once again demonstrates a high degree of consistency with the ground truth in its overall statistical distribution and is the only model that reasonably captures the probability of extreme deep ocean warming and cooling events. In sharp contrast, the distribution of FourcastNet is severely distorted, exhibiting extremely heavy tails, which indicates that the model generates a large amount of unrealistic and physically implausible extreme temperature noise. The other two models, AI-GOMS and ORCA\_DL, show a consistent conservative behavior. Their distributions exhibit a typical thin-tailed shape with an overly sharp peak, which means they severely underestimate the frequency and intensity of extreme deep-sea temperature events. This result suggests that when simulating the more complex internal ocean dynamics, TritonCast maintains excellent stability and physical consistency, while the other models reveal significant deficiencies.

\clearpage

\section{In-Depth Validation of TritonCast for Ocean Current Forecasting}
\label{appendix:full_ocean_stream_exp}
This appendix provides a comprehensive and in-depth evaluation to validate the superior performance of the TritonCast model in the long-term prediction of ocean circulation, particularly mesoscale eddy dynamics. Our analysis proceeds from a global, coarse-resolution assessment to a regional, high-resolution examination. First, we assess the stability and spatial accuracy of the global 0.25° resolution model over an extended one-month integration to establish its reliability as a foundational model. Subsequently, we shift our focus to the most challenging task: employing the 0.125° high-resolution model to perform eddy-resolving simulations for several months in three of the world's most energetic western boundary current regions. This is done to demonstrate the model's ultimate capability in capturing high-fidelity physical processes.

\subsection{Supplementary Evaluation of the Global $0.25\degree$ Resolution Model}

In this section, we aim to answer a core question: Is TritonCast a physically self-consistent and long-term stable forecasting model at the global scale? To this end, we conduct our evaluation from two perspectives: first, whether the model adheres to the fundamental principles of fluid dynamics, specifically how energy is distributed and transferred across different scales; and second, whether the model can accurately reproduce the well-known spatial structures of global ocean circulation.

\subsubsection{Long-Term Stability of the Kinetic Energy Spectrum: A Test of Physical Consistency}

In any long-term autoregressive forecasting task, a central challenge is the accumulation of errors. In physical systems, this error often manifests as an unrealistic accumulation or excessive dissipation of energy at specific scales (typically small scales), ultimately causing the simulation to "crash" or become overly smooth. Kinetic energy spectrum analysis is the gold standard for diagnosing this issue. It quantitatively illustrates the distribution of energy across all spatial scales. A physically reliable model should produce a predicted kinetic energy spectrum that closely matches the spectral structure of the ground truth over extended periods.

\begin{figure}[h!]
\centering
\includegraphics[width=1\linewidth]{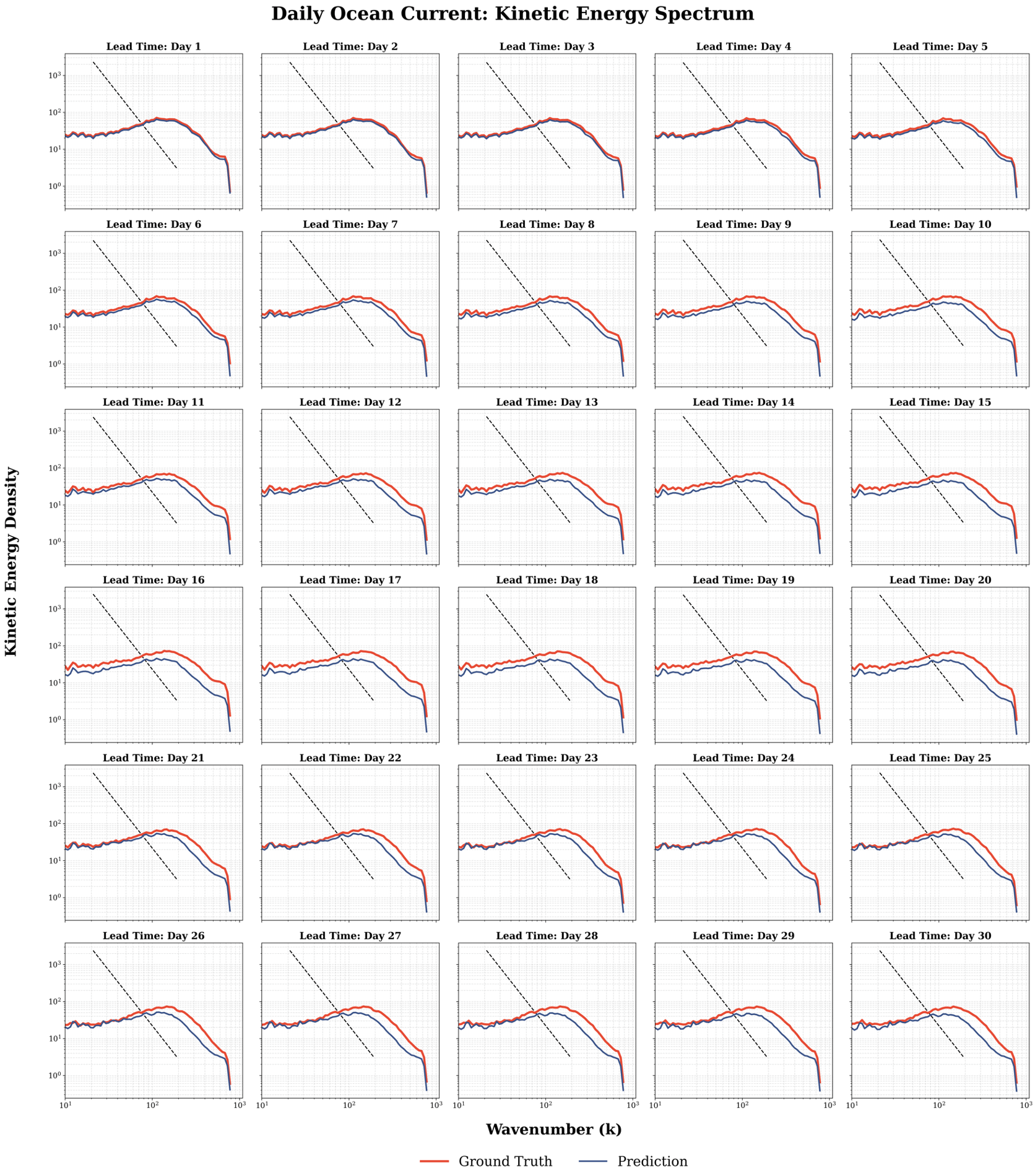}
\caption{\textbf{Daily evolution of the Kinetic Energy Spectrum for the global $0.25\degree$ ocean model from day 1 to day 30.} The prediction from TritonCast (blue solid line) is compared with the ground truth (red solid line), while the black dashed line indicates a reference spectral slope. It is evident that over a continuous 30-day forecast, the kinetic energy spectrum predicted by TritonCast maintains excellent agreement with the ground truth across all wavenumbers (spatial scales), demonstrating the model's superior long-term integration stability and physical fidelity.}
\label{fig:esd_ocean_global_025}
\end{figure}

\textbf{Fig.~\ref{fig:esd_ocean_global_025}} presents a remarkable result. Over a continuous 30-day autoregressive forecast, the kinetic energy spectrum predicted by TritonCast (blue solid line) almost perfectly coincides with that of the ground truth (red solid line). This indicates that TritonCast not only performs excellently in the initial days but also exhibits no discernible spectral deviation for up to a month. This implies that the model's internal dynamical evolution is highly physically self-consistent. It correctly handles the transfer and dissipation of energy across scales, thereby laying a solid physical foundation for long-term, stable integration.

\subsubsection{Spatial Accuracy of Global Currents and Eddy Fields: Reproducing the Ocean's Structure}
Having established the model's physical consistency, we now examine the accuracy of its spatial predictions. \textbf{Figs.~\ref{fig:global_speed_comparison_kuroshio_case1_leadtime10d}} and\textbf{~\ref{fig:global_speed_comparison_kuroshio_case4_leadtime30d}} show the spatial distribution of the global sea current velocity field at 10-day and 30-day forecast lead times, respectively. These figures clearly illustrate that TritonCast accurately captures all major global current systems, such as the Kuroshio, the Gulf Stream, the equatorial current systems, and the Antarctic Circumpolar Current. Crucially, even after 30 days, the core locations, strengths, and widths of these current systems remain in high agreement with the ground truth.

However, the ocean is characterized not only by stable mean currents but also by vibrant mesoscale eddies. Eddy Kinetic Energy (EKE) is a key metric for quantifying the intensity of eddy activity. \textbf{Fig.~\ref{fig:ocean_current_eke_comparison}} compares the global EKE distribution between the ground truth and the TritonCast forecast. This represents a more demanding test, as it requires the model to predict not only 'where the currents are' but also 'where the variability is strongest'. As shown, TritonCast precisely identifies the global 'hotspots' of EKE, including the extensions of major western boundary currents and the Southern Ocean storm tracks. This result provides strong evidence that TritonCast possesses the capability to accurately delineate the spatiotemporal distribution of global mesoscale eddies.

\begin{figure}[h!]
\centering
\includegraphics[width=1\linewidth]{Appendix_figures/global_speed_comparison_kuroshio_case1_leadtime10d.png}
\caption{\textbf{Comparison of the global sea current velocity field at a 10-day forecast lead time.} The left panel shows the ground truth, and the right panel presents the prediction from TritonCast. The model accurately captures the spatial structure of the major global current systems, particularly in western boundary current regions such as the Kuroshio, where its velocity, path, and width are precisely reproduced.}
\label{fig:global_speed_comparison_kuroshio_case1_leadtime10d}
\end{figure}

\begin{figure}[h!]
\centering
\includegraphics[width=1\linewidth]{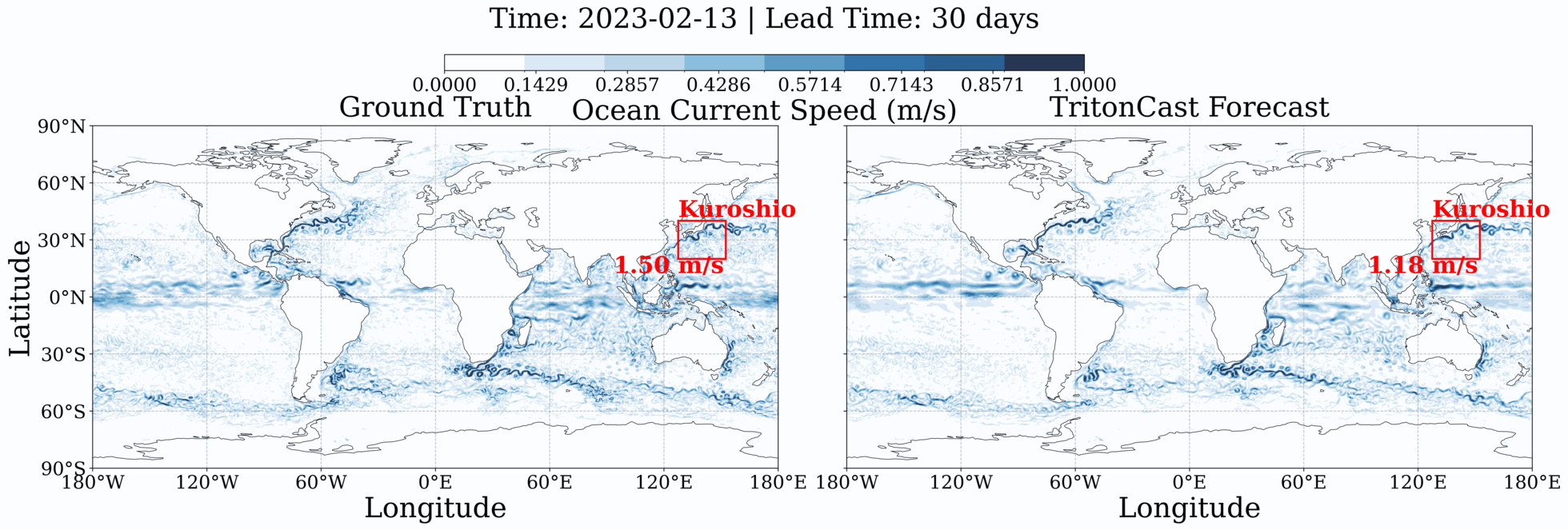}
\caption{\textbf{Comparison of the global sea current velocity field at a 30-day forecast lead time.} Even after a forecast extending to one month, the prediction from TritonCast (right panel) maintains a high degree of consistency with the ground truth (left panel) in its macroscopic structure, demonstrating the model's powerful capability for large-scale dynamical prediction.}
\label{fig:global_speed_comparison_kuroshio_case4_leadtime30d}
\end{figure}

\begin{figure}[h!]
\centering
\includegraphics[width=0.65\linewidth]{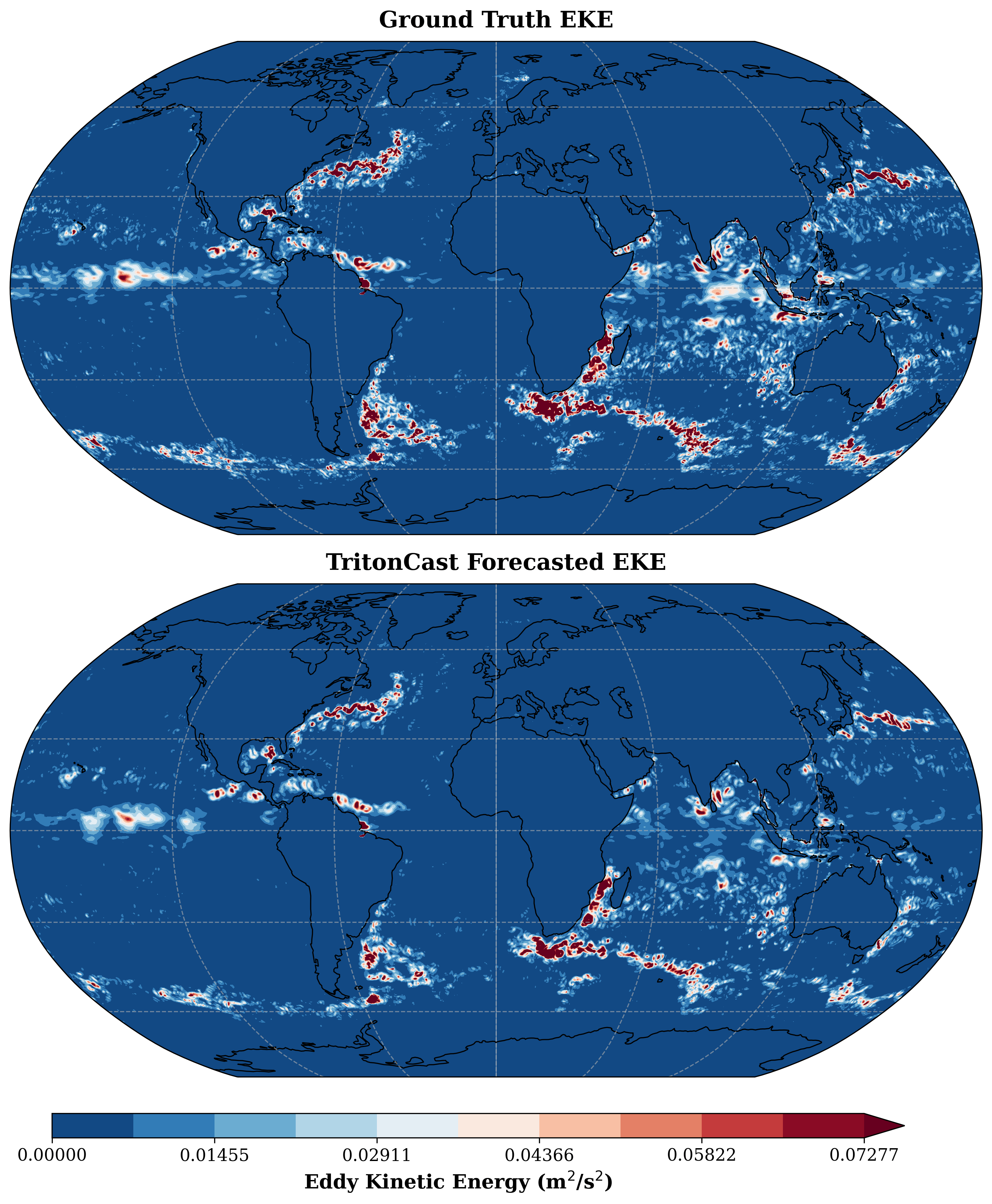}
\caption{\textbf{Climatological comparison of the global Eddy Kinetic Energy (EKE) spatial distribution, averaged over a 30-day forecast period across multiple initial cases.} The top panel shows the ground truth, and the bottom panel presents the prediction from TritonCast. EKE characterizes the activity level of mesoscale eddies, representing regions of high oceanic variability. The figure clearly demonstrates that TritonCast accurately reproduces this climatological pattern, identifying the 'hotspots' of intense eddy activity globally, such as the Kuroshio Extension, the Gulf Stream, the Agulhas Current, and the Antarctic Circumpolar Current. This capability to capture the statistical properties of ocean turbulence is crucial for robust climate projections and for studying ocean energy transport.}
\label{fig:ocean_current_eke_comparison}
\end{figure}

\subsection{High-Fidelity Eddy-Resolving Forecasts with the $0.125\degree$ High-Resolution Model in Key Regions}
\label{appendix:Key_Regions}
The success of the global model demonstrates the robustness of TritonCast, but its true transformative potential lies in its ability to capture fine-scale physical processes. To this end, we conducted the most rigorous test in this study: performing continuous forecasts for up to 180 days at a high resolution of $0.125\degree$ for three of the most dynamically complex and energetic regions of the global ocean—the Agulhas Current, the Gulf Stream, and the Kuroshio Extension. This task demands that the model not only remains stable but also precisely captures every instance of eddy generation, evolution, merging, and decay.

\begin{figure}[h!]
\centering
\includegraphics[width=1\linewidth]{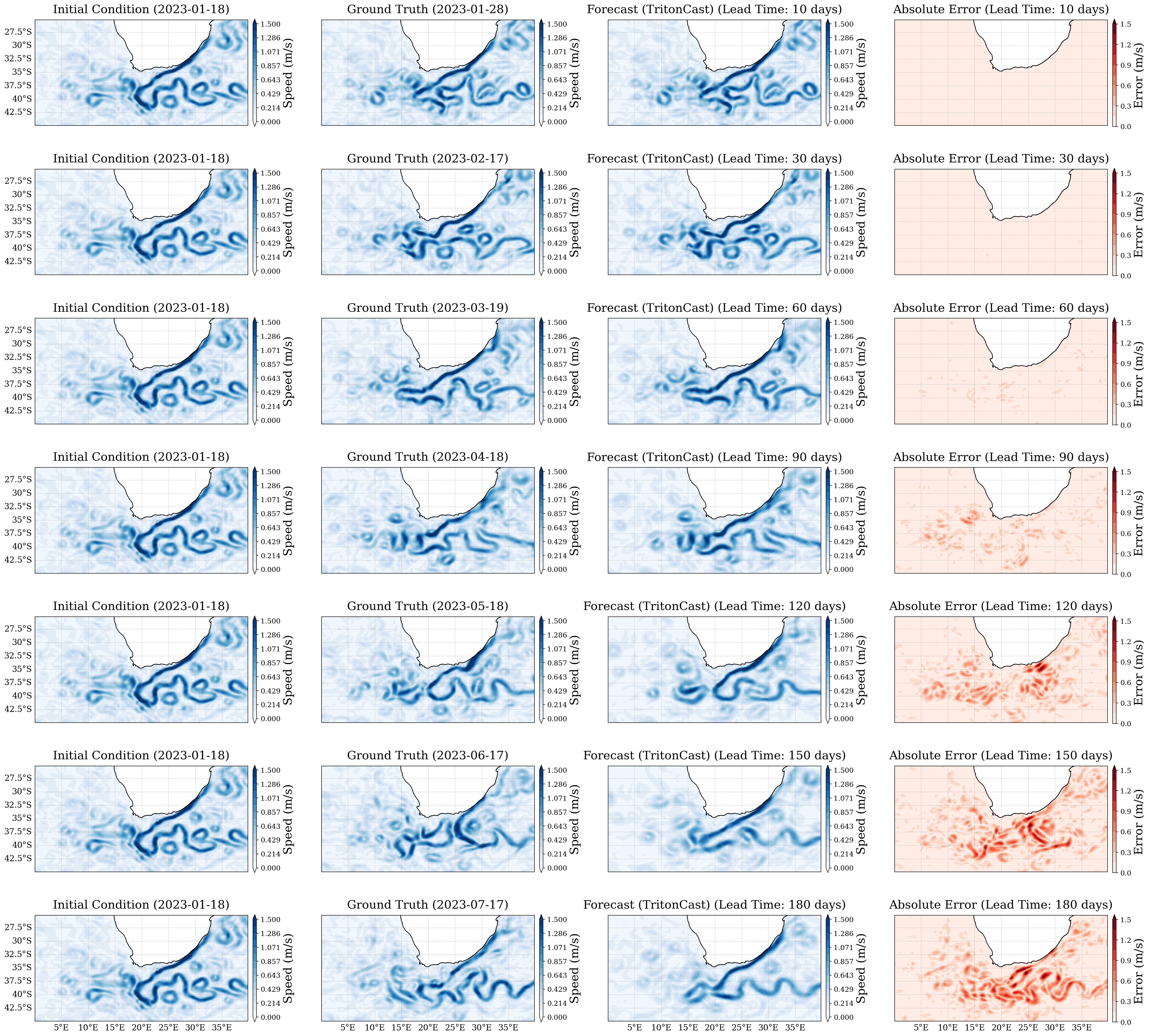}
\caption{\textbf{High-fidelity forecast for the Agulhas Current region at different lead times (from 10 to 180 days).} Each row shows a forecast time step, displaying, from left to right: the initial field, the ground truth, the TritonCast forecast, and the absolute error. The figure clearly demonstrates the model's capability for long-term, precise tracking of complex processes such as the Agulhas retroflection and eddy shedding.}
\label{fig:agu}
\end{figure}

\begin{figure}[h!]
\centering
\includegraphics[width=1\linewidth]{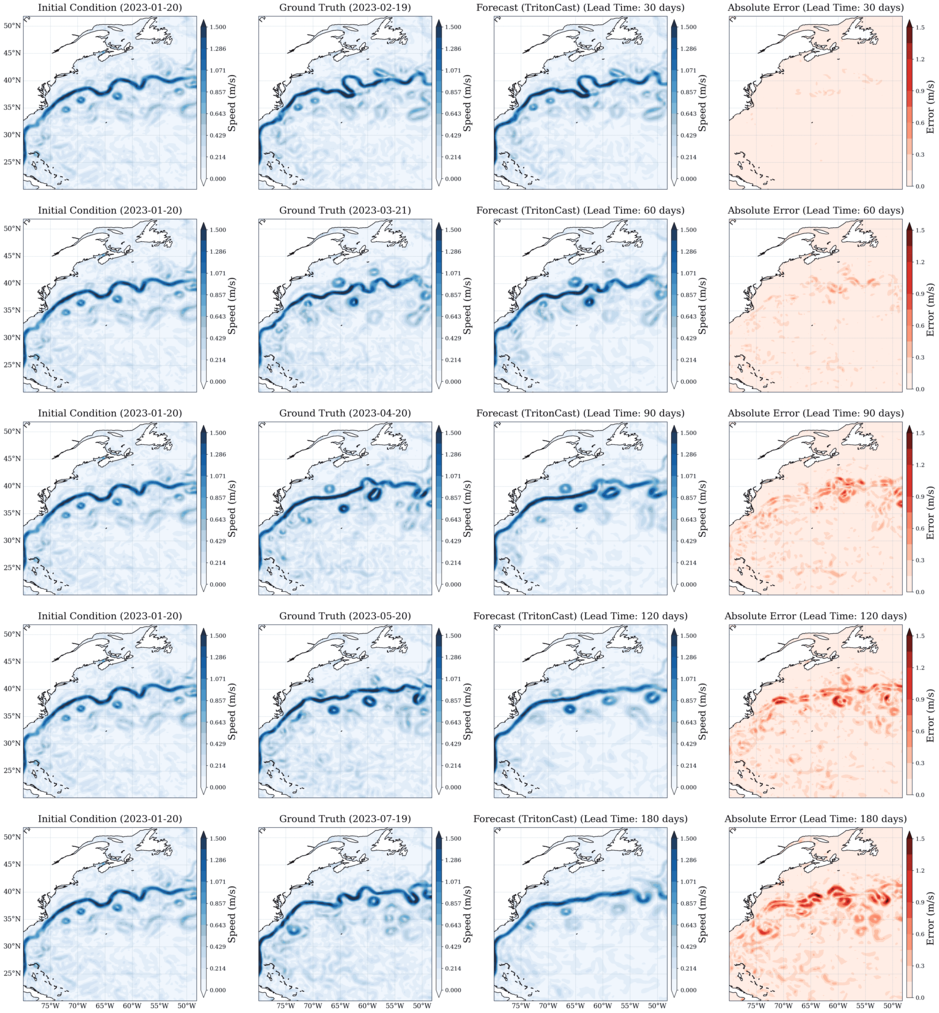}
\caption{\textbf{High-fidelity forecast for the Gulf Stream region at different lead times (from 30 to 180 days).} The figure illustrates TritonCast's precise simulation of the meandering path of the Gulf Stream and the generation and evolution of its warm- and cold-core eddies. Even after several months of integration, the predicted eddy morphology and position remain in high agreement with the ground truth.}
\label{fig:gulf}
\end{figure}

\begin{figure}[h!]
\centering
\includegraphics[width=1\linewidth]{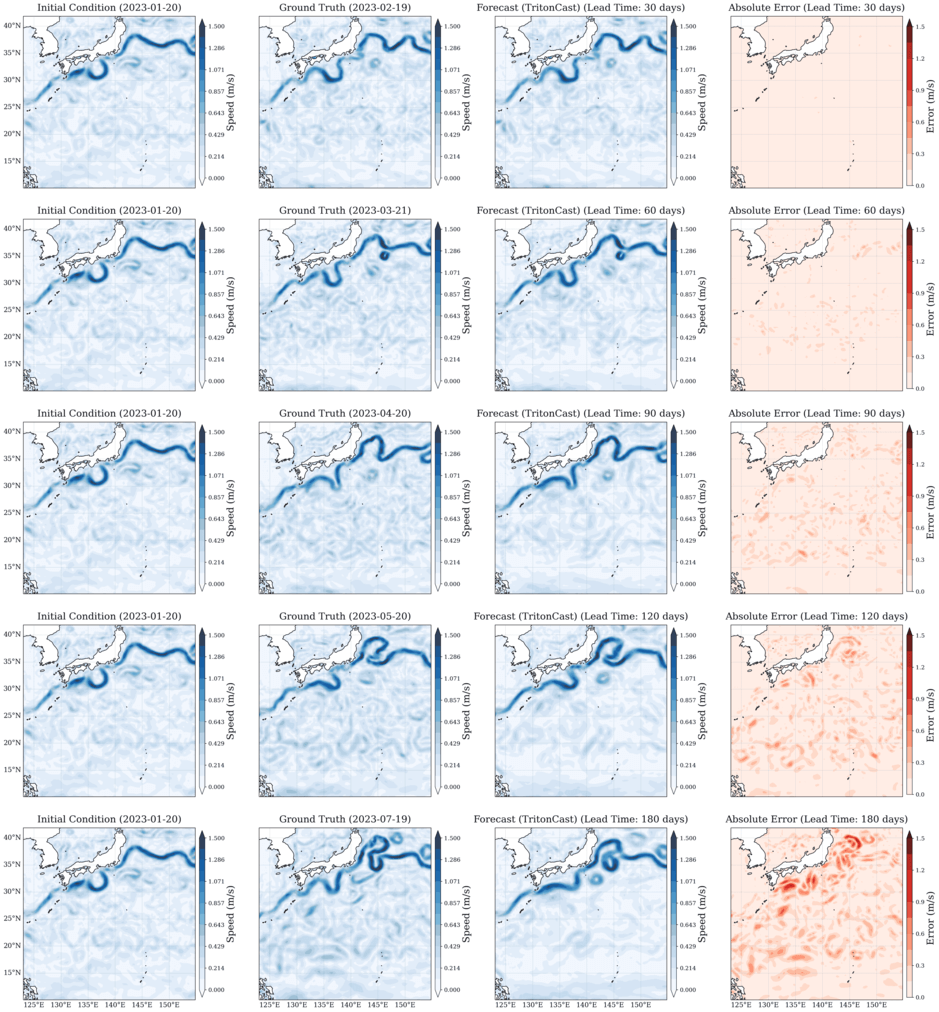}
\caption{\textbf{High-fidelity forecast for the Kuroshio Extension region at different lead times (from 30 to 180 days).} As one of the world's most active regions for mesoscale eddies, forecasting the Kuroshio is extremely challenging. The results show that TritonCast successfully captures fine-scale dynamical processes such as eddy merging, splitting, and interaction, demonstrating its great potential as a next-generation tool for eddy-resolving forecasting.}
\label{fig:kuroshio}
\end{figure}

\textbf{Figs.~\ref{fig:agu}},\textbf{~\ref{fig:gulf}}, and\textbf{~\ref{fig:kuroshio}} provide direct visual evidence of this capability. These sets of figures display, with unprecedented detail, TritonCast's long-term, high-fidelity simulation of the complex eddy fields in these three regions. Taking the Gulf Stream as an example (\textbf{Fig. \ref{fig:gulf}}), we can clearly observe the meandering path of the current and the generation of warm- and cold-core eddies from fluid instabilities. Remarkably, even after 120 or even 180 days, the morphology, position, and intensity of the eddies predicted by TritonCast remain strikingly similar to the ground truth, with very slow error growth. This marks a significant leap forward in the effective forecast horizon for AI models.

\begin{figure}[h!]
\centering
\includegraphics[width=1\linewidth]{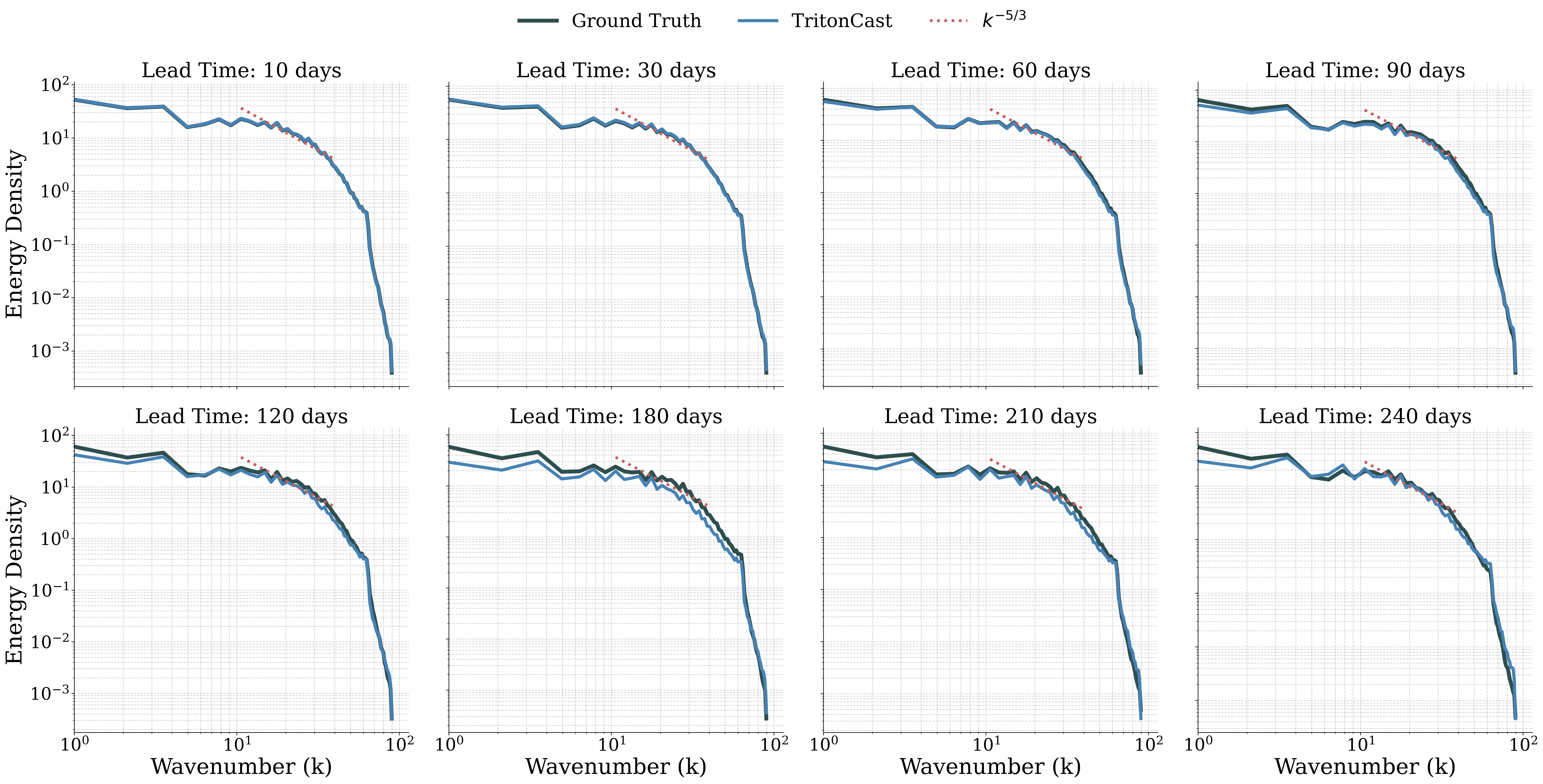}
\caption{\textbf{Evolution of the kinetic energy spectrum in the Gulf Stream region over a 240-day forecast.} The figure compares the spectra at multiple forecast lead times. The results show that the spectrum predicted by TritonCast remains remarkably consistent with the spectral structure of the ground truth (including the theoretical $k^{-5/3}$ slope) throughout the forecast period, with no issues of small-scale energy dissipation.}
\label{fig:gulf_esd}
\end{figure}

\begin{figure}[h!]
\centering
\includegraphics[width=1\linewidth]{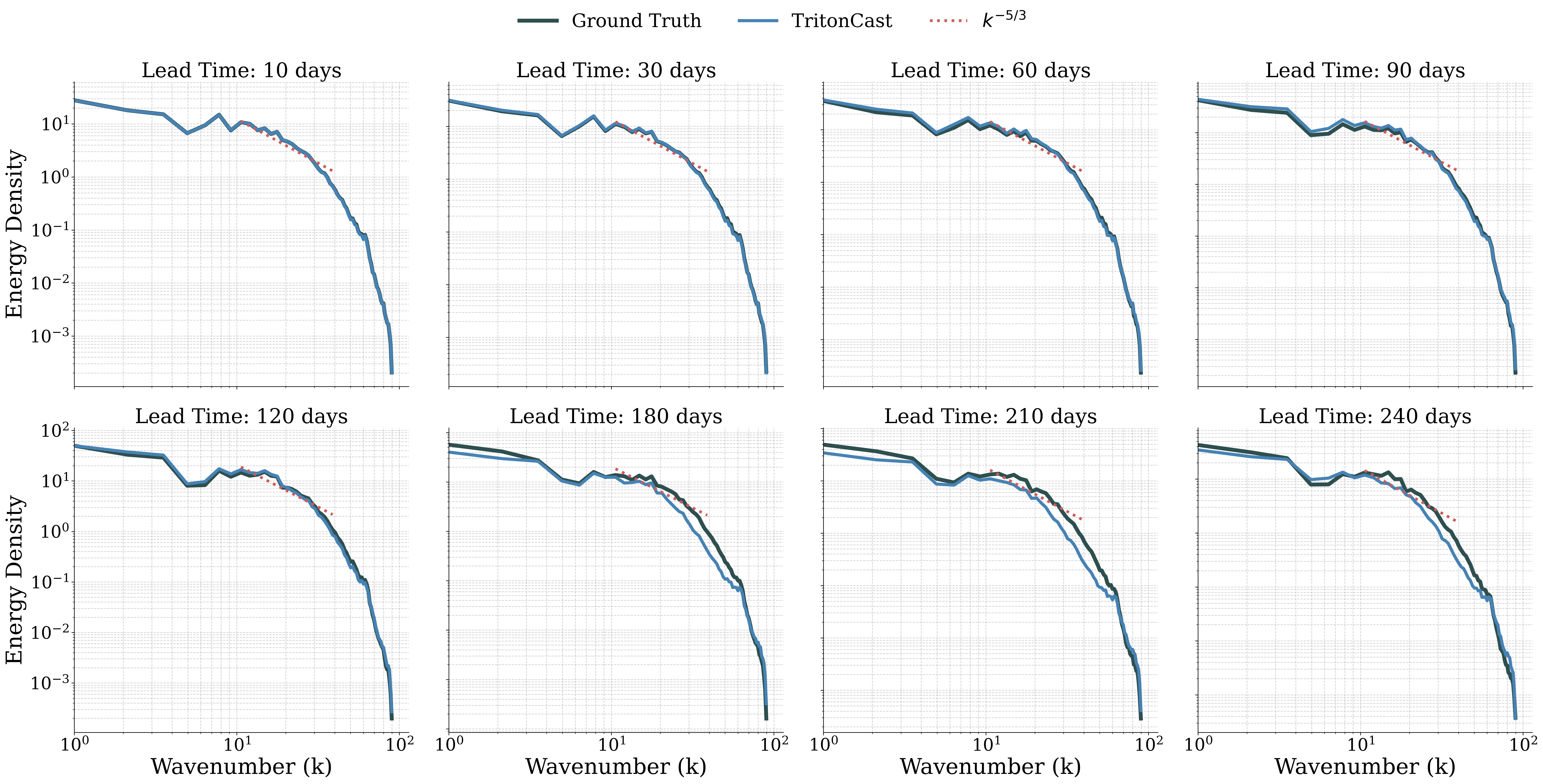}
\caption{\textbf{Evolution of the kinetic energy spectrum in the Kuroshio Extension region over a 240-day forecast.} Consistent with the results for the Gulf Stream region, this figure again demonstrates that TritonCast is physically self-consistent and can correctly simulate the energy cascade process of ocean turbulence, which is the fundamental reason for its ability to achieve long-term, high-fidelity forecasts.}
\label{fig:kuroshio_esd}
\end{figure}

\begin{figure}[h!]
\centering
\includegraphics[width=1\linewidth]{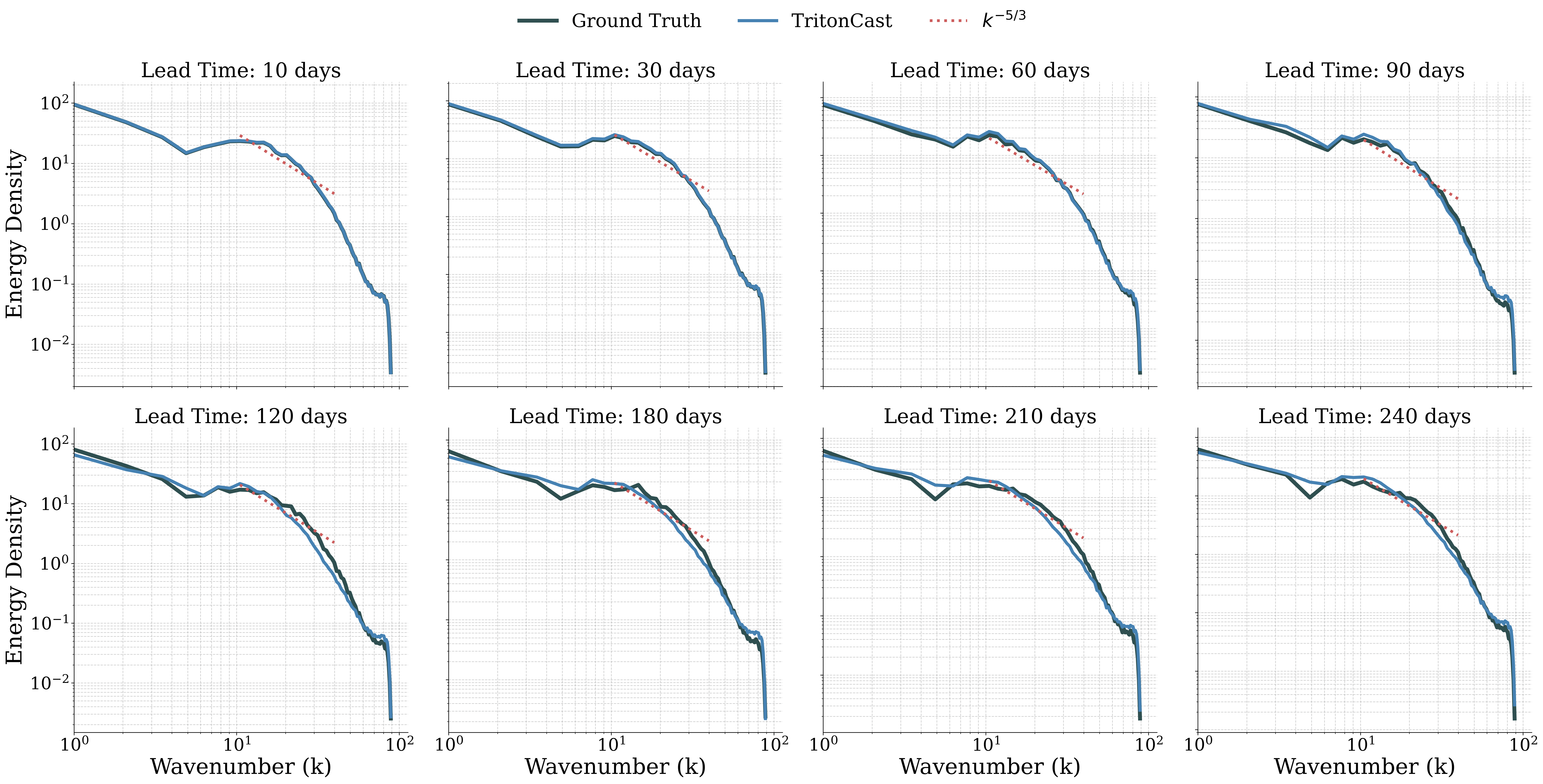}
\caption{\textbf{Evolution of the kinetic energy spectrum in the Agulhas Current region over a 240-day forecast.} Together, the analyses of the kinetic energy spectra from these three regions form a strong chain of evidence, proving that TritonCast maintains a high degree of physical consistency in high-resolution, long-lead-time, eddy-resolving forecasting.}
\label{fig:agu_esd}
\end{figure}

\begin{figure}[h]
\centering
\includegraphics[width=0.80\textwidth]{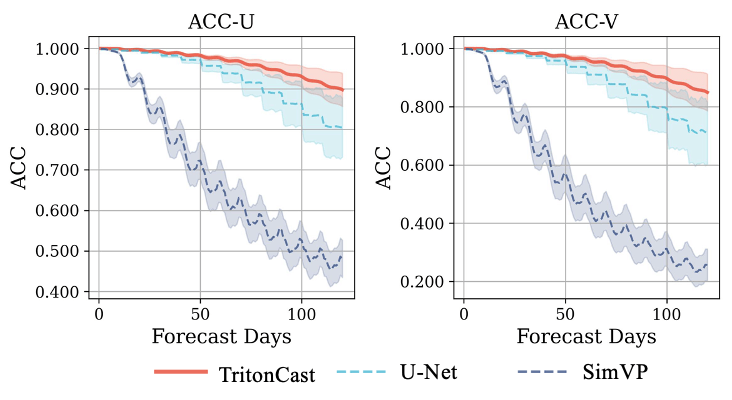}
\caption{\textbf{Comparison of Anomaly Correlation Coefficient (ACC) for the U and V components of sea surface velocity in the 0.125° Kuroshio region.} The plot compares the performance of TritonCast, U-Net, and SimVP over a 120-day forecast period. The solid lines represent the mean performance, and the shaded areas indicate the standard deviation across multiple runs.}
\label{fig:acc_comparison_kuroshio}
\end{figure}

Underpinning this exceptional visual fidelity is the model's robust physical consistency. We again use kinetic energy spectrum analysis to quantitatively verify this. \textbf{Figs.~\ref{fig:gulf_esd}},\textbf{~\ref{fig:kuroshio_esd}}, and\textbf{~\ref{fig:agu_esd}} show the evolution of the kinetic energy spectra in the three regions over forecasts of up to 240 days. The results are conclusive: across all three regions and all forecast lead times, the spectral structure predicted by TritonCast is in high agreement with the ground truth and correctly reproduces the theoretical spectral slopes (e.g., $k^{-5/3}$). This proves that TritonCast correctly simulates the energy cascade process in high-resolution turbulence. It is this intrinsic physical correctness that allows the model to avoid the distortion of small-scale information, thereby extending the effective forecast horizon for mesoscale eddies to several months. As shown in \textbf{Fig.~\ref{fig:acc_comparison_kuroshio}}, TritonCast maintains the highest Anomaly Correlation Coefficient (ACC) for both the U and V components of the velocity over a 120-day forecast period, with a much slower decay in performance compared to the other models. 

\subsection{Zero-Shot Generalization Across Resolutions}
\label{appendix:zeroshot}
This section is dedicated to showcasing one of the most remarkable capabilities of the TritonCast model: zero-shot generalization across resolutions. The core experiment involves applying a model trained \textbf{exclusively on 0.25° coarse-resolution data} to generate 0.125° high-resolution forecasts. Crucially, the model undergoes \textbf{no fine-tuning on any high-resolution data} in this process. This rigorous test aims to examine whether the model has learned scale-invariant, universal physical dynamics rather than merely memorizing patterns specific to a particular resolution. We validate this extraordinary capability from both qualitative and quantitative perspectives.

\subsubsection{Qualitative Validation: Generating Physically Realistic Fine-Scale Structures}

The first step in evaluating generalization is a visual inspection: do the fine-scale structures generated by the coarse-resolution model appear physically plausible at a higher resolution? We address this question by comparing the spatial distribution of Eddy Kinetic Energy (EKE), as it effectively reveals the fine morphology and energy distribution of mesoscale eddies.

\begin{figure}[h!]
\centering
\includegraphics[width=0.65\linewidth]{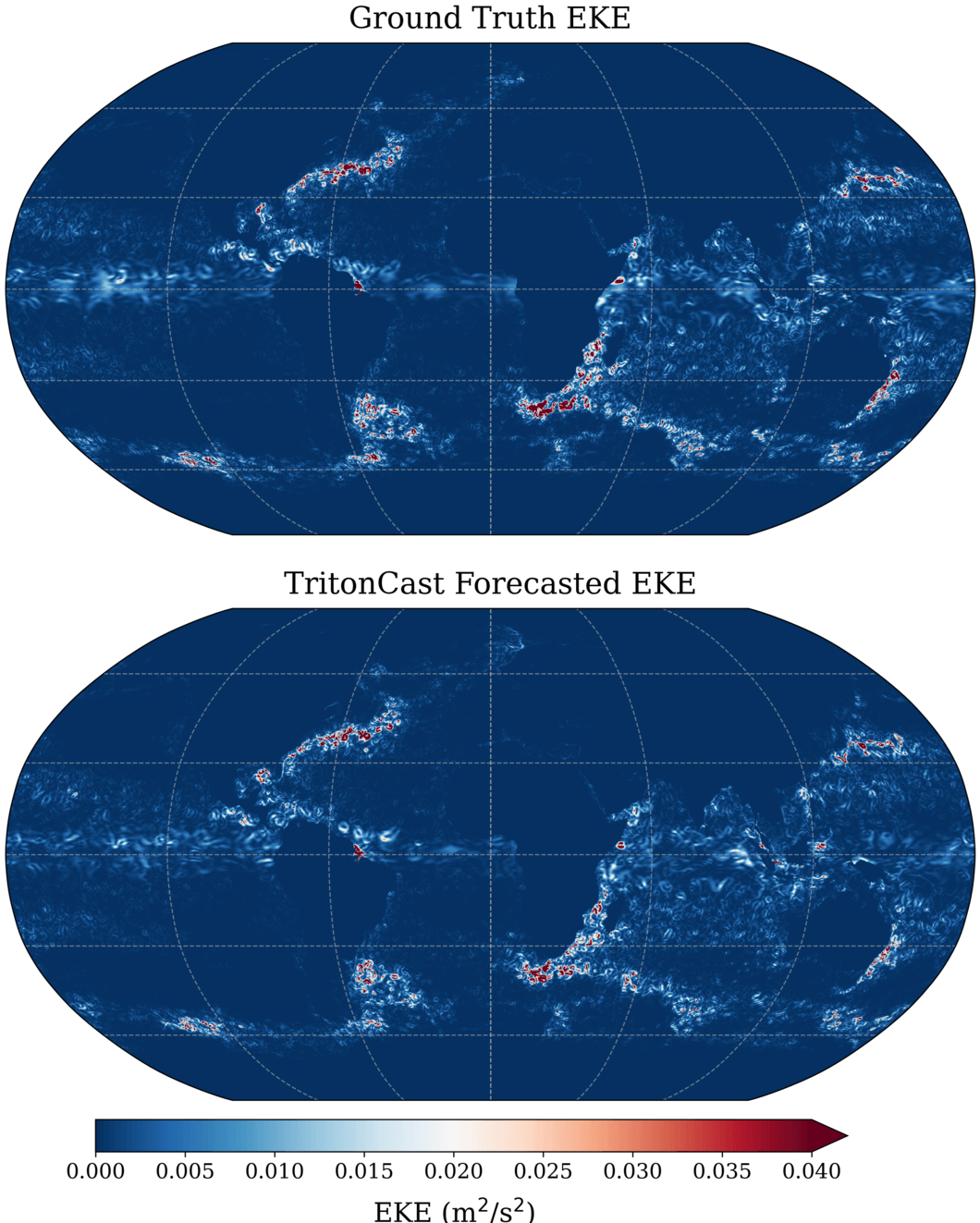}
\caption{
    \textbf{Comparison of the spatial distribution of Eddy Kinetic Energy (EKE, in m\textsuperscript{2}/s\textsuperscript{2}) at $0.125\degree$ horizontal resolution, demonstrating zero-shot generalization.} 
    (Top) Ground truth EKE from high-resolution data. 
    (Bottom) EKE from a 10-day forecast by a TritonCast model trained \textbf{only on 0.25° data}. 
    The model not only reproduces the macroscopic EKE hotspots but also successfully generates the fine-scale filamentary structures inherent to the high-resolution dataset, showcasing its high-fidelity predictive capability.
}
\label{fig:ocean_eke_comparison_case_0_no_land_flipped}
\end{figure}

\textbf{Fig.~\ref{fig:ocean_eke_comparison_case_0_no_land_flipped}} presents a compelling result. The EKE field generated by the TritonCast zero-shot forecast (bottom panel) aligns perfectly with the high-resolution ground truth (top panel) in its macroscopic structure, accurately capturing the eddy "hotspots" such as the Kuroshio Extension and the Gulf Stream. More importantly, the model successfully \textbf{generates} the rich, \textbf{fine-scale filamentary structures} inherent to the ground truth data. These details are not directly provided by the coarse-resolution input but are instead "inferred" by the model based on its learned physical principles. This strongly demonstrates that TritonCast's generalization capability far exceeds that of traditional interpolation methods; it performs a physically consistent "super-resolution" generation.

\subsubsection{Quantitative Validation: Physical Consistency and Error Diagnostics}

Visual realism must be supported by rigorous quantitative analysis. We assess the physical consistency of the zero-shot forecasts from three perspectives: the kinetic energy spectrum, forecast bias, and key transect analysis.

\begin{figure}[h!]
\centering
\includegraphics[width=1\linewidth]{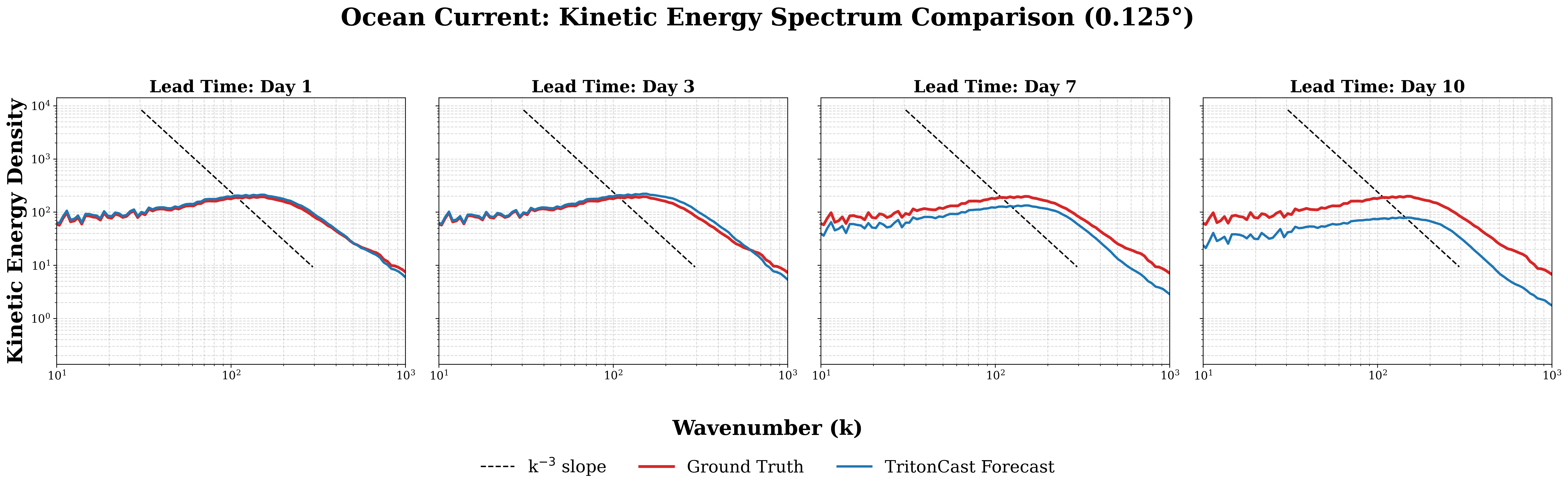}
\caption{
    \textbf{Comparison of the globally-averaged Kinetic Energy Spectrum at a $0.125\degree$ resolution for forecast lead times of 1, 3, 7, and 10 days.} The TritonCast forecast (blue) shows excellent agreement with the ground truth (red) at large spatial scales (low wavenumbers). While a progressive energy deficit emerges at smaller scales over time, the model's ability to maintain the correct spectral slope (referenced by the dashed $k^{-3}$ line) demonstrates a robust representation of the underlying ocean dynamics.
}
\label{fig:KE_spectrum_comparison_0125deg}
\end{figure}

\textbf{Kinetic Energy Spectrum Analysis:} \textbf{Fig.~\ref{fig:KE_spectrum_comparison_0125deg}} compares the globally-averaged kinetic energy spectrum. At low wavenumbers (large scales), the TritonCast forecast shows excellent agreement with the ground truth. At high wavenumbers (small scales), a progressive energy deficit emerges over time. This is a "smoothing effect" common in neural network models, where small-scale eddy energy is gradually dissipated. However, it is crucial to note that the spectral slope of the forecast remains consistent with the theoretical $k^{-3}$ slope. This indicates that although energy is dissipated, the fine-scale structures generated by the model follow the correct \textbf{physical laws of the energy cascade}, demonstrating the model's intrinsic physical robustness.

\begin{figure}[h!]
\centering
\includegraphics[width=1\linewidth]{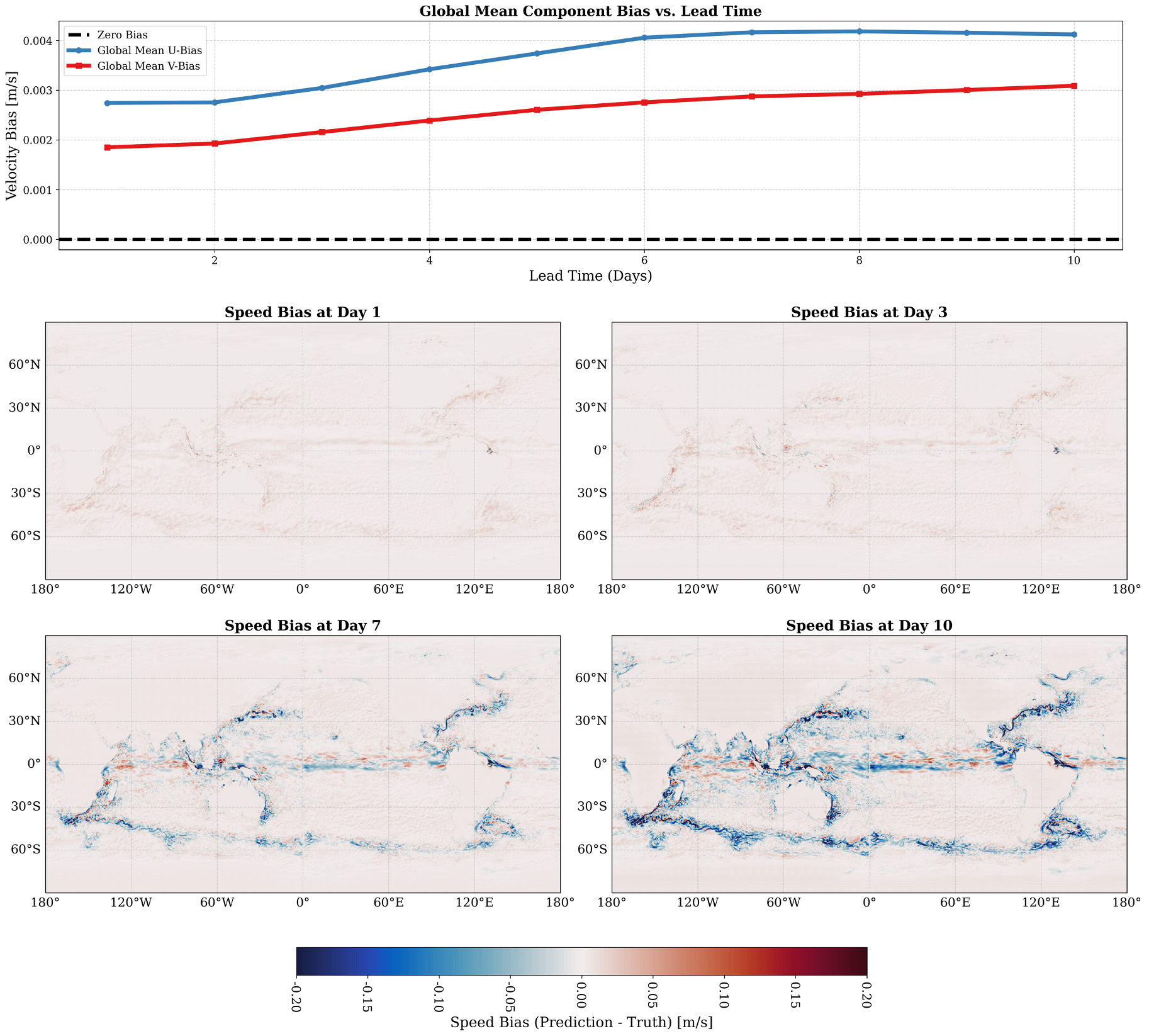}
\caption{
    \textbf{Forecast bias analysis of the zero-shot generalization.} 
    (Top) Time series of the globally-averaged zonal (U-Bias) and meridional (V-Bias) velocity bias. The biases remain small and stable, indicating no systematic drift. 
    (Bottom) Spatial distribution of the sea surface speed bias (Prediction - Truth). A distinct pattern emerges, characterized by a negative bias in the core of major currents and a positive bias on their flanks, consistent with the smoothing effect observed in the spectral analysis.
}
\label{fig:bias_analysis_0125deg_ocean_only_CORRECTLY_flipped}
\end{figure}

\textbf{Forecast Bias Analysis:} \textbf{Fig.~\ref{fig:bias_analysis_0125deg_ocean_only_CORRECTLY_flipped}} reveals the structure and origin of the forecast error. The top panel shows that the globally-averaged velocity biases (U-Bias and V-Bias) remain small and stable over the 10-day period, indicating no significant systematic drift. The spatial bias maps below reveal a pattern consistent with the spectral analysis: a negative bias (underestimation) in the core of strong currents like the Gulf Stream, and a positive bias (overestimation) on their flanks. This is the spatial manifestation of the smoothing effect—the model tends to slightly reduce the peak velocity of jets while broadening their structure. This error pattern is physically interpretable and not random noise.

\begin{figure}[h!]
\centering
\includegraphics[width=0.76\linewidth]{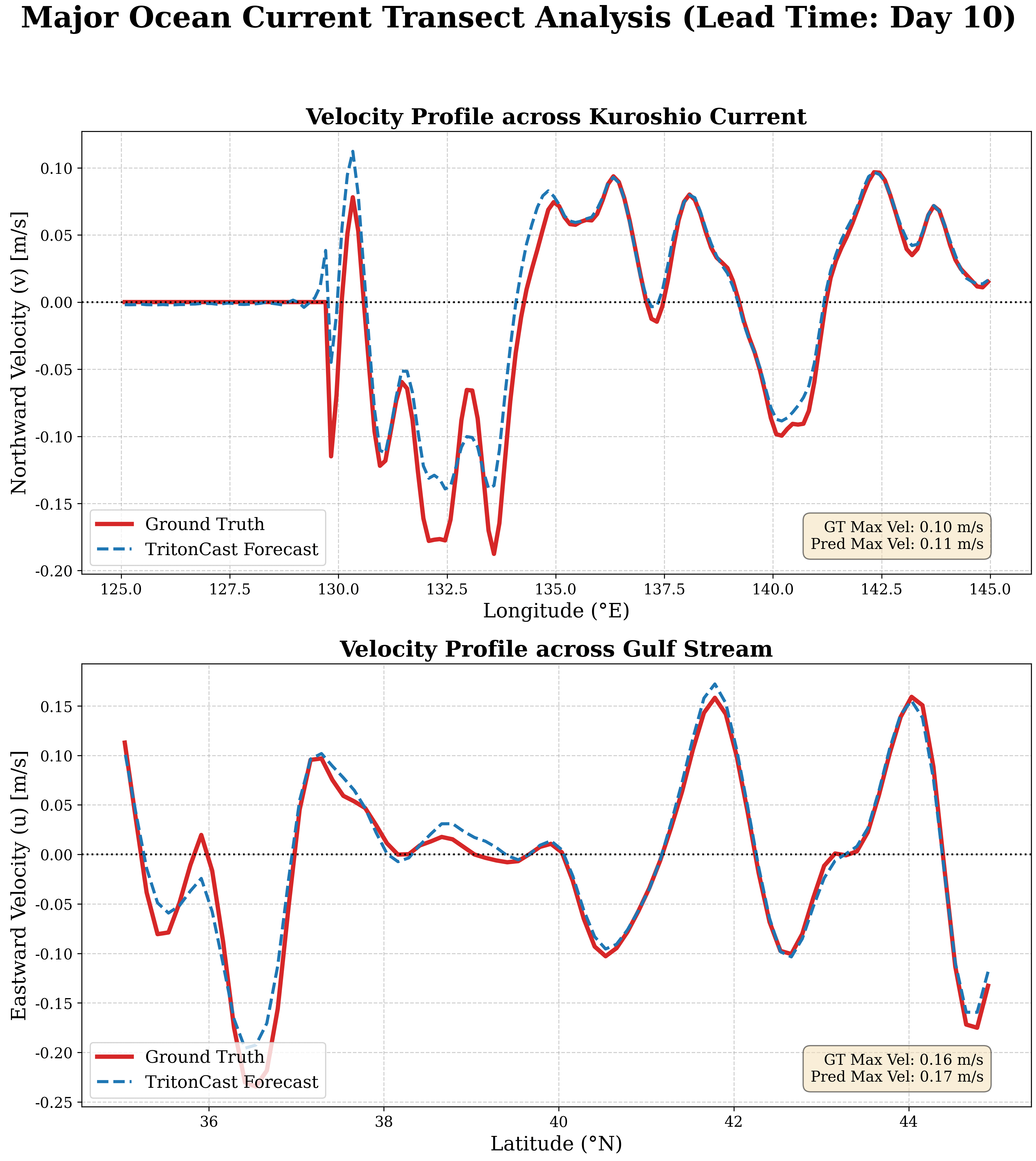}
\caption{
    \textbf{Analysis of velocity profiles across major ocean currents for a 10-day zero-shot forecast.} 
    (Top) Northward velocity profile across the Kuroshio Current at 30°N. 
    (Bottom) Eastward velocity profile across the Gulf Stream at 295°E (-65°W). 
    In both cases, the TritonCast forecast (dashed blue line) demonstrates excellent agreement with the ground truth (solid red line) in both the phase (location) and amplitude (strength) of the current jets and associated eddies.
}
\label{fig:transect_analysis_0125deg_legend_adjusted}
\end{figure}

\textbf{Key Transect Analysis:} Finally, in \textbf{Fig.~\ref{fig:transect_analysis_0125deg_legend_adjusted}}, we examine the model's ability to resolve the fine structure of two key western boundary currents: the Kuroshio and the Gulf Stream. By taking transects perpendicular to the main flow, we compare the velocity profiles. The results are excellent: the TritonCast forecast (dashed blue line) shows high agreement with the ground truth (solid red line) in both the phase (location) and amplitude (strength) of the current jets and their associated eddies. This demonstrates that the generalized forecast is sufficient to resolve the detailed internal structure of these highly unstable currents, confirming the model's advanced capability in capturing key ocean dynamics.

\clearpage

\section{Extended Results in Turbulence Forecasting and Baseline Comparisons}
\label{sec:appendix_turbulence_en}

This supplementary provides supplementary material for the long-range forecasting task of 2D decaying turbulence. Through detailed quantitative and qualitative analyses, we further validate the superiority of our proposed model (TritonCast) over mainstream baselines. We follow a logical progression from macro to micro, quantitative to qualitative, and phenomenon to essence to systematically demonstrate the model's performance.

\subsection{Quantitative Performance Evaluation}
\label{ssec:quantitative_eval_en}

To comprehensively evaluate the long-range forecasting performance, we first calculate the mean relative L2 error for all models at several key time steps. As Tab.~\ref{tab:relative_l2_error_en} shows, TritonCast consistently maintains the lowest error level throughout the entire forecast horizon, and its error accumulation rate is significantly slower than all baselines. This numerically demonstrates its exceptional long-term forecasting stability.

\begin{table}[htbp]
\centering
\caption{\textbf{Comparison of the mean Relative L2 Error between model predictions and the ground truth.} The error is evaluated at five key forecast time steps (T=1, 25, 50, 75, and 99). The lowest error at each time step is highlighted in bold.}
\label{tab:relative_l2_error_en}
\begin{tabular}{@{}lrrrrrrr@{}}
\toprule
\textbf{Time Step (T)} & \textbf{CNO} & \textbf{FNO} & \textbf{LSM} & \textbf{PDE-Refiner} & \textbf{SimVP} & \textbf{TritonCast} & \textbf{UNet} \\
\midrule
1  & 9.63\%  & 7.80\%  & 3.21\%  & 0.98\%  & 2.09\%  & \textbf{0.52\%}  & 6.44\%  \\
25 & 106.10\% & 50.35\% & 64.88\% & 20.92\% & 37.64\% & \textbf{5.76\%}  & 67.38\% \\
50 & 140.93\% & 79.52\% & 103.30\%& 42.55\% & 89.85\% & \textbf{13.21\%} & 97.95\% \\
75 & 167.66\% & 95.89\% & 122.55\%& 62.05\% & 120.89\%& \textbf{22.76\%} & 115.28\%\\
99 & 196.39\% & 104.10\%& 133.14\%& 76.77\% & 132.15\%& \textbf{32.96\%} & 126.46\%\\
\bottomrule
\end{tabular}
\end{table}

\subsection{Statistical Robustness Verification}
\label{ssec:statistical_robustness_en}

\begin{figure}[htbp]
\centering
\includegraphics[width=\linewidth]{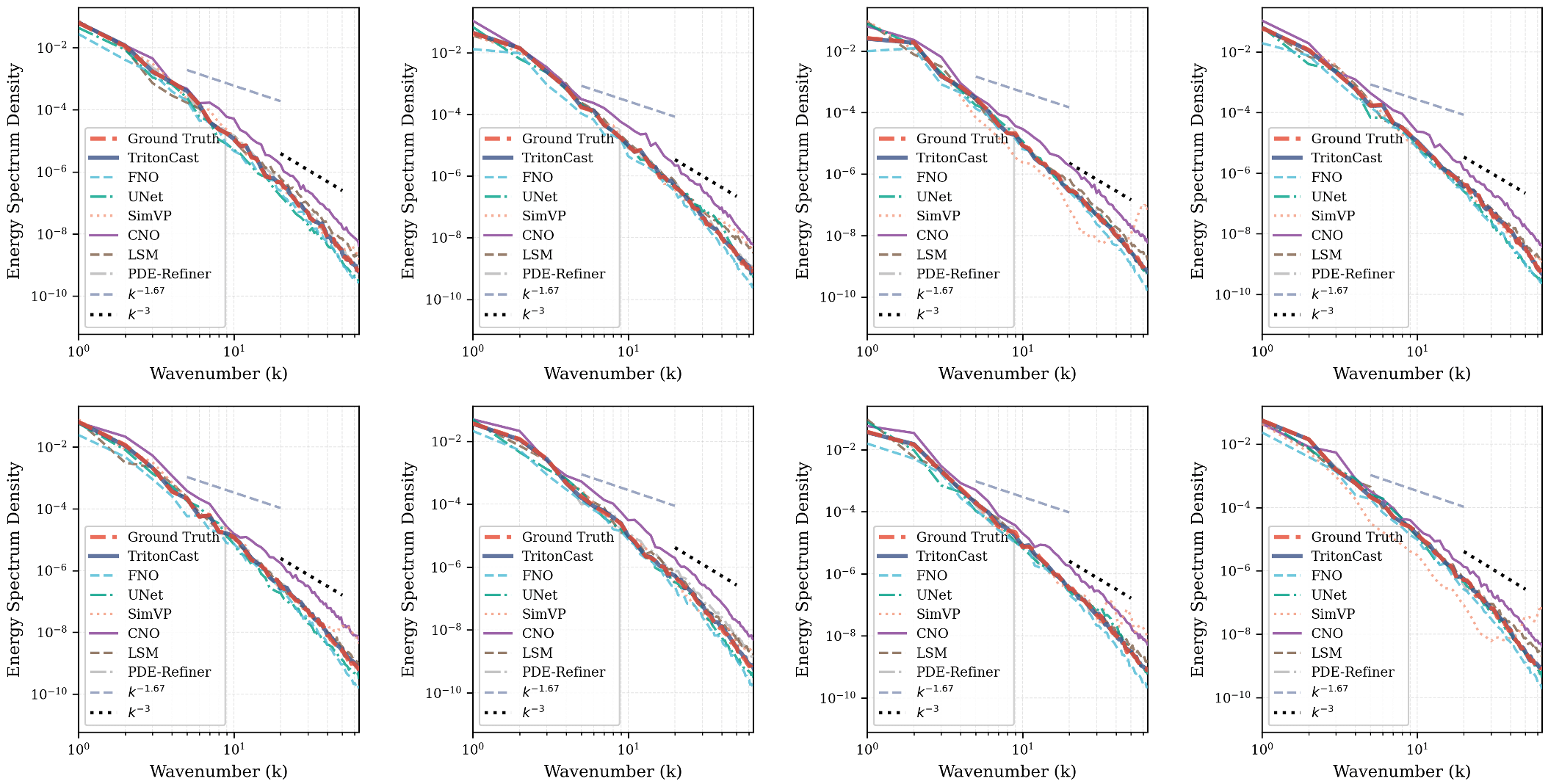}
\caption{\textbf{Comparison of energy spectral density for all models at the final forecast time step (T=99) across eight randomly selected initial conditions.} Each subplot represents an independent forecast sample.}
\label{fig:esd_grid_en}
\end{figure}

To verify the consistency of the model's performance, we randomly select several different initial conditions for testing. \textbf{Fig.~\ref{fig:esd_grid_en}} shows the energy spectral density for eight independent samples at the final time step (T=99). In all cases, the energy spectrum of TritonCast (solid dark blue line) most closely matches that of the ground truth (dashed red line). This indicates that the model's superior performance is statistically robust and not dependent on specific initial conditions.

\subsection{Physical Field Fidelity and Error Analysis}
\label{ssec:field_fidelity_en}

\begin{figure}[htbp]
\centering
\includegraphics[width=\linewidth]{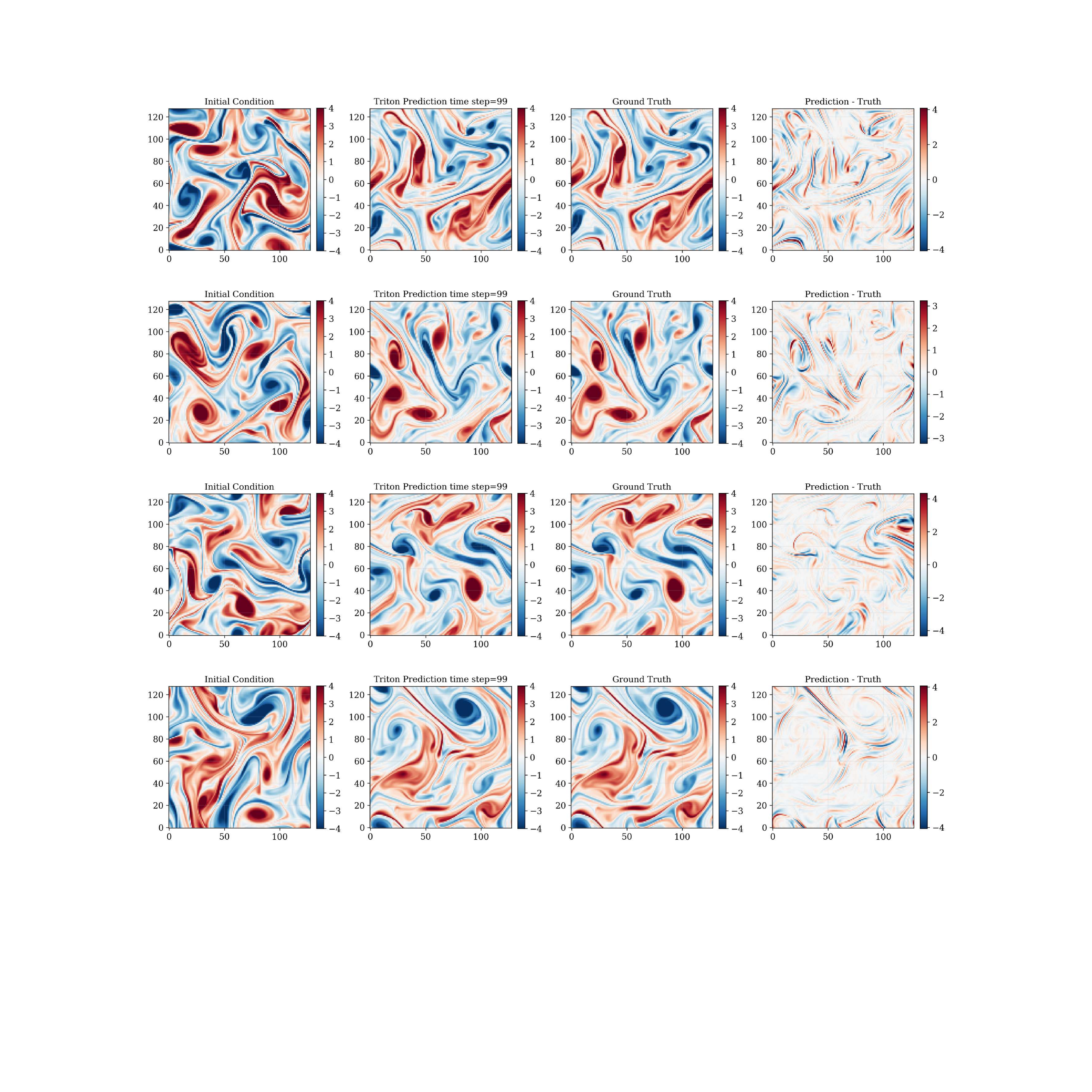}
\caption{\textbf{Visualization of TritonCast's long-range forecast at the final time step (t=99).} Each row corresponds to an independent forecast task initiated from a different random initial condition. From left to right: the initial condition (t=0), TritonCast's 99-step autoregressive prediction (t=99), the corresponding ground truth (t=99), and the pointwise error map (Prediction - Truth).}
\label{fig:triton_vis_en}
\end{figure}

\textbf{Fig.~\ref{fig:triton_vis_en}} presents a detailed visualization of TritonCast's long-range forecasts for four independent initial conditions. The figure shows that even after 99 autoregressive steps, the model accurately captures complex vortical dynamics, including the evolution of fine filamentary structures. The error maps (rightmost column) reveal that errors are primarily localized in regions of high gradients and are of controlled magnitude, without any systematic biases that would corrupt the overall flow structure. This demonstrates the high fidelity of the model's predictions.

\subsection{Failure Mode Analysis and the Root Cause of Spectral Bias}
\label{ssec:failure_modes_en}

\begin{figure}[htbp]
\centering
\includegraphics[width=\linewidth]{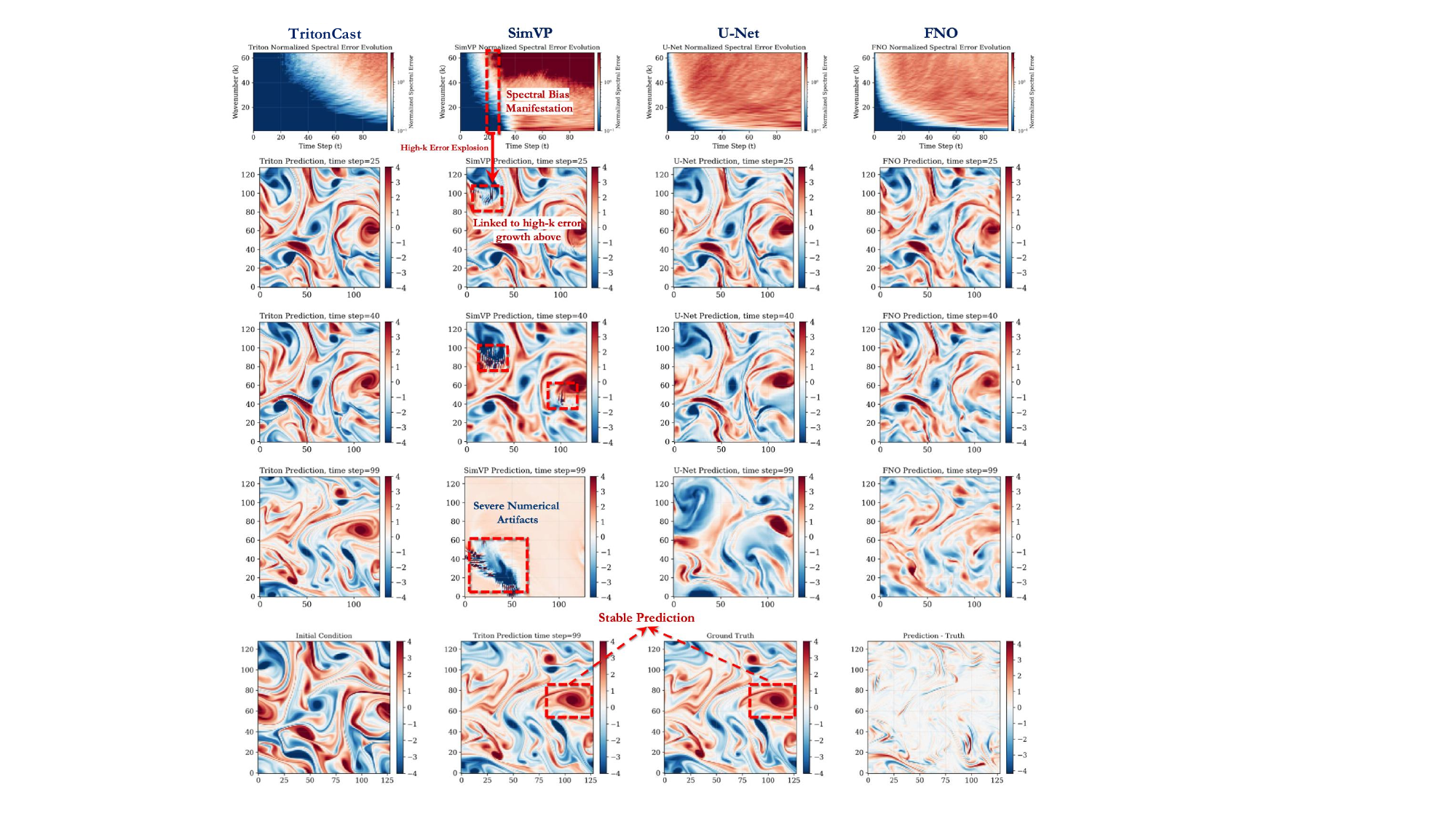}
\caption{\textbf{Visual and spectral comparison of long-term turbulence forecasting.} This figure juxtaposes the evolution of normalized spectral error (top row) with the corresponding physical field predictions (middle and bottom rows) for different models. The results show that Triton's long-term stability stems from its effective suppression of spectral error. In contrast, baseline models exhibit distinct failure modes: SimVP's high-wavenumber error explosion leads to numerical artifacts, while U-Net and FNO lose fine-scale details due to excessive smoothing.}
\label{fig:failure_analysis_en}
\end{figure}

To investigate the root cause of TritonCast's long-term stability, we perform an analysis that links the evolution of the physical field with its error evolution in the spectral domain (\textbf{Fig.~\ref{fig:failure_analysis_en}}). This figure clearly reveals two typical failure modes in the baseline models. SimVP suffers from a numerical collapse and produces severe artifacts due to its inability to control high-wavenumber errors. U-Net and FNO, on the other hand, produce overly smooth physical fields and lose critical small-scale details due to excessive dissipation of high-wavenumber components. In sharp contrast, TritonCast successfully suppresses error accumulation across the entire spectrum, particularly avoiding the catastrophic growth of small-scale errors. This provides strong evidence that the ability to correctly handle spectral bias is key for data-driven models to achieve long-range, high-fidelity physical forecasting.

\clearpage

\section{Ablation Studies: Synergistic Roles of Core Components in TritonCast}
\label{app:ablation}
\begin{figure}[h!]
    \centering
    \includegraphics[width=1\textwidth]{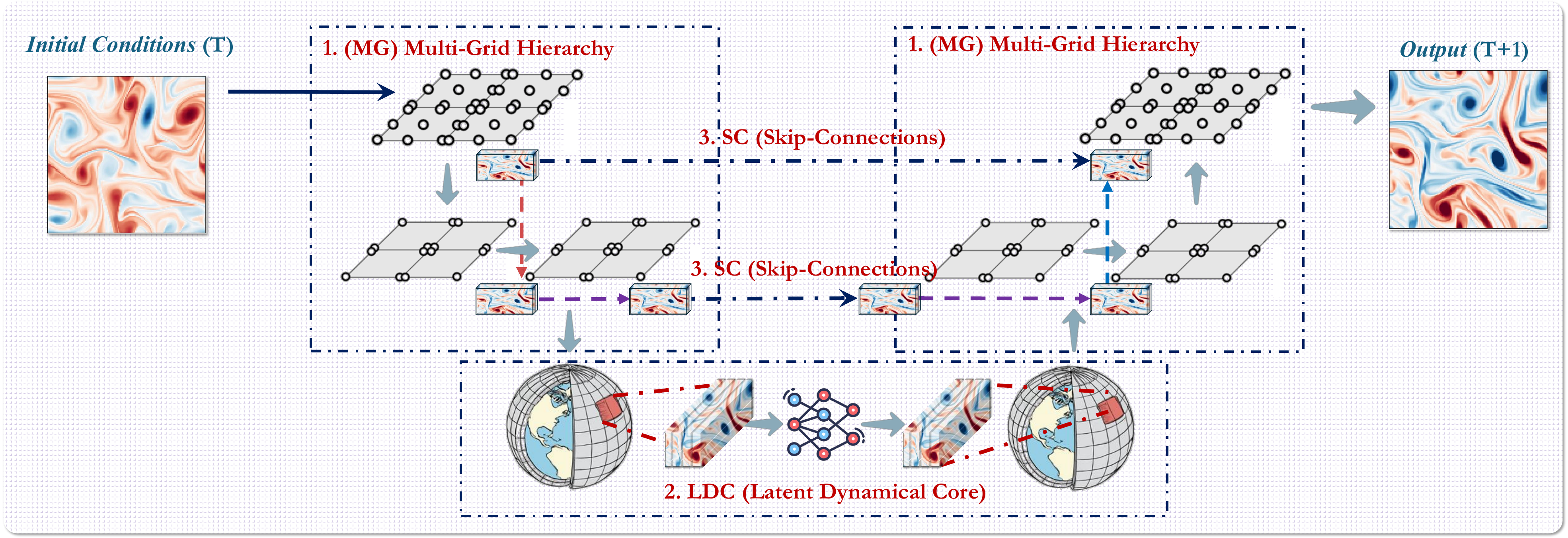} 
    \caption{
        \textbf{Schematic of TritonCast's three core components discussed in the ablation studies.} 
        This simplified diagram illustrates the interplay between:
        \textbf{(1) Multi-Grid (MG) Hierarchy:} The structural backbone that processes information across different resolution grids.
        \textbf{(2) Latent Dynamical Core (LDC):} The stability engine that evolves the system's state at the coarsest, most abstract level.
        \textbf{(3) Skip-Connections (SC):} The pathways for injecting fine-scale details from the encoder (down-sampling path) to the decoder (up-sampling path). 
        The ablated models (No MG, No LDC, No SC) are conceptually defined by the removal of the corresponding numbered component from this architecture.
    }
    \label{fig:appendix_schematic}
\end{figure}

To systematically investigate the distinct contribution of each core component within the TritonCast architecture, we design three ablated model variants. Each variant is constructed by precisely removing one of the three foundational components the Multi-Grid hierarchy, the Latent Dynamical Core, or the Skip-Connections from the full model. This approach allows for a rigorous assessment of their individual necessity and synergistic interplay. The functional roles of these three components are visually summarized in Fig.~\ref{fig:appendix_schematic}.

\subsection{Definition of Ablated Model Variants (No MG, No LDC, No SC)}

\paragraph{No MG (Without Multi-Grid Hierarchy)} The \textbf{No MG} variant removes the hierarchical multi-grid (MG) architecture. This modification transforms the model into a non-hierarchical, single-scale framework, thereby eliminating the V-cycle computational pattern that facilitates the decomposition and reconstruction of dynamics across multiple scales. The absence of the MG structure prevents the model from effectively separating and learning multi-scale physical processes. Consequently, this variant exhibits severe, non-physical energy dissipation at small scales and fails to reproduce theoretical energy cascades, which leads to systematic biases and a loss of physical realism in long-term forecasts.

\paragraph{No LDC (Without Latent Dynamical Core)} The \textbf{No LDC} variant is defined by the removal of the Latent Dynamical Core (LDC), which operates at the coarsest level of the model's hierarchy. The LDC functions as the primary engine for long-term stability by exclusively simulating the dominant, large-scale dynamics in a spectrally simplified latent space, thereby shielding the integration process from high-frequency instabilities. Its removal deprives the model of a stable anchor for long-term autoregressive rollouts. As a result, the No LDC model experiences catastrophic numerical failure within a few forecast steps, confirming that a dedicated core for large-scale evolution is indispensable for long-term stability.

\paragraph{No SC (Without Skip-Connections)} The \textbf{No SC} variant is constructed by ablating the skip-connections (SC) that link corresponding resolution levels between the encoder and decoder pathways. In the full TritonCast architecture, these connections are crucial for re-injecting high-frequency details from the input state into the forecasting process. This mechanism directly corrects for deviations in fine-scale features and suppresses the accumulation of small-scale errors. The absence of skip-connections results in a significant degradation of forecast fidelity. The model exhibits severe climate drift and fails to maintain the structural integrity and geographical positioning of key weather systems, such as the polar vortex, over year-long simulations. This underscores the critical role of skip-connections in ensuring the physical consistency of detailed structures in long-term forecasting.




\subsection{Spectral Bias Analysis: Validating Model Stability through Frequency Filtering}
\label{appendix:ablation_Spectral}

\begin{figure}[h!]
    \centering
    \includegraphics[width=\textwidth]{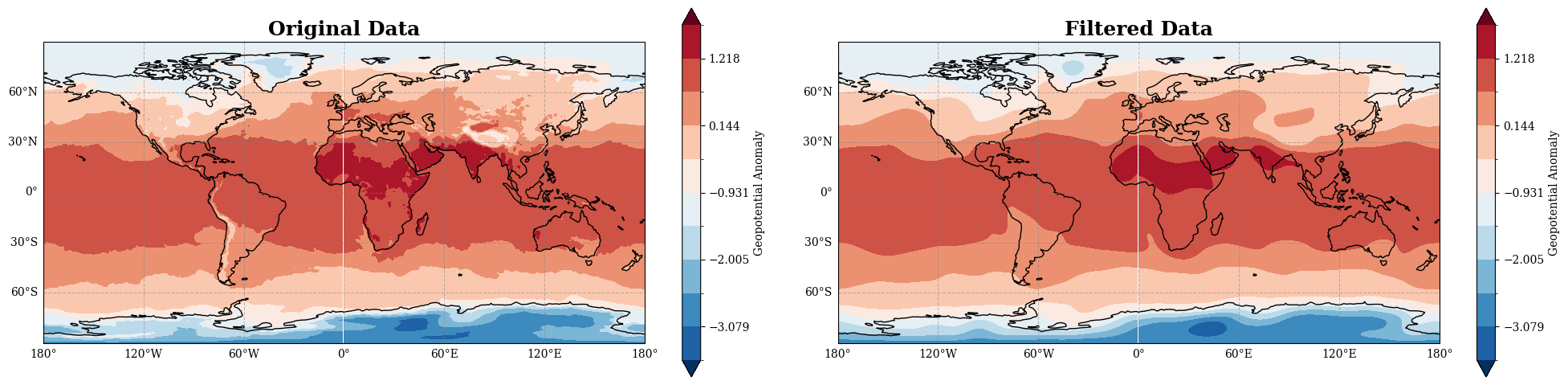}
    \caption{\textbf{Visualization of the low-pass filtering process.} An example 2-meter temperature (T2M) field from the forecast is shown before (Original Data, left) and after (Filtered Data, right) the application of the 2D low-pass filter. The filtering preserves the large-scale structure of the field while removing the high-frequency, small-scale spatial variations, which results in a noticeably smoother output. This process is applied at each autoregressive step for the TritonCast (w/o High Freq.) model.}
    \label{fig:filtering_example_t2m}
\end{figure}

The prevalent instability of AI models in long-term autoregressive forecasting is often attributed to their inherent spectral bias, namely an inadequate representation of high-frequency, small-scale signals such as the fronts and eddies that drive weather evolution~\cite{rahaman2019spectral}. To directly test this central hypothesis, we design a controlled experiment that isolates and evaluates the precise role of these small-scale dynamics in a long-term forecast. The experiment comprises two parallel, year-long global weather forecasts: (1) the standard TritonCast model, and (2) a version, denoted in the figures as \textbf{TritonCast (w/o High Freq.)}, where we apply a 2D low-pass filter to the model's output at each autoregressive step to actively remove high-frequency spatial components. \textbf{Fig.~\ref{fig:filtering_example_t2m}} visually illustrates the effect of this filtering operation on a sample 2-meter temperature field. This experiment reveals that the accurate representation of small-scale physical processes is fundamental to maintaining the model's long-term stability and large-scale climate realism.

\begin{figure}[h!]
    \centering
    \includegraphics[width=0.95\textwidth]{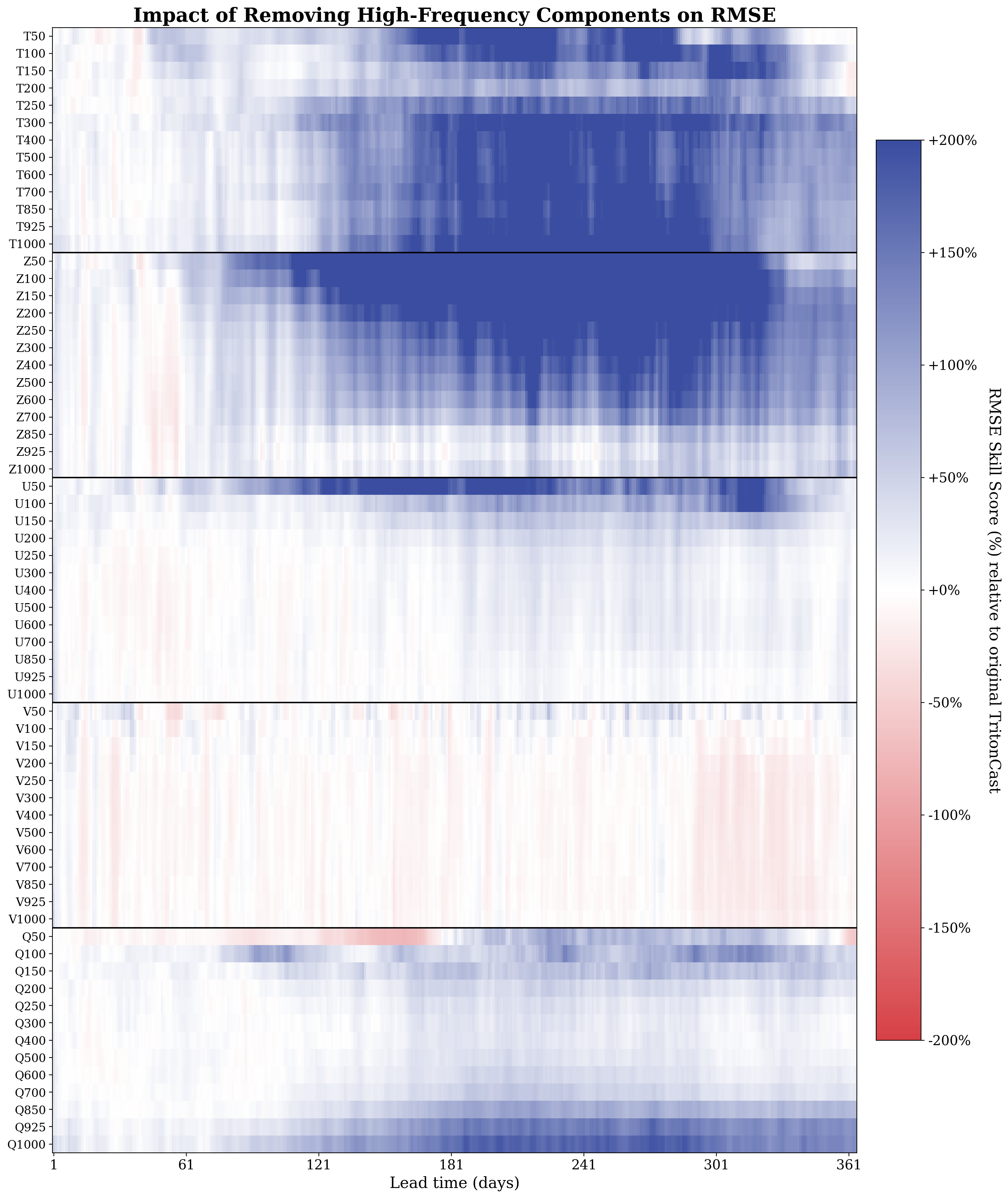} 
    \caption{\textbf{Quantitative impact of removing high-frequency components on long-term forecast error.} The color indicates the percentage change in RMSE skill score for the TritonCast (w/o High Freq.) model relative to the standard TritonCast. Blue denotes a significant increase in error. The plot clearly shows a transition from comparable short-term (<90 days) error to catastrophic long-term (>90 days) error divergence.}
    \label{fig:rmse_card_fft}
\end{figure}

\begin{figure}[h!]
    \centering
    \includegraphics[width=1\textwidth]{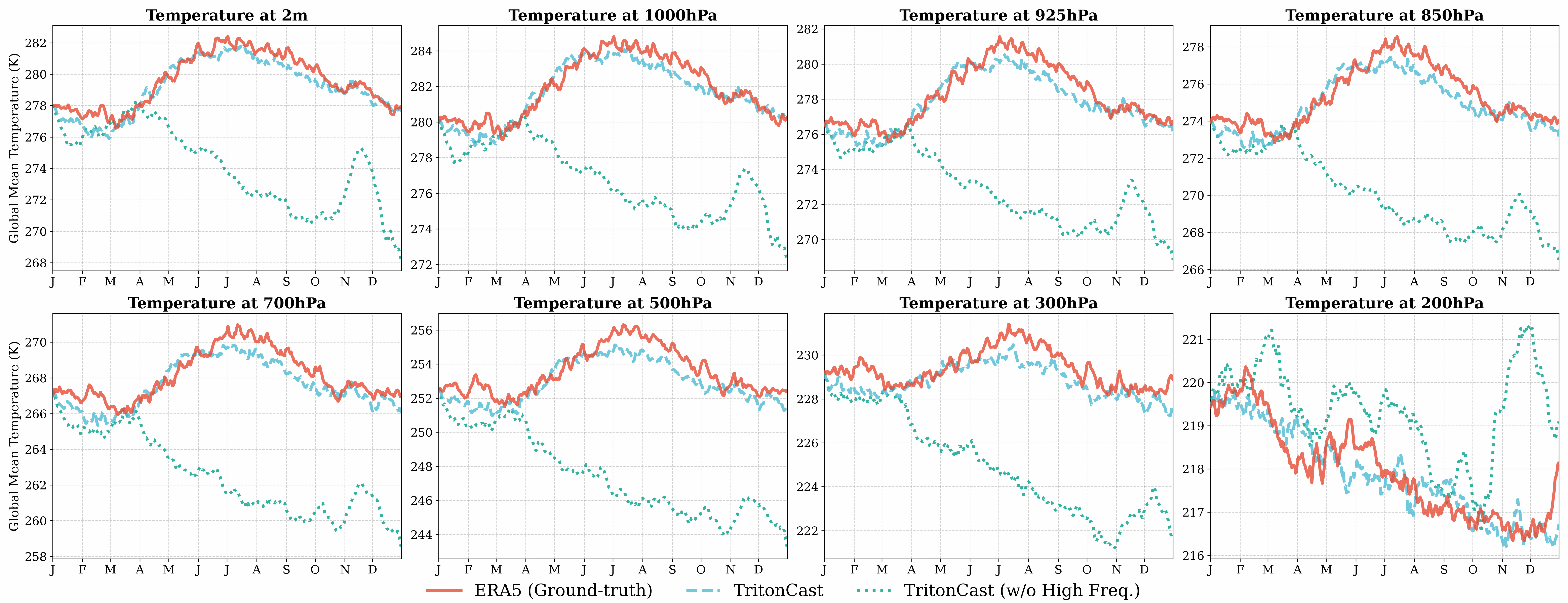} 
    \caption{\textbf{Long-term evolution of global mean temperature.} The comparison shows that the standard TritonCast (light blue line) successfully forecasts a seasonal cycle consistent with observations (ERA5, red line). In contrast, the TritonCast (w/o High Freq.) model (green dotted line) undergoes a severe systematic climate drift in the long-term forecast.}
    \label{fig:t_fft}
\end{figure}

Quantitative analysis first reveals a striking divergence in prediction error. As \textbf{Fig.~\ref{fig:rmse_card_fft}} illustrates, the Root Mean Square Error (RMSE) of TritonCast (w/o High Freq.) is comparable to the standard model during the initial forecast phase (up to approximately 90 days). Following this brief period, however, its error catastrophically increases, exceeding that of the standard model by over 200\% for several key variables. This error accumulation manifests as a severe, systematic climate drift rather than random noise. The global mean temperature evolution in \textbf{Fig.~\ref{fig:t_fft}} provides a stark visualization of this phenomenon: the standard TritonCast successfully reproduces the seasonal cycle of Earth's climate, whereas the forecast from TritonCast (w/o High Freq.) completely deviates from reality after about three months, entering a physically erroneous, continuous cooling state.

\begin{figure}[h!]
    \centering
    \includegraphics[width=0.95\textwidth]{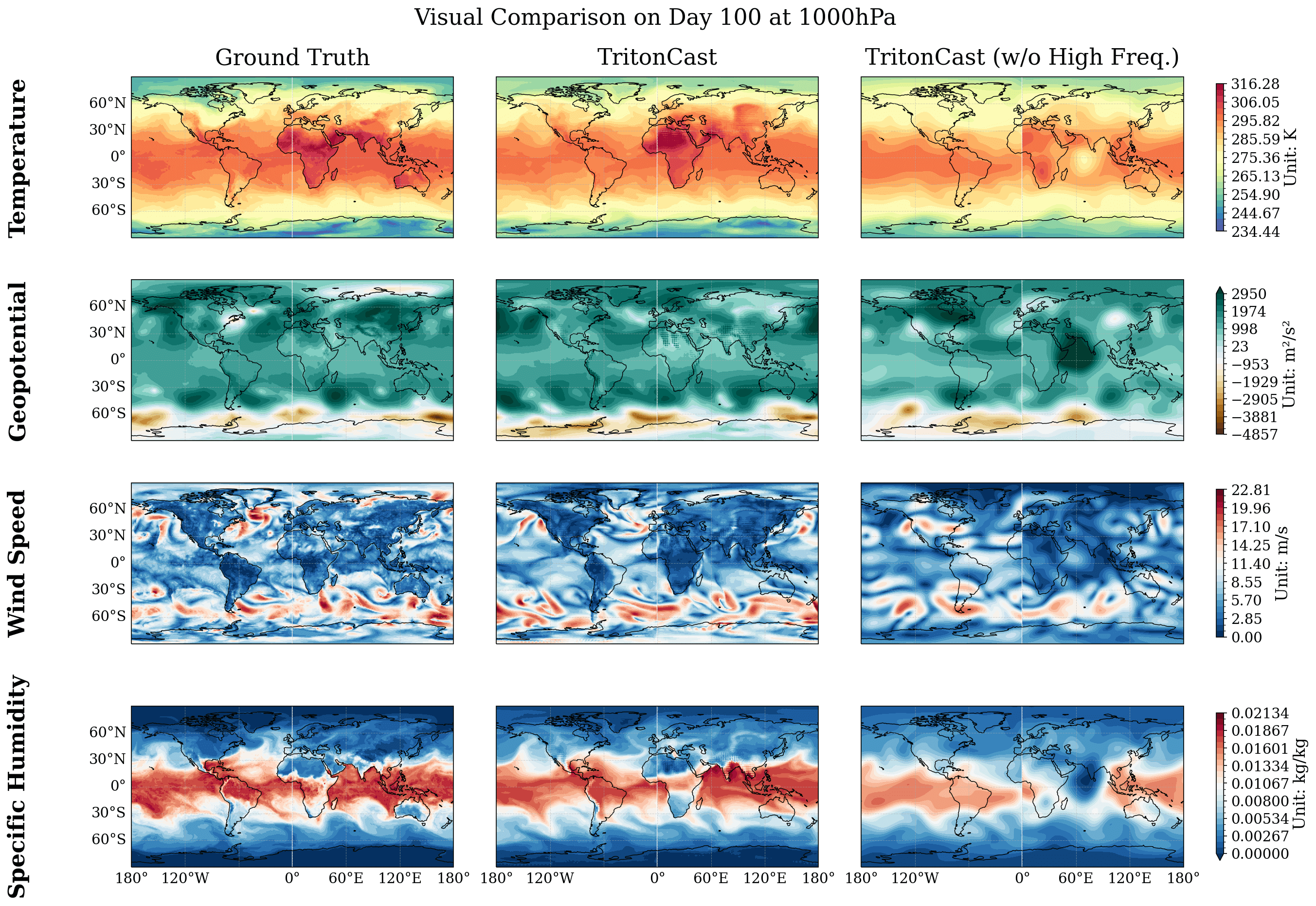} 
    \caption{\textbf{Comparison of physical field fidelity at forecast day 100.} Compared to the ground truth (left) and the standard TritonCast (middle), the output from TritonCast (w/o High Freq.) (right) appears overly smoothed due to the removal of high-frequency, small-scale information, losing critical physical details.}
    \label{fig:compaer_day100}
\end{figure}

\begin{figure}[h!]
    \centering
    \includegraphics[width=0.95\textwidth]{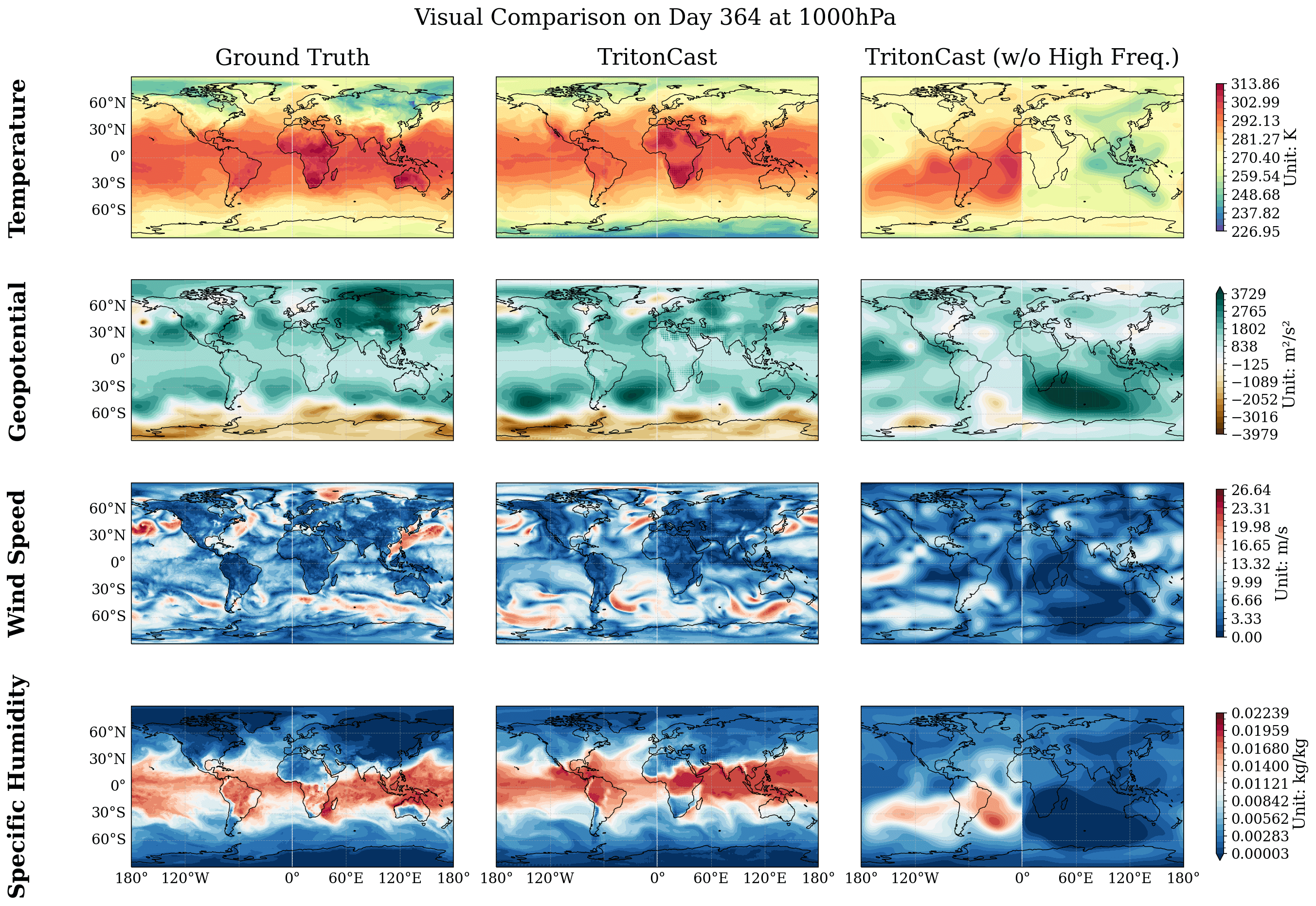} 
    \caption{\textbf{Comparison of physical field structure at forecast day 364.} After a one-year autoregressive forecast, the standard TritonCast (middle) still maintains a physically plausible atmospheric circulation structure, whereas the TritonCast (w/o High Freq.) model (right) completely collapses into a physically unrealistic state.}
    \label{fig:compaer_day364}
\end{figure}

\begin{figure}[h!]
    \centering
    \includegraphics[width=0.95\textwidth]{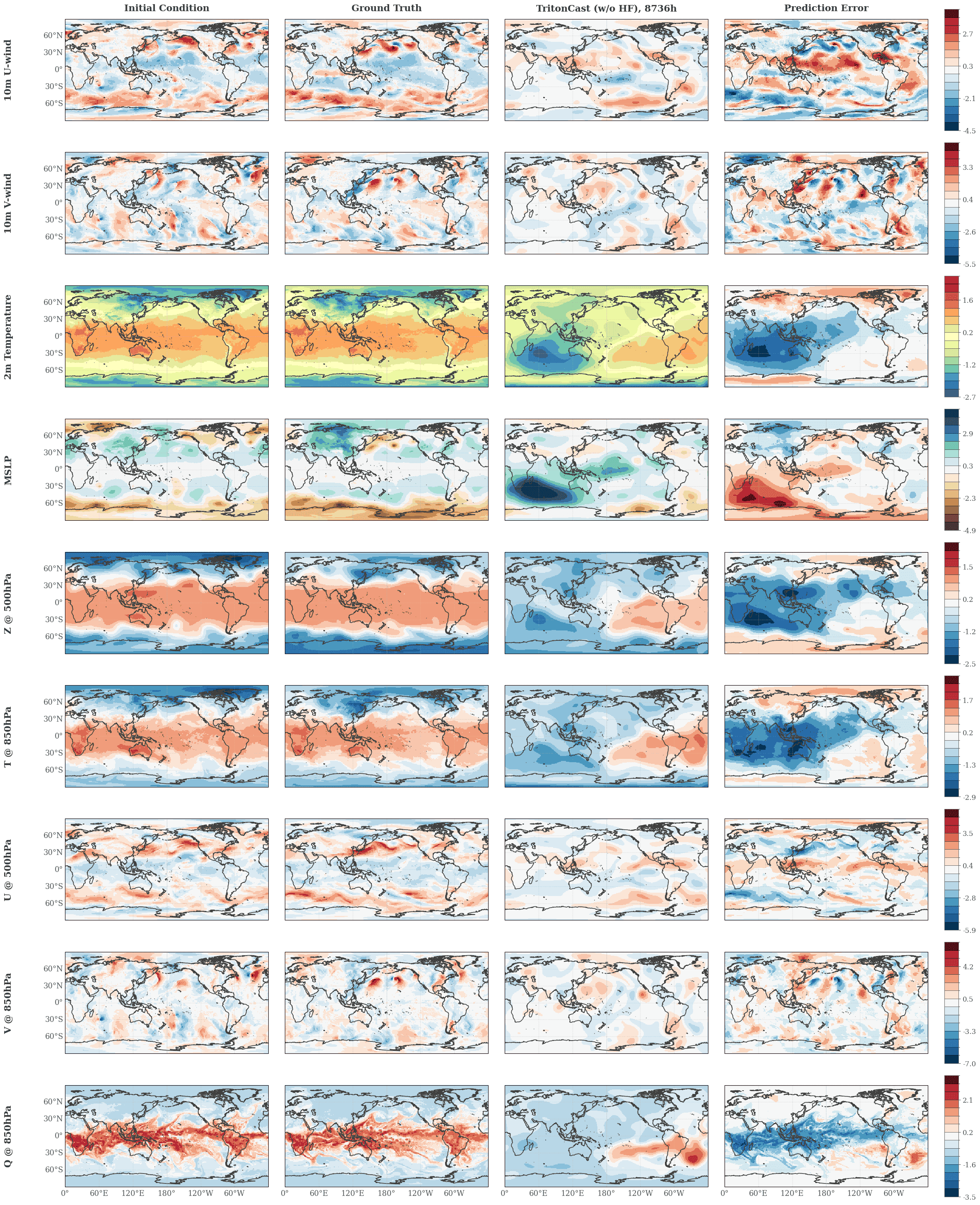} 
    \caption{\textbf{Detailed view of variable fields and their errors at forecast day 364.} The fourth column (Prediction Error) clearly shows the immense and structurally systematic forecast errors of the TritonCast (w/o HF) model, confirming its systemic collapse.}
    \label{fig:364day_global}
\end{figure}

To investigate the physical mechanisms behind this systematic drift, we analyze the spatial fields of the forecasts. In the medium range (day 100, \textbf{Fig.~\ref{fig:compaer_day100}}), the output from TritonCast (w/o High Freq.) already loses many of the small-scale dynamical processes crucial for weather evolution. For instance, fine-scale structures of jet streams and mid-latitude eddies in the wind field appear overly smoothed, and the intricate patterns of tropical convection in the humidity field become blurred. This progressive erosion of small-scale physical details ultimately culminates in the structural collapse of the large-scale atmospheric circulation by the end of the forecast (day 364, \textbf{Fig.~\ref{fig:compaer_day364}}). At this point, the standard TritonCast forecast maintains a physically self-consistent and plausible atmospheric state. In contrast, the output from TritonCast (w/o High Freq.) degenerates into physically meaningless fields, an outcome further confirmed by the massive, systematic error structures shown in \textbf{Fig.~\ref{fig:364day_global}}.

In summary, this experiment establishes a clear causal chain from the microscopic to the macroscopic: the continuous removal of high-frequency, small-scale components in an autoregressive forecast causes a progressive loss of physical fidelity. This, in turn, disrupts the system's crucial nonlinear dynamical equilibrium. The energy carried by small-scale processes cannot be correctly transferred to the large-scale circulation, and vice versa. This breakdown in cross-scale energy transport leads to catastrophic error accumulation and systematic drift. This finding robustly demonstrates that high-frequency, small-scale dynamics are not negligible "noise" but are, in fact, fundamental to the entire energy cascade of the Earth system, collectively maintaining the stability and realism of the large-scale climate. Therefore, the superior long-term forecast performance of TritonCast stems directly from its novel hierarchical architecture, which accurately captures, maintains, and propagates these dynamically critical high-frequency, small-scale details.

\clearpage
\section{Training Strategy Scaling Law Exploration}

To identify an optimal training paradigm for the lightweight TritonCast (0.02b) model that balances both efficiency and accuracy, a series of preliminary experiments on scaling laws is conducted. These experiments systematically investigate the impact of two key dimensions: the time span of the training data and the amount of training computation (measured in total epochs) on the final model performance.

The results of these experiments are summarized in \textbf{Fig.~\ref{fig:scaling_law}}. The figure plots the Mean Squared Error (MSE) on the validation set for various configurations; a lower MSE value indicates better model performance. Two clear trends emerge from the analysis:

\begin{figure}[h!]
    \centering
    \includegraphics[width=0.85\textwidth]{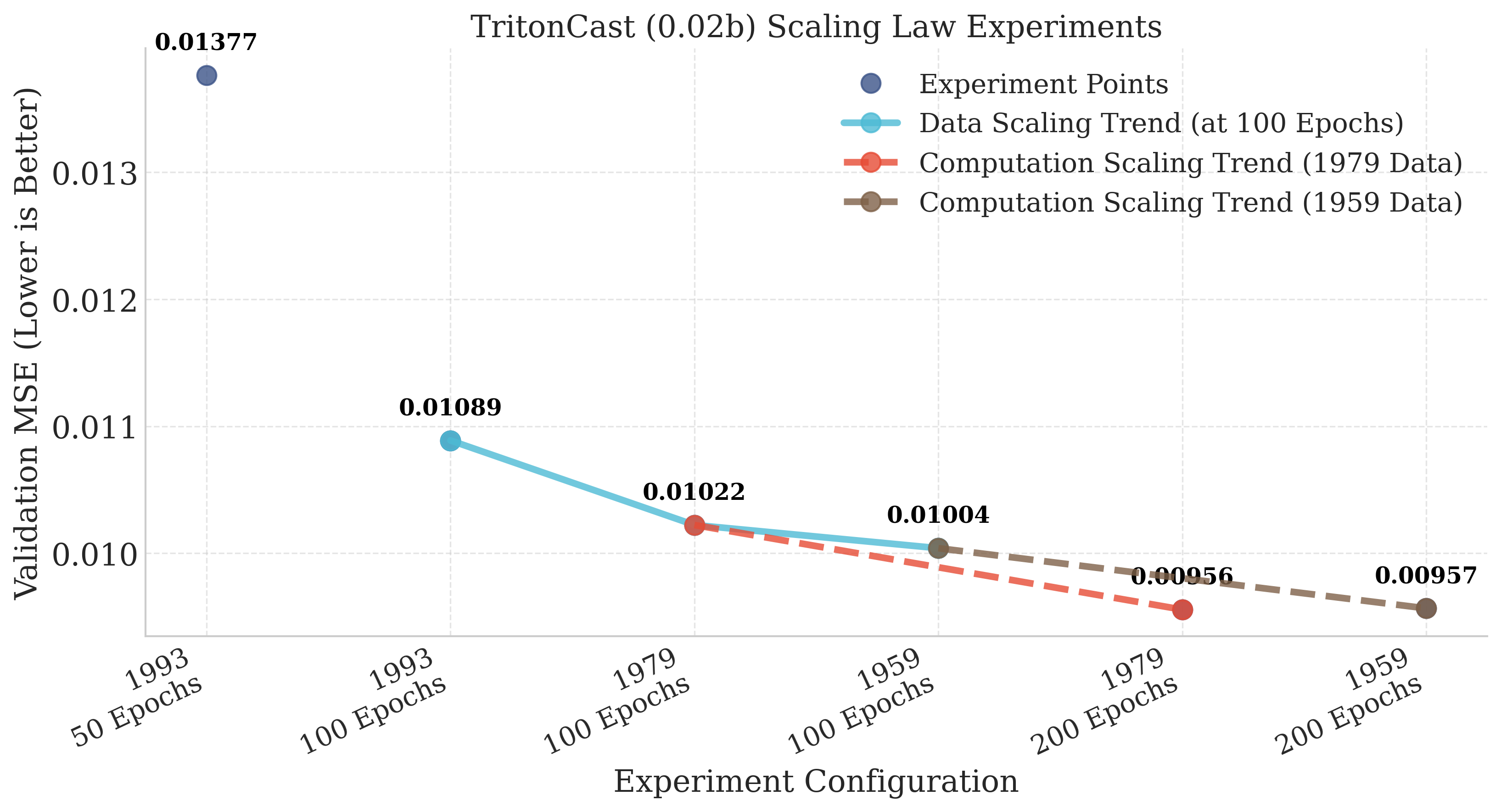}
    \caption{\textbf{Scaling law experiments for the TritonCast (0.02b) training strategy.} The figure illustrates the impact of different training data start years and numbers of epochs on the validation MSE. The solid line (blue) indicates the data scaling trend under a fixed budget of 100 epochs, while the dashed lines (red and brown) indicate the computation scaling trend for fixed datasets.}
    \label{fig:scaling_law}
\end{figure}

\textbf{Data Scaling Trend}:
As shown by the solid blue line in \textbf{Fig.~\ref{fig:scaling_law}}, under a fixed budget of 100 training epochs, extending the training data from a 1993 start year to a 1979 start year significantly reduces the validation MSE from 0.01089 to 0.01022. This indicates that increasing the amount of data, particularly to cover longer climate periods, is crucial for improving model performance. However, when the data is further extended from 1979 to 1959, the performance improvement diminishes sharply (MSE only drops from 0.01022 to 0.01004), exhibiting clear diminishing returns.

\textbf{Computation Scaling Trend}:
As illustrated by the dashed red and brown lines, increasing the training computation consistently yields significant performance gains. For the dataset starting from 1979, increasing the epochs from 100 to 200 reduces the MSE by approximately 6.5\%, from 0.01022 to 0.00956. A similar improvement is observed for the dataset starting from 1959. Notably, the result from training with 1979 data for 200 epochs (MSE 0.00956) is slightly better than that from training with more data (starting from 1959) for the same duration (MSE 0.00957). This may suggest that the data quality from earlier periods, or its divergence from modern climate patterns, could introduce minor noise into the model.

\textbf{Conclusion and Future Outlook}:
Based on this analysis, the conclusion is that for the TritonCast 0.02b model, training with \textbf{data from 1979 onwards for 200 epochs} is an effective strategy that strikes an optimal balance between data cost, computational expense, and final model accuracy. We emphasize that this is a preliminary exploration. Identifying the optimal trade-off between computational efficiency and model performance remains an important direction for future research, which is crucial for the operational deployment of AI models.

\clearpage
\section{Details for Tropical Cyclone Tracking}
\label{typhoon}

This appendix provides details on how the comparison against baselines for tropical cyclone tracking.

\subsection{Baseline and Ground-truth}

We compare \method{} with ECMWF-HRES, a strong cyclone tracking method based on high-resolution (9 km × 9 km) operational weather forecasting. The position of cyclone for ECMWF-HRES can be found in~\url{https://confluence.ecmwf.int/display/TIGGE/Tools#Tools-tc}. Due to our \method{} model for medium-range weather forecasting is 1.5 degree in spatial resolution, we use bilinear interpolation to interpolate it to 0.25 degrees. Specifically, due to our data (with height 120 and width 240) doesn't include Antarctica, we first use the last row of data to fill in the Antarctic region to create a global grid (with height 121 and width 240), and then use bilinear interpolation to a resolution of 0.25 degree (with height 721 and width 1440, 25 km × 25 km). And we use the results from the International Best Track Archive for Climate Stewardship (IBTrACS) project~\cite{knapp2010international,kenneth2019international} as the ground-truth , which contains the best available estimations for tropical cyclones.

\subsection{Algorithm for Tropical Cyclones Tracking}

We use the tropical cyclones tracking algorithm released by Aurora~\cite{bodnar2025foundation}, which is a multi-step procedure that resorts to Z700 when detection of the MSLP local minimum fails. This procedure relieves the issues that MSLP may be plagued by multiple local minima due to local topological features like mountains, thus improving the tacking accuracy. Further details refer to Aurora. The tracking step used in this work is 6 hours. Specifically, the tracking follows below procedure:
    \begin{itemize}
        \item When the tracker searches for the local minimum closest to a reference position within an $x^{\circ}\times x^{\circ}$ box, it first extracts an $x^{\circ}\times x^{\circ}$ region centered at the reference latitude--longitude. A grid point inside this region is classified as a \emph{local minimum} if it achieves the smallest value within an additional $2^{\circ}\times 2^{\circ}$ neighborhood centered at that point. The tracker then according to this classification to find the local minimum in the $x^{\circ} \times x^{\circ}$ box closest to the reference position. Any local minimum on the boundary is discarded, and if no interior local minimum remains, the step is deemed a failure and the tracker reports that no valid local minimum is found.
        \item The tracker is initialised with the current latitude--longitude position of the tropical cyclone (TC), which is taken as the reference position. If the surrounding $5^{\circ}\times 5^{\circ}$ region is free of land, the tracker first attempts to update the reference position using the closest local minimum of the 6 h MSLP forecast within this box. If no update is obtained, it repeats the same operation with progressively smaller boxes of $4^{\circ}\times 4^{\circ}$, $3^{\circ}\times 3^{\circ}$, $2^{\circ}\times 2^{\circ}$, and finally $1.5^{\circ}\times 1.5^{\circ}$. If the position still cannot be updated, the tracker attempts to update the position using the closest local minimum of the 6 h Z700 forecast within a $5^{\circ}\times 5^{\circ}$ box, and then repeats the same sequence of decreasing box sizes for MSLP. The resulting current position is the track forecast at lead time 6 h.
        \item For each subsequent lead time, the tracker first forms a guessed next-step latitude-longitude by linearly extrapolating the previously tracked positions using the last eight estimates (corresponding to two days). This extrapolated location is treated as the new current position. Using fields predicted a further 6 h into the future, the tracker then updates this current position by applying exactly the same update procedure as in the initialisation step. The updated position is recorded as the track estimate 6 h further ahead. This procedure is iterated for future position tracking.

    \end{itemize}

\subsection{Results for Tropical Cyclones Tracking}
\begin{figure}[h!]
    \centering
    \includegraphics[width=0.95\textwidth]{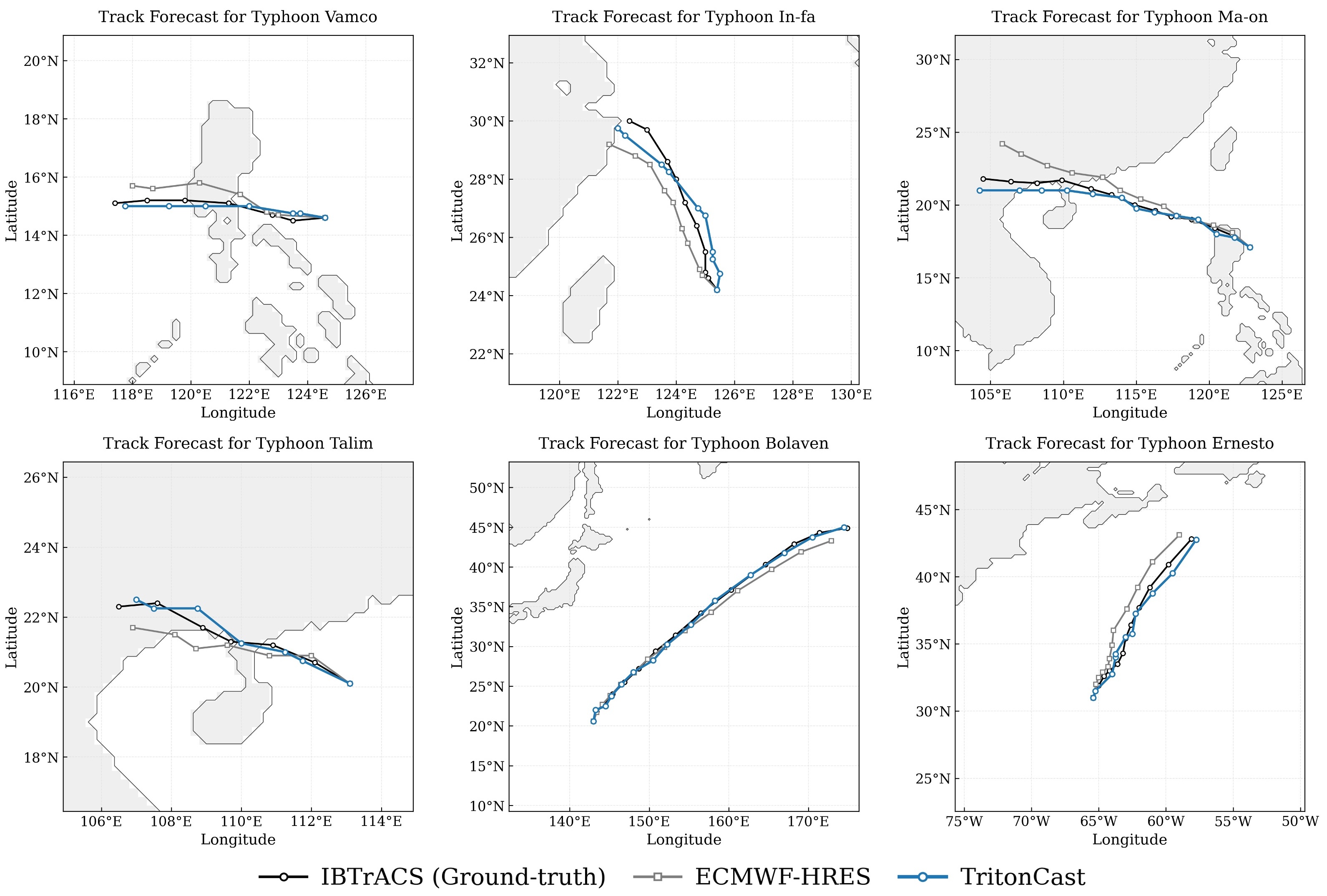} 
    \caption{\textbf{TritonCast achieves competitive performance in tropical cyclone tracking compared with ECMWF-HRES.}}
    \label{fig:typhoon}
\end{figure}

We assess the tropical cyclone tracking results of \method{} and ECMWF-HRES for serval representative typhoons, which ranges from 2020 to 2024. Both \method{} and HRES are initialized from 00:00 UTC. The initial condition for different typhoon can be found in Table~\ref{tab:typhoon}. As shown in Fig.~\ref{fig:typhoon}, although the spatial resolution of \method{} medium-range forecasting model (150 km × 150 km) is hugely lower than HRES (9 km × 9 km), it produces competitive tracking results compared with ECMWF-HRES. This demonstrates that \method{} balances the trade-off between efficiency and accuracy, and has potential for extreme event forecasting. Further, as mentioned in the Aurora~\cite{bodnar2025foundation}, the tracker used for cyclone tracking is much simpler than those used operationally, which slightly disadvantages \method{} in the comparison, as the ECMWF-HRES and IBTrACS use state-of-the-art trackers. Despite \method{} achieves satisfactory tracking results, we acknowledge that direct comparison between \method{} and ECMWF-HRES is somewhat unfair, because ECMWF-HRES uses the IFS initial condition data as its input, whereas \method{} uses reanalysis data.


\begin{table}[h!]
\centering
\caption{\textbf{Initial time for different tropical cyclones.}}
\label{tab:typhoon}
\setlength{\tabcolsep}{4.5pt}
\begin{tabular}{l|cccccc}
\toprule
Typhoon           & Vamco          & In-fa           & Ma-on          & Talim          & Bolaven        & Ernesto        \\ \midrule
Initial condition & 2020-11-11 & 2021-07-23 & 2022-08-23 & 2023-07-17 & 2023-10-12 & 2024-08-17 \\ \bottomrule
\end{tabular}
\end{table}

\clearpage
\end{appendices}

\end{document}